%% file: main.tex
\documentclass[Afour,sageh,times]{./style/sagej}

\usepackage{moreverb,url}

\usepackage[colorlinks,bookmarksopen,bookmarksnumbered,citecolor=red,urlcolor=red]{hyperref}

\input{preamble_ijrr}

\newboolean{blind}
\setboolean{blind}{true}%

\newcommand{\rmcaffiliation}{German Aerospace Center (DLR), Institute of Robotics and Mechatronics (RM), M\"unchener Str. 20, 82234 We\ss ling, Germany}
\newcommand{\tumaffiliation}{Technical University of Munich (TUM); School of Computation, Information and Technology; Boltzmannstr. 3, 85748 Garching, Germany}

\runninghead{Schneyer et al.}

\title{ErgoSurf: Ergodic Control for the Coverage of Unknown Surfaces}

\author{Stefan Schneyer\affilnum{1}\affilnum{2},
Timo Bachmann\affilnum{1},
Maged Iskandar\affilnum{1},
Korbinian Nottensteiner\affilnum{1},
Alin Albu-Schäffer\affilnum{1}\affilnum{2},
Freek Stulp\affilnum{1} and
Jo\~ao Silv\'erio\affilnum{1}}

\affiliation{\affilnum{1} \rmcaffiliation\\
\affilnum{2} \tumaffiliation}
\corrauth{Stefan Schneyer, Institute of Robotics and Mechatronics, German Aerospace Center (DLR), Münchener Str. 20, Wessling 82234, Germany.}
\email{Stefan.Schneyer@dlr.de}

\newcommand{\x}{\bm{x}}%
\newcommand{\y}{\bm{y}}%
\newcommand{\z}{\bm{z}}%
\newcommand{\q}{\bm{q}}%
\newcommand{\n}{\bm{n}}%
\renewcommand{\u}{u}%

\newcommand{\setS}{\mathcal{S}}
\newcommand{\setSt}{\mathcal{S}_k}

\newcommand{\R}{\mathbb{R}}
\newcommand{\D}{\mathcal{D}}

\newcommand{\ergTime}{t}

\newcommand{\iter}[1]{_{#1}}%

\begin{document}

\begin{abstract}
Contact-centric tasks on surfaces, ranging from inspection and cleaning to sanding and polishing, require robots to systematically cover the surface while maintaining stable contact.
Ergodic control generates trajectories that spend time at a location proportional to a desired, task-specific spatial distribution, enabling efficient information gathering and coverage.
However, traditional ergodic control methods rely on prior knowledge of surface geometry or require a vision sensory input to scan the geometry beforehand, limiting their applicability in real-world scenarios with unknown or dynamic environments.
This paper introduces a novel online ergodic control framework that achieves systematic surface coverage while simultaneously reconstructing unknown surface geometry.
We employ a Gaussian Process Implicit Surface (GPIS) model that learns global surface geometry from intrinsic tactile sensing during execution.
For efficient online planning, we approximate the surface locally using point clouds sampled from tangent planes at observed contact points and iteratively fit them to the Gaussian Process.
This approximation simultaneously serves as the sampling domain for both the target and the coverage distributions.
We employ a heat-diffusion analogy to compute potential fields that guide ergodic exploration, translating spatial coverage objectives into smooth robot trajectories.
We demonstrate our framework through simulation and real-robot experiments, validating simultaneous ergodic coverage and online surface geometry learning with reconstruction errors approaching the ground truth.
\end{abstract}


\maketitle

\section{Introduction}
In many practical applications, from surface finishing to inspection and maintenance tasks, robots must operate on surfaces whose exact geometry is unavailable beforehand.
Systematically covering a surface is a core requirement of such tasks, and ergodic control~\citep{Mathew2011, Ivic2017} provides a principled framework for it, generating trajectories that distribute coverage effort according to task-specific spatial objectives rather than relying on heuristic approaches such as spiral patterns or random walks.
This approach has proven effective across diverse domains, including search problems~\citep{Miller2016}, manipulation~\citep{Shetty2022, Sun2024, Bilaloglu2023}, and surface cleaning~\citep{Bilaloglu2025}.
As a concrete example, consider an operator preparing a surface finishing or inspection task on a workpiece for which no CAD model or prior scan is available.
Rather than programming a trajectory offline, the operator kinesthetically guides the robot by hand to a few points, delimiting the workspace region to be treated and marking where coverage effort should concentrate.
These taught points specify the task directly in workspace coordinates, independent of the still-unknown surface, while the robot discovers the geometry through tactile contact and adapts its coverage online.
However, conventional ergodic control methods assume that surface geometry is known beforehand, treating geometric learning and trajectory generation as separate offline phases.
While vision-based approaches~\citep{Berger2017} can provide geometric information, they suffer from well-known limitations, including occlusions, transparent materials, specular reflections, and varying lighting conditions, which compromise reconstruction quality.
Applying ergodic control to unknown surfaces creates a circular dependency: planning an ergodic exploration strategy requires knowledge of the surface geometry, yet acquiring that geometry requires executing an exploration strategy---a problem that offline planning cannot resolve.
Instead, robots must learn geometry and control exploration concurrently, adapting decisions online as new information becomes available.
Beyond the necessity of online adaptation, several additional challenges arise.
Real-time control demands computational efficiency---surface learning and planning must remain tractable at each time step.
Furthermore, sensor measurements are inherently noisy and incomplete, requiring robust estimation that accounts for uncertainty without overfitting to sparse or ambiguous observations.
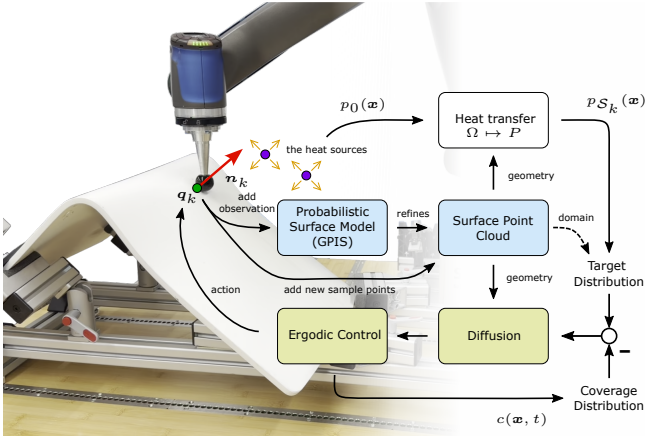
\begin{figure}
	\centering
	\begin{tikzpicture}
		\node[anchor=north west, inner sep=0] (image) at (0,0) {
			\includegraphics[width=\columnwidth]{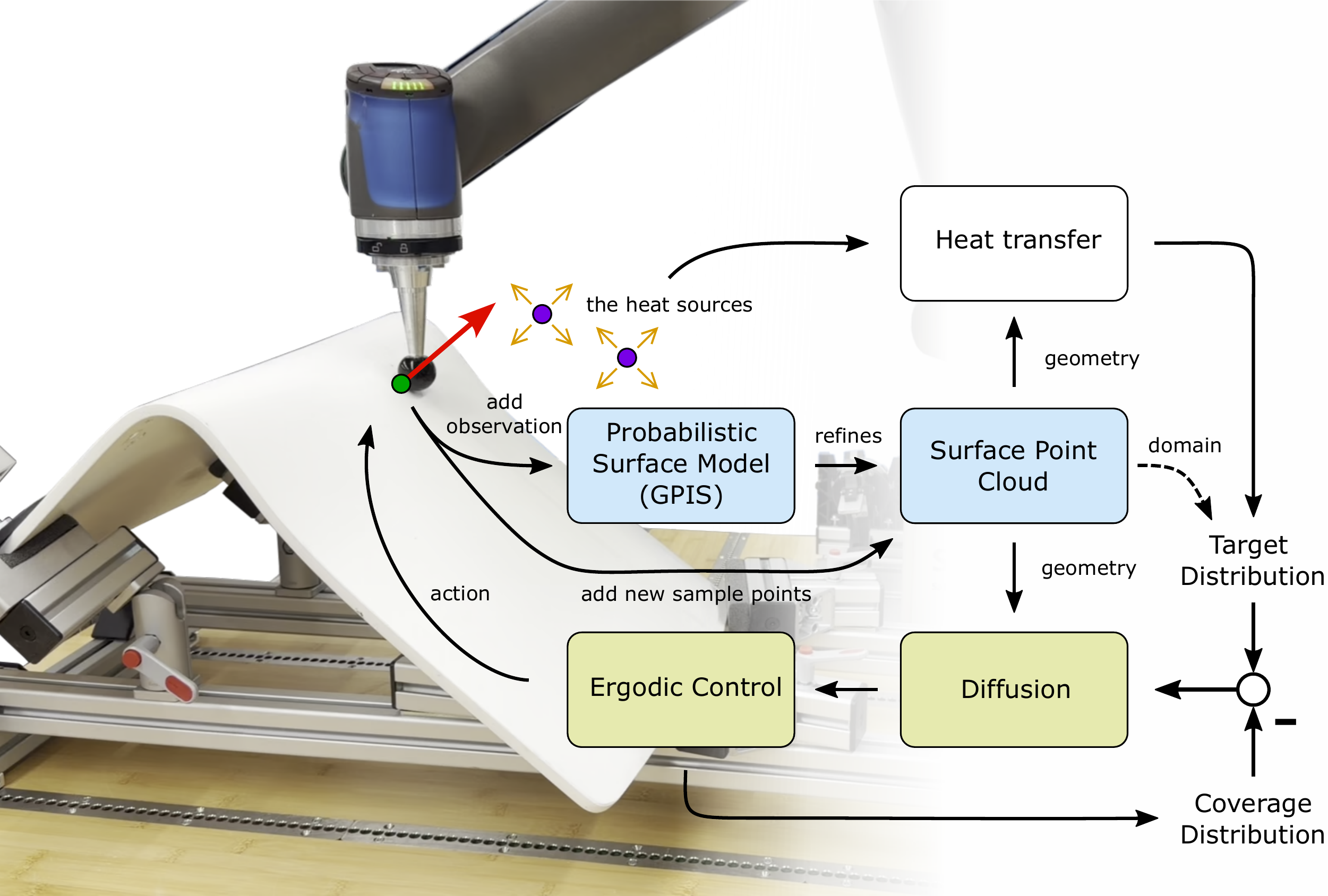}
		};
		\begin{scope}[x={(image.north east)}, y={(image.south west)}]
			\node[black, text opacity=1, fill=white, fill opacity=0, anchor=north west, font=\tiny] at (1130.23/1600, 290/1080) {$\Omega \mapsto P$};%
		\end{scope}
		\begin{scope}[x={(image.north east)}, y={(image.south west)}]
			\node[black, text opacity=1, fill=white, fill opacity=0, anchor=north west, font=\tiny] at (410/1600, 450/1080) {$\q_k$};%
		\end{scope}
		\begin{scope}[x={(image.north east)}, y={(image.south west)}]
			\node[black, text opacity=1, fill=white, fill opacity=0, anchor=north west, font=\tiny] at (530/1600, 410/1080) {$\n_k$};%
		\end{scope}
		\begin{scope}[x={(image.north east)}, y={(image.south west)}]
			\node[black, text opacity=1, fill=white, fill opacity=0, anchor=north west, font=\tiny] at (816/1600, 215/1080) {$p_0(\x)$};%
		\end{scope}
		\begin{scope}[x={(image.north east)}, y={(image.south west)}]
			\node[black, text opacity=1, fill=white, fill opacity=0, anchor=north west, font=\tiny] at (1430/1600, 205/1080) {$p_{\setSt}(\x)$};%
		\end{scope}
		\begin{scope}[x={(image.north east)}, y={(image.south west)}]
			\node[black, text opacity=1, fill=white, fill opacity=0, anchor=north west, font=\tiny] at (1205/1600, 1000/1080) {$c(\x, t)$};%
		\end{scope}
	\end{tikzpicture}
	\caption{Framework combining \colorbox{dlrgreen5}{ergodic control} and \colorbox{dlrblue5}{online surface learning}. Tactile contact observations (contact points $\q_k$ and normals $\n_k$) incrementally refine a probabilistic surface model. The user specifies task goals as heat sources $p_0(\x)$ in workspace coordinates, independent of the unknown geometry; these radiate onto the current surface estimate to form the target distribution $p_{\setSt}(\x)$. Heat diffusion on the learned geometry then drives the coverage distribution $c(\x, t)$ toward high-target, low-coverage regions.}%
	\label{fig:visual_abstract}%
\end{figure}

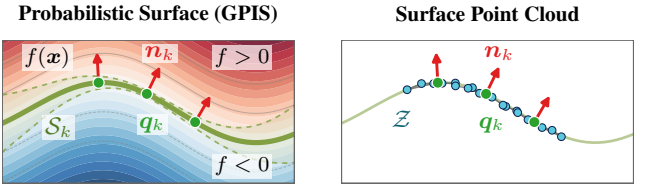
\begin{figure}[t]
	\centering
	\begin{tikzpicture}
		\tikzset{figlbl/.style={anchor=center, font=\small, fill=white, fill opacity=0.7, text opacity=1, inner sep=1pt}}
		\tikzset{figttl/.style={anchor=center, align=center, font=\footnotesize\bfseries, inner sep=1pt}}
		\node[anchor=south west, inner sep=0] (img) at (0,0)
			{\includegraphics[width=\columnwidth]{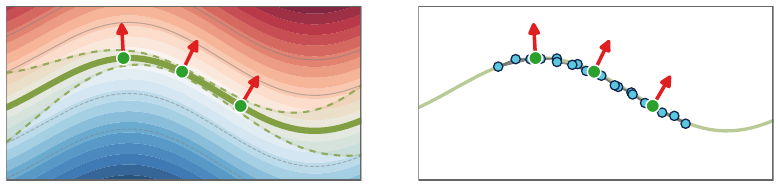}};
		\begin{scope}[x={(img.south east)}, y={(img.north west)}]
			\node[figttl] at (0.236,1.13) {Probabilistic Surface (GPIS)};
			\node[figttl] at (0.764,1.13) {Surface Point Cloud};
			\node[figlbl, font=\footnotesize] at (0.079,0.852) {$f(\x)$};
			\node[figlbl] at (0.097,0.399) {\textcolor[HTML]{4F7A1F}{$\setSt$}};
			\node[figlbl, font=\footnotesize] at (0.382,0.148) {$f<0$};
			\node[figlbl, font=\footnotesize] at (0.382,0.852) {$f>0$};
			\node[figlbl] at (0.630,0.455) {\textcolor[HTML]{0E5F70}{$\mathcal{Z}$}};
			\node[figlbl] at (0.254,0.879) {\textcolor[HTML]{E02020}{$\n_k$}};
			\node[figlbl] at (0.242,0.416) {\textcolor[HTML]{2CA02C}{$\q_k$}};
			\node[figlbl] at (0.782,0.879) {\textcolor[HTML]{E02020}{$\n_k$}};
			\node[figlbl] at (0.770,0.416) {\textcolor[HTML]{2CA02C}{$\q_k$}};
		\end{scope}
	\end{tikzpicture}
	\caption{Dual surface representation: \emph{Left}: a probabilistic GPIS, a continuous implicit function $f(\x)$ (negative inside, positive outside) with zero-level set $\setSt$ and a dashed uncertainty band. \emph{Right}: a discrete point cloud $\mathcal{Z}$, sampled in the contact tangent planes and projected onto the surface, forming the ergodic control domain. Both cover the observed surface, differing as a continuous model versus a discrete point set.}%
	\label{fig:dual_representation}%
\end{figure}

To overcome these limitations, we propose a dual representation approach that combines the strengths of global and local surface modeling, see \Cref{fig:visual_abstract} and \Cref{fig:dual_representation}.
As illustrated in \Cref{fig:visual_abstract}, the user specifies the task goal as heat sources $p_0(\x)$ in fixed workspace coordinates, independent of the unknown geometry.
These sources remain fixed while the target distribution $p_{\setSt}(\x)$ they induce on the surface changes as the surface estimate $\setSt$ is progressively refined from tactile contacts, so that the coverage objective continuously adapts to the current geometry.
For global surface modeling, we use a Gaussian Process Implicit Surface (GPIS)~\citep{Williams2007} to continuously learn the surface geometry from tactile sensing, providing a coherent, uncertainty-aware estimate of the entire surface.
For local surface modeling, we compute local tangent plane approximations at observed contact points, which provide a sampling domain for ergodic control guidance. 
This two-level representation (\Cref{fig:dual_representation}) enables fast, robust ergodic control updates while avoiding the computational expense of global surface inference.
Building on the Heat Equation Driven Area Coverage (HEDAC) \citep{Ivic2017} ergodic control method, we employ heat diffusion to compute potential fields that guide systematic exploration while respecting surface geometry and maintaining contact constraints.
We review related work on ergodic control and surface modeling in Section~\ref{sec:related_work} and provide background on Gaussian processes and the HEDAC ergodic control framework in Section~\ref{sec:background}.
\noindent
\paragraph*{Key Contributions}
To our knowledge, this is the first ergodic control framework that operates on a domain that is itself being reconstructed online; existing ergodic methods assume the domain (rectangle, mesh, or pre-scanned surface) is known before execution.
The contributions that make this possible are:
\begin{enumerate}
  	\item A framework that combines online surface reconstruction with ergodic control, enabling concurrent geometric learning and task-specific coverage without offline planning or prior geometric knowledge (see \Cref{fig:visual_abstract}).

  	\item A dual surface representation (see \Cref{fig:dual_representation}): a Gaussian Process Implicit Surface (GPIS) provides a global probabilistic model with surface normals, and local tangent plane samples iteratively projected onto the GPIS zero-level set form a lightweight discrete domain for ergodic control; both are built from fully dynamically decoupled contact wrenches (Sections~\ref{sec:contact_obs}, \ref{sec:gpis_surface}, and \ref{sec:local_surface}).

  	\item A two-stage potential field formulation that lets users specify task-relevant regions in workspace coordinates before the surface geometry is known: a first stage spreads the user-defined goals from $\R^3$ onto the current surface estimate to obtain a surface-supported target distribution, and a second stage diffuses this target along the surface to drive ergodic control (Sections~\ref{sec:potential_field}).

  	\item A direct point cloud gradient via local linear least squares on unstructured neighbors (Section~\ref{sec:gradient}), together with a selective coverage update that recomputes only newly sampled or projected points and updates static points incrementally (Section~\ref{sec:erg_ctl}).
\end{enumerate}

We further validate our approach experimentally in simulation and on real robot hardware (Section~\ref{sec:experiments}).
Section~\ref{sec:discussion} discusses impact, limitations, and directions for future work. 
Finally, Section~\ref{sec:conclusion} concludes with a summary of our work.

\section{Related Work}
\label{sec:related_work}

\subsection{Ergodic Control}
A principled approach to designing coverage tasks is provided through ergodic control, which ensures that time-averaged spatial statistics of a trajectory match a desired spatial distribution~\citep{Mathew2011}.
This property distinguishes ergodic methods from heuristic strategies such as spiral motions~\citep{Shetty2022} or random walks~\citep{Ketchum2024}, which lack guarantees on coverage quality.
Applications span mobile robot exploration~\citep{Miller2016, Dong2023, Sartoretti2022, Lanca2025}, energy-aware planning~\citep{Seewald2024}, rehabilitation robotics~\citep{Fitzsimons2022}, active learning~\citep{Kalinowska2021, Ketchum2024}, shape estimation~\citep{Abraham2017}, manipulation~\citep{Shetty2022, Sun2024}, surface cleaning~\citep{Bilaloglu2025, Sun2025}, surface finishing~\citep{Schneyer2026}, and drawing~\citep{Loew2022, Sun2025}.
Various approaches to realize ergodic control have been developed over the years.

The Spectral Multiscale Coverage (SMC) method~\citep{Mathew2011} utilizes Fourier series over rectangular domains to match time-averaged agent trajectories with spatial distributions, decomposing them into spectral components across multiple scales~\citep{Mathew2009, Calinon2020}.
Extensions include heterogeneous multi-agent teams with Gaussian footprints~\citep{Sartoretti2022}, arbitrary and time-varying tool imprints for surface finishing~\citep{Schneyer2026}, and volumetric ergodic control~\citep{Kwon2025}, which quantifies the efficiency gains from footprint awareness compared to point-mass models.

The Heat Equation Driven Area Coverage (HEDAC) method~\citep{Ivic2017} offers an alternative approach to SMC by formulating coverage as a heat diffusion process, where potential fields guide agents toward underexplored regions.
Extensions of HEDAC have addressed whole-body coverage by decomposing the robot's body into multiple kinematically constrained agents for tactile exploration~\citep{Bilaloglu2023} and for applications to curved-surface tasks such as cleaning~\citep{Bilaloglu2025}.

Trajectory optimization methods compute entire paths before execution, with formulations addressing ergodic constraints via iLQR \citep{Miller2013}, minimum-time objectives \citep{Dong2023, Seewald2024}, arbitrary domains \citep{Hughes2025}, and flow-matching metrics \citep{Sun2025}.

A common assumption across the abovementioned ergodic control methods is that the state space geometry is fully known prior to execution. 
In contrast, our work removes the requirement for prior geometric knowledge: we simultaneously learn the surface geometry while performing ergodic coverage, enabling deployment on unknown surfaces without offline reconstruction or pre-scanning, which delays deployment and increases operational overhead.

\subsection{Surface Modeling and Reconstruction}
Surface reconstruction from point data has been extensively studied, spanning classical geometric methods~\citep{Berger2017,Curless1996,Kazhdan2006} to recent deep learning approaches~\citep{Farshian2024}.
Implicit representations define surfaces as level sets of continuous functions.
When combined with Gaussian Processes (GPs)~\citep{Rasmussen2006}, they provide uncertainty quantification, smoothness determined by kernel choice, and closed-form gradients for computing surface normals and curvature.
Gaussian Process Implicit Surfaces (GPIS)~\citep{Williams2007} model surfaces as the zero-isolevel of a scalar field.
\citep{Dragiev2011} applied GPIS to robotic shape estimation and grasping, fusing visual, tactile, and laser data, while GPs allow incorporating geometric priors to handle noisy data~\citep{Martens2017}.
Tactile exploration offers distinct advantages over vision-based reconstruction, as it avoids common issues such as occlusions, transparency, and reflectivity.
\citet{Abraham2017} demonstrated that ergodic exploration enables shape estimation with low-resolution contact sensors, exploiting non-contact motion to infer shape boundaries.
\citet{Yi2016} proposed an active touch strategy using GPIS variance to guide discrete touch-and-retract motions.
\citet{Driess2017} introduced an active learning framework for continuous sliding along informative paths using GPIS, later extended~\citep{Driess2019} to multiple end-effectors.
\citet{khadivar2023} adapts GPIS hyperparameters and optimizes hand pose based on local surface complexity.
\citet{Zhao2026} proposed a dual GPIS architecture combining local and global uncertainty-aware policies.

In contrast to these uncertainty-driven methods, our approach decouples the coverage objective from surface reconstruction, allowing operators to directly specify task-specific distributions while surface reconstruction emerges as a byproduct.

\section{Background}
\label{sec:background}

\subsection{Gaussian Process (GP)}

To model functions with uncertainty quantification in a principled probabilistic framework, we employ a \emph{Gaussian Process (GP)} \citep{Rasmussen2006}.
A GP is a non-parametric regression approach that defines a distribution over functions.
Rather than specifying a function $f(\x): \R^n \mapsto \R$ explicitly, a GP characterizes it through the property that any finite collection of function values is jointly Gaussian.
Specifically, for inputs $\mathbf{X} = \{\x_{(1)}, \dots, \x_{(n)}\}$, the corresponding function values $f_{(i)} := f(\x_{(i)})$ follow a multivariate normal distribution
\begin{equation}
	f(\mathbf{X}) = \{f_{(1)}, \dots, f_{(n)}\} \sim \mathcal{N}(\mathbf{m}, \mathbf{K}),
\end{equation}
where $\mathbf{m}$ is the mean vector with entries $m_{i} = \mathbb{E}[f_{(i)}]$ and $\mathbf{K}$ is the covariance matrix with entries $K_{ij} = k(\x_{(i)}, \x_{(j)}) = \text{Cov}(f_{(i)}, f_{(j)})$.
Compactly, we write $f(\x) \sim \mathcal{GP}(m(\x), k(\x, \x'))$.
The behavior of the GP depends entirely on the choice of kernel function $k: \R^n \times \R^n \mapsto \R$, which encodes structural assumptions such as smoothness, periodicity, and correlation length scales.
To account for sensor noise, observations are modeled as $y_{(i)} = f(\x_{(i)}) + \epsilon_{(i)}$, where $\epsilon_{(i)} \sim \mathcal{N}(0, \sigma_n^2)$ represents independent measurement noise.
Given a training set $\D = (\mathbf{X}, \mathbf{y})$ consisting of $n$ observations, where $\mathbf{X}$ are the input points and $\mathbf{y} = \{y_{(1)}, \dots, y_{(n)}\}$ are the corresponding outputs, we can perform inference at a new test point $\x_*$.
The joint distribution of the observed outputs and the function value at the test point is Gaussian, given by
\begin{equation}
	\begin{bmatrix} \mathbf{y} \\ f(\x_*) \end{bmatrix} \sim \mathcal{N}\left(\begin{bmatrix} \mathbf{m} \\ m(\x_*) \end{bmatrix}, \begin{bmatrix} \mathbf{K} + \sigma_n^2 \mathbf{I} & \mathbf{k}_* \\ \mathbf{k}_*^\top & k(\x_*, \x_*) \end{bmatrix}\right),
\end{equation}
where $\mathbf{k}_*$ is the vector with entries $k(\x_{(i)}, \x_*)$.
By conditioning on the observations, the predictive posterior distribution at $\x_*$ is Gaussian, providing both a mean estimate and uncertainty quantification, expressed as
\begin{equation}
	\begin{aligned}
		f(\x_*) \mid \D &\sim \mathcal{N}(\mu_*, \sigma^2_*),\\
		\mu_* &= m(\x_*) + \mathbf{k}_*^\top (\mathbf{K} + \sigma_n^2 \mathbf{I})^{-1} (\mathbf{y} - \mathbf{m}),\\
		\sigma^2_* &= k(\x_*, \x_*) - \mathbf{k}_*^\top (\mathbf{K} + \sigma_n^2 \mathbf{I})^{-1} \mathbf{k}_*.
	\end{aligned}
\end{equation}
\subsection{GP Implicit Surface Representation}
\label{sec:gpis}
We define an implicit surface $\setS \subset \R^3$ as the zero level set of a function $F : \R^3 \mapsto \R$
\begin{equation}
	\setS = \{ \x \in \R^3 \mid F(\x) = 0 \} \text{.}
	\label{eq:def_implicit_function}
\end{equation}
The implicit function $F(\x)$ behaves similarly to a signed distance function (SDF): it assigns positive values to points outside the object, negative values to points inside, and zero on the surface itself.
$F$ is not a true SDF, as we do not enforce the Eikonal constraint $\|\nabla F\| = 1$.
The gradient $\nabla F(\x)$ provides the normal direction at any point, computed as
\begin{equation}
	\n(\x) = \frac{\nabla F(\x)}{\| \nabla F(\x) \|}.
	\label{eq:gpis_normal_def}
\end{equation}

Following the Gaussian Process Implicit Surface (GPIS) framework~\citep{Williams2007, Dragiev2011}, we learn the unknown implicit function $F$ using a Gaussian Process, which provides a probabilistic approximation $f(\x)$ to the true surface function.
The GP models the posterior distribution $p(f(\x) \mid \D)$ conditioned on observations $\D = \{(\x_{(i)}, y_{(i)})\}$ of the implicit surface function.

When derivative observations are available, such as surface normals, GPs naturally incorporate them through the covariance structure.
The key is that, by the linearity of differentiation, the covariance between function values and their partial derivatives can be computed by differentiating the kernel function.
Specifically, the cross-covariance between function values at $\x_{(i)}$ and gradients at $\x_{(j)}$ is
$\text{Cov}(f_{(i)}, \nabla_{\x_{(j)}}  f_{(j)}) = \nabla_{\x_{(j)}} k(\x_{(i)}, \x_{(j)})$,
while the covariance between gradient observations at $\x_{(i)}$ and $\x_{(j)}$ is
$\text{Cov}(\nabla_{\x_{(i)}} f_{(i)}, \nabla_{\x_{(j)}} f_{(j)}) = \frac{\partial^2 k(\x_{(i)}, \x_{(j)})}{\partial \x_{(i)} \partial \x_{(j)}}$.
In our framework, we exploit this property to condition the GP directly on the surface normals obtained from tactile contact, constraining the gradient of the implicit function and thereby yielding a more accurate surface estimate from few observations.

Beyond conditioning on derivative observations, the gradient of the GP prediction itself can be computed.
Under the Gaussian Process model, both the function value and its spatial derivatives follow a joint Gaussian distribution, and the mean of the gradient at any point can be computed from the kernel function as
\begin{equation}
	\mathbb{E}[\nabla f(\x_*)] = \nabla m(\x_*) + \nabla \mathbf{k}_*^\top (\mathbf{K} + \sigma_n^2 \mathbf{I})^{-1} (\mathbf{y} - \mathbf{m}),
	\label{eq:gp_gradient_mean}
\end{equation}
where $\nabla$ denotes the gradient with respect to $\x_*$.

\subsection{HEDAC Ergodic Control}
The Heat Equation Driven Area Coverage (HEDAC) method \citep{Ivic2017} provides a principled approach to ergodic control by formulating coverage as a heat diffusion process. 
The method defines a virtual source term
\begin{equation}{\label{eq:virtual_source}}
	s(\x, \ergTime) = \max{\big(p(\x) - c(\x, \ergTime),0\big)}^2
\end{equation}
that captures the discrepancy between desired and achieved coverage, where $p(\x)$ is the target coverage probability distribution over a domain space $\Omega$ with $\int_{\Omega} p(\x) \, d\x = 1$, and $c(\x, \ergTime)$ is the coverage distribution at time $t$ over the domain with $\int_{\Omega} c(\x, \ergTime) \, d\x = 1$.
The coverage distribution is computed by normalizing the cumulative coverage through $c(\x, \ergTime)=\frac{\tilde{c}(\x, \ergTime)}{\int_{\Omega} \tilde{c}(\x, \ergTime) d \x}$, where $\tilde{c}(\x, \ergTime)$ represents the time-averaged sum of convolutions between the agent's footprint $\varphi(\bm{r})$ and the trajectory $\x_a(t)$ for a single agent, given by
\begin{equation}{\label{eq:coverage}}
	\tilde{c}(\x, \ergTime)=\frac{1}{\ergTime} \int_0^\ergTime \varphi\big(\x-\x_a(t^\prime)\big) d t^\prime.
\end{equation}
HEDAC computes the potential field $u(\x, \ergTime)$ by diffusing the source term through the stationary heat equation with sink term
\begin{equation}{\label{eq:HEDAC}}
	\alpha \Delta u(\x, \ergTime) - \beta u(\x, \ergTime) + s(\x, \ergTime) = 0,
\end{equation}
where $\alpha>0$ is the diffusion coefficient, $\beta > 0$ is the sink rate, and $\Delta$ is the Laplacian operator. 
At $\partial \Omega$, the zero-Neumann boundary condition $\n(\x) \cdot \nabla u(\x, \ergTime)=0, \quad \forall \x \in \partial \Omega$ is imposed, where $\n(\x)$ represents the outward unit normal vector to the boundary.
The smooth gradient field of the diffused potential $u(\x, \ergTime)$ provides control commands through first-order dynamics $\dot{\x}_a = v \frac{\nabla u(\x_a, \ergTime)}{\|\nabla u(\x_a, \ergTime)\|}$, where $v$ is the desired velocity magnitude.
The ergodicity measure $\epsilon(t)$ quantifies the total coverage deficit as the integral of the source term over the domain
\begin{equation}
	\epsilon(t) = \int_{\Omega} s(\x, t) \, d\x,
	\label{eq:ergodicity}
\end{equation}
which represents the total deficit between the target and actual coverage distributions.

When extending HEDAC ergodic control to curved surfaces, the Euclidean Laplacian $\Delta$ must be replaced by the Laplace-Beltrami operator (LBO) to respect the intrinsic geometry of the manifold~\citep{Bilaloglu2025}.
On triangular meshes, the LBO can be approximated via a graph Laplacian, $-\Delta u(\x_i) \approx (\mathbf{L}\mathbf{u})_i$, which computes weighted differences between a vertex's potential and those of its neighbors.
The simplest discretization uses uniform weights based solely on connectivity, whereas geometry-aware methods~\citep{Crane2013, Cao2010, Liu2012, Sharp2020} account for variations in triangle shapes and sizes.
An intuitive 1D example illustrating the connection between the continuous and discrete Laplacian is presented in \Cref{app:discrete_laplacian}.

\section{Proposed Approach}
\label{sec:proposed_approach}
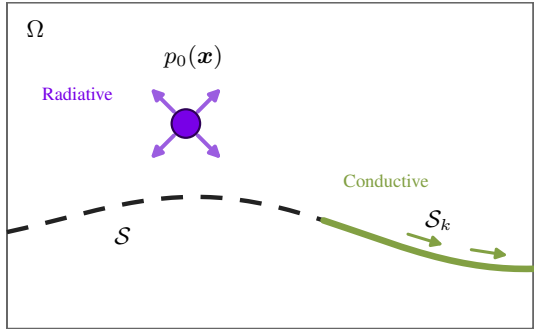
\begin{figure}[t]
	\centering
	\begin{tikzpicture}
		\definecolor{radpurple}{HTML}{7800e7}%
		\tikzset{figlbl/.style={anchor=center, font=\small, fill=white, fill opacity=0.7, text opacity=1, inner sep=1pt}}
		\node[anchor=south west, inner sep=0] (img) at (0,0)
			{\includegraphics[width=0.82\columnwidth]{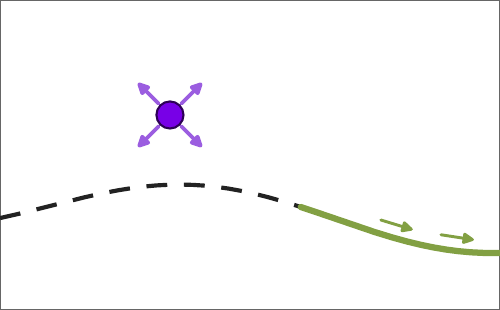}};
		\begin{scope}[x={(img.south east)}, y={(img.north west)}]
			\node[figlbl] at (0.055,0.919) {$\Omega$};%
			\node[figlbl] at (0.355,0.831) {$p_0(\x)$};%
			\node[figlbl, font=\scriptsize, text=radpurple] at (0.135,0.718) {Radiative};%
			\node[figlbl, font=\scriptsize, text=dlrgreen1] at (0.720,0.453) {Conductive};%
			\node[figlbl] at (0.221,0.285) {$\setS$};%
			\node[figlbl] at (0.820,0.332) {$\setSt$};%
		\end{scope}
	\end{tikzpicture}
	\caption{Task setup: the robot explores an unknown surface $\setS$ within the domain $\Omega$; the observed portion $\setSt$ (green) grows over time. A user-specified heat source $p_0(\x)$ radiates onto the surface, while heat also conducts along the observed surface.}%
	\label{fig:domain}%
\end{figure}

This section presents ErgoSurf, our ergodic control approach on surfaces with unknown geometry.
Let $\Omega \subset \mathbb{R}^3$ denote the domain space of interest (see \Cref{fig:domain}), and let the agent's task space be represented as the surface set $\setS \subset \Omega$.
For in-contact tasks, $\setS$ comprises all feasible contact points within $\Omega$, constraining the robot's motion to this subset during environment interaction.
Since $\setS$ is initially unknown, we propose an ergodic control framework that integrates two core components (see \Cref{fig:visual_abstract}):
(i) surface coverage by generating robot actions that track a probability distribution on arbitrary surfaces, and
(ii) simultaneous reconstruction of surface-shape models that define the ergodic agent's task space.

We adopt HEDAC~\citep{Ivic2017} as the underlying ergodic control formulation because, unlike Spectral Multiscale Coverage~\citep{Mathew2011} (fixed Fourier basis on a rectangular domain) or trajectory-optimization formulations~\citep{Miller2013} (require the full geometry and plan offline), HEDAC produces a per-step gradient on the current domain and extends to curved surfaces via the Laplace--Beltrami operator~\citep{Bilaloglu2025}.
Since the surface geometry is unknown, we first describe how contact points and normals are obtained from force measurements (\Cref{sec:contact_obs}), and then introduce a Gaussian Process Implicit Surface (GPIS) representation for incremental reconstruction from these contact observations (\Cref{sec:gpis_surface}).
To enable tractable control computation, we then describe local tangent plane sampling with projection onto the learned surface (\Cref{sec:local_surface}), and finally present the heat-diffusion-based ergodic controller that generates coverage-optimal motions on this evolving representation (Sections~\ref{sec:erg_ctl}, \ref{sec:potential_field}, and \Cref{sec:gradient}).
\paragraph{Time and indexing convention}
Throughout, we distinguish continuous time from discrete control steps.
Continuous time is denoted $t$, while the control loop advances in discrete steps indexed by $k \in \mathbb{N}$, where $t_k$ is the continuous time at step $k$ and $t_{k+1} = t_k + \Delta t$ for a fixed step size $\Delta t$.
Any time-varying quantity $x_t$ evaluated at a control step is written $x_k := x_{t_k}$.

\subsection{Contact Observation}
\label{sec:contact_obs}

During exploration, the robot maintains contact with the surface using a Cartesian impedance controller with force overlay. 
The ergodic controller commands the tangential end-effector motion along the surface, while a constant force component in the surface-normal direction ensures stable contact. 
Building the surface model requires a robust estimate of the contact force at the end-effector, but the measured joint torques also reflect the robot's own dynamics---gravity, inertia, and Coriolis effects.
To recover the contact wrench, a full dynamical decoupling method is employed to isolate the contact-induced generalized forces.
This is achieved using a momentum-based observer~\citep{iskandar2021collision} that does not require joint-acceleration measurements.
It interprets any deviation between the generalized momentum's expected and measured evolution as an external generalized force, yielding a residual
\begin{equation}
    \bm{r}(t)
    =
    \bm{K}_o
    \left(
        \bm{p}(t)
        -
        \int_{0}^{t}
        \left(
            \bm{\tau}
            -
            \bm{h}
            +
            \bm{r}
        \right)
        \mathrm{d}t
        -
        \bm{p}(0)
    \right),
\end{equation}
where $\bm{r}\in\mathbb{R}^{m}$ is the estimated external generalized-force vector, $\bm{K}_o$ is a positive diagonal observer-gain matrix, $\bm{p}=\bm{B}(\bm{\theta})\dot{\bm{\theta}}$ is the generalized momentum, $\bm{\theta}\in\mathbb{R}^{n}$ are the joint positions and $\bm{B}(\bm{\theta})$, and $\bm{\tau}$ is the sensed-force vector.
The term
\begin{equation}
    \bm{h}(\bm{\theta},\dot{\bm{\theta}})
    =
    \bm{g}(\bm{\theta})
    +
    \bm{C}(\bm{\theta},\dot{\bm{\theta}})\dot{\bm{\theta}}
    -
    \dot{\bm{B}}(\bm{\theta},\dot{\bm{\theta}})
    \dot{\bm{\theta}},
\end{equation} 
collects the gravity $\bm{g}(\bm{\theta})$, Coriolis/centrifugal $\bm{C}(\bm{\theta},\dot{\bm{\theta}})$, and time derivative of the inertia matrix $\dot{\bm{B}}(\bm{\theta},\dot{\bm{\theta}})$ contributions.
By incorporating the end-effector force--torque sensor measurements into the momentum observer, the external end-effector wrench is obtained directly as
\begin{equation}
    \bm{r}_{\mathrm{ee}}
    =
    \begin{bmatrix}
        \hat{\bm{f}}_{\mathrm{ee}}^{\mathrm{ext}} \\
        \hat{\bm{m}}_{\mathrm{ee}}^{\mathrm{ext}}
    \end{bmatrix},
	\label{eq:ext_ee_wrench}
\end{equation}
where $\hat{\bm{f}}_{\mathrm{ee}}^{\mathrm{ext}}\in\mathbb{R}^{3}$ and $\hat{\bm{m}}_{\mathrm{ee}}^{\mathrm{ext}}\in\mathbb{R}^{3}$ denote the estimated contact force and moment, respectively.
Since no Jacobian inversion is required, the wrench estimate is independent of the robot configuration. Here, $\bm{r}_{\mathrm{ee}}\in\mathbb{R}^{6}$ is a subvector of the augmented residual vector $\bm{r}\in\mathbb{R}^{m}$, where $m$ depends on the robot's number of degrees of freedom and the number of additional force-torque sensors with which the system is equipped.

At each time step, two quantities are extracted from the estimated contact: the contact point $\q_k \in \mathbb{R}^3$ and the corresponding surface normal $\n_k \in \mathbb{S}^2$, as illustrated in \cref{fig:visual_abstract}). 
Assuming rigid bodies, negligible tangential friction, and a single-point contact, both quantities can be determined from the line of force action, or wrench axis, of the measured external wrench~\cite{Salisbury1984,iskandar2021collision}. 
The wrench axis provides the surface-normal direction, subject to the adopted sign convention, while the contact point is obtained by intersecting the line of force action with the known tool geometry. The resulting wrench measurements provide an intrinsic tactile sensing modality for reliable real-time contact-point estimation. For further details on force-based contact localization, see~\cite{iskandar2024intrinsic}.

\subsection{Surface Representation}
\label{sec:gpis_surface}

To model the unknown surface geometry online, we employ a GPIS representation, where the surface is defined by an implicit function $f(\x)$ (see \eqref{eq:def_implicit_function}) on the domain $\Omega$.
Each contact yields two measured quantities at the contact point $\q_k$: the implicit surface value $f(\q_k)$ and its gradient $\nabla f(\q_k)$.
Since the point lies on the surface, its implicit value is known to be zero \eqref{eq:def_implicit_function}, and since the gradient defines the surface normal \eqref{eq:gpis_normal_def}, the gradient at that point is known to follow the measured contact normal $\n_k$. We therefore feed the GP two observations per contact, modeling the function value as $f(\q_k) \sim \mathcal{N}(0, \sigma_n^2)$ and the gradient as $\nabla f(\q_k) \sim \mathcal{N}(\n_k, \sigma_n^2 \mathbf{I})$, to incrementally update the GPIS model using a zero mean prior, $\mathbf{m}=\mathbf{0}$.
New observations are only added if their distance to existing observations exceeds a threshold $d_{\text{obs}}$, avoiding redundant nearby samples and bounding the kernel-matrix size.
To efficiently compute the inverse of the kernel matrix $\mathbf{K}$ as new observations arrive, we employ Cholesky updates, which avoid recomputing the full matrix inverse at each step and reduce the complexity from $\mathcal{O}(n^3)$ to $\mathcal{O}(n^2)$ per update \citep{Rasmussen2006}.
Gradient observations are essential for reconstruction: while multiple surfaces could pass through the same contact points, the observed normals uniquely constrain the surface's local orientation, enabling robust reconstruction from sparse data.

\subsection{Point Cloud for GPIS Discretization}
\label{sec:local_surface}
The ergodic control framework requires evaluating target and coverage distributions over the surface domain.
Since the surface geometry is unknown, these distributions cannot be computed directly on the implicit surface; instead, they must be discretized over a point cloud $\mathcal{Z}_k$ that approximates the observed surface $\setSt$.
While the GPIS provides a global surface model for visualization and high-level reasoning, its direct use in the ergodic control update \eqref{eq:HEDAC} presents two challenges.
First, when observations are spatially sparse (early in exploration), the global GPIS model becomes unreliable in unobserved regions, leading to inaccurate estimates of surface coverage density and exploration progress.
Second, extracting a discrete domain from the implicit surface is expensive: the standard approach, marching cubes~\citep{Lorensen1987}, requires a dense voxel grid, and evaluating the GP at every one of its $\mathcal{O}(n^3)$ voxels scales cubically in the resolution $n$.
We instead sample incrementally around contact points, so GP inference is needed only at those samples, and the domain scales with observations rather than workspace volume.

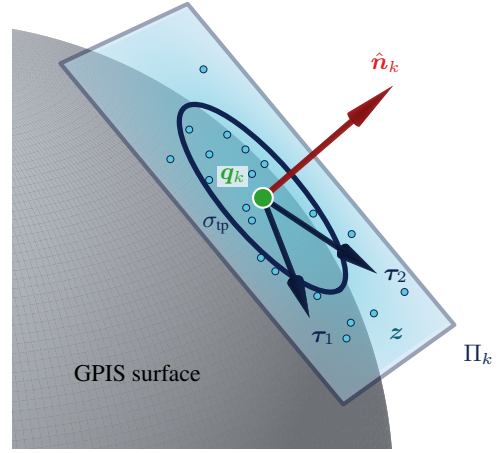
\begin{figure}[t]
	\centering
	\begin{tikzpicture}
		\tikzset{figlbl/.style={anchor=center, font=\small, fill=white, fill opacity=0.7, text opacity=1, inner sep=1pt}}
		\tikzset{figlbl2/.style={anchor=center, font=\small, fill=white, fill opacity=0.0, text opacity=1, inner sep=1pt}}
		\node[anchor=south west, inner sep=0] (img) at (0,0)
			{\includegraphics[width=0.78\columnwidth]{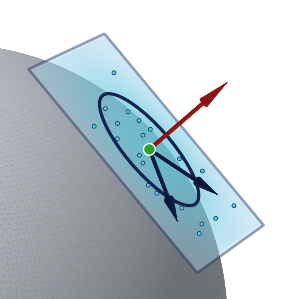}};
		\begin{scope}[x={(img.south east)}, y={(img.north west)}]
			\node[figlbl] at (0.44,0.543) {\textcolor[HTML]{2CA02C}{$\q_k$}};
			\node[figlbl2] at (0.745,0.77) {\textcolor[HTML]{E02020}{$\hat{\n}_k$}};
			\node[figlbl2] at (0.62,0.221) {\textcolor[HTML]{10224F}{$\bm{\tau}_1$}};
			\node[figlbl2] at (0.765,0.345) {\textcolor[HTML]{10224F}{$\bm{\tau}_2$}};
			\node[figlbl2] at (0.930,0.188) {\textcolor[HTML]{10224F}{$\Pi_k$}};
			\node[figlbl2] at (0.407,0.446) {\textcolor[HTML]{10224F}{$\sigma_\text{tp}$}};
			\node[figlbl2] at (0.766,0.230) {\textcolor[HTML]{0E5F70}{$\z$}};
			\node[figlbl2] at (0.25,0.15) {GPIS surface};
		\end{scope}
	\end{tikzpicture}
	\caption{Tangent plane sampling at a contact point $\q_k$ (green) on the current GPIS zero-level set estimate (gray surface). The GPIS normal $\hat{\n}_k$ defines the local tangent plane $\Pi_k$ with orthonormal basis $\mathbf{T}_k = \left[\bm{\tau}_1, \bm{\tau}_2\right]$. Points $\z$ are drawn from a Gaussian in the tangent plane \eqref{eq:tangent_sampling} (spread $\sigma_\text{tp}$ shown as the circle) and collected in the point cloud $\mathcal{Z}$.}%
	\label{fig:tangent_sampling}
\end{figure}

We employ a local surface approximation using tangent planes at the observed contact points.
Rather than using the instantaneous sensor-measured normals directly, we use normals $\hat{\n}_k$ inferred from the GPIS model at each observed position $\q_k$ (see \Cref{sec:gpis}).
Instantaneous sensor-measured normals can be corrupted by noise or tilted due to friction. In contrast, GPIS normals integrate information across all previously observed data points, yielding smoother, more reliable estimates.
Each observation thus provides a local linear approximation in the tangent plane at contact point $\q_k$, defined as
\begin{equation}
	\Pi_k = \{ \x \in \mathbb{R}^3 \mid (\x - \q_k)^\top \hat{\n}_k = 0 \}.
\end{equation}
For each contact point $\q_k$, we sample $N_s$ points in the local tangent space (illustrated in \Cref{fig:tangent_sampling}); each newly sampled point is generated as
\begin{equation}
	\z_k^{(j)} = \q_k + \mathbf{T}_k \boldsymbol{\eta}, \quad
	\boldsymbol{\eta} \sim \mathcal{N}(\mathbf{0}, \sigma_\text{tp}^2 \mathbf{I}),
	\label{eq:tangent_sampling}
\end{equation}
where $\mathbf{T}_k \in \mathbb{R}^{3 \times 2}$ consists of an orthonormal basis spanning the tangent plane $\Pi_k$, $\boldsymbol{\eta} \in \mathbb{R}^2$ is an independent random offset drawn in the tangent plane per sample, and $\sigma_\text{tp}$ is a user-defined parameter setting the tangent plane sampling standard deviation.
The point cloud is built incrementally as $\mathcal{Z}_k = \mathcal{Z}_{k-1} \cup \mathcal{Z}_k^{\text{new}}$, with $\mathcal{Z}_0 = \emptyset$ and $\mathcal{Z}_k^{\text{new}}$ the points newly sampled around $\q_k$. Its members are $\mathcal{Z}_k = \{\z_k^{(j)}\}_{j=1}^{M_k}$, where the bracketed superscript $(j)$ indexes an element of the cloud.
The sampling range $\sigma_\text{tp}$ must exceed the agent's coverage footprint, whose extent is set by its standard deviation $\sigma_\varphi$ (the footprint $\varphi$ in \eqref{eq:coverage}), so that the point cloud captures the transition from covered to uncovered regions, enabling the HEDAC framework to accurately compute the coverage discrepancy and identify exploration frontiers.
To control the density of sampled points and give a finite upper bound on the sampling, candidate points are only added to the point cloud if their closest existing neighbor is beyond a specified distance threshold $d_{\min}$.

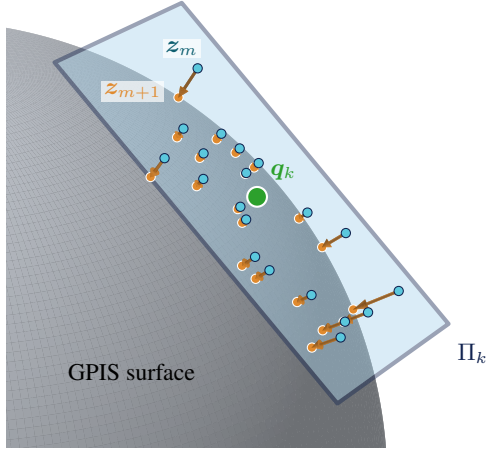
\begin{figure}[t]
	\centering
	\begin{tikzpicture}
		\tikzset{figlbl/.style={anchor=center, font=\small, fill=white, fill opacity=0.7, text opacity=1, inner sep=1pt}}
		\tikzset{figlbl2/.style={anchor=center, font=\small, fill=white, fill opacity=0.0, text opacity=1, inner sep=1pt}}
		\node[anchor=south west, inner sep=0] (img) at (0,0)
			{\includegraphics[width=0.78\columnwidth]{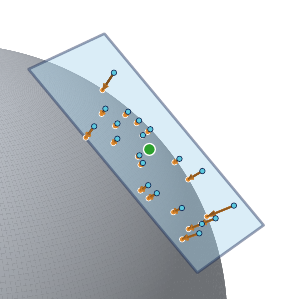}};
		\begin{scope}[x={(img.south east)}, y={(img.north west)}]
			\node[figlbl] at (0.350,0.796) {\textcolor[HTML]{0E5F70}{$\z\iter{m}$}};
			\node[figlbl] at (0.252,0.710) {\textcolor[HTML]{E08A2E}{$\z\iter{m+1}$}};
			\node[figlbl2] at (0.552,0.55) {\textcolor[HTML]{2CA02C}{$\q_k$}};
			\node[figlbl2] at (0.930,0.188) {\textcolor[HTML]{10224F}{$\Pi_k$}};
			\node[figlbl2] at (0.25,0.15) {GPIS surface};
		\end{scope}
	\end{tikzpicture}
	\caption{Projection refinement of the tangent plane samples. One refinement step moves each point $\z\iter{m}$ from the tangent plane $\Pi_k$ (\Cref{fig:tangent_sampling}) toward the current GPIS zero-level set estimate along the estimated normal by the signed distance \eqref{eq:projection_refinement}, giving the updated point $\z\iter{m+1}$; the step is iterated until the point reaches the surface. Corrections vanish near the contact $\q_k$, where the tangent plane already matches the surface, and grow with distance as curvature bends the surface away from the plane.}%
	\label{fig:projection_refinement}
\end{figure}

Tangent plane approximations become inaccurate on high-curvature surfaces.
We therefore project each sampled point onto the GPIS zero-level set at each control update (\Cref{fig:projection_refinement}), so that the control domain respects the learned surface while retaining the cost advantage of incremental sampling.
For each sampled point $\z_k^{(j)}$, initialized as $\z\iter{0} := \z_k^{(j)}$, we iteratively refine its position as
\begin{equation}
	\z\iter{m+1} = \z\iter{m} - \mathbb{E}[f(\z\iter{m})] \hat{\n}(\z\iter{m}),
	\label{eq:projection_refinement}
\end{equation}
where $\mathbb{E}[f(\z\iter{m})]$ is the inferred signed distance at $\z\iter{m}$, and $\hat{\n}(\z\iter{m})$ is the corresponding surface normal, obtained by normalizing the GP gradient prediction $\mathbb{E}[\nabla f(\z\iter{m})]$ \eqref{eq:gp_gradient_mean} as in \eqref{eq:gpis_normal_def}.
The subscript $m$ denotes the projection-refinement iteration, and the converged iterate replaces $\z_k^{(j)}$ in the point cloud $\mathcal{Z}_k$.
Each iteration thus moves the point along the estimated normal by the estimated signed distance, driving it toward the zero-level set.
The point cloud $\mathcal{Z}_k$ is updated with the refined positions at every control step, so that the control domain continuously tracks the evolving surface estimate rather than being recomputed from scratch.
The iteration terminates when the predicted distance falls below a small tolerance, indicating convergence to the zero-level set.
This projection step is necessary rather than optional: without it, tangent plane samples in high-curvature regions (e.g., near the bunny's ears, \Cref{sec:exp_bunny}) would remain systematically offset from the true surface, distorting the coverage and target distributions defined on $\setSt$.
Points that diverge are excluded, which ensures that all retained control points lie on the learned surface estimate $\setSt$, providing geometric consistency.
Divergence is primarily an epistemic uncertainty issue: the GPIS predictions that drive the projection---the signed distance $\mathbb{E}[f(\x)]$ and normal $\hat{\n}(\x)$---are reliable only within the kernel's effective neighborhood (length scale $l$ in \eqref{eq:imq_kernel}) around observed contacts, and samples placed beyond it revert toward the GP prior and can be driven away from the surface rather than toward it.
This bounds the tangent plane sampling range from above; combined with the lower bound from the coverage footprint $\sigma_\varphi$ \eqref{eq:coverage}, we choose $\sigma_\varphi < \sigma_\text{tp} < l$ in practice.
With this choice, only a small fraction of samples are excluded, and any transient gaps are backfilled by new tangent plane samples around subsequent contacts.

\subsection{Ergodic Control on an Evolving Surface Geometry}
\label{sec:erg_ctl}
Ergodic control is applied over the entire domain $\Omega$, but the target and coverage distributions are defined within the task space of the agent, which is the observed surface subspace $\setSt \subset \Omega$ (see \Cref{fig:domain}).
This subspace evolves dynamically as new observations refine the surface estimate.
In practice, these distributions are computed and evaluated on the point cloud $\mathcal{Z}_k$ that approximates the surface $\setSt$ (as described in \Cref{sec:local_surface}).
This discretization enables tractable control updates while ensuring the ergodic framework respects the observed surface geometry.
The target distribution $p_{\setSt}(\x)$ and coverage distribution $c(\x,\ergTime)$ are defined as mappings on the surface subspace as
\begin{equation}
	\begin{aligned}
		p_{\setSt}(\x) &: \mathcal{Z}_k \mapsto \R \text{, }\\
		c(\x, \ergTime) &: \mathcal{Z}_k \times \R \mapsto \R \text{.}
	\end{aligned}
\end{equation}
These distributions encode the desired exploration priorities and the agent's progress over time $\ergTime$, respectively, with practical evaluation occurring on the point cloud representation.
The target distribution $p_{\setSt}(\x)$ adapts dynamically as the surface geometry is refined: as new observations update the GPIS model and the point cloud $\mathcal{Z}_k$ is resampled, the target distribution is recomputed to reflect the evolving surface shape.
The coverage distribution $c(\x, \ergTime)$ is updated at each control step using a selective strategy: points (newly sampled on the surface or moved due to projection) have their coverage values recomputed from the control trajectory, while static reference points (from previous observations) are updated incrementally as in the original HEDAC formulation~\citep{Ivic2017}.
This selective approach keeps the per-step coverage update cost proportional to the number of newly added or moved points rather than the full point cloud size, which is essential given that the point cloud grows monotonically over the course of exploration (\Cref{tab:experiments}).
We refer the reader to \Cref{alg:ergodic_control} for the algorithmic perspective.

\subsection{From Heat Sources to Potential Field of Temperatures on the Surface}
\label{sec:potential_field}
Inspired by \citep{Ivic2017} and \citep{Bilaloglu2025}, we construct the control potential via two successive heat-diffusion processes.
First, radiative heat transfer spreads the user-specified goal distribution from $\Omega$ onto the current surface estimate to define the surface-supported target distribution $p_{\setSt}(\x)$.
Second, conductive heat transfer diffuses a coverage-deficit signal along the surface to compute the potential field that drives ergodic control.

\subsubsection{Radiative Heat Transfer: Heat Source to Surface}
We consider a heat source distributed within the domain $\Omega$, which is not necessarily located on the surface (see \Cref{fig:visual_abstract} and \Cref{fig:domain}), and energy radiates from this source throughout the domain onto the surface.
This decouples task specification from the unknown geometry: the user defines goals as heat sources $p_0$ in workspace coordinates, and the surface target $p_{\setSt}$ emerges by radiating these sources onto the current surface estimate $\setSt$ (\Cref{fig:radiative_heat}).
To model this process, we use an analogy to radiative heat transfer, where intensity decays with distance as $I(r) \propto \frac{1}{r^2}$ \citep{Griffiths2013}.%
To describe the transition of heat distribution across the domain, we define the radiative heat-transfer kernel $\kappa(\x)$ as
\begin{equation}
	\kappa(\x) = \left(\frac{\|\x\|_2}{r_d}\right)^{-2}\text{,}
	\label{eq:kernel}
\end{equation}
which quantifies how heat intensity decays with distance. The parameter $r_d > 0$ controls the influence range of the heat source and can be tuned based on task requirements.
Given a source distribution $p_0: \Omega \to \R$ measuring the radiative power emitted at each location in the domain, the heat distribution on the surface is then computed as the convolution of the source distribution with the kernel
\begin{equation}
	p_{\setSt}(\x) = \left( p_0 * \kappa \right)(\x)\text{.}
	\label{eq:rad_conv}
\end{equation}
This formulation assumes uniform surface properties and absorption across the surface while neglecting shadowing effects, focusing solely on proportional heat distribution rather than absolute physical values.
The analysis is restricted to the observed portion of the surface $\setSt$, where the heat transfer is calculated using the above convolution.

For example, if the source distribution $p_0$ consists of $N_0$ point masses at locations $\bm{\xi}_i$ with strengths $w_i$, the resulting distribution on the surface becomes
\begin{equation}
	p_{\setSt}(\x) = \sum_{i=1}^{N_0} w_i \kappa(\x - \bm{\xi}_i).
	\label{eq:heat_points}
\end{equation}
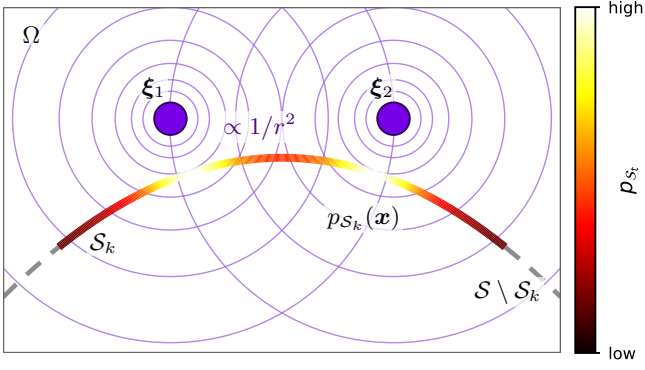
\begin{figure}[t]
	\centering
	\begin{tikzpicture}
		\tikzset{figlbl/.style={anchor=center, font=\small, fill=white, fill opacity=0.7, text opacity=1, inner sep=1pt}}
		\node[anchor=south west, inner sep=0] (img) at (0,0)
			{\includegraphics[width=\columnwidth]{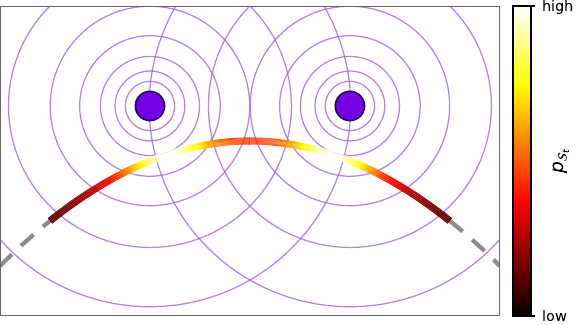}};
		\begin{scope}[x={(img.south east)}, y={(img.north west)}]
			\node[figlbl] at (0.235,0.758) {$\bm{\xi}_1$};
			\node[figlbl] at (0.590,0.758) {$\bm{\xi}_2$};
			\node[figlbl] at (0.4,0.65) {\textcolor[HTML]{4A0088}{$\propto 1/r^2$}};
			\node[figlbl] at (0.045,0.905) {$\Omega$};
			\node[figlbl] at (0.157,0.331) {$\setSt$};
			\node[figlbl] at (0.786,0.196) {$\setS \setminus \setSt$};
			\node[figlbl] at (0.562,0.4) {$p_{\setSt}(\x)$};
		\end{scope}
	\end{tikzpicture}
	\caption{Radiative heat transfer from sources to surface (2D illustration). Two point heat sources $\bm{\xi}_1, \bm{\xi}_2$ of equal weight in the domain $\Omega$ radiate with intensity decaying as $1/r^2$ (iso-intensity contours). The induced target distribution on the observed surface $\setSt$ \eqref{eq:heat_points} is shown as the color along the surface, peaking where the surface is closest to a strong source. The dashed continuation $\setS \setminus \setSt$ marks the not-yet-observed remainder of the surface, for which no target density is defined.}%
	\label{fig:radiative_heat}
\end{figure}

\subsubsection{Conductive Heat Transfer: Diffusion on the Surface}
Heat diffusion on a surface is constrained to tangential directions following \citep{Bilaloglu2025}.
We adapt the HEDAC heat equation (see \eqref{eq:HEDAC}) to the surface $\setSt$ by replacing the Euclidean Laplacian with the Laplace-Beltrami operator $\Delta_{\setSt}$, yielding
\begin{equation}
	\scalebox{0.97}{$\frac{\partial}{\partial t} \u(\x,t) = \alpha \, \Delta_{\setSt} \u(\x,t) - \beta \u(\x,t) + s(\x,t), \x \in \setSt$}
	\label{eq:heat_surface}
\end{equation}
where $\u(\x,t)$ is the temperature distribution on the surface, and $s(\x,t)$ is the source term from \Cref{eq:virtual_source} with target and coverage distributions defined on $\setSt$.
The time $t$ in $s(\x, t)$ and $\setSt$ represents exploration progress (robot motion and surface reconstruction), which is distinct from the thermal diffusion time. 
Within each control step, the thermal potential is solved in quasi-static equilibrium ($\frac{\partial}{\partial t} \u = 0$), yielding a stationary heat equation with $s(\x, t)$ and $\setSt$ treated as fixed parameters.
On the discrete point cloud with $\mathbf{u}$ being the vector of temperatures at each point, the quasi-static heat equation
\begin{equation}
	\alpha \Delta_{\setSt} \mathbf{u} - \beta \mathbf{u} + \mathbf{s} = 0
\end{equation}
is discretized via $\Delta_{\setSt} \approx -\mathbf{M}^{-1}\mathbf{L}$ using the strong Laplacian \citet{Sharp2020}. The solution is
\begin{equation}
	\mathbf{u} = (\alpha \mathbf{L} + \beta \mathbf{M})^{-1}\mathbf{M}\mathbf{s},
	\label{eq:discrete_solution}
\end{equation}
where $\mathbf{L}$ is the Laplacian matrix, $\mathbf{M}$ is the mass matrix, and $\mathbf{s}$ is the discrete source vector.

\subsection{Deriving Ergodic Actions from Temperature Gradients}
\label{sec:gradient}
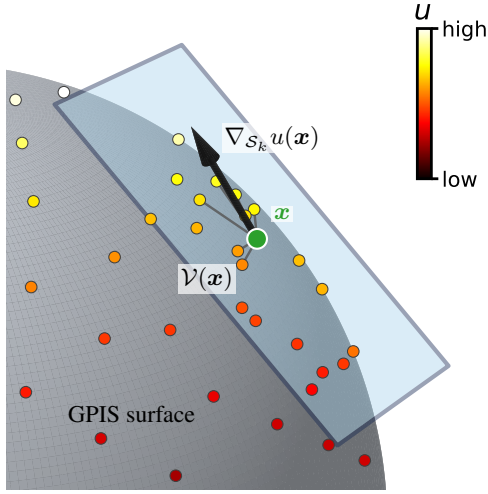
\begin{figure}[t]
	\centering
	\begin{tikzpicture}
		\tikzset{figlbl/.style={anchor=center, font=\small, fill=white, fill opacity=0.7, text opacity=1, inner sep=1pt}}
		\tikzset{figlbl2/.style={anchor=center, font=\small, fill=white, fill opacity=0.0, text opacity=1, inner sep=1pt}}
		\node[anchor=south west, inner sep=0] (img) at (0,0)
			{\includegraphics[width=0.78\columnwidth]{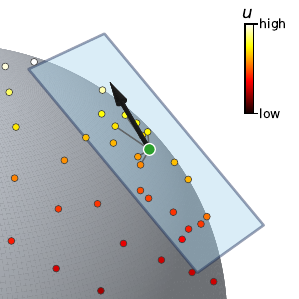}};
		\begin{scope}[x={(img.south east)}, y={(img.north west)}]
			\node[figlbl] at (0.53,0.7) {$\nabla_{\setSt}\u(\x)$};
			\node[figlbl] at (0.55,0.55) {\textcolor[HTML]{2CA02C}{$\x$}};
			\node[figlbl] at (0.4,0.42) {$\mathcal{V}(\x)$};
			\node[figlbl2] at (0.25,0.15) {GPIS surface};
		\end{scope}
	\end{tikzpicture}
	\caption{Surface gradient estimation of a scalar field $\u$ on the current GPIS zero-level set estimate. The field is known at the points (colored by $\u$). At the query point $\x$ (green), a local linear least squares fit over the neighborhood $\mathcal{V}(\x)$ (the $M_\mathcal{V}$ nearest points, connected by thin lines) estimates $\nabla \u(\x)$, which is projected onto the tangent plane $\Pi_k$ to give the surface gradient $\nabla_{\setSt}\u(\x)$ pointing toward increasing $\u$.}%
	\label{fig:surface_gradient}
\end{figure}

On the discrete point cloud, we compute the surface gradient using local linear least squares (\Cref{fig:surface_gradient}).
Assume that the scalar field $\u(\x)$ behaves approximately linearly in a neighborhood of a given point $\x \in \setSt$.
Using a first-order Taylor expansion, we approximate the temperature at any nearby point $\z \in \setSt$ as
\begin{equation}
\u(\z) \approx \u(\x)
+ \nabla \u(\x)^{\mathsf{T}} \left( \z - \x \right)\text{,}
\end{equation}
where $\nabla \u(\x)$ is the gradient of $\u$ at $\x$.
Let $\mathcal{V}(\x):=\left\{\z^{(1)},\cdots,\z^{(M_\mathcal{V})}\right\}$ denote the set of $M_\mathcal{V}$ neighboring points of $\x$.
We seek the gradient that satisfies the first-order Taylor expansion for all neighbors.
In matrix form with $\mathbf{r} \in \mathbb{R}^{M_\mathcal{V}}$ and $\mathbf{A} \in \mathbb{R}^{M_\mathcal{V} \times 3}$, the $j$-th entries are
\begin{equation}
r^{(j)} := \u(\z^{(j)}) - \u(\x), \quad
\mathbf{A}^{(j)} := (\z^{(j)} - \x)^{\mathsf{T}}.
\end{equation}
The gradient $\nabla \u(\x) \in \mathbb{R}^3$ is then obtained by solving the least squares problem
\begin{equation}
\nabla \u(\x) = \arg\min_{\mathbf{b} \in \mathbb{R}^3} \| \mathbf{A} \mathbf{b} - \mathbf{r} \|^2 \text{,}
\end{equation}
where the solution is given by the Moore-Penrose pseudoinverse $\mathbf{A}^+ = (\mathbf{A}^T \mathbf{A})^{-1} \mathbf{A}^T$ as
\[
\nabla \u(\x) = \mathbf{A}^+ \mathbf{r}.
\]
This provides an estimate of the gradient $\nabla \u(\x)$ at the point $\x$, leveraging local temperature measurements from neighboring points.
The method is robust to noise and irregular sampling, making it suitable for real-world applications where surface data may be sparse or unevenly distributed.
Finally, the computed gradient $\nabla \u(\x)$ must be projected onto the tangent plane of the surface $\setSt$ at $\x$ to obtain the surface gradient $\nabla_{\setSt} \u(\x)$.
This projection ensures that the gradient lies entirely within the surface and aligns with the direction of heat flow.
The projection is achieved by removing the component of $\nabla \u(\x)$ that is normal to the surface.
If $\bm{n}(\x)$ is the unit normal vector at $\x$, then
\begin{equation}
	\bm{P}_{\bm{n}} (\x) = \bm{I}_3 - \bm{n}(\x)\bm{n}(\x)^\intercal \text{,}
\end{equation}
\begin{equation}
	\nabla_{\setSt} \u(\x) = \bm{P}_{\bm{n}}(\x) \nabla \u(\x) \text{.}
\end{equation}
This surface gradient, after normalization, directly provides the control command
\begin{equation}
	\dot{\x}_a = v \frac{\nabla_{\setSt} \u(\x)}{\|\nabla_{\setSt} \u(\x)\|}\text{,}
\end{equation}
where $a$ indexes the agent and $v$ is the desired velocity magnitude, enabling navigation along the thermal field on the surface at a consistent speed.
\Cref{alg:ergodic_control} summarizes how the preceding components combine into a single control loop.
At each control step $k$, the contact observation updates the GPIS surface estimate, tangent space samples are drawn and projected onto the learned surface, the potential field is recomputed, and its surface gradient yields the next ergodic action, iterating until a maximum of $K_{\max}$ steps is reached.

\begin{algorithm}[H]
	\caption{Ergodic Control with Unknown Surface Geometries}
	\begin{algorithmic}[1]
		\State Initialize step $k \gets 1$
		\While{$k \le K_{\max}$}
		\State Observe agent state $\x$ from robot contact ($\q_k$, $\n_k$)
		\State Add observation to GPIS surface (\Cref{sec:gpis_surface})
		\State Sample surface points $\z \in \mathcal{Z}_k$ in tangent space \eqref{eq:tangent_sampling}
		\State Project points $\mathcal{Z}_k$ onto surface $\setSt$ \eqref{eq:projection_refinement}
		\State Transfer heat source from $p_0(\x)$ to $p_{\setSt}(\x)$ \eqref{eq:rad_conv}
		\State Recompute spatial coverage $c(\x, t)$ \eqref{eq:coverage}
		\State Potential field computation (\Cref{sec:potential_field})
		\State Gradient calculation ($\nabla_{\setSt} \u$) (\Cref{sec:gradient})
		\State Ergodicity measure evaluation $\epsilon_k$ \eqref{eq:ergodicity}
		\State Move agent by commanding robot
		\State Advance step $k \gets k + 1$
		\EndWhile
	\end{algorithmic}
	\label{alg:ergodic_control}
\end{algorithm}

\section{Experiments and Results}
\label{sec:experiments}
We validate the proposed framework through simulation and real-world experiments with a single agent, demonstrating simultaneous surface reconstruction and ergodic coverage of unknown surfaces without prior geometric knowledge.
The bunny, chair, and backpanel experiments are shown in the supplementary video.

We assess performance using three different metrics across the experiments. 
The ergodic cost $\epsilon_k$ \eqref{eq:ergodicity} quantifies coverage quality relative to the target distribution. 
The \textit{Chamfer distance} $d_{\text{C}}(S_1, S_2)$~\citep{Achlioptas2018} measures the discrepancy between the reconstructed GPIS surface $S_1$ and ground truth $S_2$ using 10000 points sampled uniformly on each surface and is computed as
\begin{equation}
	\begin{aligned}
		d_{\text{C}}(S_1, S_2) = {} & \frac{1}{|S_1|}\!\sum_{\x \in S_1}\!\min_{\y \in S_2}\!\|\x - \y\|_2 \\
		& + \frac{1}{|S_2|}\!\sum_{\y \in S_2}\!\min_{\x \in S_1}\!\|\y - \x\|_2.
	\end{aligned}
	\label{eq:chamfer}
\end{equation}
The \textit{target Chamfer distance} $d_{\text{TC}}(S_1, S_2)$ is defined as a variant that weighs each point's contribution by the target distribution $p_{\setSt}(\x)$; this emphasizes reconstruction quality in regions of interest.
All experiments use GPIS with an inverse multiquadric kernel
\begin{equation}
	k(\x, \x') = \left(\|\x - \x'\|_2^2 + l^2\right)^{-1/2},
	\label{eq:imq_kernel}
\end{equation}
where $l$ is the length scale.
We choose this kernel, rather than the more common squared-exponential (RBF) kernel, following established prior work on tactile GPIS reconstruction~\citep{Driess2017}: its slower decay of correlation with distance propagates surface information further from sparse contact observations, which is particularly advantageous in the early, poorly-observed stage of exploration when contacts are sparse.
The parameters used in the following experiments are listed in \Cref{tab:parameters} (\Cref{app:parameters}).

\subsection{Bunny Experiment (Simulation)}
\label{sec:exp_bunny}
\begin{figure}
	\centering
	\newcommand{\colwidthbunny}{0.40\columnwidth}
	\newcommand{\colwidthbunnycbar}{0.09\columnwidth}
	\setlength{\tabcolsep}{2pt}
    \begin{tabular}{cc c}
		\multicolumn{2}{c}{\small\textit{Target Distribution}} & \\
        \includegraphics[width=\colwidthbunny]{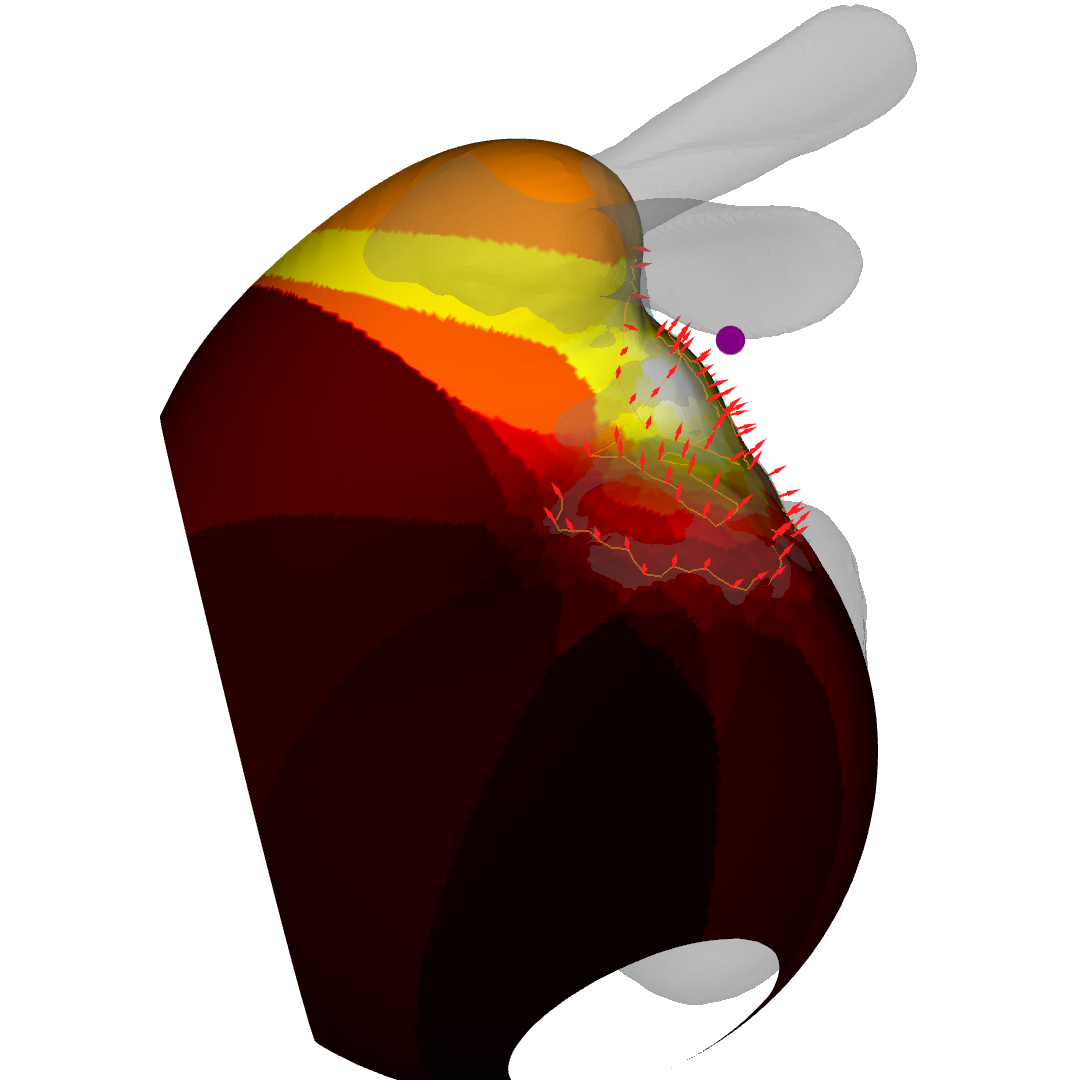} &
        \includegraphics[width=\colwidthbunny]{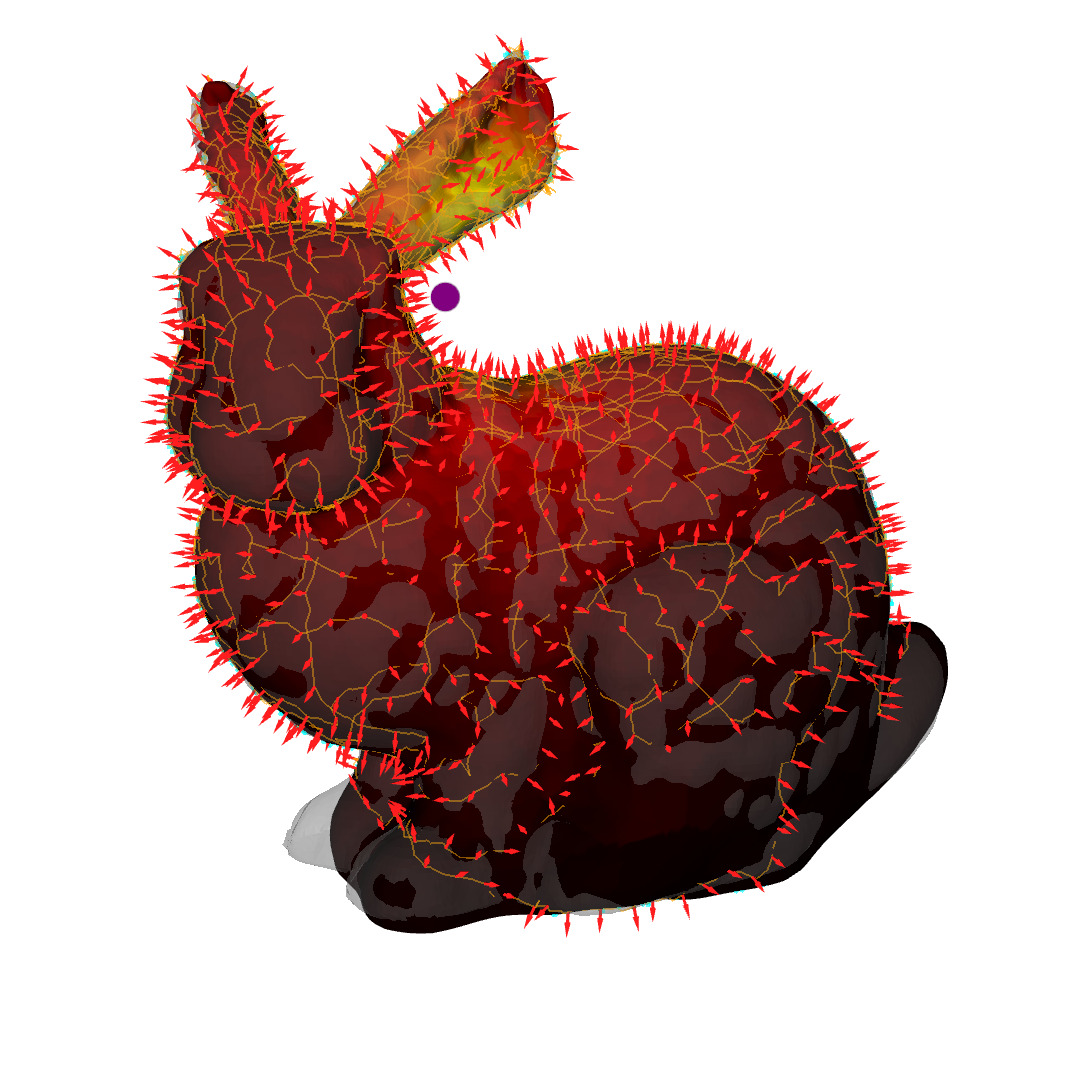} &
        \includegraphics[width=\colwidthbunnycbar]{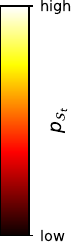} \\
		\multicolumn{2}{c}{\small\textit{GPIS Uncertainty}} & \\
        \includegraphics[width=\colwidthbunny]{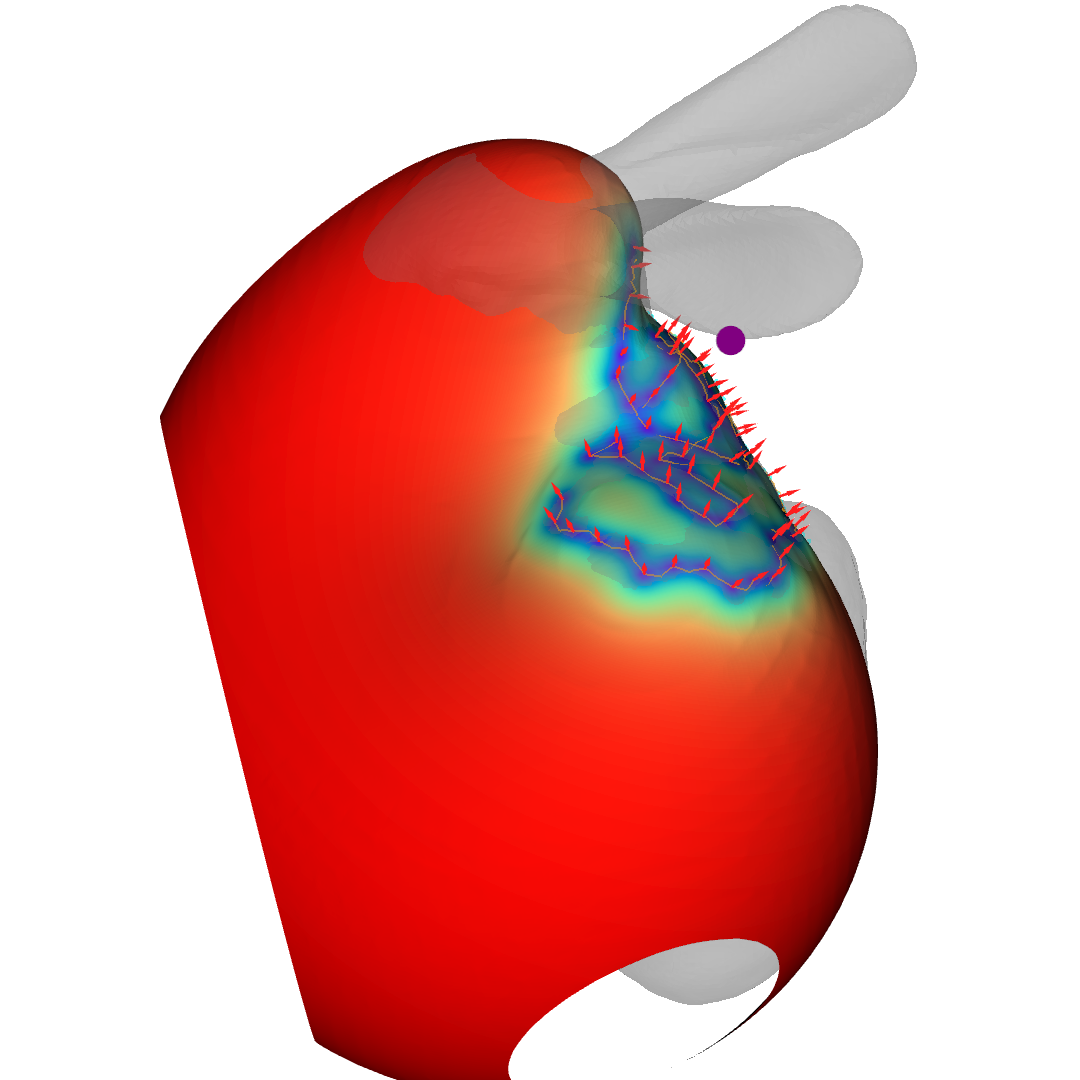} &
        \includegraphics[width=\colwidthbunny]{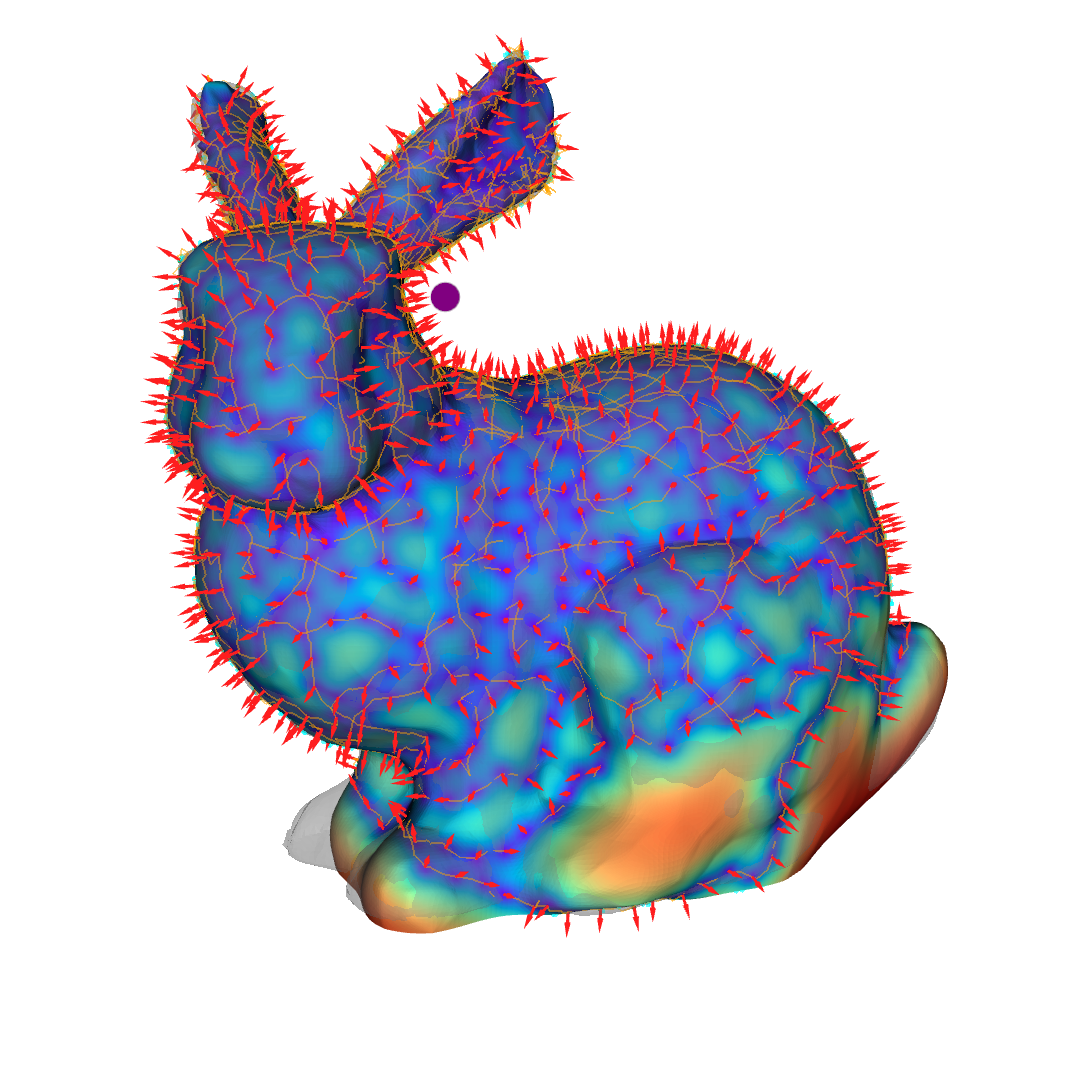} &
        \includegraphics[width=\colwidthbunnycbar]{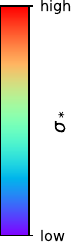} \\
    \end{tabular}
	\caption{Bunny surface reconstruction at early (time step 150) and late (time step 6570) exploration, showing target distribution (warmer colors indicate proximity to heat source; the purple point marks the heat source location) and GPIS prediction uncertainty (blue indicates lower uncertainty, red indicates higher uncertainty), with colorbars at right.}%
	\label{fig:surface_reconstruction}%
\end{figure}
The Stanford bunny model \citep{Stanford3DRepo} serves as simulation benchmark, scaled to fit within a $1\,\text{m}$ cubic domain $\Omega$ (see \Cref{tab:parameters}).
Its geometry presents challenging features: high curvatures, non-convex regions including protruding ears and indentations, and rapid changes in surface normal orientation (see \Cref{fig:surface_reconstruction}).
In our experiment, the target distribution is generated from a single point heat source placed between the ears and the back (see \Cref{fig:surface_reconstruction}), and the coverage footprint in \eqref{eq:coverage} is modeled as $\varphi(\x) = \mathcal{N}(\mathbf{0}, \sigma_\varphi^2 \mathbf{I})$.
In simulation, observations are queried directly from the ground-truth triangle mesh, where at its current location the agent reads off the exact surface position and normal vector, yielding an idealized observation.
Unlike the real robot experiments, the simulation imposes no workspace or collision constraints, allowing us to apply the method to the entire shape without restricting the exploration domain, and enabling the longer 10000-step exploration horizon used here compared to the real-robot trials (\Cref{tab:experiments}).

\begin{figure}
	\centering
	\includegraphics[width=0.95\columnwidth]{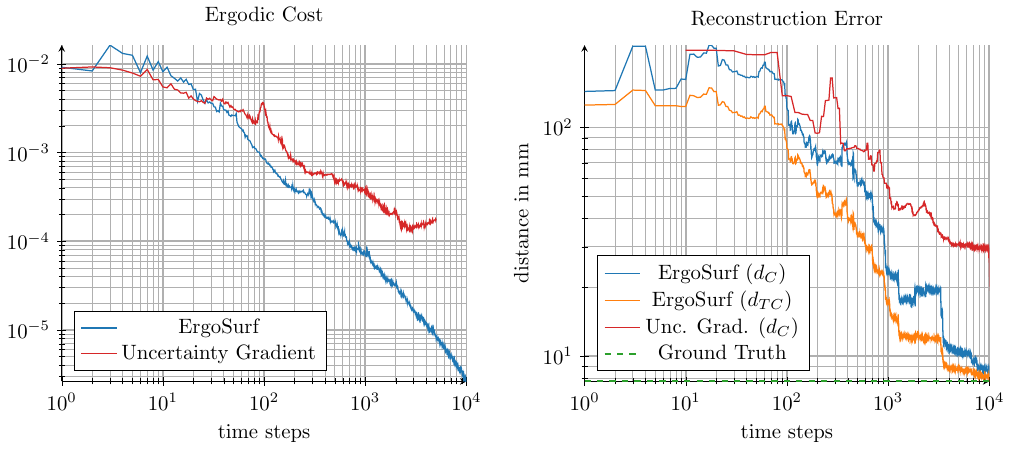}
	\caption{\textbf{Bunny Experiment:} Ergodic cost $\epsilon_k$ (left, log-log scale) converges toward coverage target. Reconstruction error (right) with Chamfer distance (blue) and target-weighted variant (orange) approaching ground truth baseline (green dashed). Uncertainty gradient (red) is the active learning policy for tactile GPIS from related work \citep{Driess2017}. Both metrics stabilize with accumulated observations.}%
	\label{fig:ergodicity_plot}%
\end{figure}

The Chamfer distance decreases throughout exploration, with the target Chamfer distance converging even faster and approaching ground truth (see \Cref{tab:experiments} and \Cref{fig:ergodicity_plot}).
Our method achieves comparable reconstruction quality to the uncertainty gradient policy \citep{Driess2017} while additionally satisfying ergodicity.
Ergodicity, however, is where the two methods part.
The uncertainty gradient policy converges faster initially, yet its ergodic cost is already about $5\times$ higher than ours at step 1000 ($3.4\times10^{-4}$ vs.\ $6.9\times10^{-5}$), and by step 5000 the gap has widened to roughly $21\times$ ($1.8\times10^{-4}$ vs.\ $8.4\times10^{-6}$).
More tellingly, it does not merely stagnate: its cost bottoms out near $1.4\times10^{-4}$ around step 2800 and then rises over the following 2000 steps, whereas ours continues to fall through step 10000.
Note that on the log-log axes of \Cref{fig:ergodicity_plot} both the widening gap and this subsequent rise appear far less pronounced than they are.
This reversal is not incidental to the run: the uncertainty gradient acts on a purely local signal and becomes confined once nearby uncertainty is reduced, and as it never references the task distribution, it has no mechanism that drives the coverage deficit $\epsilon_k$ \eqref{eq:ergodicity} toward zero.
It reduces uncertainty, but it cannot achieve ergodicity.
In contrast, the steady-state heat equation \eqref{eq:heat_surface} couples the coverage deficit globally across the entire surface domain: distant underexplored regions raise the potential $\u$ in their vicinity, and the resulting smooth field propagates this influence through the surface so that the agent is attracted toward them even when no informative gradient is locally present (see \Cref{fig:local_optima} in \Cref{app:local_optima}).
Our diffusion-based formulation thus matches the uncertainty gradient policy in reconstruction quality, while being the only one of the two whose coverage deficit actually converges.
Both the observation accumulation rate and the point cloud growth rate decrease over time (see \Cref{fig:ammount_plots} (left) in \Cref{sec:progression}), as new observations increasingly fall below the minimum distance threshold $d_{\text{obs}}$.
Figure~\ref{fig:surface_reconstruction} illustrates the exploration outcome: the top row shows the target distribution, which---unlike in classical ergodic control (see related work in \Cref{sec:related_work})---evolves over time as it is defined on the reconstructed surface geometry. As the GPIS representation expands, the target distribution progressively concentrates around the heat source location, while the bottom row shows the corresponding reduction in GPIS prediction uncertainty.
During early exploration, sparse observations initially misrepresent the surface topology and shape; however, the method remains stable and converges to the correct geometry as contact points accumulate.
The trajectory exhibits smooth adaptation to the complex geometry, with the GPIS successfully capturing non-convex features and intricate details despite the high local curvature.
Visual progression of the coverage distribution, target distribution, and GP prediction uncertainty are presented in \Cref{fig:bunny_coverage_progression} and in the accompanying supplementary video.
\begin{figure*}[htb]
	\centering
	\includegraphics[width=0.85\textwidth]{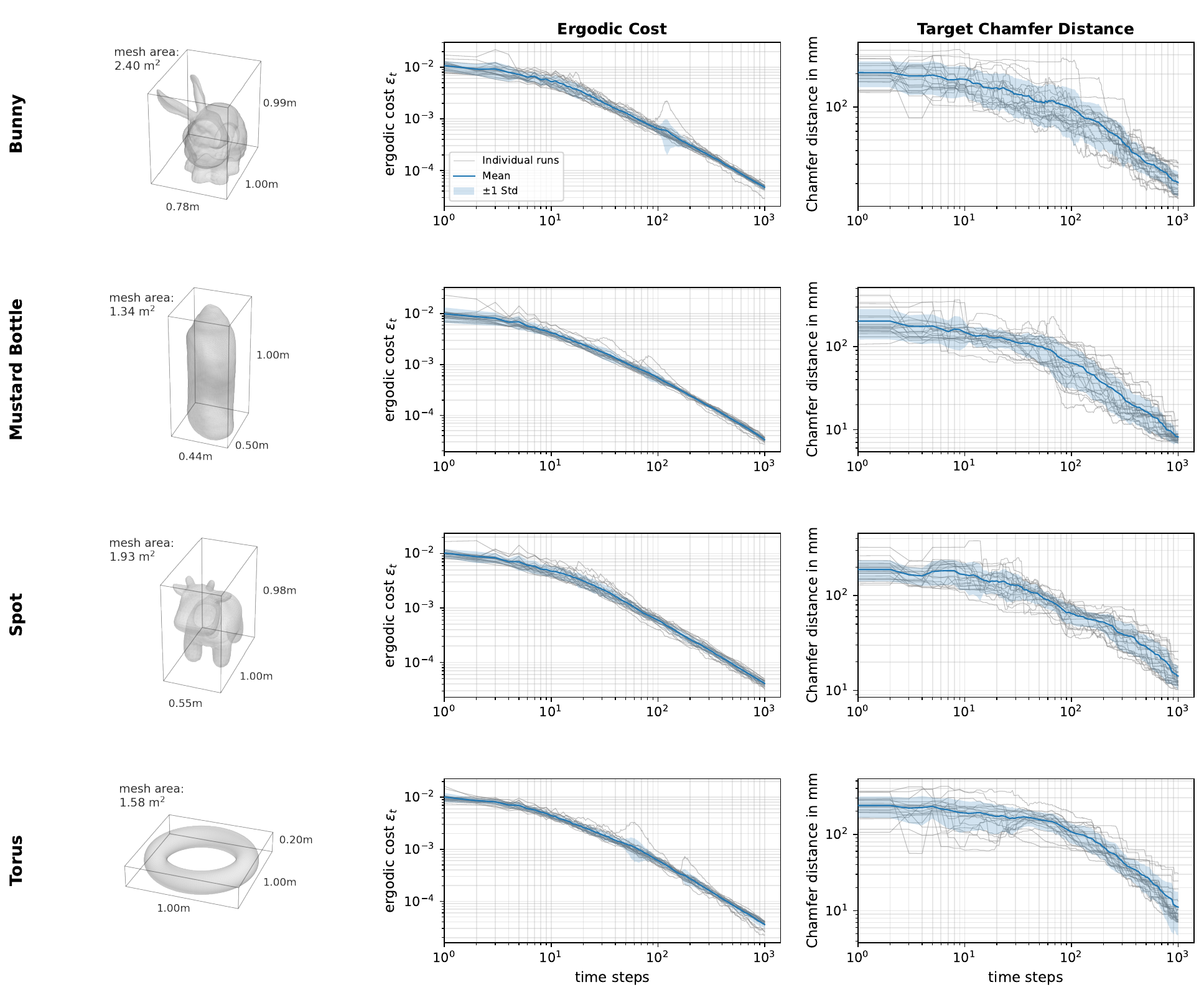}
	\caption{\textbf{Multi-Run Analysis (20 runs per object):} Each row shows one object's geometry (left) alongside its mean ergodic cost (center) and mean target Chamfer distance (right), each with a ±1 standard deviation error band over the individual runs (gray). Across all objects (Stanford bunny, YCB mustard bottle, spot, and torus), both metrics converge consistently with narrow standard-deviation bands under randomized start points and heat configurations.}%
	\label{fig:multi_run}%
\end{figure*}
\begin{table}[htb]
	\centering
	\small
	\setlength{\tabcolsep}{4pt}
	\caption{Experimental Results Summary}%
	\label{tab:experiments}%
	\resizebox{\columnwidth}{!}{%
	\begin{tabular}{lcccc}
		\toprule
		& \textbf{Steps} & \textbf{Point Cloud} & \textbf{Obs. Points} & \textbf{Traj.} \\
		\midrule
		Bunny     & 10000 & 3249 & 1651 & 10000 \\
		Chair     & 3000  & 874  & 293  & 3000 \\
		Backpanel & 3000  & 1472 & 417  & 3000 \\
		\bottomrule
	\end{tabular}}
	\\[0.5em]
	\resizebox{\columnwidth}{!}{%
	\begin{tabular}{lccc}
		\toprule
		& $\epsilon_k$ & $d_{\text{C}}$ & $d_{\text{TC}}$ \\
		\midrule
		Bunny     & $9.2\text{e-}3 \to 2.6\text{e-}6$ & $143.6 \to 8.7$ & $125.2 \to 8.1$ \\
		Chair     & $1.0\text{e-}2 \to 1.4\text{e-}5$ & $44.1 \to 2.5$  & $35.5 \to 2.5$ \\
		Backpanel & $8.3\text{e-}3 \to 1.2\text{e-}5$ & $35.5 \to 3.3$  & $29.2 \to 2.7$ \\
		\bottomrule
	\end{tabular}}
	\\[2pt]
	\scriptsize
	$\epsilon_k$: ergodic cost~\eqref{eq:ergodicity}; $d_{\text{C}}$, $d_{\text{TC}}$: Chamfer and target Chamfer distance in mm~\eqref{eq:chamfer}.
	$a \to b$: metric's evolution from start ($a$) to end ($b$) of the experiment.
\end{table}

\subsection{Multi-Run Robustness Analysis}
\label{sec:multi_run_analysis}
While the preceding experiment demonstrates the framework on a single representative task, a single run cannot establish that this performance is consistent rather than incidental to a favorable configuration. We therefore conduct a statistical robustness study in which the framework is repeatedly evaluated under randomized task conditions. Each run randomizes three independent factors: (i) the number of heat sources, drawn uniformly from one to ten; (ii) the positions of the heat sources, sampled within the domain space bounded by the object's bounding box; and (iii) the initial contact point on the mesh. Together, these factors induce a broad distribution of target distributions and exploration starting conditions, allowing us to characterize both the typical convergence behavior and its variability across tasks.

We perform this analysis on four objects chosen to span a range of geometric and topological characteristics: the \emph{Stanford bunny} of the detailed single-run experiment (\Cref{fig:ergodicity_plot}), here re-evaluated under randomized tasks to confirm that its reported performance is not incidental to a favorable configuration; the \emph{mustard bottle} from the YCB object set~\citep{Calli2015}, a familiar manipulation benchmark; the \emph{spot} (cow) model of \citet{Crane2013robust}, which exhibits richer surface detail and protruding features; and a \emph{torus}, whose genus-one topology probes the framework's behavior on a non-simply-connected surface. For each object we execute 20 independent runs and record the ergodic cost $\epsilon_k$~\eqref{eq:ergodicity} and the Chamfer distances~\eqref{eq:chamfer} throughout exploration. Aggregating across runs yields the mean trajectory together with its standard-deviation band, summarizing both expected performance and run-to-run variability. The per-object parameters used in this study are listed in \Cref{tab:parameters_multi} (\Cref{app:parameters}).

As shown in \Cref{fig:multi_run}, the ergodic cost and reconstruction error converge consistently with narrow standard-deviation bands across diverse object geometries, topologies, and randomized tasks, demonstrating that the framework reliably achieves both ergodic coverage and accurate surface reconstruction beyond any single configuration. A detailed single-run analysis of a representative case where the starting point and the heat source lie on opposite sides of the object is provided in \Cref{app:bunny_run7}.

\subsection{Hardware Platform and Experimental Configuration}
\begin{figure*}
	\centering
    \resizebox{\textwidth}{!}{%
	\large
    \begin{tabular}{cc}
        \begin{tabular}{c}
            \includegraphics[width=0.30\textwidth]{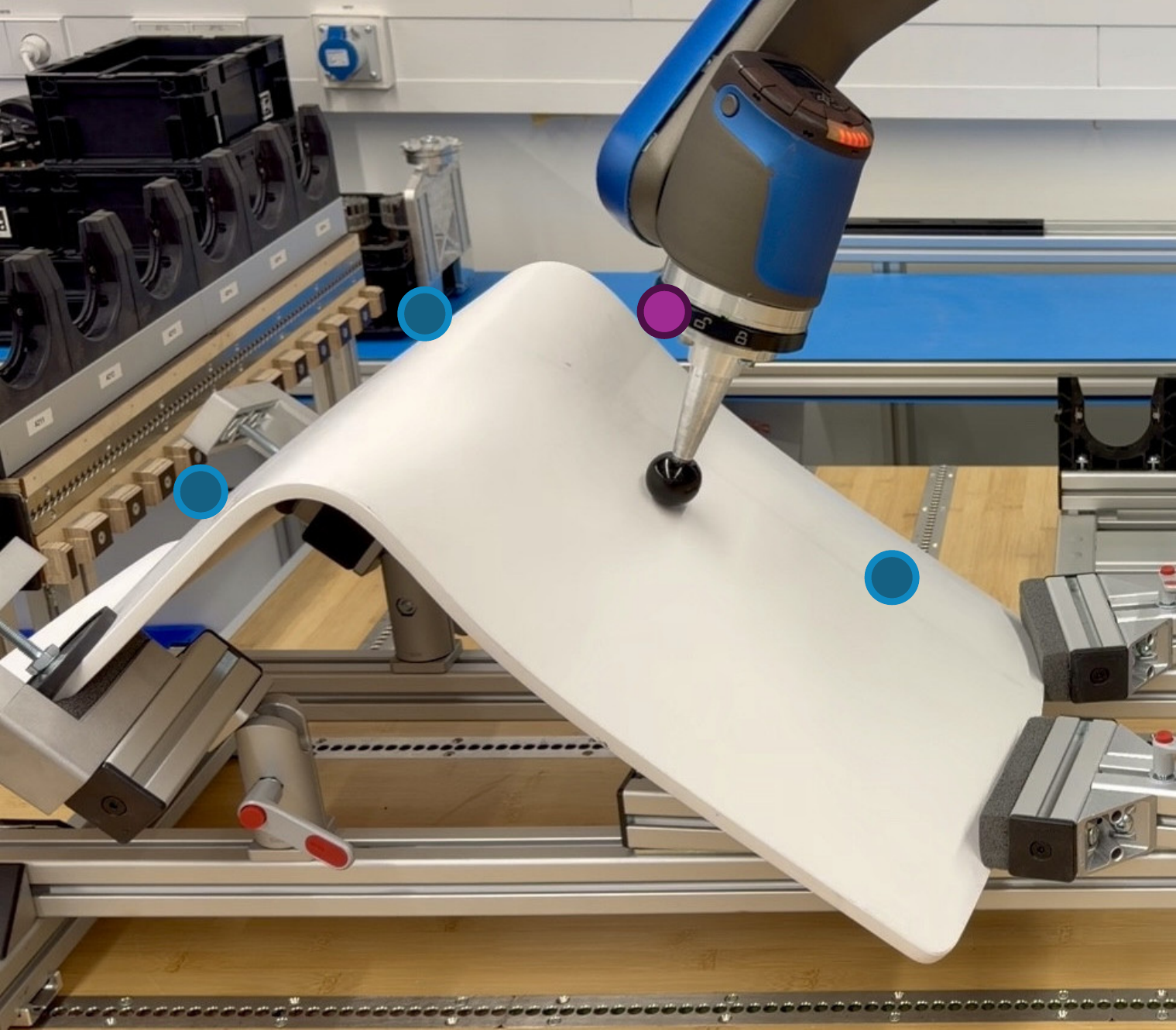}
        \end{tabular}
        &
		\begin{tabular}{c}
		\resizebox{0.66\textwidth}{!}{%
            \includegraphics{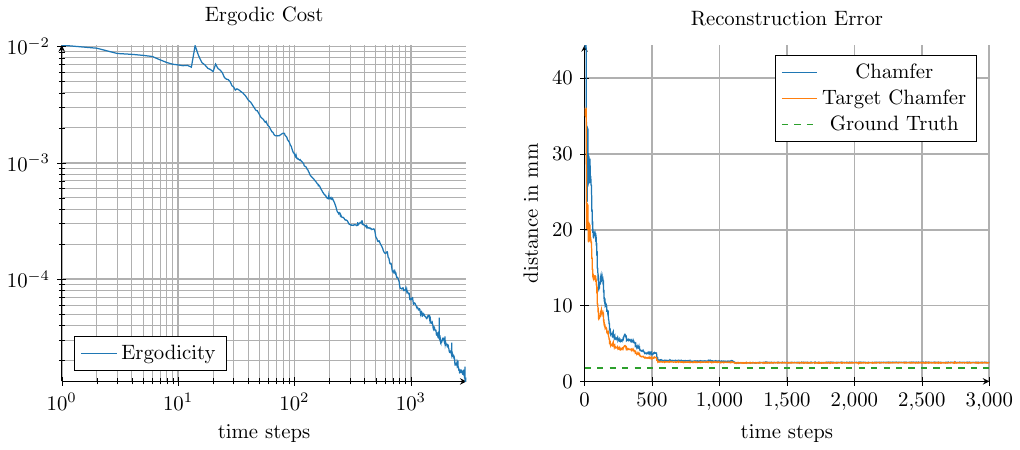}%
		}
		\end{tabular}
    \end{tabular}
	}%
	\caption{\textbf{Chair Experiment:} Physical setup (left), where the blue points mark the boundary of the domain space and the purple point shows the heat source; ergodic cost convergence (middle); and reconstruction error (right) showing Chamfer distance (blue), target-weighted variant (orange), and statistical floor from ground truth sampling (green).}%
	\label{fig:chair_results}%
\end{figure*}
\ifthenelse{\boolean{blind}}{
The following experiments are conducted using a 7-DoF robot manipulator equipped with a probing tool featuring a spherical touching head for tactile exploration.
The robot incorporates multiple force sensors, and we use the fully dynamically decoupled external end-effector wrench of \eqref{eq:ext_ee_wrench} (\Cref{sec:contact_obs}) as the contact measurement in our experiments.
}{
The following experiments are conducted using the DLR SARA robot, a 7-DoF manipulator equipped with a probing tool featuring a spherical touching head for tactile exploration. 
SARA incorporates multiple force sensors, and we use the fully dynamically decoupled external end-effector wrench of \eqref{eq:ext_ee_wrench} (\Cref{sec:contact_obs}) as the contact measurement in our experiments.
}
We make use of the spherical geometry of the tool and assume a frictionless contact where the moment exerted in the center of the sphere is set to zero to increase the robustness of the contact point calculation subject to measurement noise.
The full pipeline is implemented in Python and executed on a workstation with an Intel Core i7-10700K CPU and 32\,GB of RAM.
The ergodic control updates run at a maximum of 4\,Hz to maintain consistent update rates, while the low-level controller operates at a significantly higher frequency above 1\,kHz. 
Given the comparatively small exploration domain of the chair and backpanel workpieces (\Cref{tab:parameters}), fewer steps are required to reach ergodic coverage and reconstruction convergence than for experiments over larger domains (\Cref{tab:experiments}).
\Cref{tab:runtime} reports a per-component runtime breakdown at step 1000 for the bunny and chair experiments. Surface projection and Laplacian assembly dominate the cost; the total stays within the $250\,$ms budget of the $4\,$Hz control rate.
Turning to the start of each experiment, the exploration must be initialized: this requires an initial contact point to seed the GPIS and point cloud $\mathcal{Z}$.
Each real-robot run begins with a brief pre-phase: the end-effector moves at constant velocity along a fixed direction until contact is detected, providing the seed contact point and normal.
With a single contact, the controller is active but locally effective; as observations accumulate, effectiveness grows and domain coverage expands.
Having already shown in simulation that the uncertainty-gradient policy fails to achieve ergodic coverage under idealized, best-case conditions, we do not repeat that comparison here, as it would only reproduce the same failure.
Instead, the following chair and backpanel experiments target a complementary question: whether our method deploys on a real physical system, under real force/torque sensing, real dynamics, and without simulation's simplifying assumptions.

\begin{table}[htb]
	\centering
	\small
	\caption{Per-component runtime of one ergodic control step at exploration step 1000 (Python, Intel i7-10700K). $M_k = |\mathcal{Z}_k|$ denotes the point cloud size; percentages are of the total step time. Laplacian computation follows~\citet{Sharp2020}.}%
	\label{tab:runtime}%
	\resizebox{\columnwidth}{!}{%
	\begin{tabular}{lcc}
		\toprule
		& \textbf{Bunny (sim)} & \textbf{Chair (real)} \\
		\midrule
		Point Cloud Size $M_k$                       & 165               & 379               \\
		\midrule
		Projection + GP inference                    & 29\,ms (59\,\%)   & 61\,ms (57\,\%)   \\
		Laplacian computation                           & 5\,ms (10\,\%)    & 22\,ms (21\,\%)   \\
		Other                                        & 16\,ms (31\,\%)   & 24\,ms (22\,\%)   \\
		\midrule
		\textbf{Total}                               & \textbf{50\,ms}   & \textbf{107\,ms}  \\
		\bottomrule
	\end{tabular}}
\end{table}

\subsection{Chair Experiment}
\label{sec:exp_chair}
This experiment uses a wooden chair with smooth geometry and a central bending region (see \Cref{fig:chair_results}).
The target distribution is defined by a single heat source point at the top of the chair.
Its smooth geometry and moderate curvature allow rapid GPIS convergence and consistent ergodicity coverage, confirming that ergodic coverage extends to real robot operation on unknown surfaces (see \Cref{tab:experiments}).
Results are shown in \Cref{fig:chair_results}, with visual progression in \Cref{fig:chair_coverage_progression} and in the accompanying supplementary video.

\subsection{Backpanel Experiment}
\begin{figure*}[htbp]
    \centering
	\resizebox{\textwidth}{!}{%
	\large
    \begin{tabular}{cc}
		\begin{tabular}{c}
			\centering
			\includegraphics[width=0.30\textwidth]{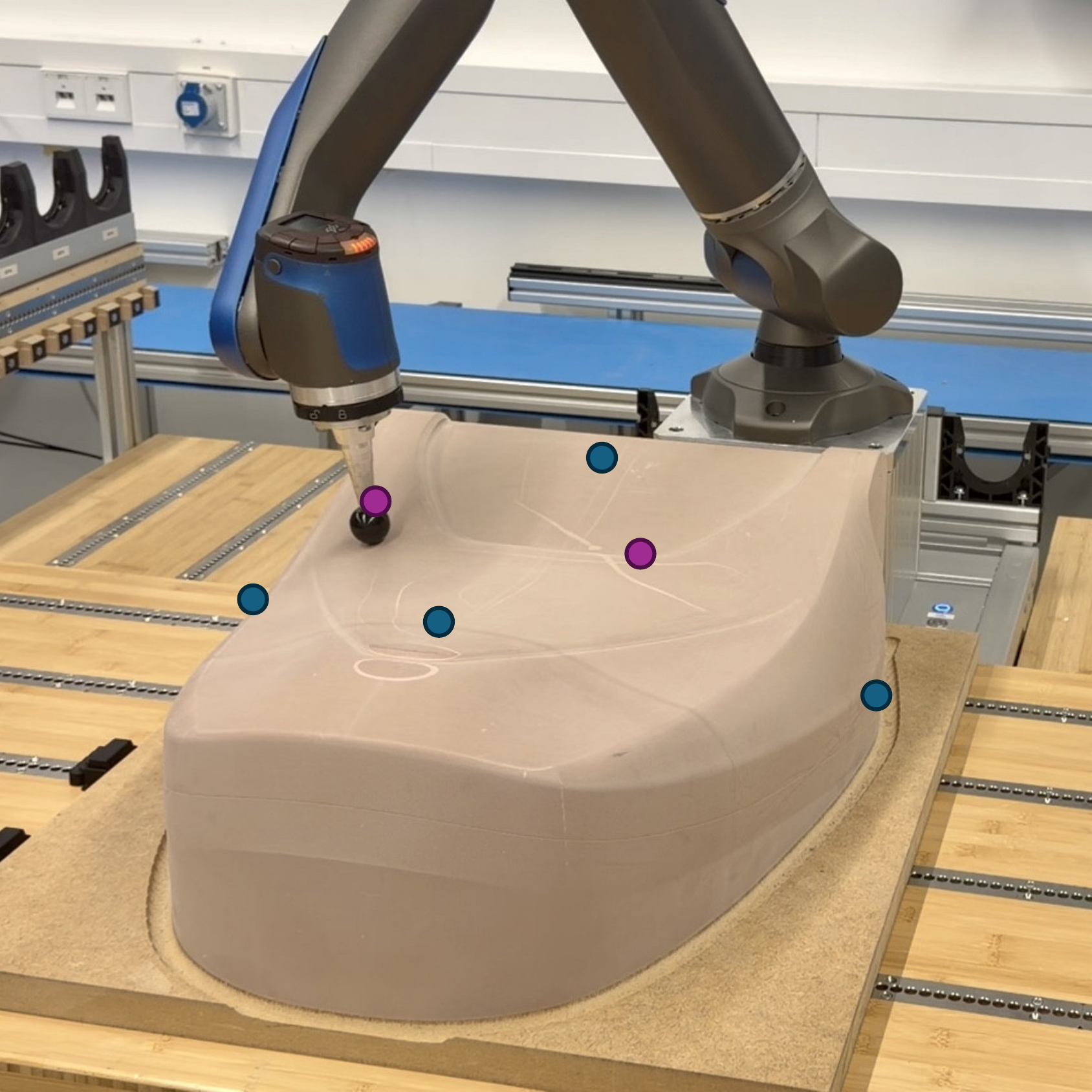}
		\end{tabular}
		&
		\begin{tabular}{c}
			\resizebox{0.66\textwidth}{!}{%
				\includegraphics[width=0.95\textwidth]{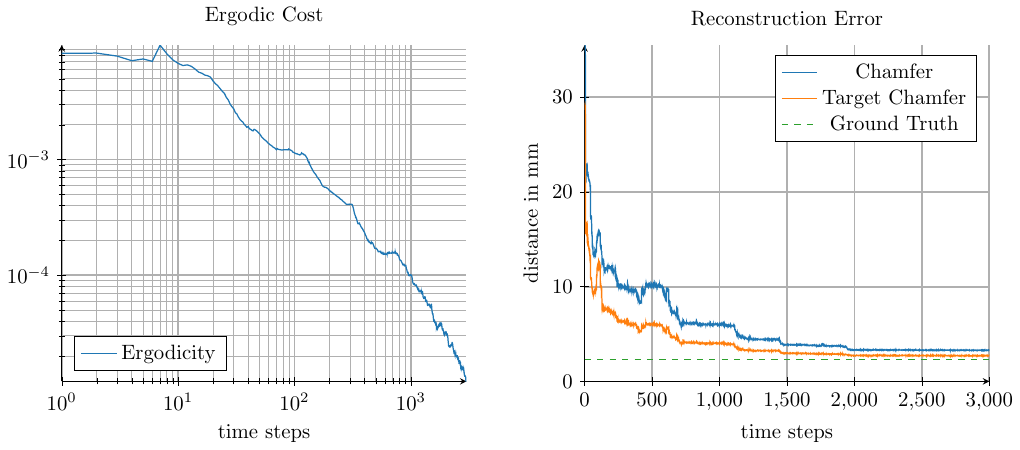}
			}
		\end{tabular}
    \end{tabular}
	}%
    \caption{\textbf{Backpanel Experiment:} Physical setup (left), where the blue points mark the boundary of the domain space and the purple points show the two heat sources; ergodic cost convergence (middle); and reconstruction error (right) with Chamfer distance (blue), target-weighted variant (orange), and ground truth floor (green).}
    \label{fig:backpanel_metrics}
\end{figure*}

\begin{figure}
	\centering
    \begin{tabular}{cc}
		Time step 110 & Time step 1585 \\
        \includegraphics[width=0.22\textwidth,clip,trim=500 355 500 240]{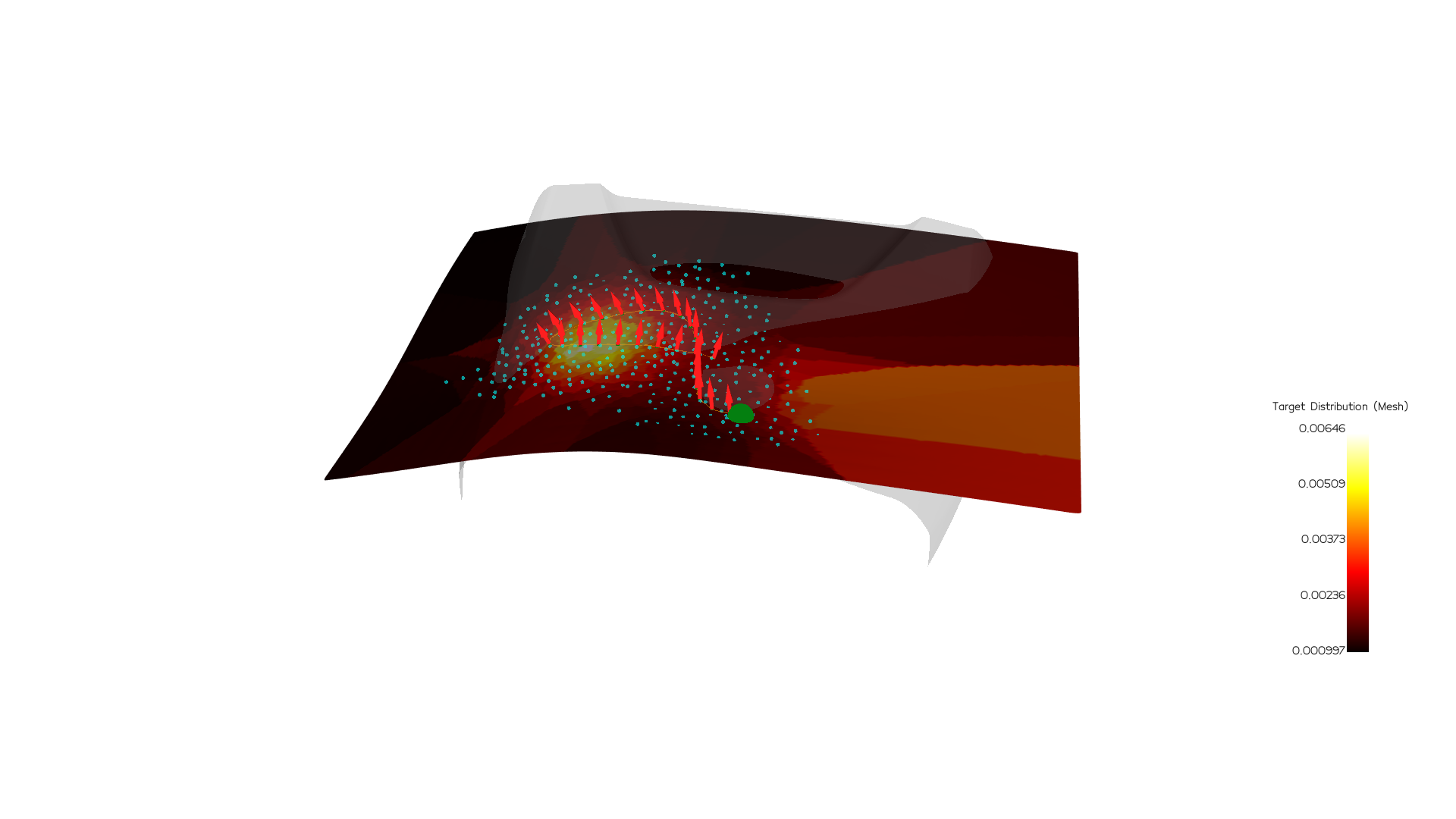} &
        \includegraphics[width=0.22\textwidth,clip,trim=500 355 500 240]{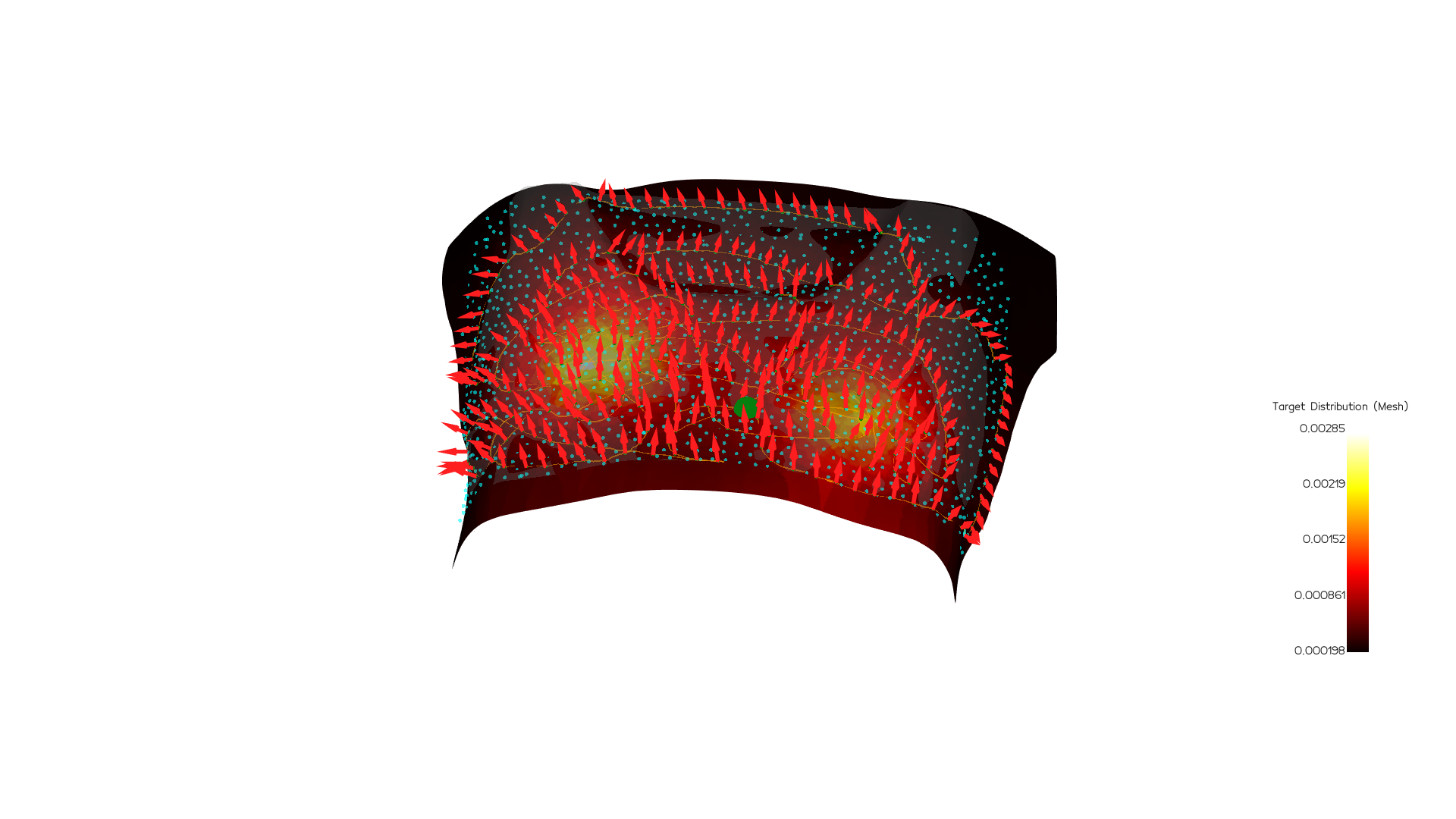} \\
    \end{tabular}
	\caption{Backpanel exploration at early (time step 110) and late (time step 1585) stages, showing target distribution (warmer colors indicate proximity to heat sources).}%
	\label{fig:backpanel_surfaces}%
\end{figure}
The second object is a positive mold of a backpanel of a car seat made of synthetic material, featuring stronger curvatures and more detailed geometric features than the chair (see \Cref{fig:backpanel_metrics} (left)).
The target distribution consists of two heat points, one on each side of the panel.
The Chamfer distance decreases throughout exploration, approaching the ground truth derived from CAD geometry, with the target Chamfer reflecting effective coverage of both target regions (see \Cref{tab:experiments}).
Despite the higher curvature and more intricate surface, the framework maintains stable exploration and achieves consistent reconstruction quality throughout the trajectory.
As in the bunny experiment, the observation and point cloud growth rates flatten over time while the trajectory keeps growing linearly (see \Cref{fig:ammount_plots} (right) in \Cref{sec:progression}), as new observations increasingly fall below the minimum distance threshold $d_{\text{obs}}$.
Surface reconstruction quality is shown in \Cref{fig:backpanel_surfaces}, with performance metrics in \Cref{fig:backpanel_metrics} and visual progression in \Cref{fig:backpanel_coverage_progression} and in the accompanying supplementary video.

\section{Discussion}
\label{sec:discussion}

The results in \Cref{sec:experiments} demonstrate that our framework successfully balances surface learning with ergodic coverage. 
By decoupling the coverage objective from surface reconstruction, the latter emerges as a byproduct of task-driven exploration, as shown by the coupled ergodic and reconstruction convergence across all experiments (see \Cref{fig:ergodicity_plot}, ~\ref{fig:multi_run}, ~\ref{fig:chair_results}, ~\ref{fig:backpanel_metrics}).
This task-specific coverage contrasts with active learning approaches, which do not satisfy ergodic convergence. In the bunny experiment (see \Cref{fig:ergodicity_plot}), the gradient-based method~\citep{Driess2017} achieves comparable reconstruction quality, yet its ergodic cost stagnates at a markedly higher level than ours, since it optimizes only local uncertainty and never drives the time-averaged trajectory toward a task distribution.
While such a policy is not designed to be ergodic, this comparison makes explicit what is sacrificed by uncertainty-driven exploration: the coverage guarantee (\Cref{fig:ergodicity_plot}) as well as global awareness, the latter leaving the agent prone to becoming trapped in local uncertainty minima (\Cref{fig:local_optima}).
Heat diffusion, in contrast, inherently integrates global structure and yields both accurate reconstruction and ergodic coverage.

\begin{table}[htb]
	\centering
	\caption{Comparison of Related Work and Our Approach}%
	\label{tab:comparison}%
	\footnotesize
	\setlength{\tabcolsep}{3pt}
	\resizebox{\columnwidth}{!}{%
	\begin{tabular}{@{}lccccc@{}}
		\toprule
		\textbf{Approach} & \textbf{Online} & \textbf{Ergodic} & \textbf{Coverage} & \textbf{Prior} & \textbf{Surface} \\
		\textbf{} & \textbf{Planning} & \textbf{Policy$^*$} & \textbf{Domain} & \textbf{Geometry} & \textbf{Learning} \\
		\midrule
		\citeauthor{Driess2017} & Yes & No & - & No & Yes \\
		\citeauthor{Bilaloglu2025} & Yes & Yes & Points & From Vision & No \\
		\citeauthor{Hughes2025} & No & Yes & Points & Yes & No \\
		\citeauthor{Seewald2024} & No & Yes & Global & Yes & No \\
		\textbf{Ours} & \textbf{Yes} & \textbf{Yes} & \textbf{GPIS/Points} & \textbf{No} & \textbf{Yes} \\
		\bottomrule
	\end{tabular}}
	\\[2pt]
	\scriptsize
	$^*$ Task-specific ergodic policy independent from surface reconstruction
	\vspace{-1\baselineskip}
\end{table}
Table~\ref{tab:comparison} positions our approach relative to existing methods. 
Compared to \citet{Driess2017}, we add task-specific ergodic control; unlike \citet{Bilaloglu2025}, we eliminate vision-based pre-scanning; and in contrast to \citet{Seewald2024}, we enable online planning without prior geometry. 
Our contribution uniquely combines online planning, ergodic control on reconstructed surfaces, and simultaneous learning without prior geometric assumptions.

Despite the positive results, some limitations are worth highlighting.
First, our approach assumes smooth, continuous surfaces; extending it to handle discontinuities and sharp edges would require modifications to the GPIS kernel or explicit edge detection.
Second, GPIS kernel hyperparameters are currently fixed; adaptive online learning of these parameters based on local surface complexity could improve reconstruction efficiency. 
Third, the point cloud surface representation limits applicability to thin objects where nearest-neighbor edges in the discrete Laplacian can connect across the two sides, mixing distinct target distributions.

Finally, computational scaling for very large surfaces or high-resolution requirements remains an open question, though our approach already improves data efficiency compared to standard GPs through dual representation, point density constraints, and Cholesky updates.

\section{Conclusion}
\label{sec:conclusion}
This paper presented a novel framework for ergodic control on unknown surfaces that integrates online surface reconstruction with task-driven coverage. 
By combining Gaussian Process Implicit Surfaces for global geometry learning with local tangent plane approximations for efficient ergodic computation, the approach enables robots to systematically cover surfaces without prior geometric knowledge or vision-based pre-scanning.  
The heat diffusion formulation, incorporating radiative energy transport from the target distribution to the reconstructed surface and conductive diffusion along the surface geometry, generates smooth potential fields that guide exploration while respecting contact constraints.
Experimental validation on real robot hardware and in simulation demonstrated that the framework achieves consistent ergodic convergence while simultaneously reconstructing surface geometry.
The decoupling of coverage objectives from surface reconstruction distinguishes this work from prior uncertainty-driven exploration methods, enabling principled coverage guarantees for task-specific spatial distributions.

\ifthenelse{\boolean{blind}}{
}{
	\begin{acks}
		This work was supported in part by the German Federal Ministry of Education and Research (BMBF) through the "The Future of Value Creation --- Research on Production, Services and Work" Program under Grant 02K20D032 and in part by the DLR internal Project "ASPIRO".
	\end{acks}
}

\bibliographystyle{style/SageH}
\bibliography{bibliography}

\appendix
\section{Appendix}

\subsection{Discrete Laplacian Derivation}
\label{app:discrete_laplacian}

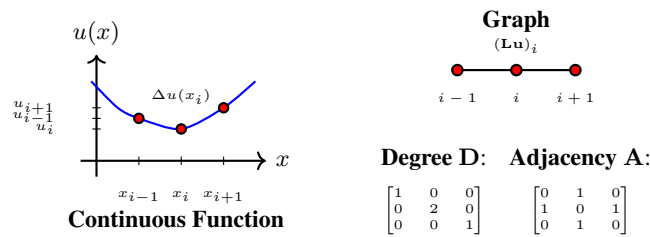
\begin{figure}[H]
	\centering
	\begin{tabular}{@{}c@{\hspace{2.5em}}c@{}}
		\begin{tikzpicture}[scale=0.7, baseline=(current bounding box.center)]
			\draw[thick, ->] (-0.3, 0) -- (3.2, 0) node[right, font=\small] {$x$};
			\draw[thick, ->] (0, -0.3) -- (0, 2) node[above, font=\small] {$u(x)$};
			\draw (0.8, -0.08) -- (0.8, 0.08);
			\draw (1.6, -0.08) -- (1.6, 0.08);
			\draw (2.4, -0.08) -- (2.4, 0.08);
			\node[below=8pt, font=\tiny] at (0.8, 0) {$x_{i-1}$};
			\node[below=8pt, font=\tiny] at (1.6, 0) {$x_i$};
			\node[below=8pt, font=\tiny] at (2.4, 0) {$x_{i+1}$};
			\draw (-0.08, 0.8) -- (0.08, 0.8);
			\draw (-0.08, 0.6) -- (0.08, 0.6);
			\draw (-0.08, 1.0) -- (0.08, 1.0);
			\node[left=12pt, font=\tiny] at (0, 0.8) {$u_{i-1}$};
			\node[left=12pt, font=\tiny] at (0, 0.6) {$u_i$};
			\node[left=12pt, font=\tiny] at (0, 1.0) {$u_{i+1}$};
			\draw[thick, blue, smooth] plot coordinates {(-0.1, 1.5) (0.4, 1) (0.8, 0.8) (1.6, 0.6) (2.4, 1) (3, 1.5)};
			\draw[fill=red, thick] (0.8, 0.8) circle (2.5pt);
			\draw[fill=red, thick] (1.6, 0.6) circle (2.5pt);
			\draw[fill=red, thick] (2.4, 1) circle (2.5pt);
			\node[above=6pt, font=\tiny] at (1.6, 0.6) {$\Delta u(x_i)$};
			\node[font=\small] at (1.5, -1.1) {\textbf{Continuous Function}};
		\end{tikzpicture}
		&
		\begin{tabular}[c]{c}
			\begin{tikzpicture}[scale=0.65, baseline=(current bounding box.center)]
				\node[font=\small] at (1.2, 2.2) {\textbf{Graph}};
				\draw[thick] (0, 1.2) -- (1.2, 1.2);
				\draw[thick] (1.2, 1.2) -- (2.4, 1.2);
				\draw[fill=red, thick] (0, 1.2) circle (3pt) node[below=5pt, font=\tiny] {$i-1$};
				\draw[fill=red, thick] (1.2, 1.2) circle (3pt) node[below=5pt, font=\tiny] {$i$};
				\draw[fill=red, thick] (2.4, 1.2) circle (3pt) node[below=5pt, font=\tiny] {$i+1$};
				\node[above=3pt, font=\tiny] at (1.2, 1.2) {$(\mathbf{L}\mathbf{u})_i$};
			\end{tikzpicture}
			\\[1cm]
			\begin{tabular}[c]{@{}c@{\hspace{0.3cm}}c@{}}
				{\small \textbf{Degree} $\mathbf{D}$:} & {\small \textbf{Adjacency} $\mathbf{A}$:} \\[0.15cm]
				{\tiny $\begin{bmatrix} 1 & 0 & 0 \\ 0 & 2 & 0 \\ 0 & 0 & 1 \end{bmatrix}$} & {\tiny $\begin{bmatrix} 0 & 1 & 0 \\ 1 & 0 & 1 \\ 0 & 1 & 0 \end{bmatrix}$}
			\end{tabular}
		\end{tabular}
	\end{tabular}
	\caption{The Laplacian bridges continuous and discrete representations: (left) a continuous function sampled at three equally-spaced points; (right) the corresponding graph topology showing node connectivity with the degree matrix $\mathbf{D}$ and adjacency matrix $\mathbf{A}$ whose difference defines the graph Laplacian $\mathbf{L} = \mathbf{D} - \mathbf{A}$.}
	\label{fig:discrete_laplacian}
\vspace{-1\baselineskip}
\end{figure}

The connection between continuous and discrete Laplacians is illustrated by the canonical 1D case with equally-spaced samples ($h=1$), giving
\begin{align}
	-\Delta u(x_i) &= -\frac{\mathrm{d}^2}{\mathrm{d} x_i^2}u(x_i) \\
	&\approx - \frac{u(x_{i}-h) - 2u(x_{i}) + u(x_{i}+h)}{h^2} \quad \Big|_{h=1} \\
	&= \sum_{j \in \{i-1, i+1\}} w_{ij} (u_i - u_j) \quad \Big|_{u(x_i) =: u_i,\, w_{ij} = 1}  \\
	&= D_{ii} u_i - \sum_j A_{ij} u_j = ((\mathbf{D} - \mathbf{A})\mathbf{u})_i \\
	&= (\mathbf{L}\mathbf{u})_i ,
\end{align}
where $\mathbf{D}$ is the degree matrix and $\mathbf{A}$ is the adjacency matrix (see \Cref{fig:discrete_laplacian}).

\subsection{Experimental Parameters}
\label{app:parameters}

This appendix collects the parameter settings used throughout the experiments.
\Cref{tab:parameters} lists the values for the single-run simulation and real-robot experiments of \Cref{sec:experiments} (bunny, chair, and backpanel), while \Cref{tab:parameters_multi} lists the values for the multi-run robustness study of \Cref{sec:multi_run_analysis}.
Each table reports the domain extent $|\Omega|_{x,y,z}$, the heat equation coefficients $\alpha$ and $\beta$~\eqref{eq:HEDAC}, the kernel length scale $l$~\eqref{eq:imq_kernel} and observation noise $\sigma_n$, and the sampling and coverage parameters ($N_s$, $d_{\text{obs}}$, $d_{\text{min}}$, $\sigma_{\text{tp}}$~\eqref{eq:tangent_sampling}, $\sigma_\varphi$~\eqref{eq:coverage}, $N_0$~\eqref{eq:heat_points}, and $r_d$~\eqref{eq:kernel}).
The equation references in the table headers indicate where each parameter is introduced.
In the multi-run study, the number of heat sources $N_0$, their positions, and the initial contact point are randomized per run, so \Cref{tab:parameters_multi} reports the shared settings and the sampling range for $N_0$.

\begin{table}[htbp]
	\centering
	\small
	\caption{Experimental Parameters (all distances in meters)}
	\label{tab:parameters}
	\resizebox{\columnwidth}{!}{%
	\begin{tabular}{lccccc}
		\toprule
		& $|\Omega|_{x,y,z}$ & $\alpha$, $\beta$ \eqref{eq:HEDAC} & $l$ \eqref{eq:imq_kernel} & $\sigma_n$ & $N_s$ \eqref{eq:tangent_sampling} \\
		\midrule
		Bunny     & 1, 1, 1 & 0.1, 1 & 0.05 & 0.001 & 60 \\
		Chair     & 0.33, 0.34, 0.2 & 0.01, 1 & 0.05 & 0.1 & 60 \\
		Backp.    & 0.36, 0.54, 0.16 & 0.01, 1 & 0.05 & 0.1 & 60 \\
		\bottomrule
	\end{tabular}}
	\\[0.5em]
	\resizebox{\columnwidth}{!}{%
	\begin{tabular}{lcccccc}
		\toprule
		& $d_{\text{obs}}$ & $d_{\text{min}}$ & $\sigma_{\text{tp}}$ \eqref{eq:tangent_sampling} & $\sigma_\varphi$ \eqref{eq:coverage} & $N_0$ \eqref{eq:heat_points} & $r_{d}$ \eqref{eq:kernel} \\
		\midrule
		Bunny     & 0.024 & 0.025 & 0.04 & 0.02 & 1 & 0.2 \\
		Chair     & 0.016 & 0.01 & 0.03 & 0.015 & 1 & 0.03 \\
		Backp.    & 0.016 & 0.01 & 0.03 & 0.015 & 2 & 0.03 \\
		\bottomrule
	\end{tabular}}
\end{table}

\begin{table}[htbp]
	\centering
	\small
	\caption{Multi-Run Experiment Parameters (all distances in meters)}
	\label{tab:parameters_multi}
	\resizebox{\columnwidth}{!}{%
	\begin{tabular}{lccccc}
		\toprule
		& $|\Omega|_{x,y,z}$ & $\alpha$, $\beta$ \eqref{eq:HEDAC} & $l$ \eqref{eq:imq_kernel} & $\sigma_n$ & $N_s$ \eqref{eq:tangent_sampling} \\
		\midrule
		Bunny     & 1, 0.78, 0.99 & 0.1, 1 & 0.05 & 0.001 & 60 \\
		Mustard   & 0.5, 0.44, 1 & 0.1, 1 & 0.05 & 0.001 & 60 \\
		Spot      & 1, 0.55, 0.98 & 0.1, 1 & 0.05 & 0.001 & 60 \\
		Torus     & 1, 1, 0.2 & 0.1, 1 & 0.05 & 0.001 & 60 \\
		\bottomrule
	\end{tabular}}
	\\[0.5em]
	\resizebox{\columnwidth}{!}{%
	\begin{tabular}{lcccccc}
		\toprule
		& $d_{\text{obs}}$ & $d_{\text{min}}$ & $\sigma_{\text{tp}}$ \eqref{eq:tangent_sampling} & $\sigma_\varphi$ \eqref{eq:coverage} & $N_0$ \eqref{eq:heat_points} & $r_{d}$ \eqref{eq:kernel} \\
		\midrule
		Bunny     & 0.024 & 0.015 & 0.04 & 0.02 & 1--10 & 0.2 \\
		Mustard   & 0.024 & 0.015 & 0.04 & 0.02 & 1--10 & 0.2 \\
		Spot      & 0.024 & 0.015 & 0.04 & 0.02 & 1--10 & 0.2 \\
		Torus     & 0.024 & 0.015 & 0.04 & 0.02 & 1--10 & 0.2 \\
		\bottomrule
	\end{tabular}}
	\\[2pt]
	\scriptsize
	$N_0$ (number of heat sources), their positions, and the initial contact point are randomized per run (see \Cref{sec:multi_run_analysis}).
\end{table}

\subsection{Global Awareness in Exploration Policies}
\label{app:local_optima}

\begin{figure}
    \centering
    \begin{tabular}{cc}
        \includegraphics[width=0.45\columnwidth]{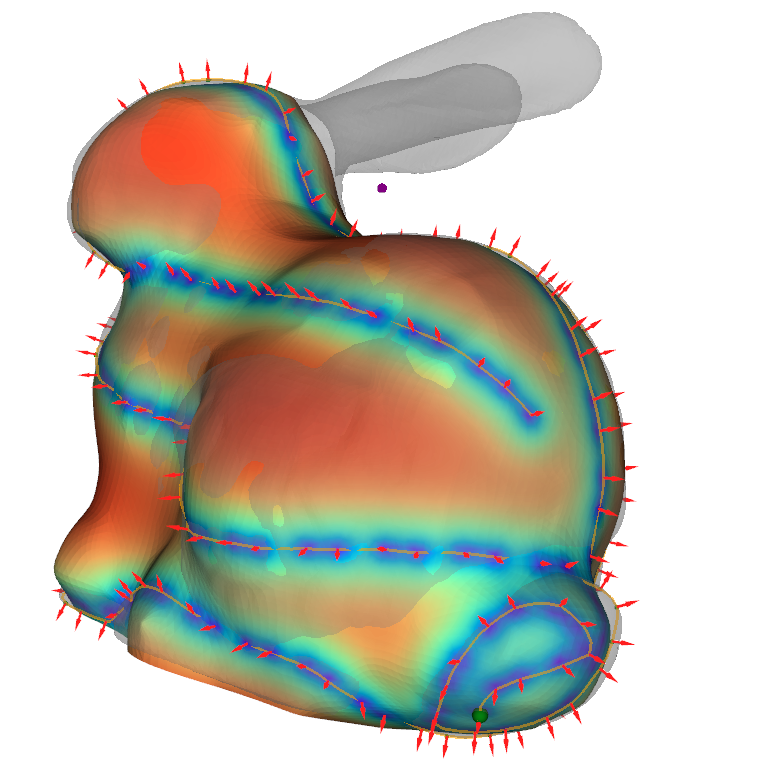} &
        \includegraphics[width=0.45\columnwidth]{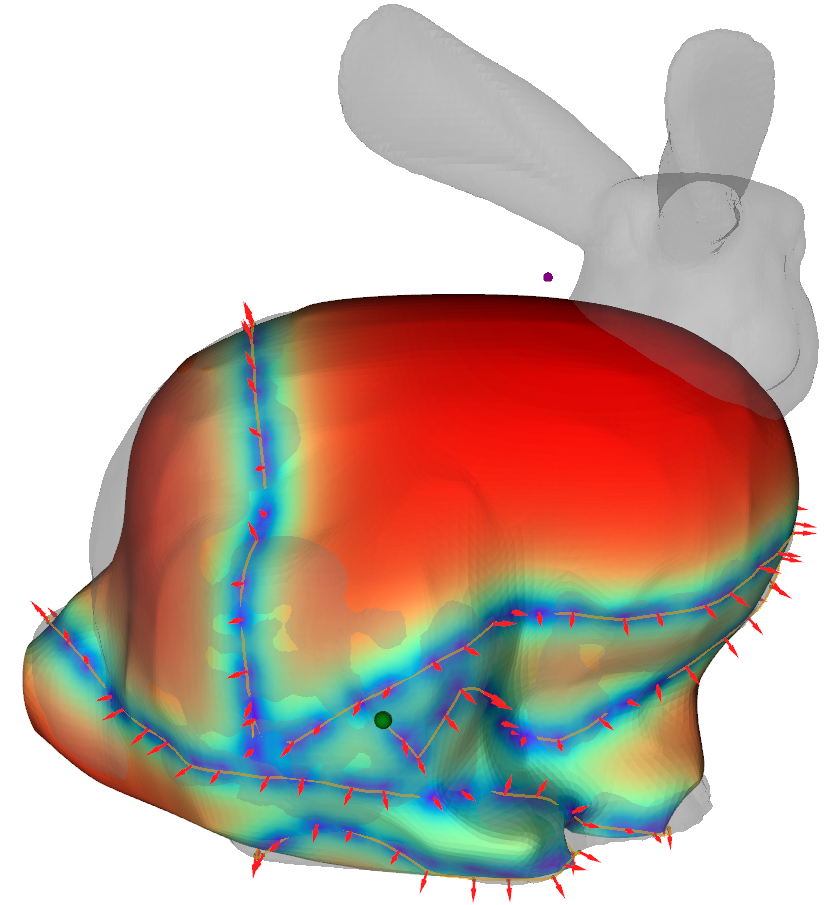} \\
    \end{tabular}
    \caption{Gradient-based policies concentrate exploration in local regions without awareness of global uncertainty, missing distant targets that heat diffusion naturally discovers.}
    \label{fig:local_optima}
	\vspace{-1\baselineskip}
\end{figure}

Gradient-based policies follow local peaks without global awareness. 
This myopic behavior (see \Cref{fig:local_optima}) contrasts with diffusion-based methods that naturally integrate global information.

\subsection{Single-Run Analysis: Bunny Run 7}
\label{app:bunny_run7}

\begin{figure}
    \centering
    \includegraphics[width=0.9\columnwidth]{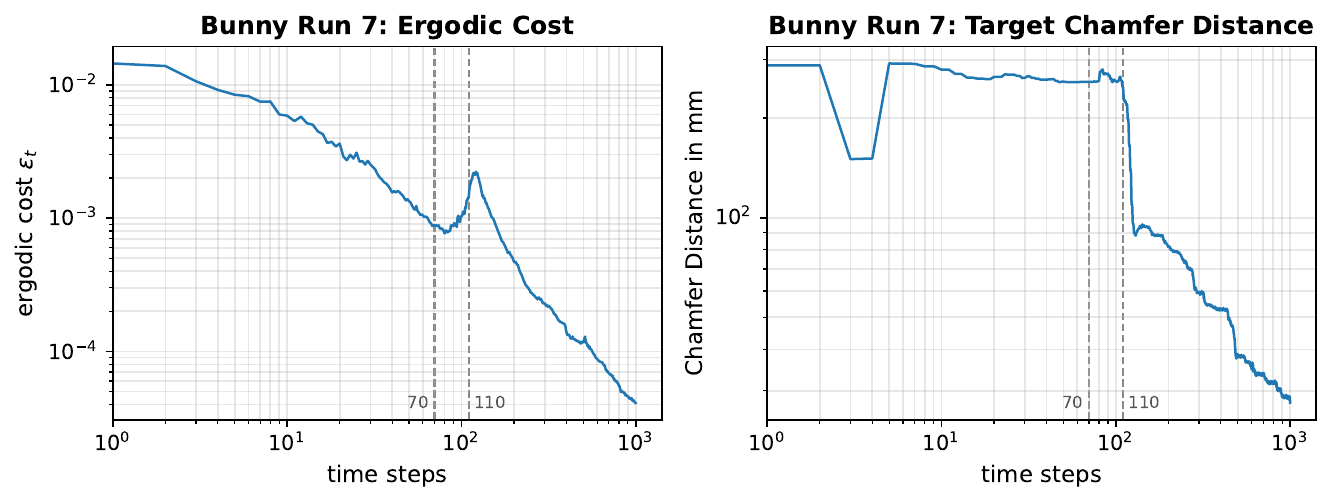}\\[0.5em]
    \begin{tabular}{@{}cc@{}}
        \includegraphics[width=0.4\columnwidth,clip,trim=300 0 540 0]{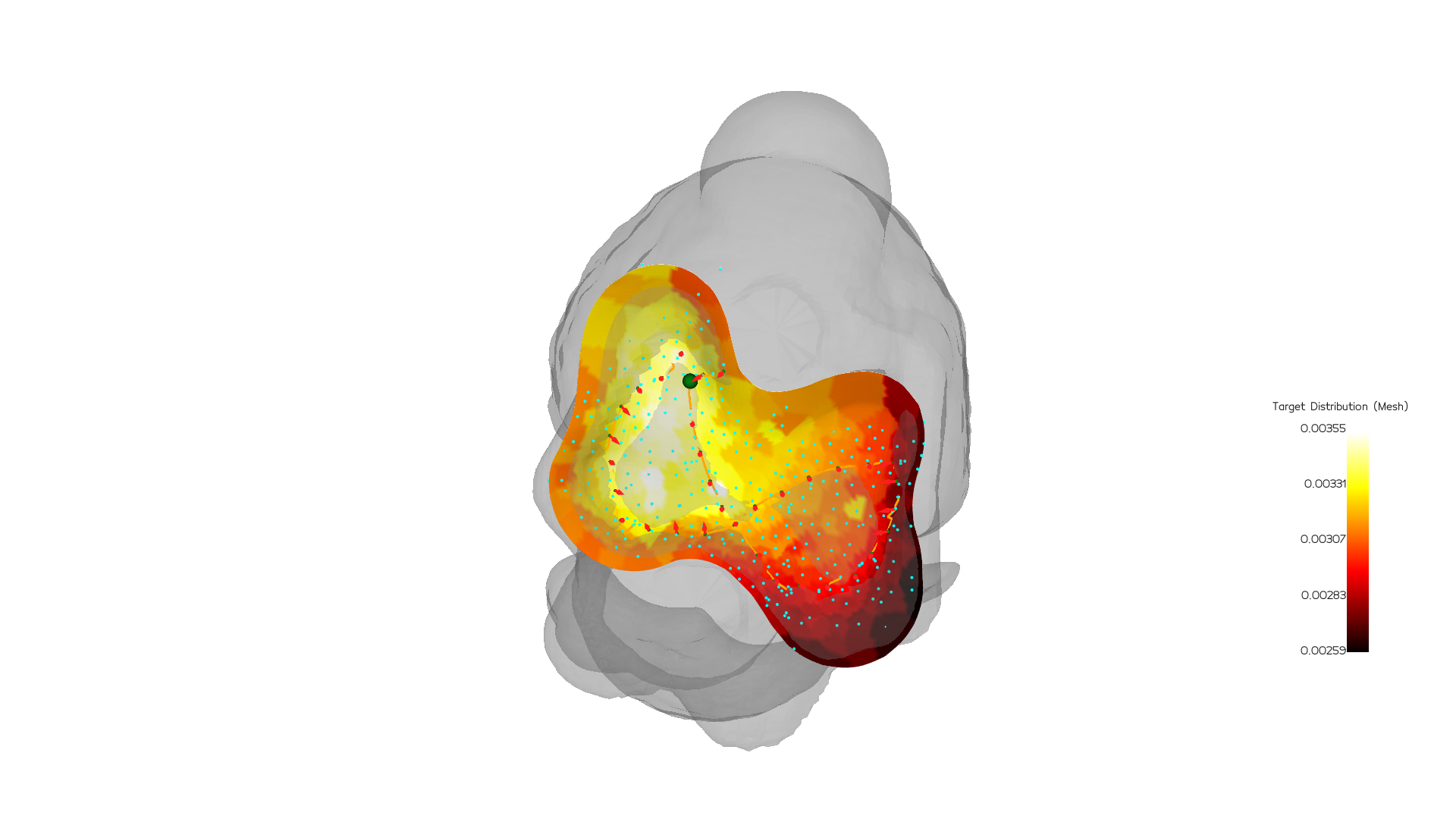} &
        \includegraphics[width=0.4\columnwidth,clip,trim=300 0 540 0]{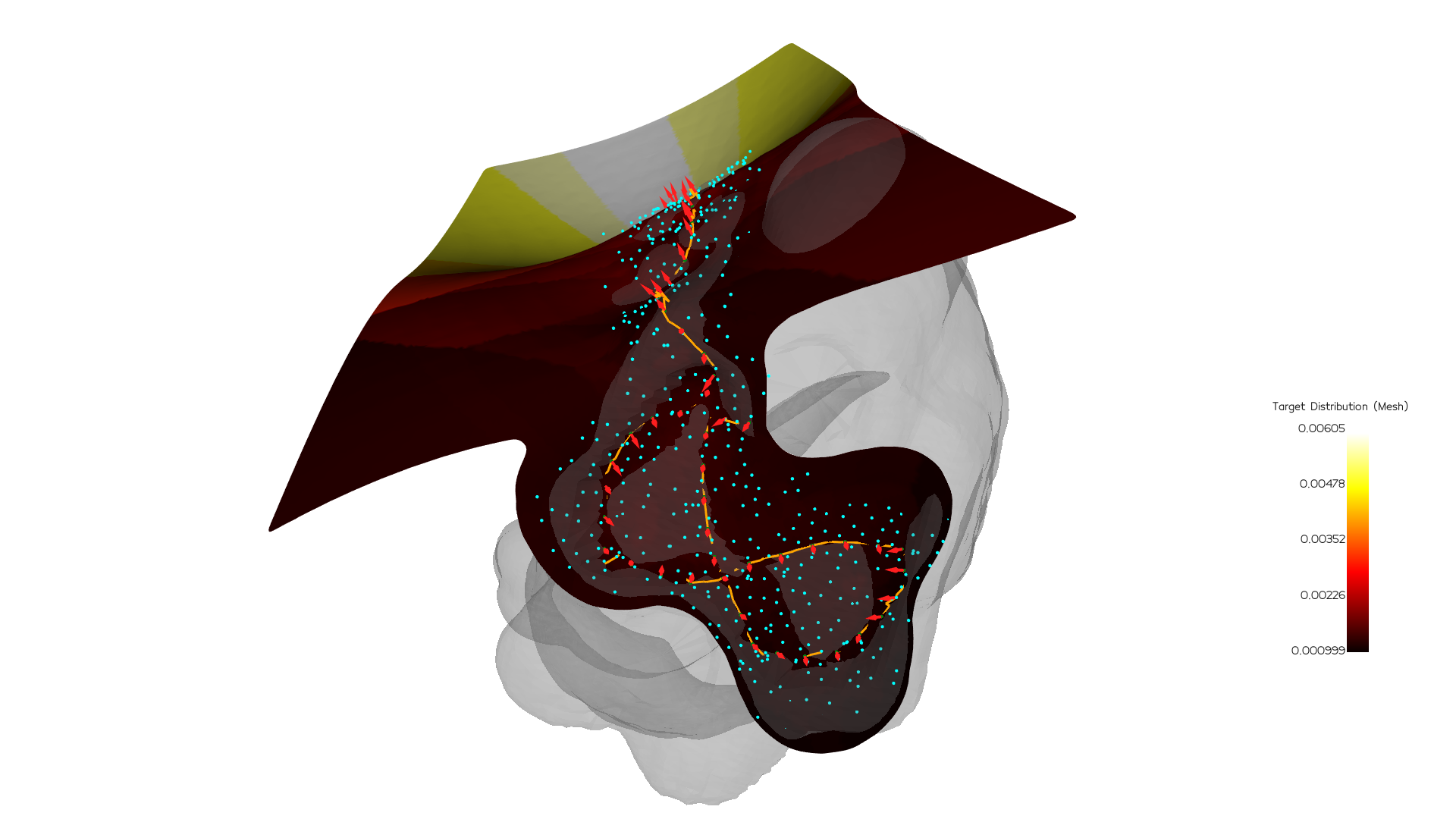} \\
        {\scriptsize Time step 70} & {\scriptsize Time step 110} \\
    \end{tabular}
    \caption{Bunny Run 7 convergence behavior: (top left) ergodic cost over 10000 time steps; (top right) target Chamfer distance demonstrating surface reconstruction improvement. The bottom row shows exploration snapshots before (time step 70) and after (time step 110) the transition at step $\approx 100$, where the agent reaches the surface near the heat source.}
    \label{fig:bunny_run7_single}
\end{figure}

\begin{figure}
	\centering
	\includegraphics[width=\columnwidth]{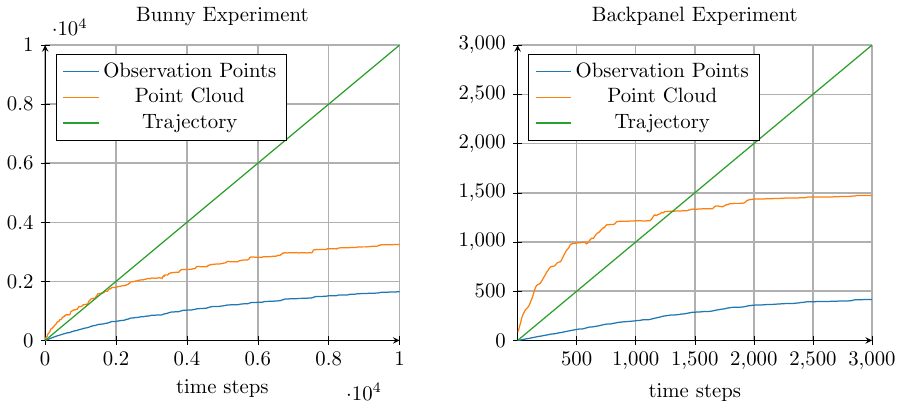}
	\caption{\textbf{Amount of Points:} Comparison of observation points, point cloud, and trajectory length over time for the bunny (left) and backpanel (right) experiments. While the trajectory grows linearly, observation and surface points converge as exploration progresses.}%
	\label{fig:ammount_plots}%
\end{figure}

This run demonstrates an interesting phenomenon in the exploration dynamics revealed in \Cref{fig:bunny_run7_single}. The heat source is injected on the opposite side of the bunny from the starting point. Initially, only the local geometry around the starting region is known, so the target heat distribution is confined to this explored neighborhood. As the agent explores and the surface estimate grows, it progresses toward the heat source location. Around step $\approx 100$ (10\% of execution), the agent reaches and begins to uncover the surface in the vicinity of the heat source, causing the target distribution to shift. This transition creates the visible hill in ergodic cost shown on the left of \Cref{fig:bunny_run7_single}, as the target heat moves from the previously explored back side to the newly discovered region around the heat source. Concurrently, the target Chamfer distance (right panel) drops sharply: it measures reconstruction quality specifically around the target heat region, which improves rapidly once that surface is being revealed --- not the entire bunny, but the part of the surface surrounding the heat source. The control law adapts to this geometric discovery and the ergodic cost converges as the relevant surface becomes well identified. Note that on the logarithmic time axis, step 100 visually appears much later than its actual position (10\%) in the execution timeline.

\subsection{Visualization of Exploration Progression}
\label{sec:progression}

\begin{figure*}[htbp]
    \centering
    \begin{tabular}{@{}r@{\hspace{0.3em}}ccccc@{}}
        \multirow{2}{*}[3em]{\rotatebox{90}{\small Coverage}} &
        \includegraphics[width=0.17\textwidth,clip,trim=420 0 420 0]{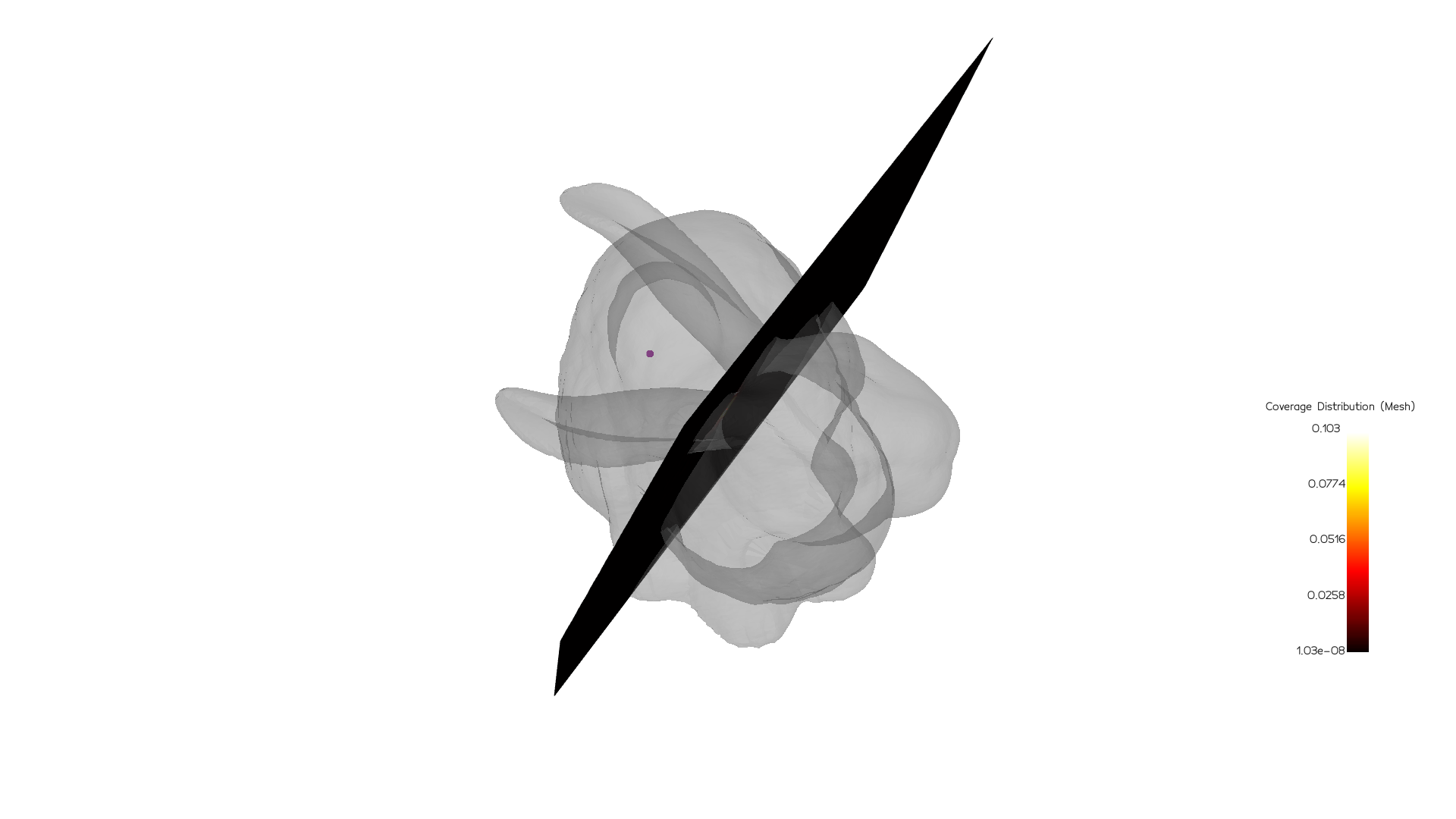} &
        \includegraphics[width=0.17\textwidth,clip,trim=420 0 420 0]{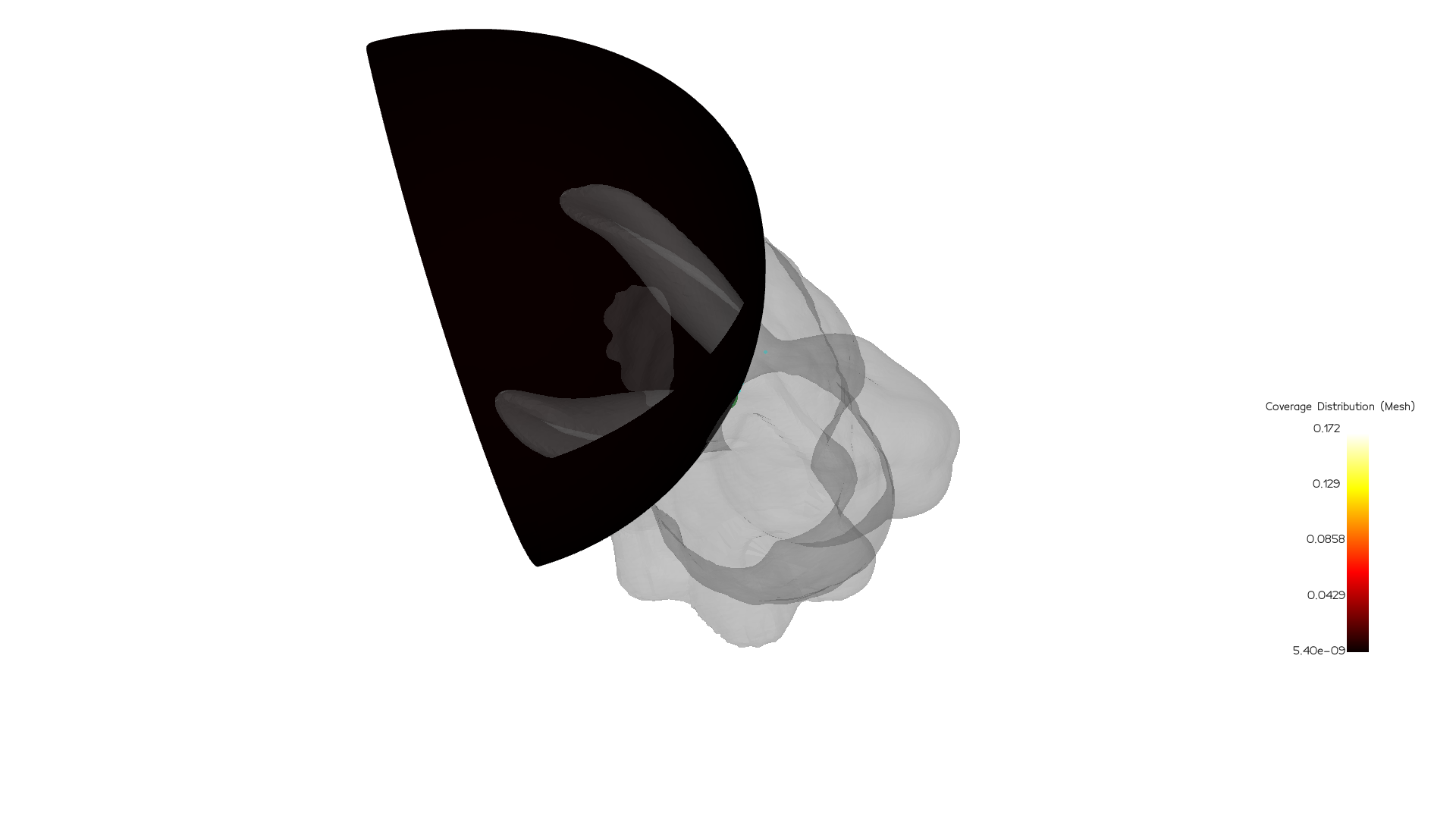} &
        \includegraphics[width=0.17\textwidth,clip,trim=420 0 420 0]{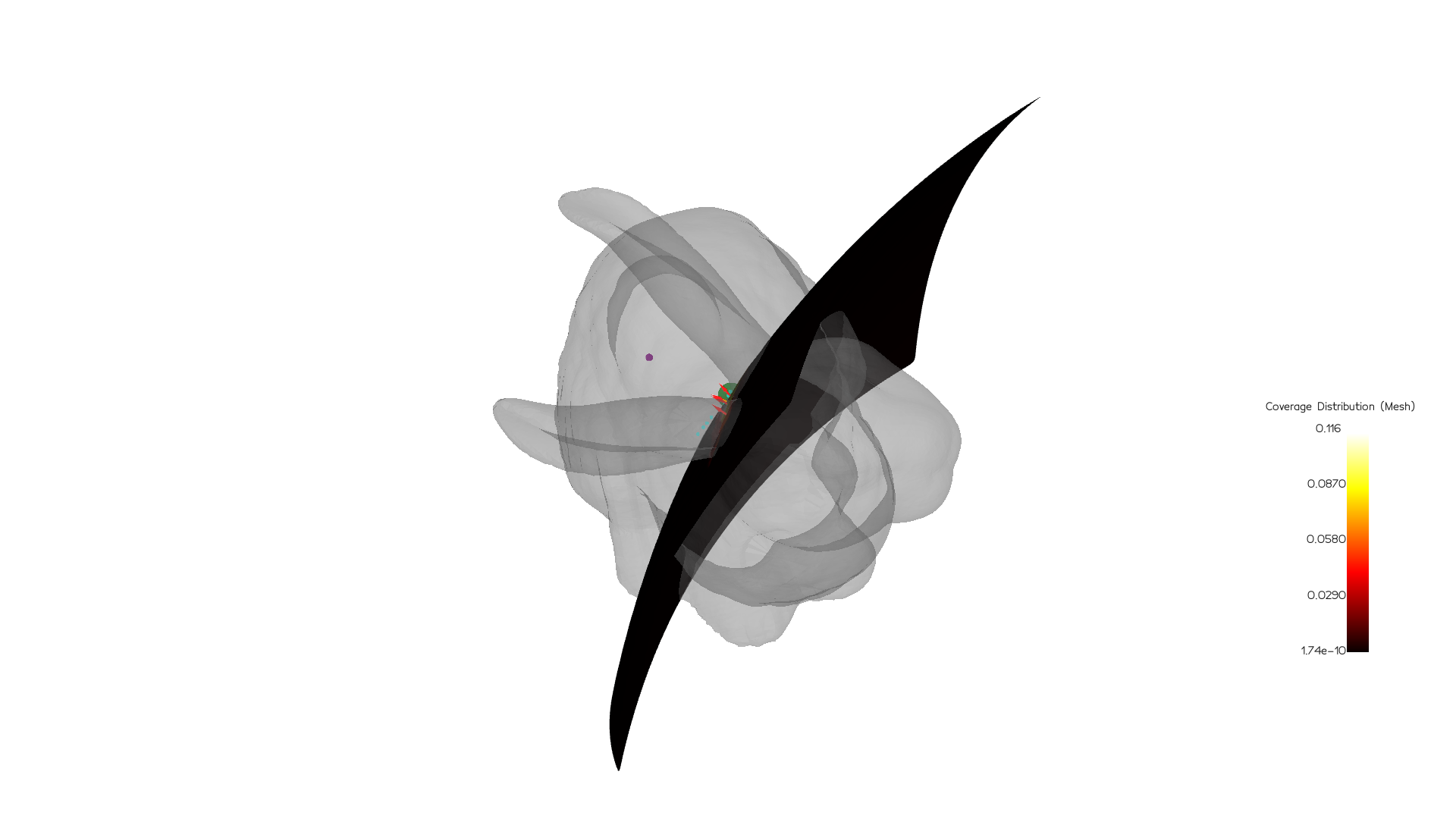} &
        \includegraphics[width=0.17\textwidth,clip,trim=420 0 420 0]{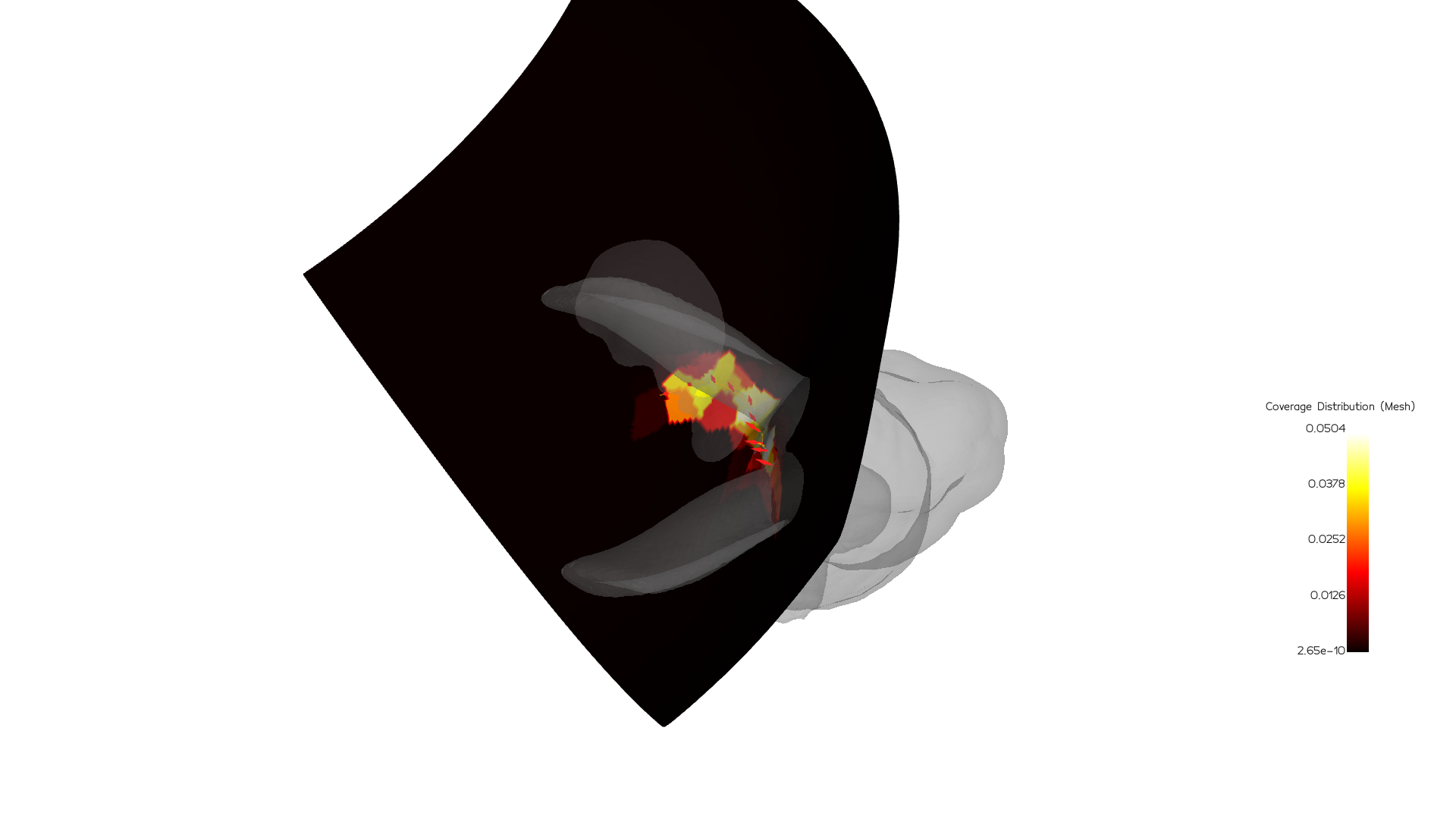} &
        \includegraphics[width=0.17\textwidth,clip,trim=420 0 420 0]{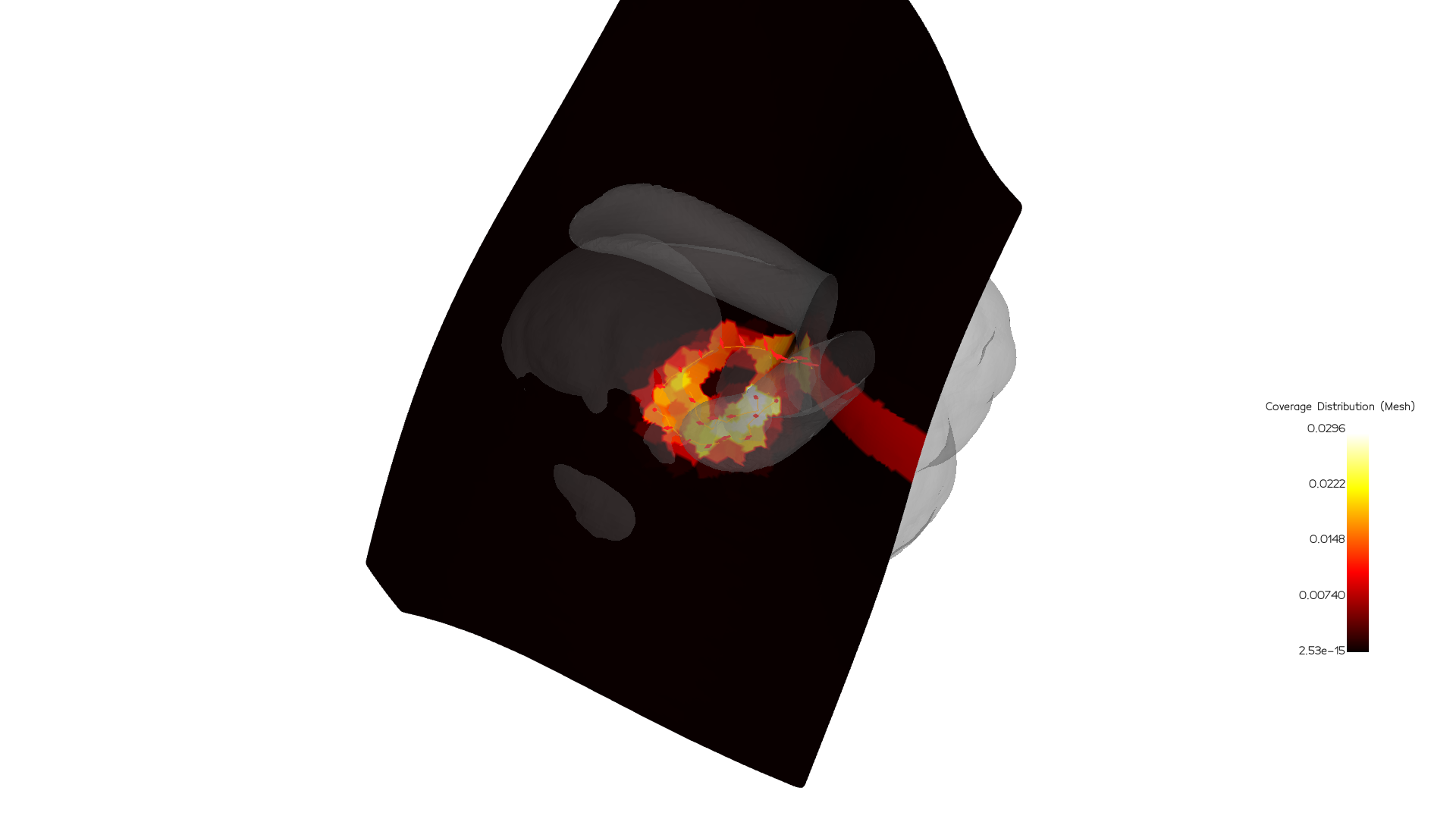} \\
        &
        \includegraphics[width=0.17\textwidth,clip,trim=420 0 420 0]{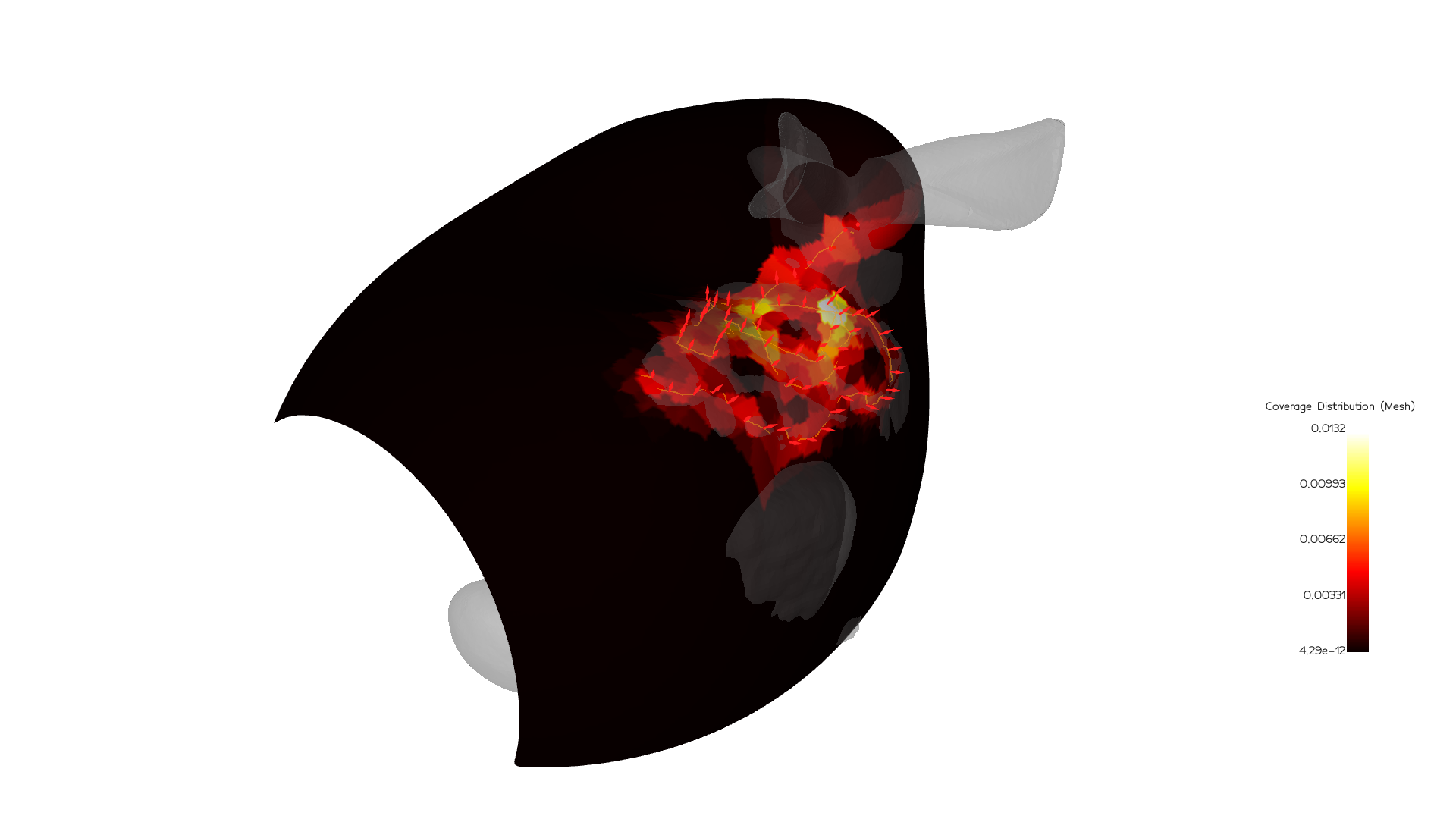} &
        \includegraphics[width=0.17\textwidth,clip,trim=420 0 420 0]{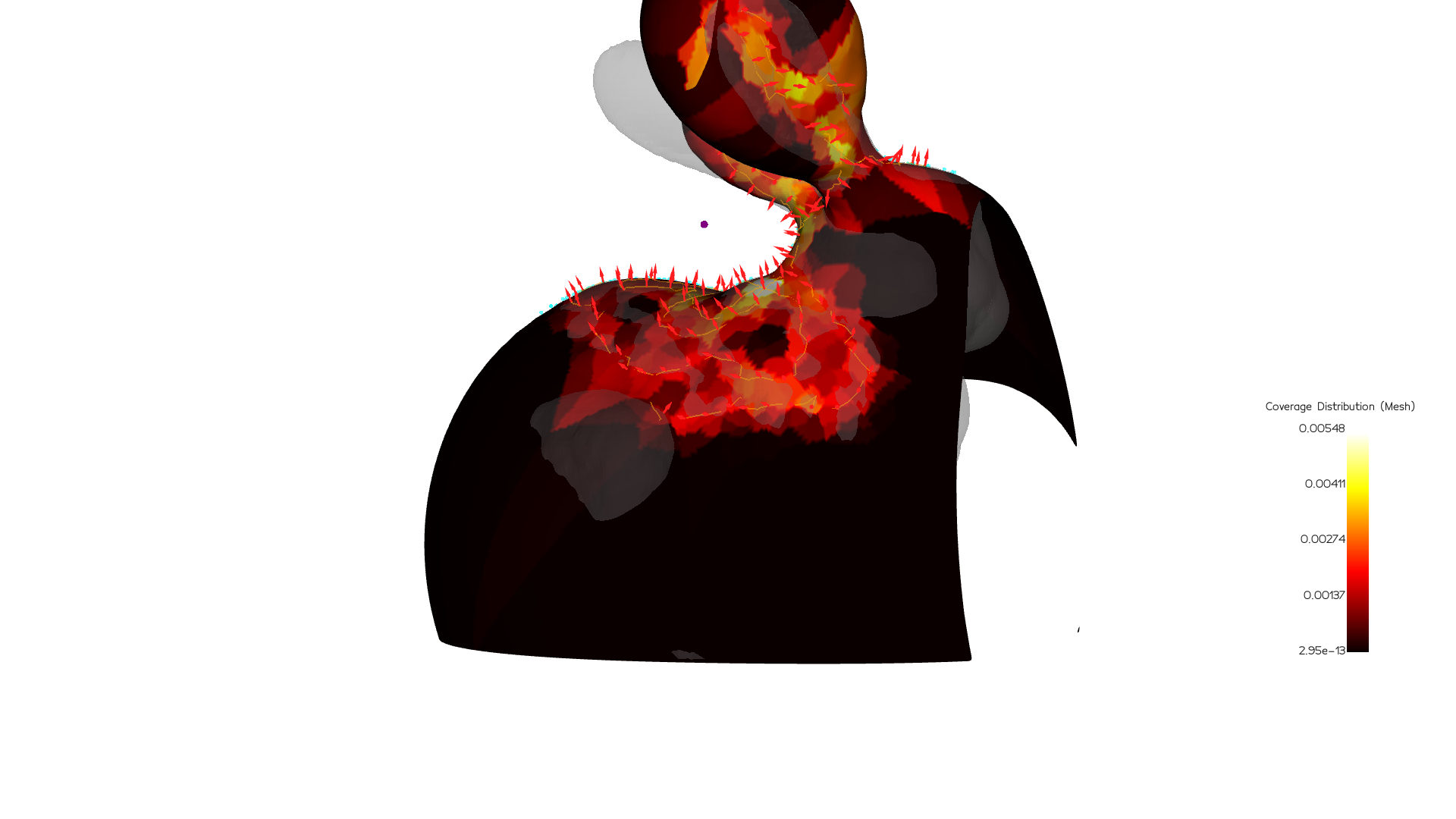} &
        \includegraphics[width=0.17\textwidth,clip,trim=420 0 420 0]{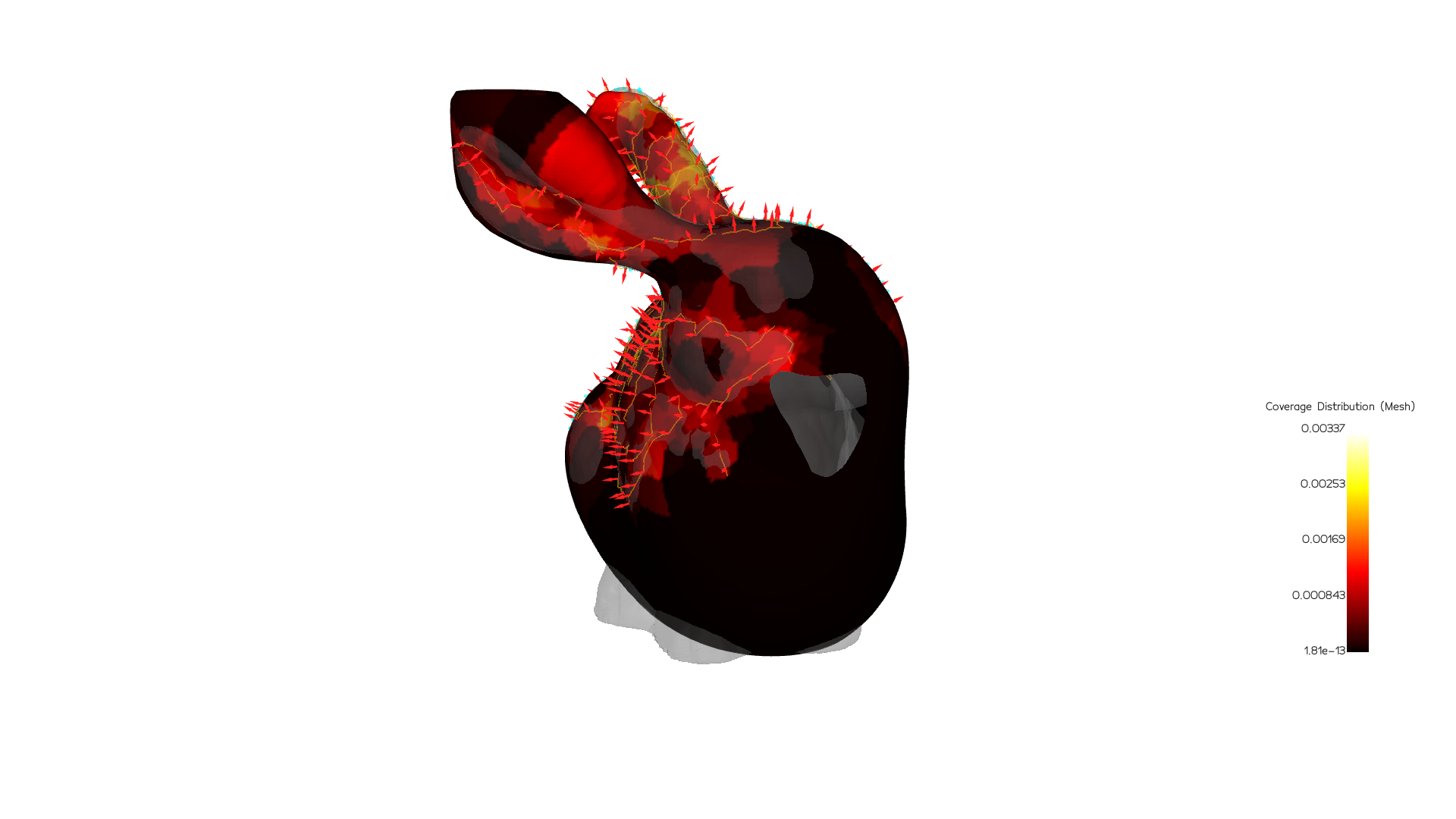} &
        \includegraphics[width=0.17\textwidth,clip,trim=420 0 420 0]{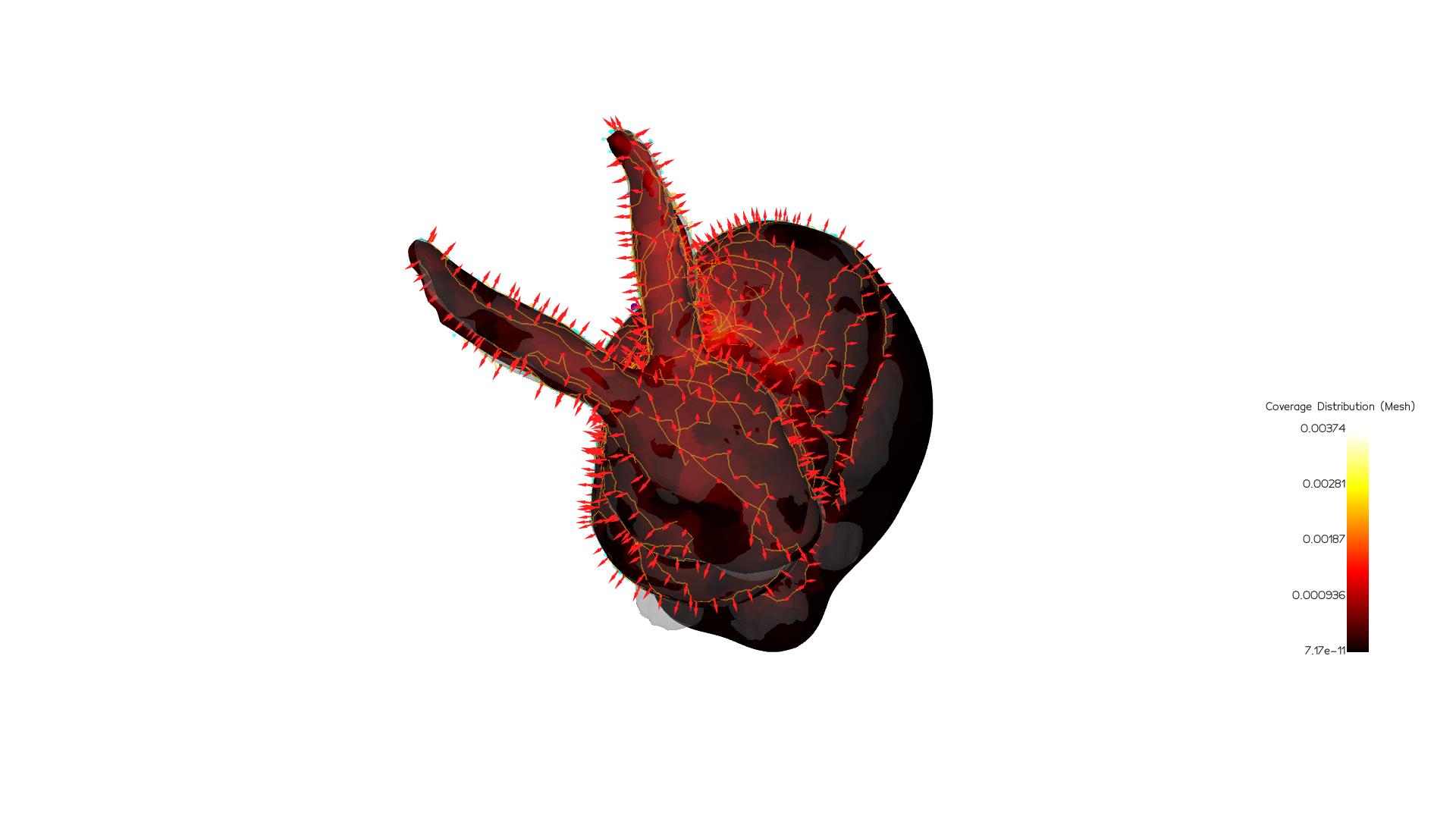} &
        \includegraphics[width=0.17\textwidth,clip,trim=420 0 420 0]{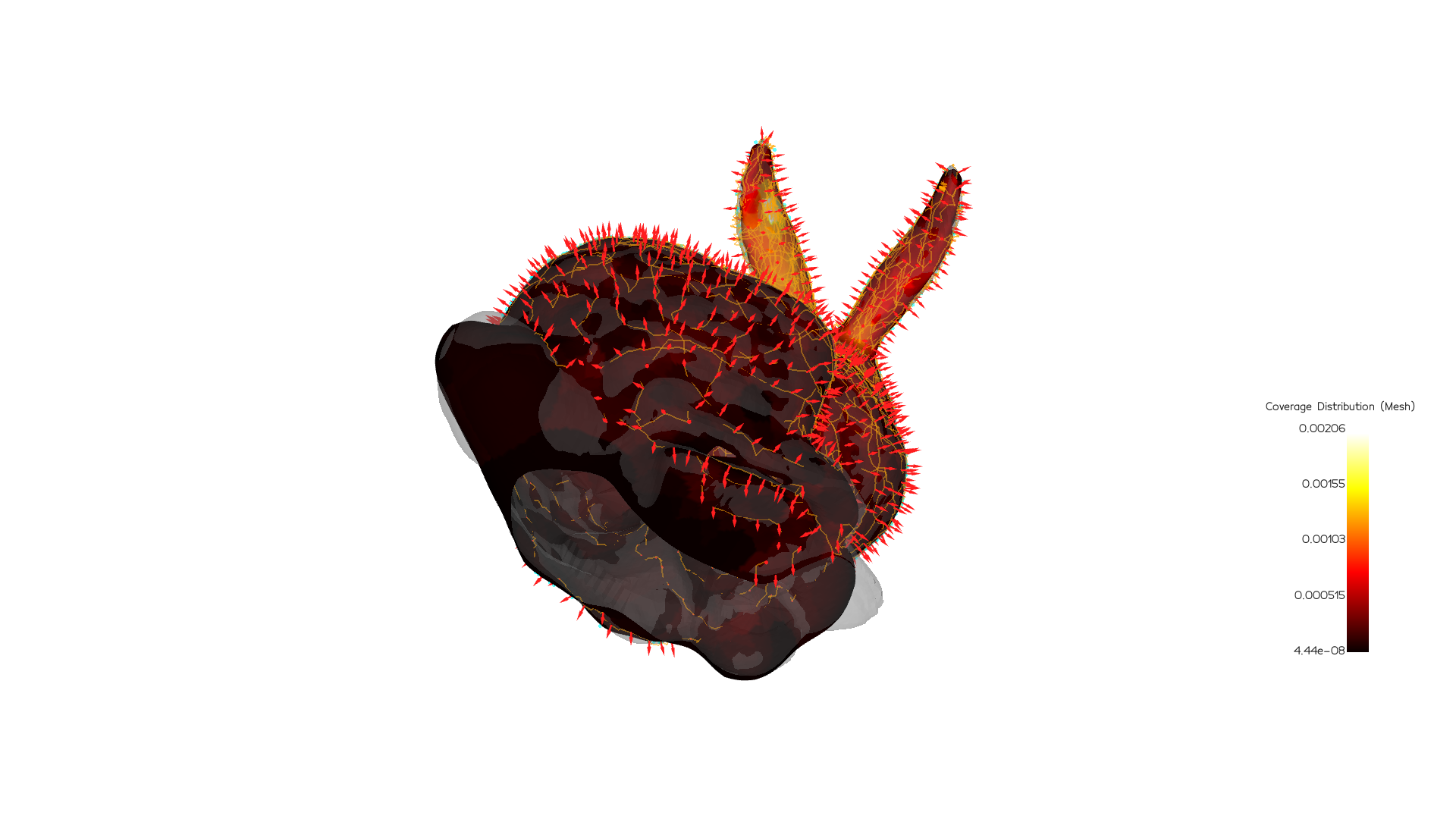} \\[0.1em]
        \cline{2-6}\\[-0.7em]
        \multirow{2}{*}[3em]{\rotatebox{90}{\small Target}} &
        \includegraphics[width=0.17\textwidth,clip,trim=420 0 420 0]{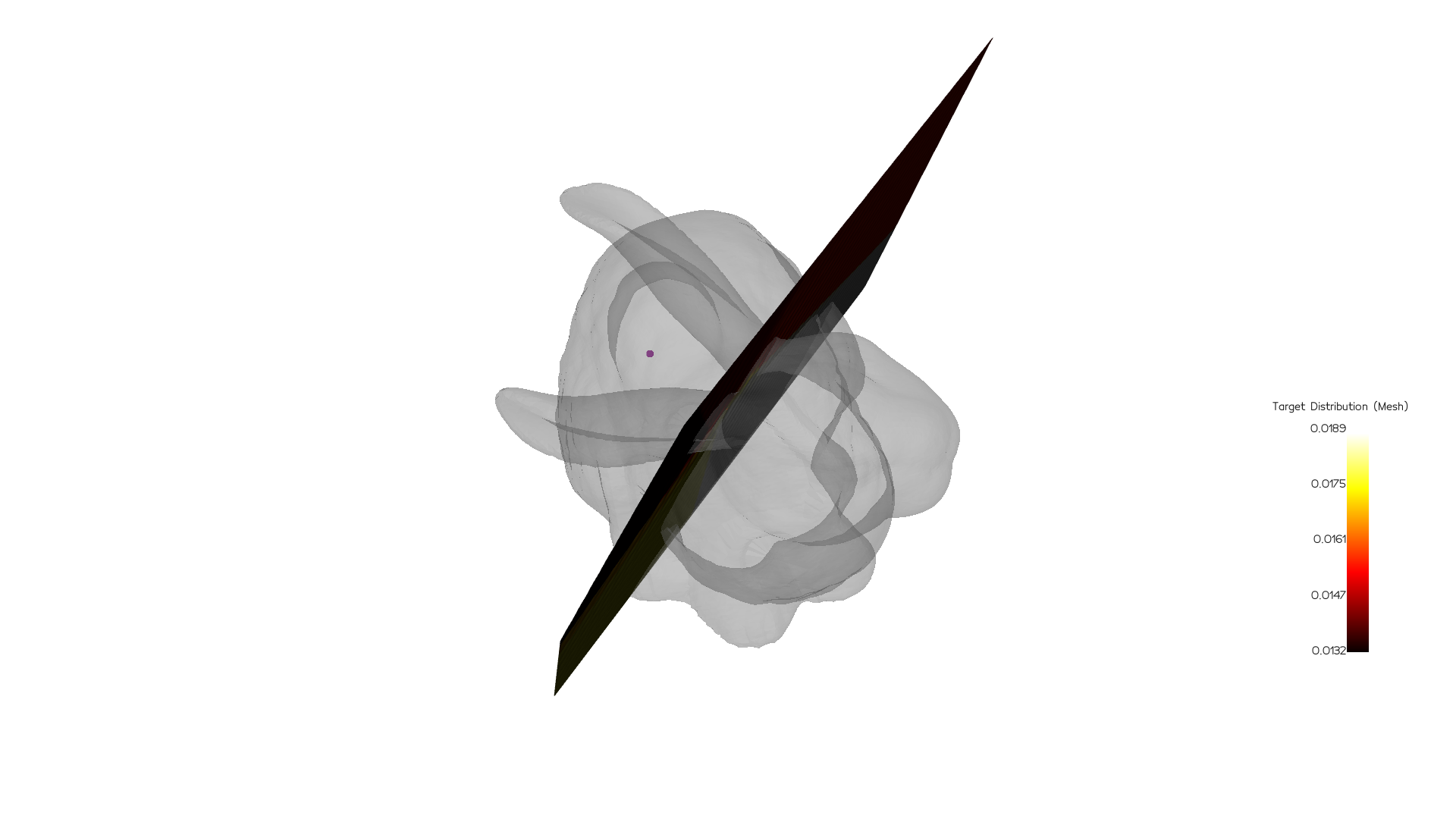} &
        \includegraphics[width=0.17\textwidth,clip,trim=420 0 420 0]{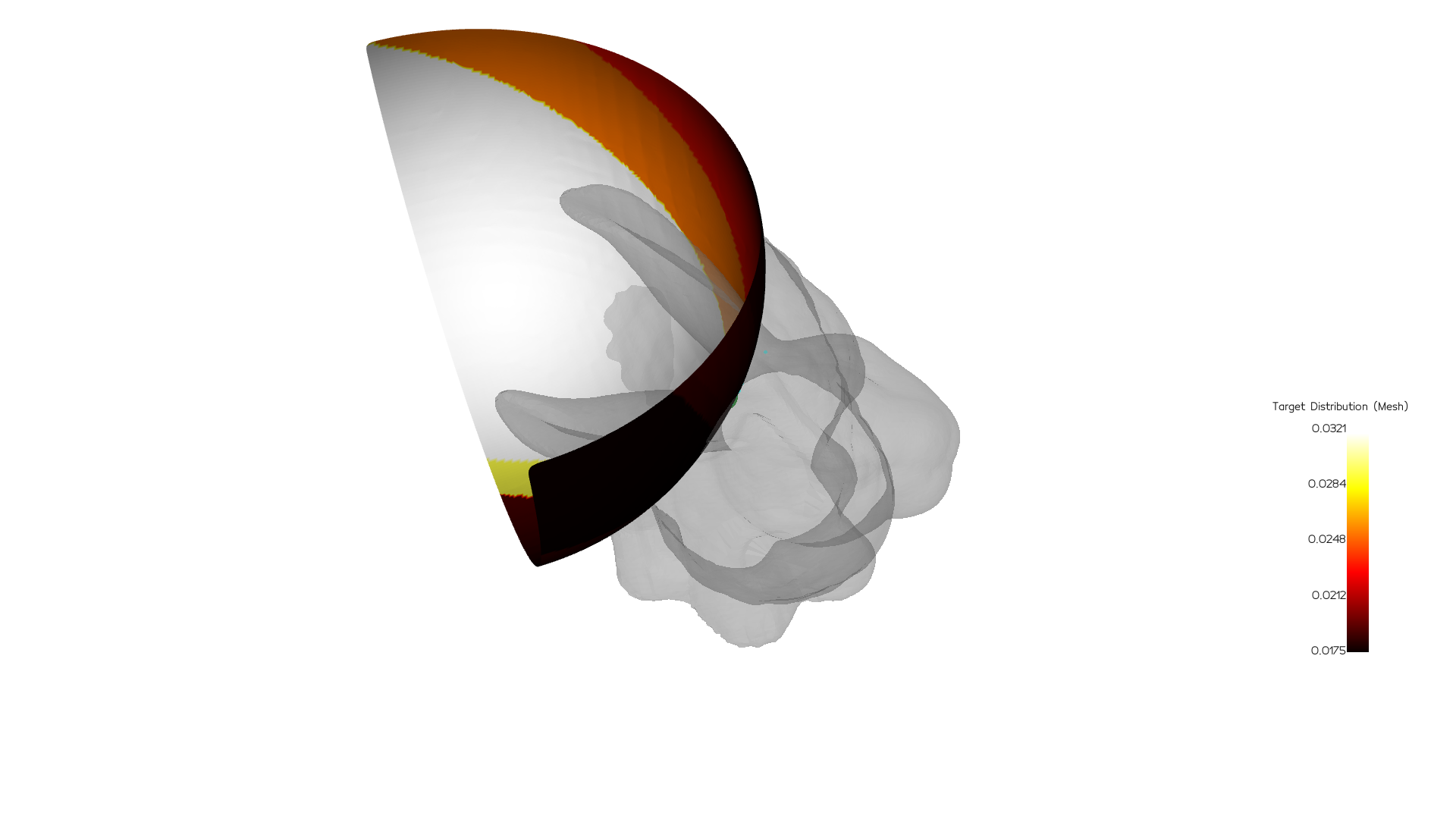} &
        \includegraphics[width=0.17\textwidth,clip,trim=420 0 420 0]{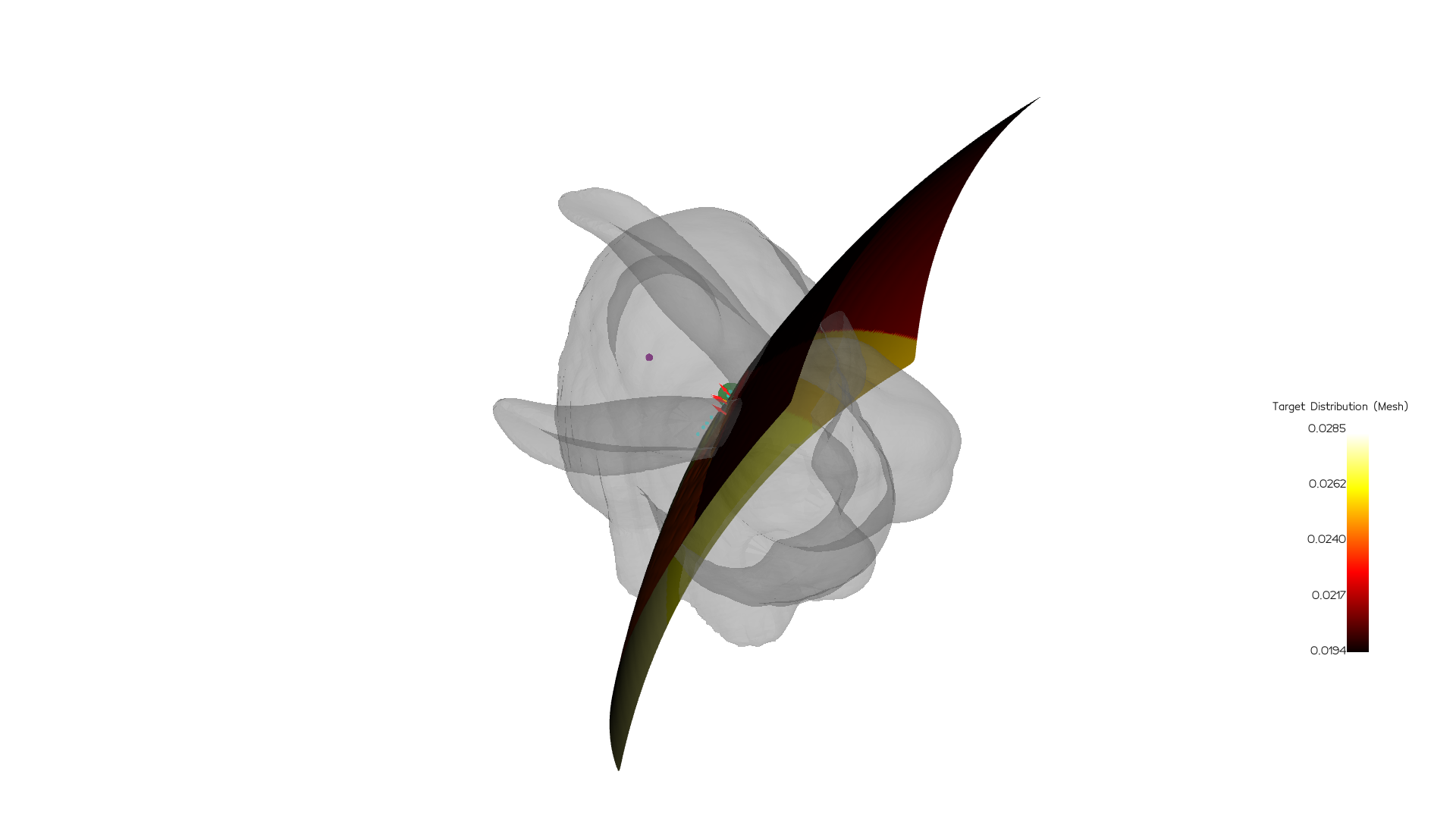} &
        \includegraphics[width=0.17\textwidth,clip,trim=420 0 420 0]{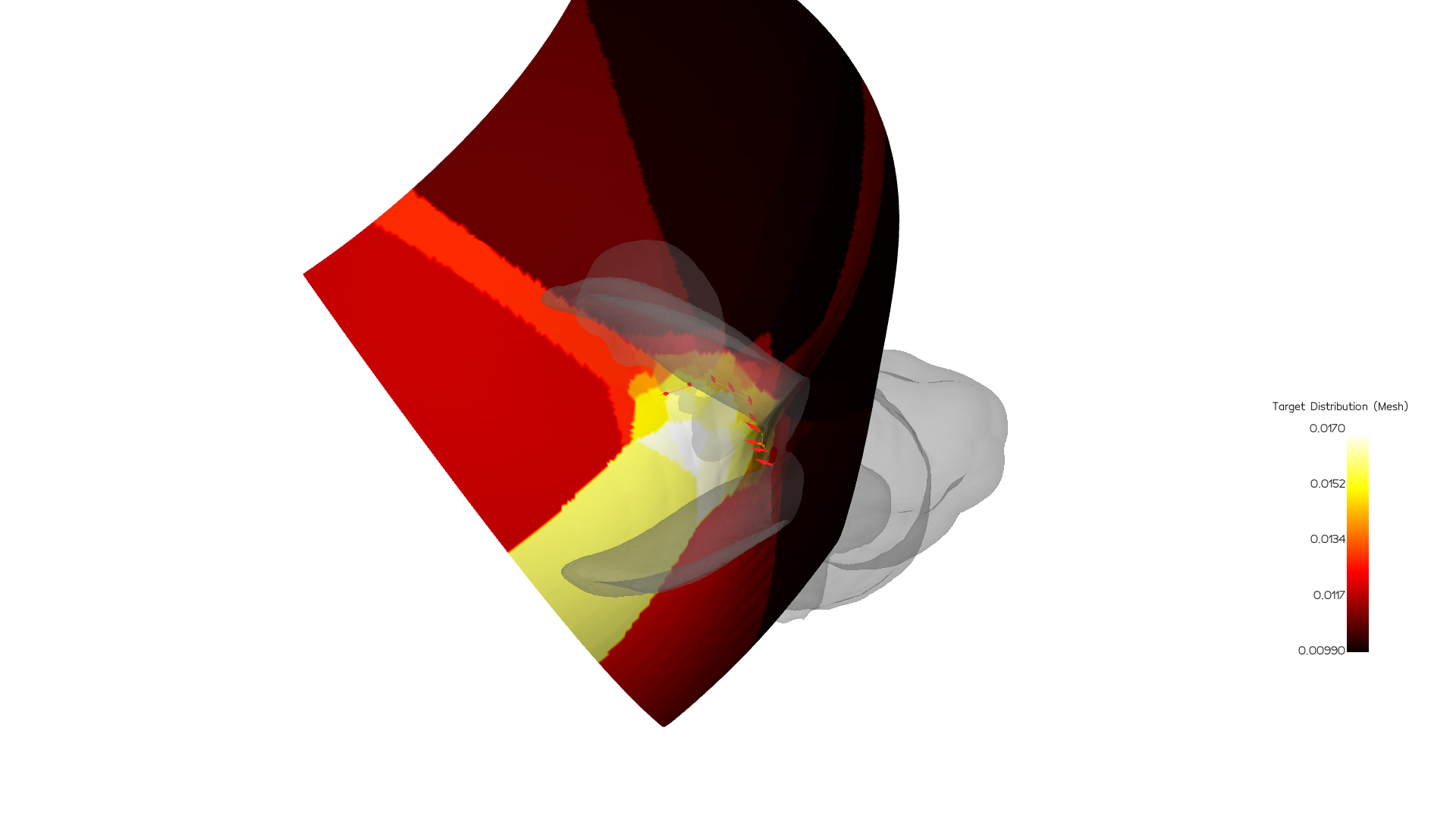} &
        \includegraphics[width=0.17\textwidth,clip,trim=420 0 420 0]{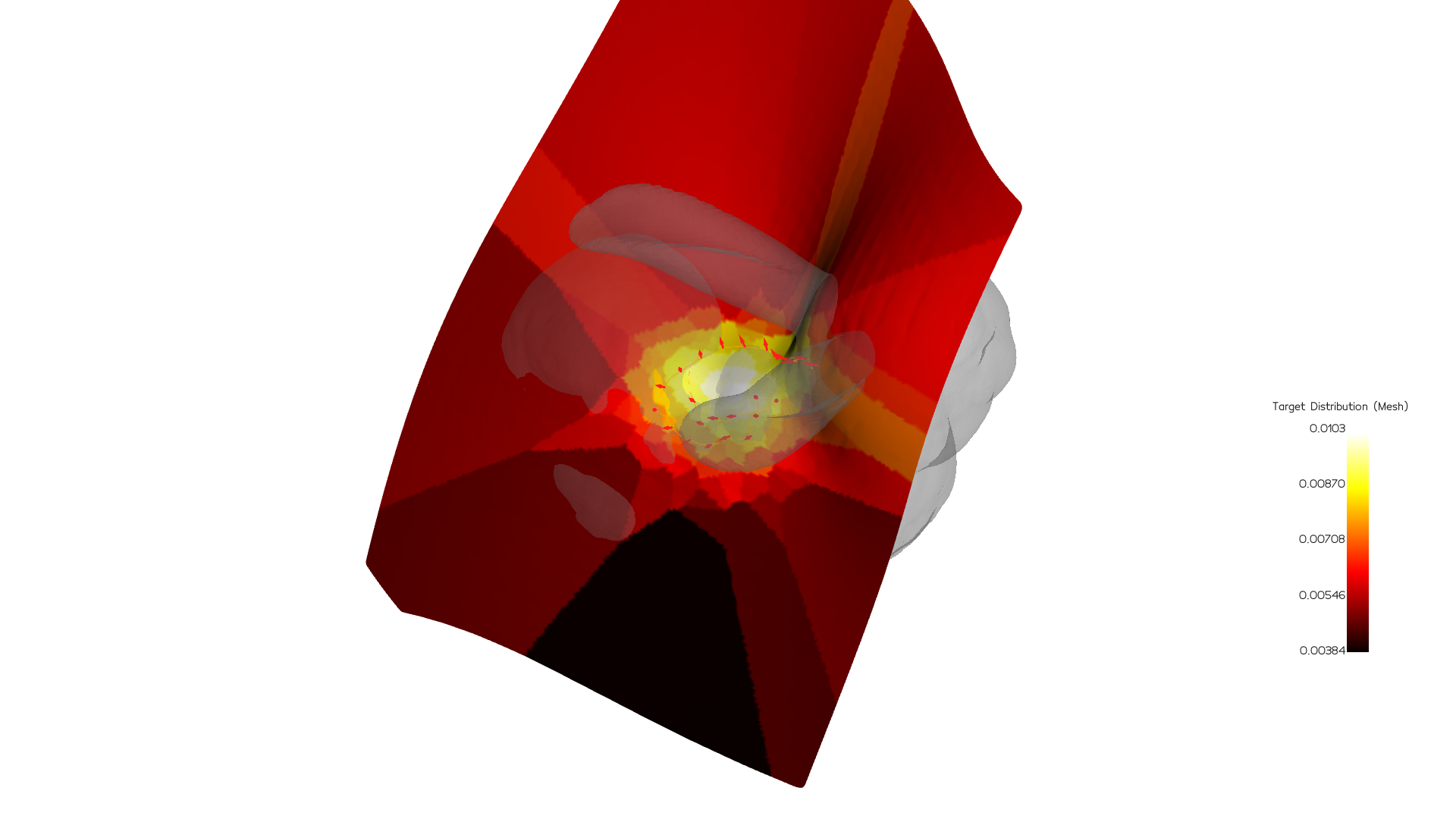} \\
        &
        \includegraphics[width=0.17\textwidth,clip,trim=420 0 420 0]{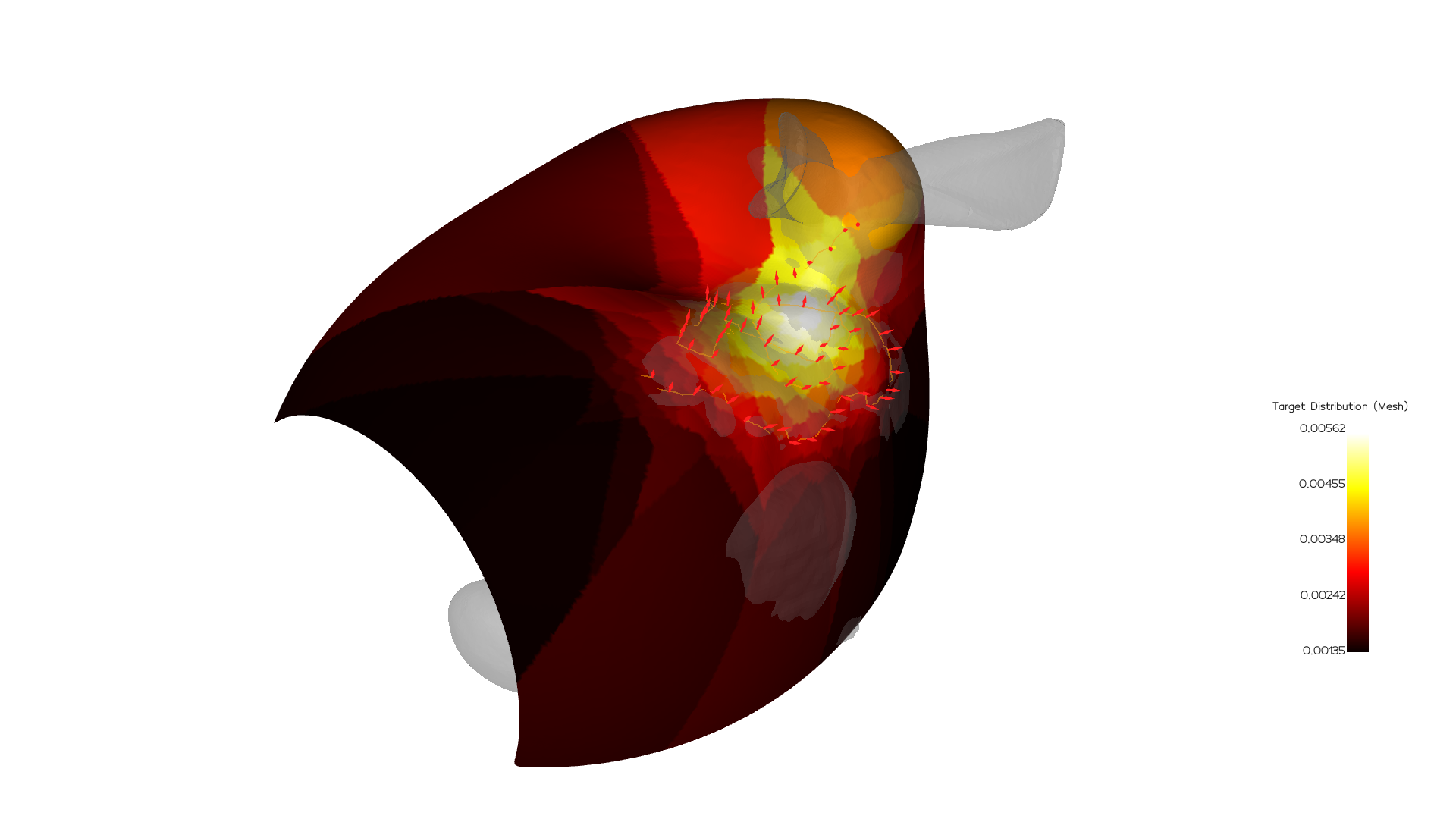} &
        \includegraphics[width=0.17\textwidth,clip,trim=420 0 420 0]{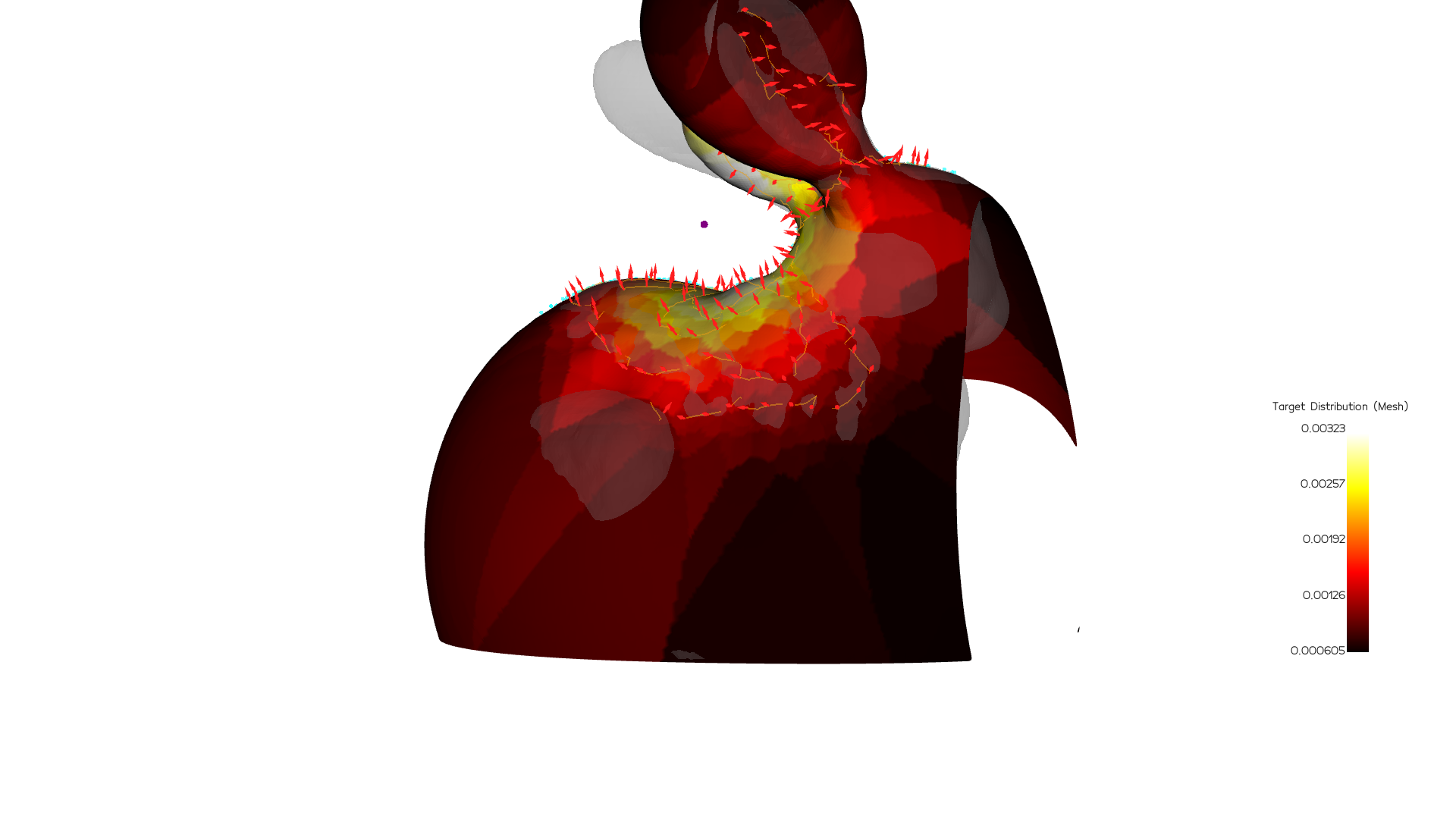} &
        \includegraphics[width=0.17\textwidth,clip,trim=420 0 420 0]{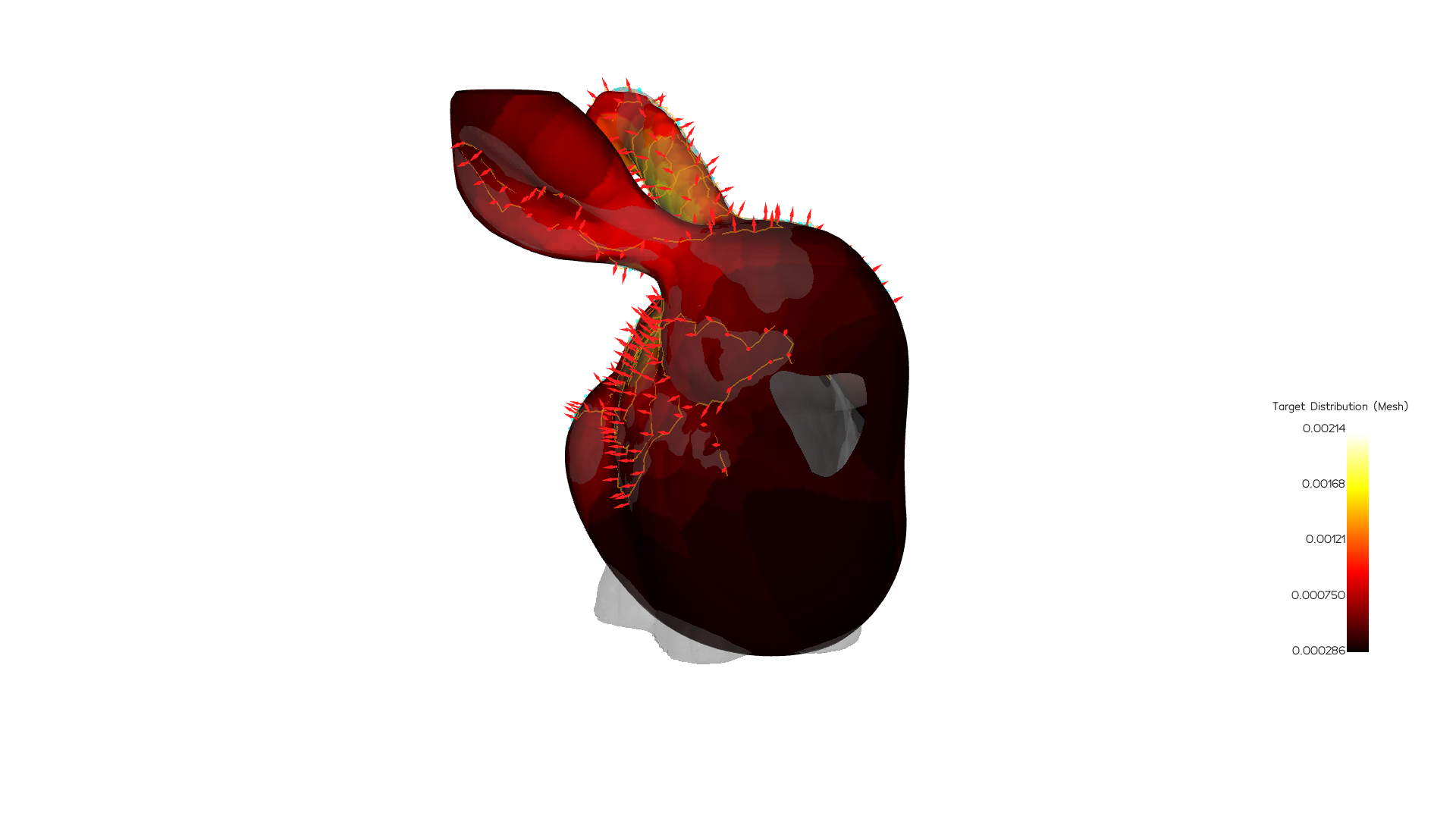} &
        \includegraphics[width=0.17\textwidth,clip,trim=420 0 420 0]{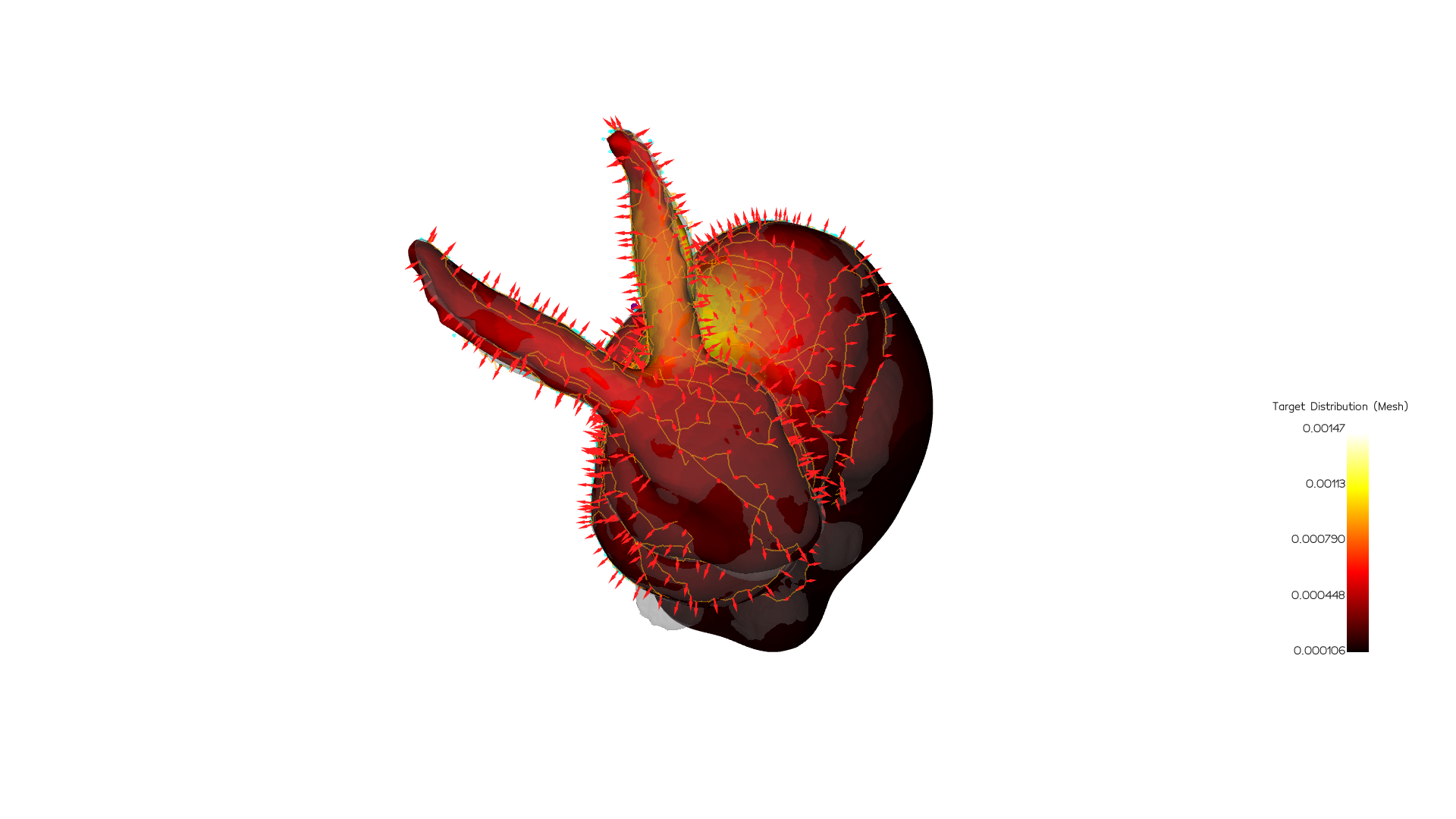} &
        \includegraphics[width=0.17\textwidth,clip,trim=420 0 420 0]{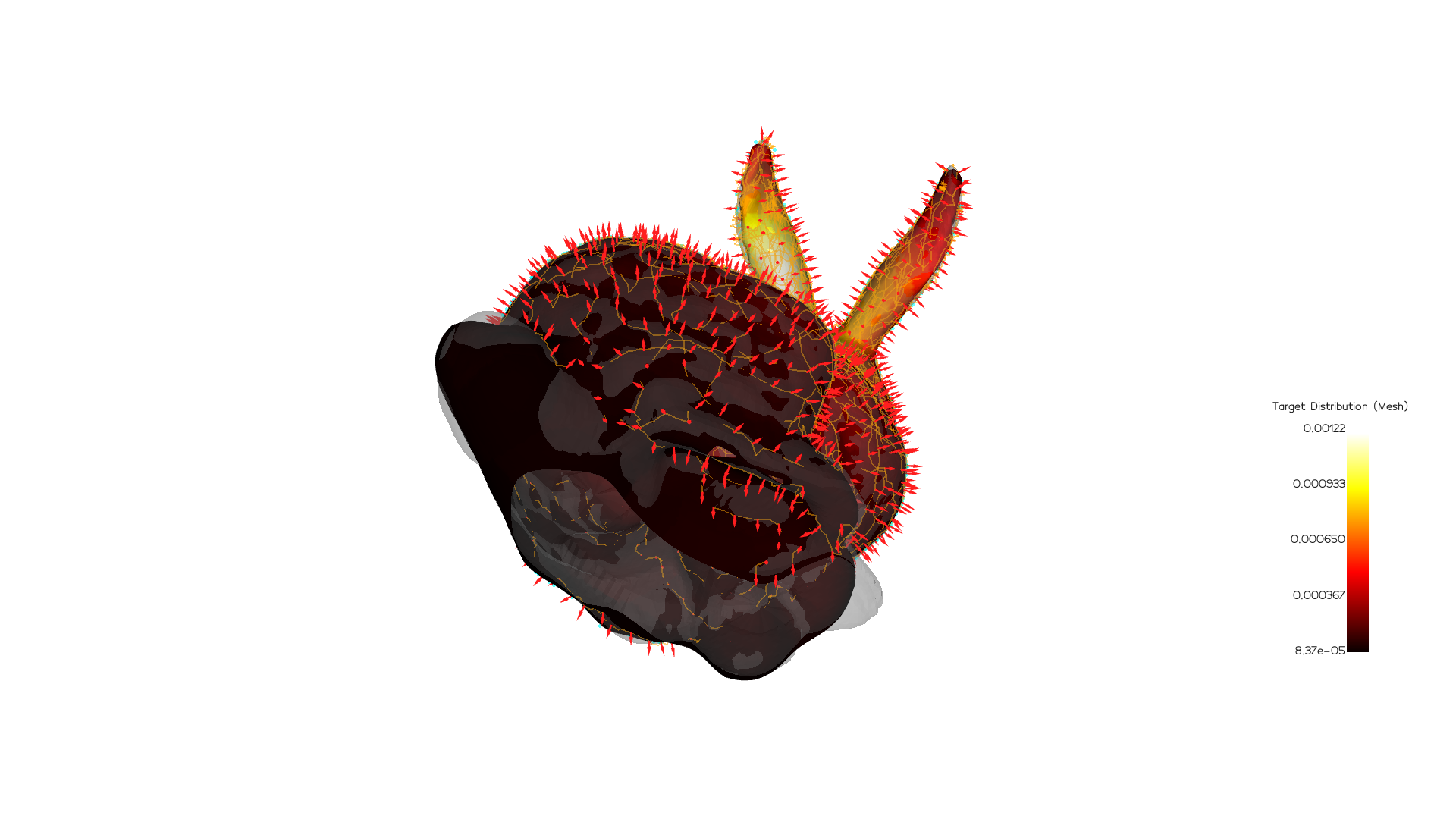} \\[0.1em]
        \cline{2-6}\\[-0.7em]
        \multirow{2}{*}[3em]{\rotatebox{90}{\small Uncertainty}} &
        \includegraphics[width=0.17\textwidth,clip,trim=420 0 420 0]{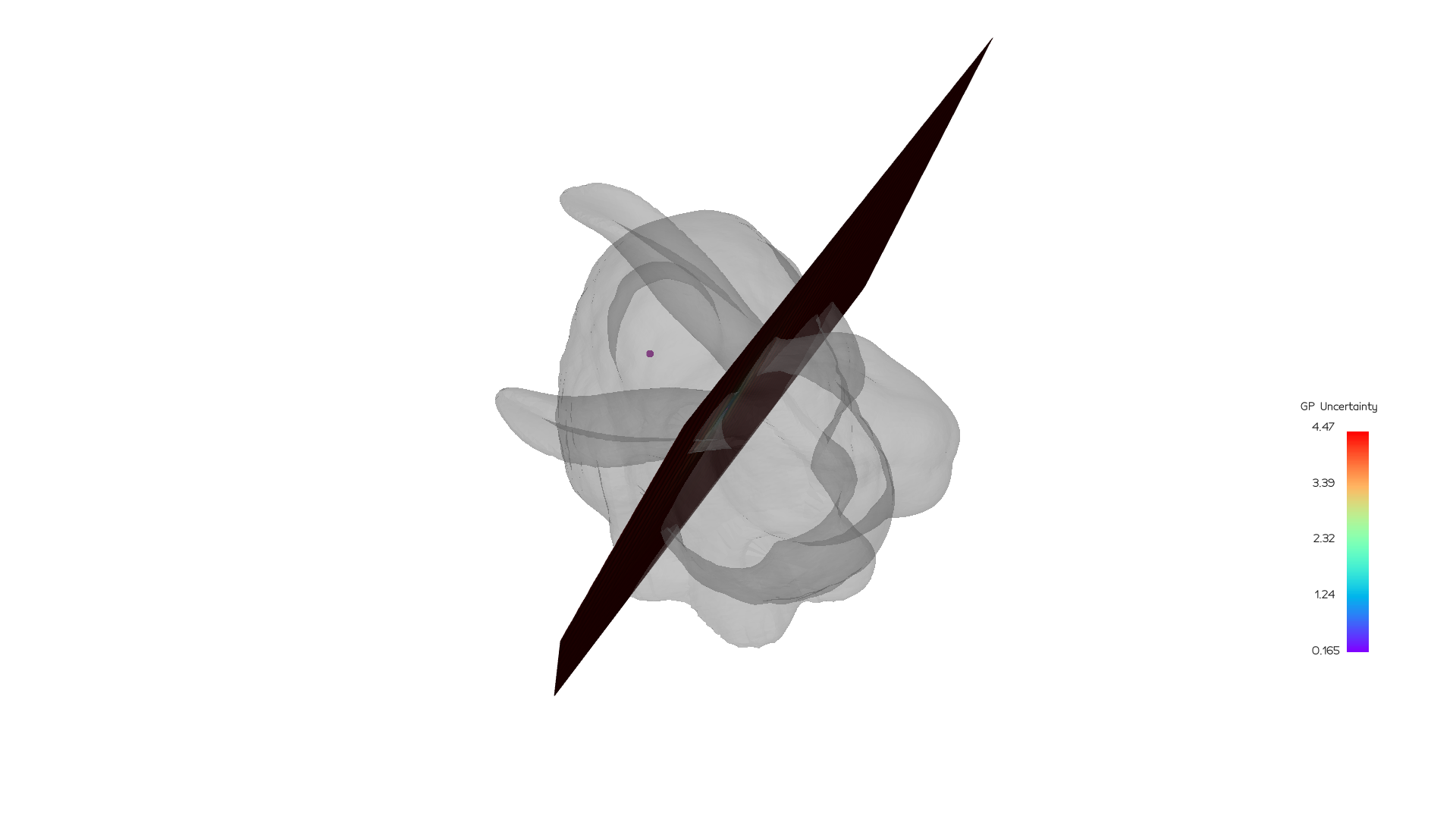} &
        \includegraphics[width=0.17\textwidth,clip,trim=420 0 420 0]{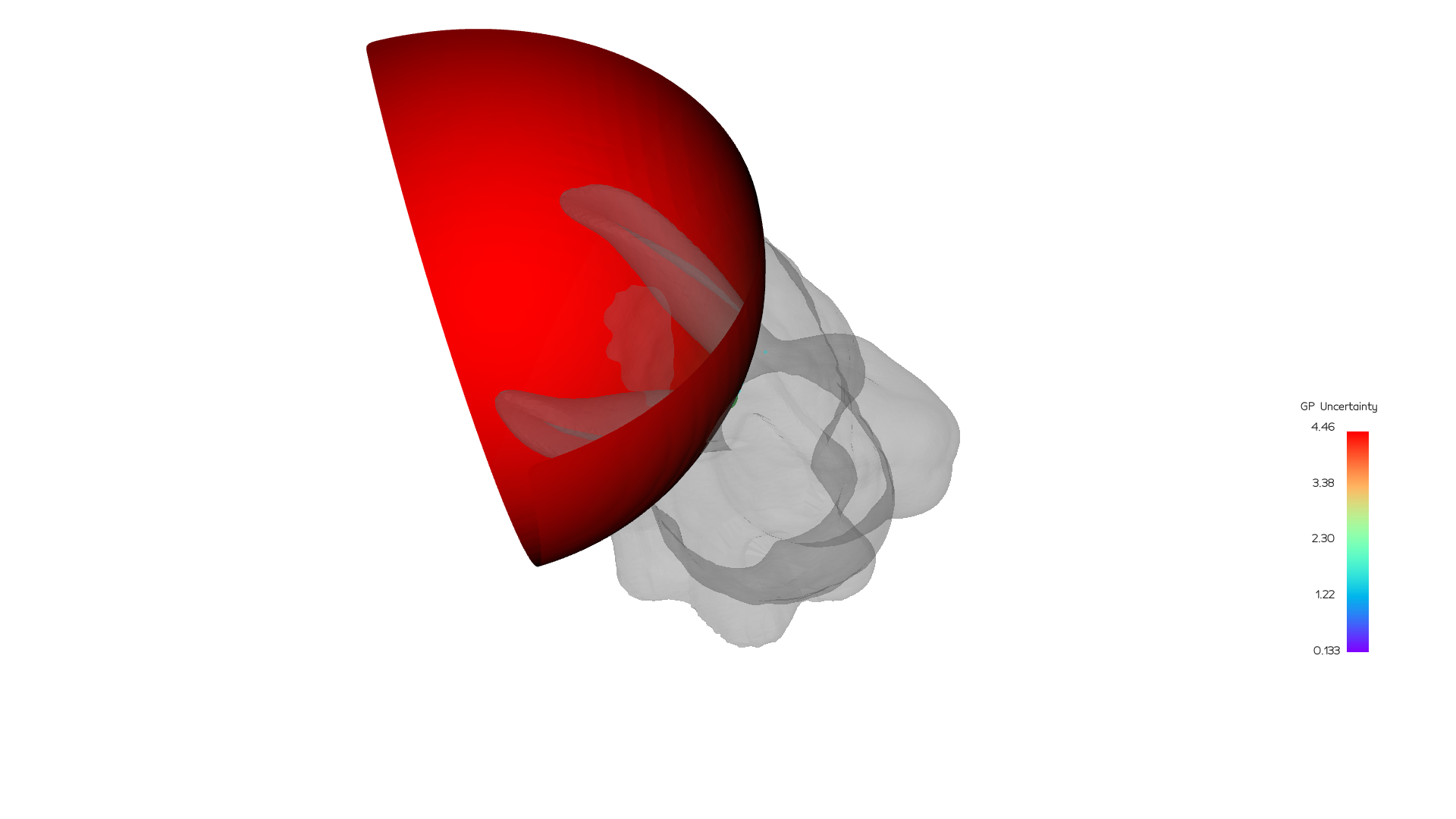} &
        \includegraphics[width=0.17\textwidth,clip,trim=420 0 420 0]{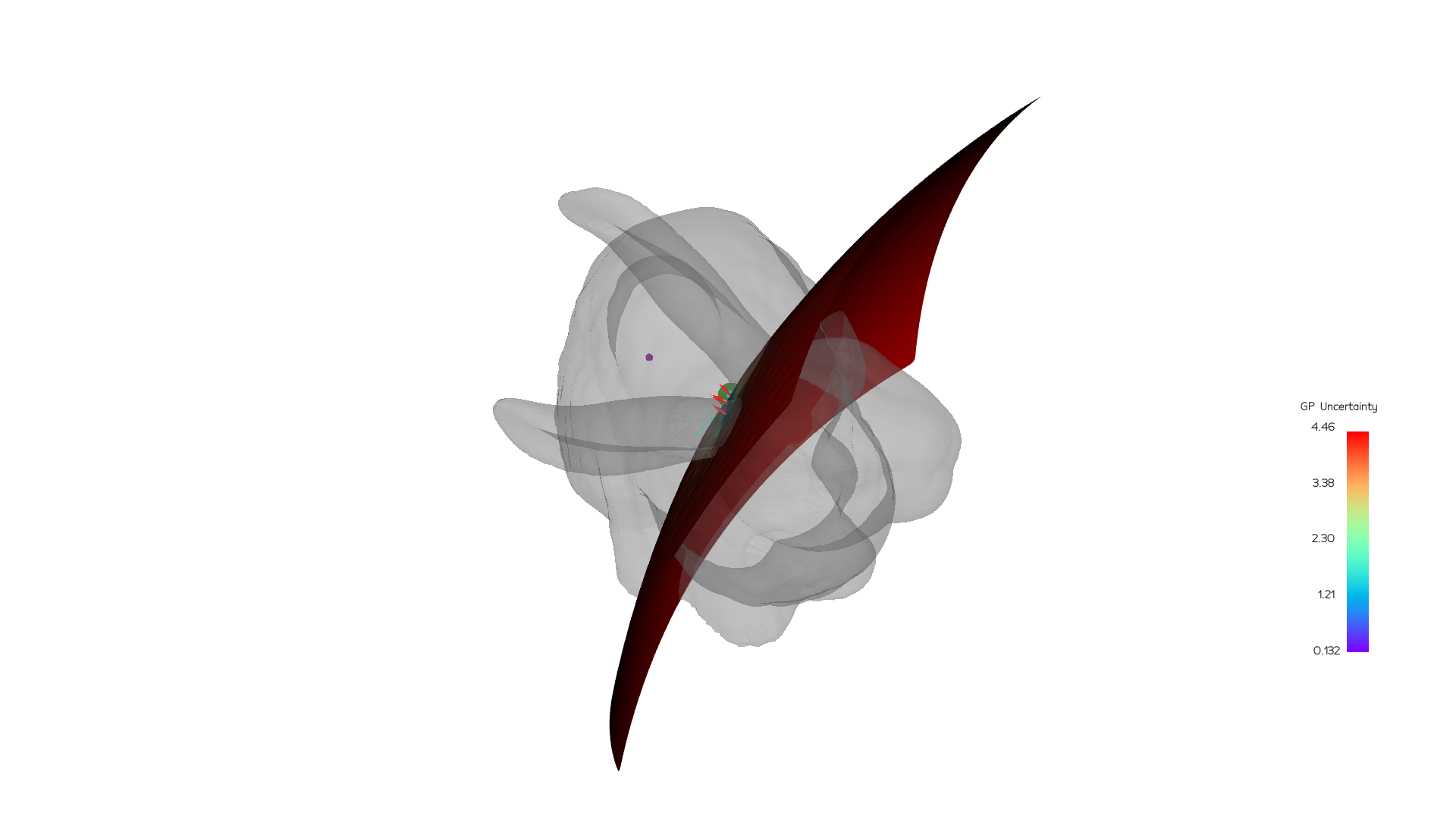} &
        \includegraphics[width=0.17\textwidth,clip,trim=420 0 420 0]{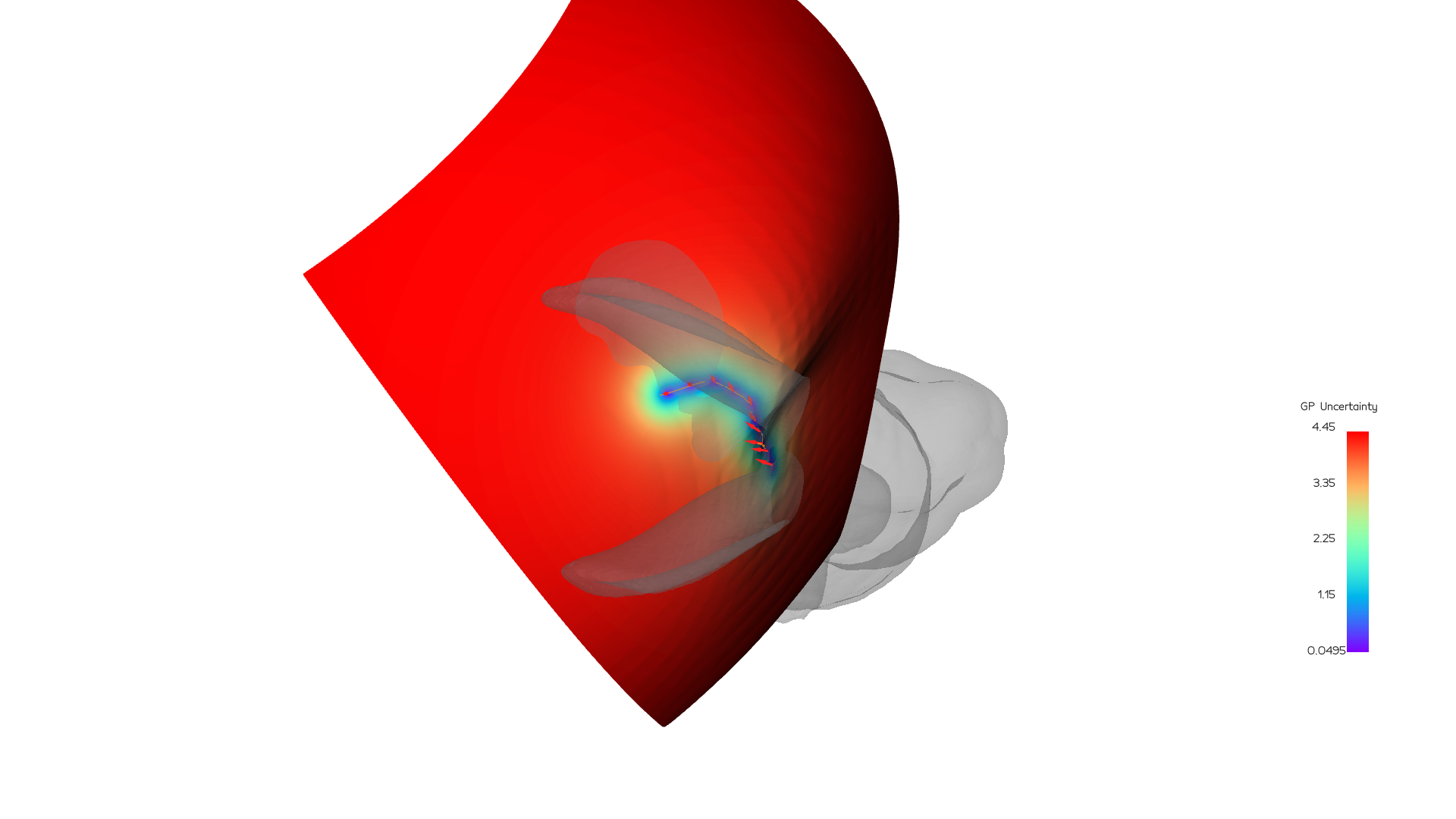} &
        \includegraphics[width=0.17\textwidth,clip,trim=420 0 420 0]{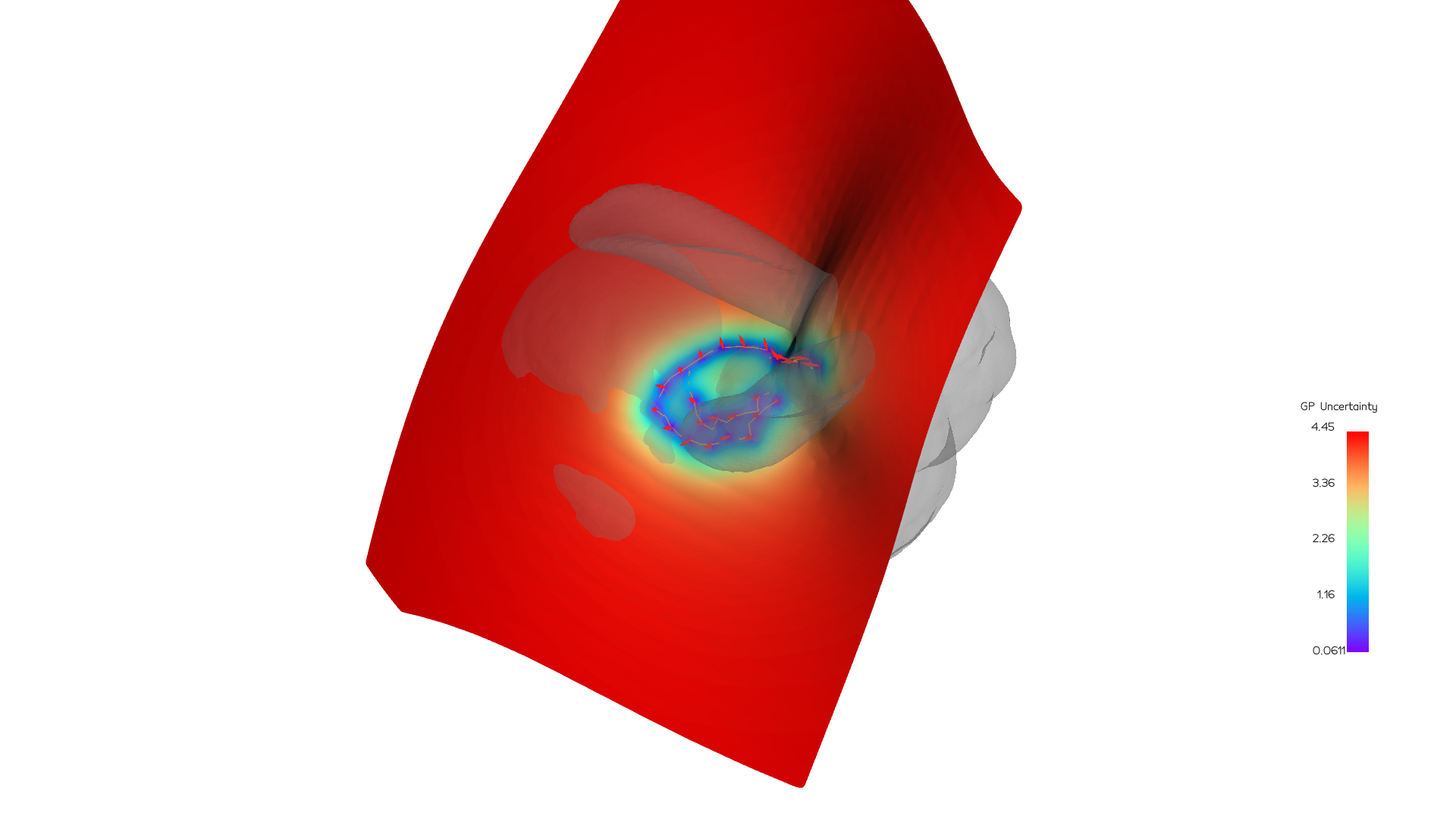} \\
        &
        \includegraphics[width=0.17\textwidth,clip,trim=420 0 420 0]{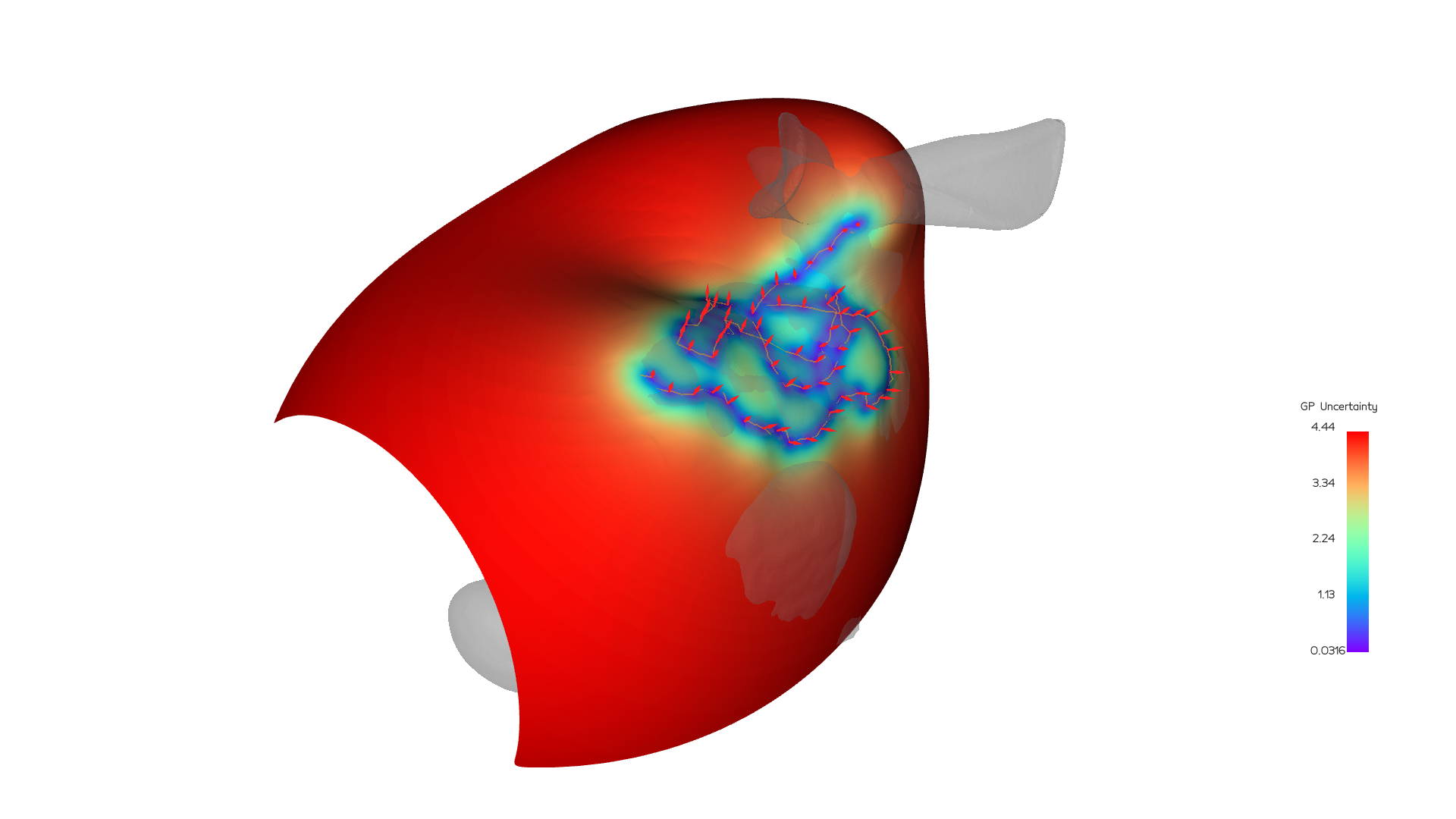} &
        \includegraphics[width=0.17\textwidth,clip,trim=420 0 420 0]{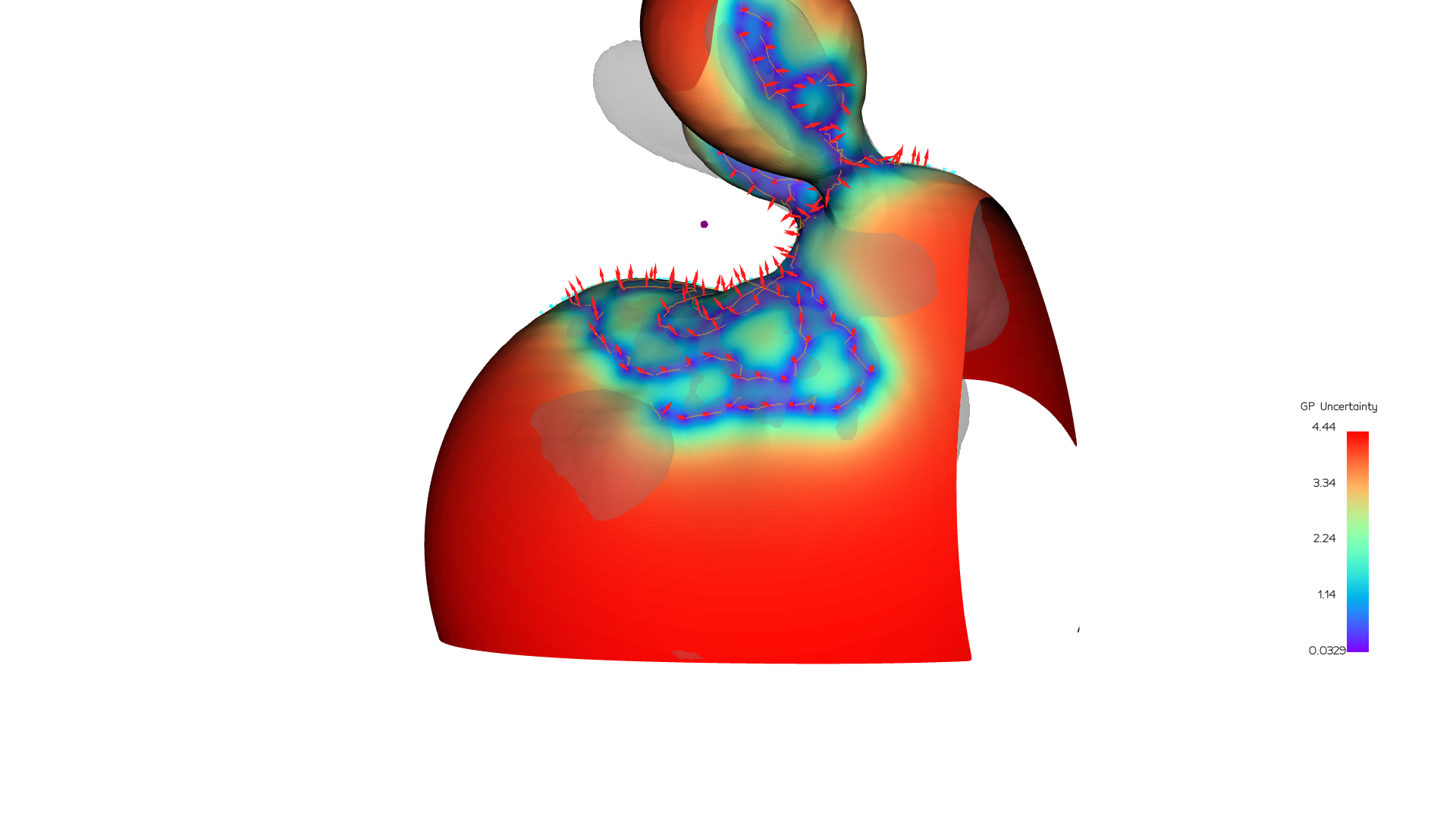} &
        \includegraphics[width=0.17\textwidth,clip,trim=420 0 420 0]{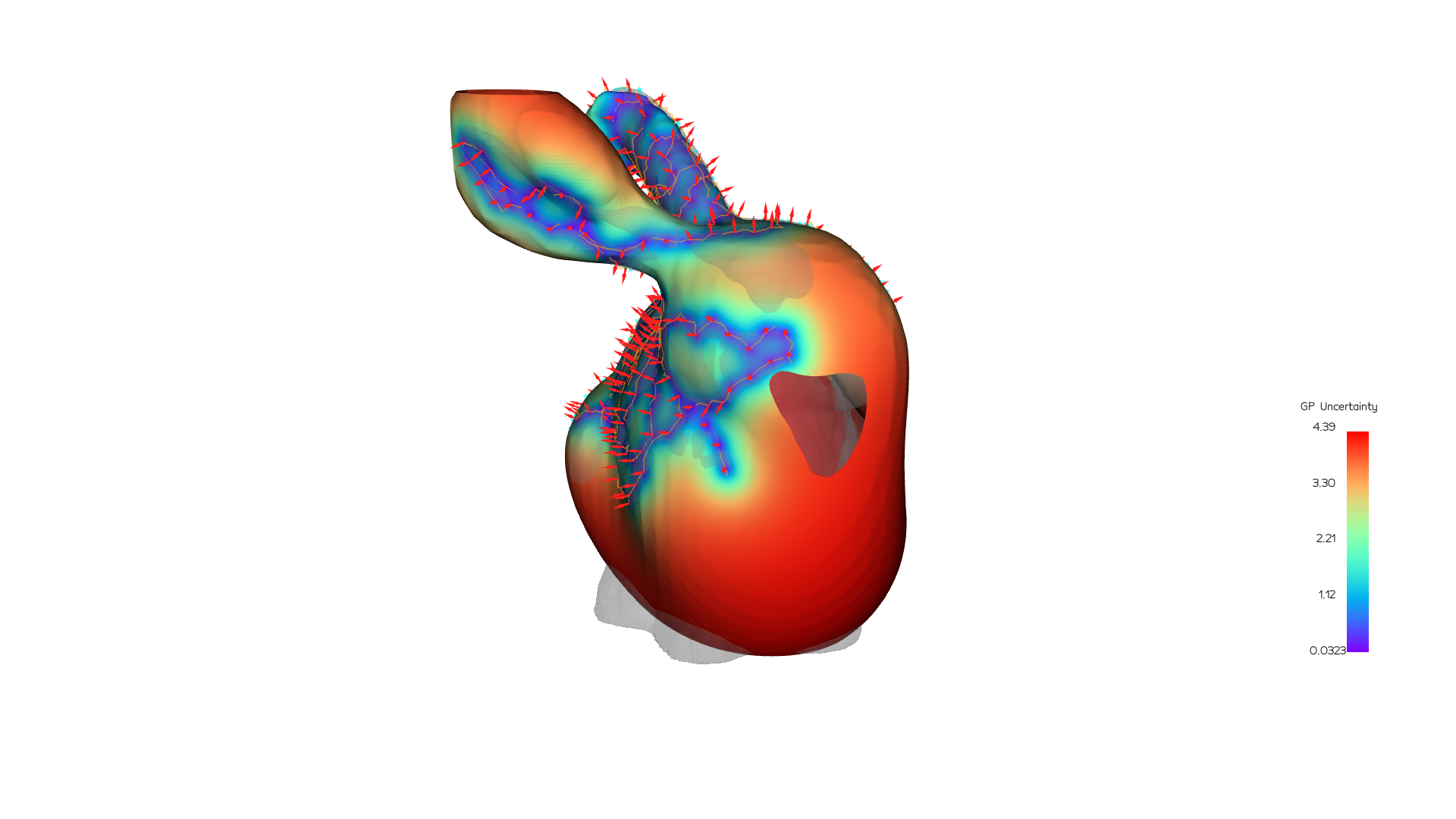} &
        \includegraphics[width=0.17\textwidth,clip,trim=420 0 420 0]{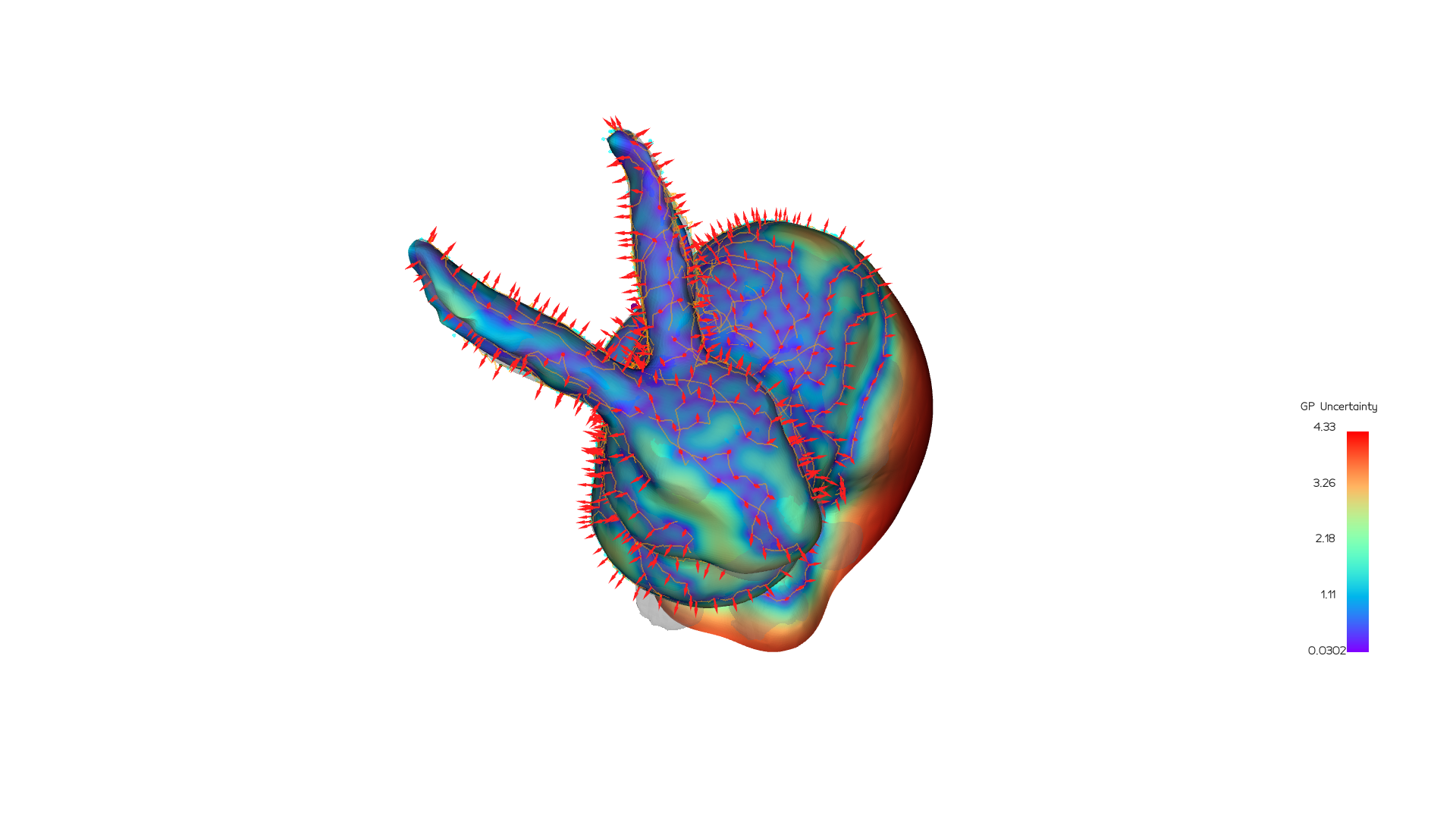} &
        \includegraphics[width=0.17\textwidth,clip,trim=420 0 420 0]{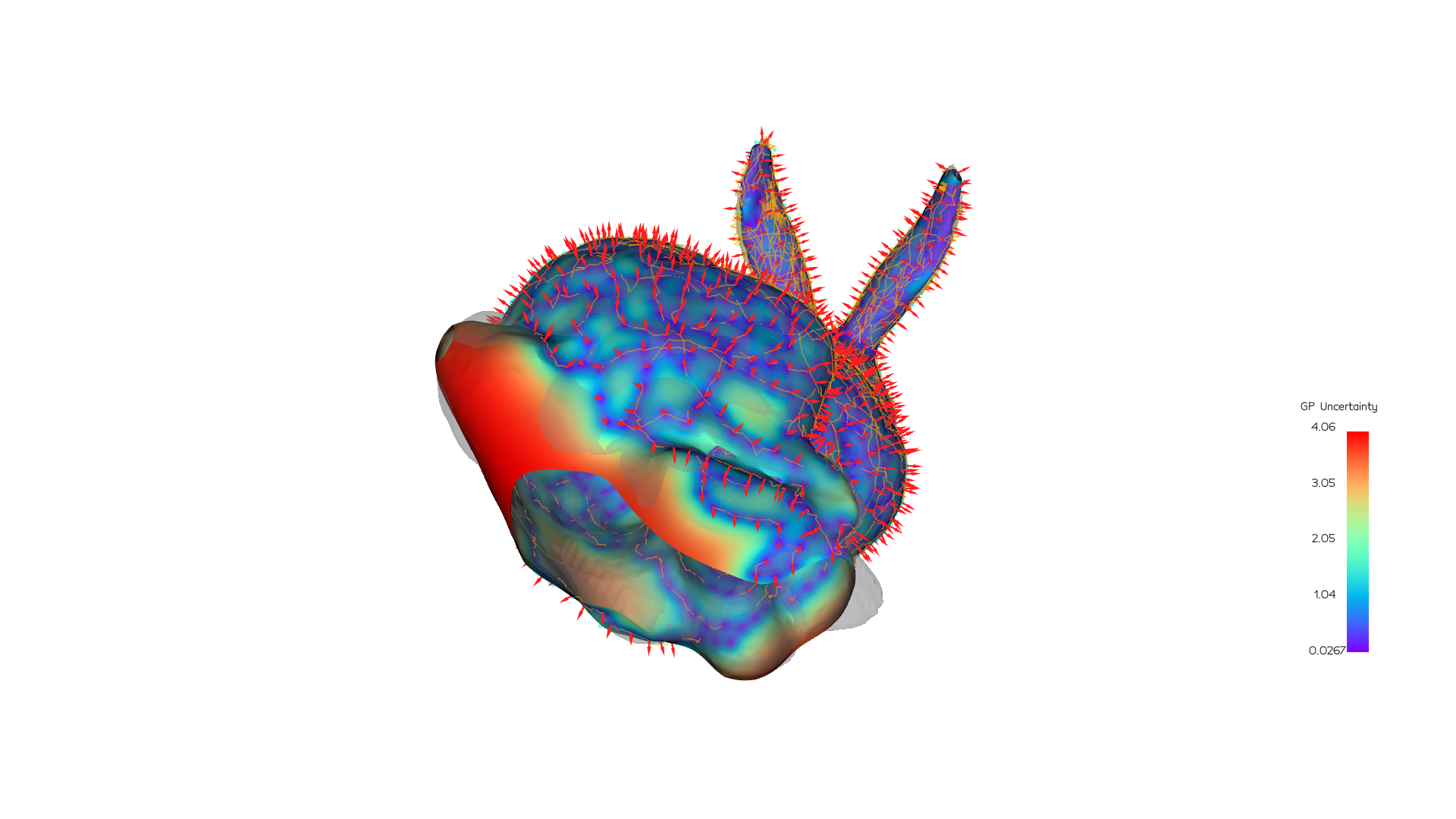} \\
        & \multicolumn{5}{c}{\scriptsize Time steps: 1, 3, 7, 19, 50 (top rows) | 133, 353, 940, 2499, 6644 (bottom rows)} \\
    \end{tabular}
    \caption{Bunny Experiment Progression over 10000 time steps (logarithmic selection). \textbf{Coverage:} Achieved exploration pattern showing systematic coverage of the Stanford bunny surface. \textbf{Target:} Task-specific target distribution from a single heat source between the ears and back. \textbf{Uncertainty:} GPIS prediction uncertainty, with warmer colors indicating higher uncertainty that progressively reduces in explored areas.}
    \label{fig:bunny_coverage_progression}
    \label{fig:bunny_target_progression}
    \label{fig:bunny_uncertainty_progression}
\end{figure*}

\begin{figure*}[htbp]
    \centering
    \begin{tabular}{@{}r@{\hspace{0.3em}}ccccc@{}}
        \multirow{2}{*}[3em]{\rotatebox{90}{\small Coverage}} &
        \includegraphics[width=0.17\textwidth,clip,trim=600 200 600 230]{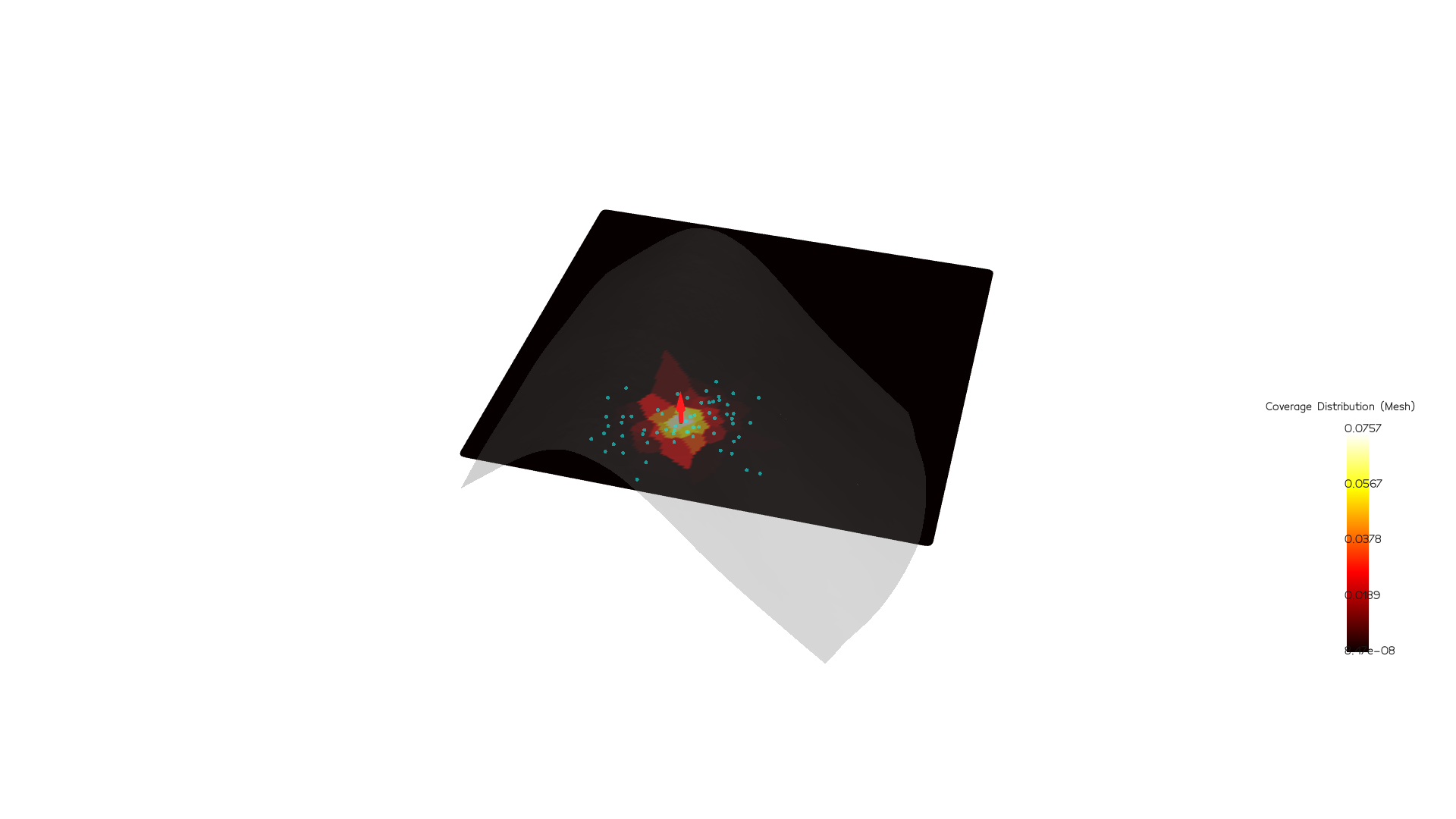} &
        \includegraphics[width=0.17\textwidth,clip,trim=600 200 600 230]{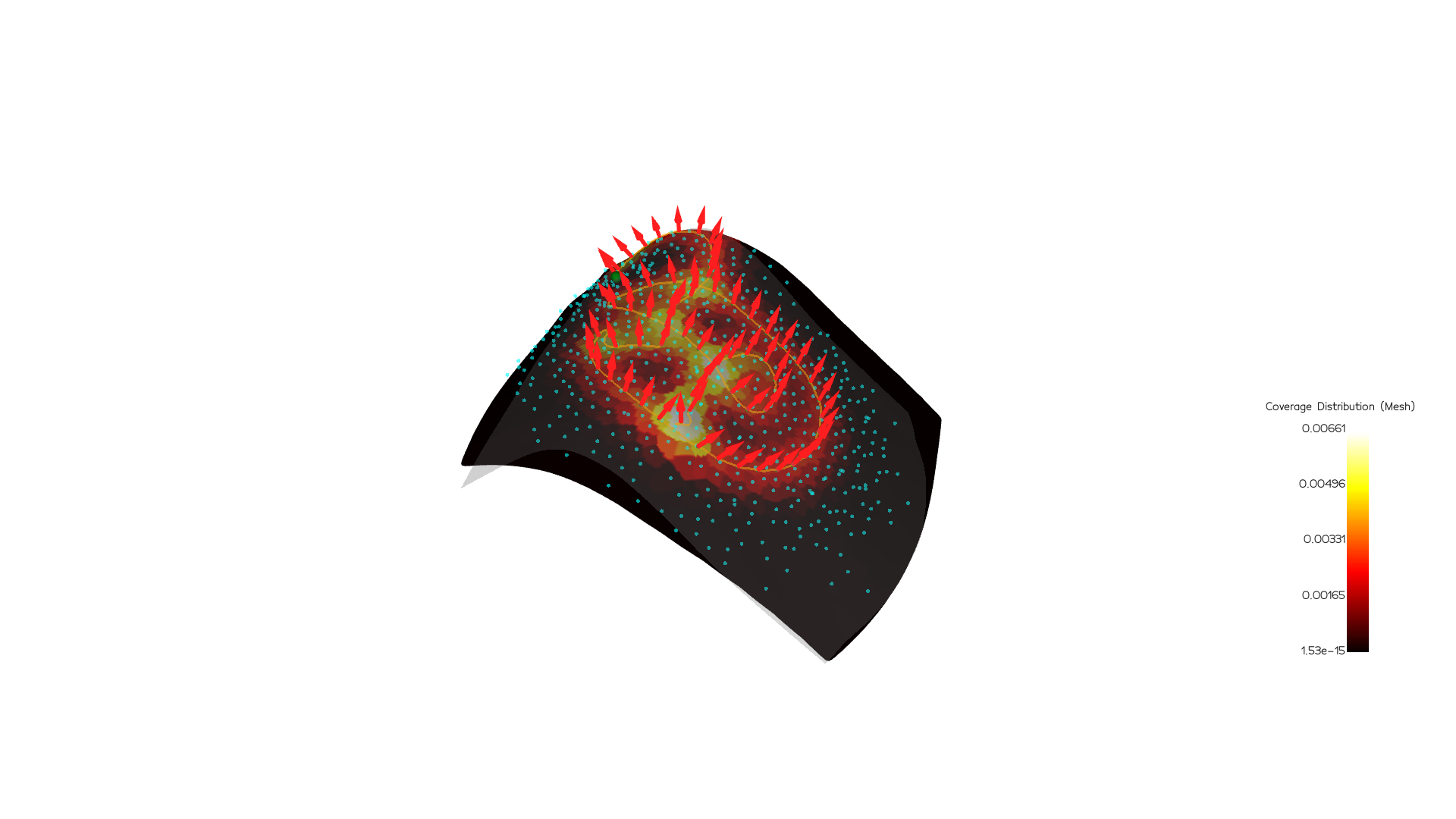} &
        \includegraphics[width=0.17\textwidth,clip,trim=600 200 600 230]{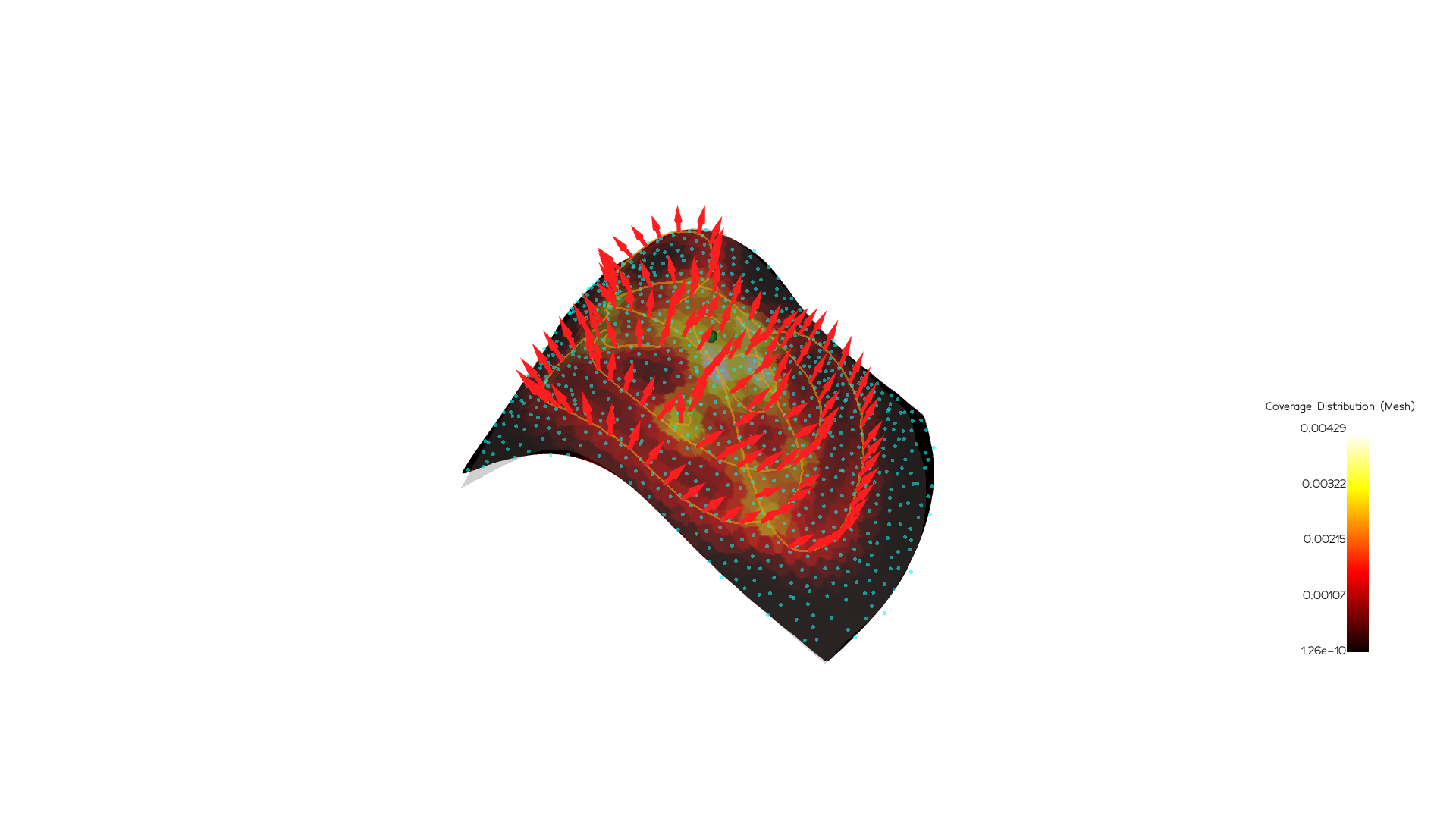} &
        \includegraphics[width=0.17\textwidth,clip,trim=600 200 600 230]{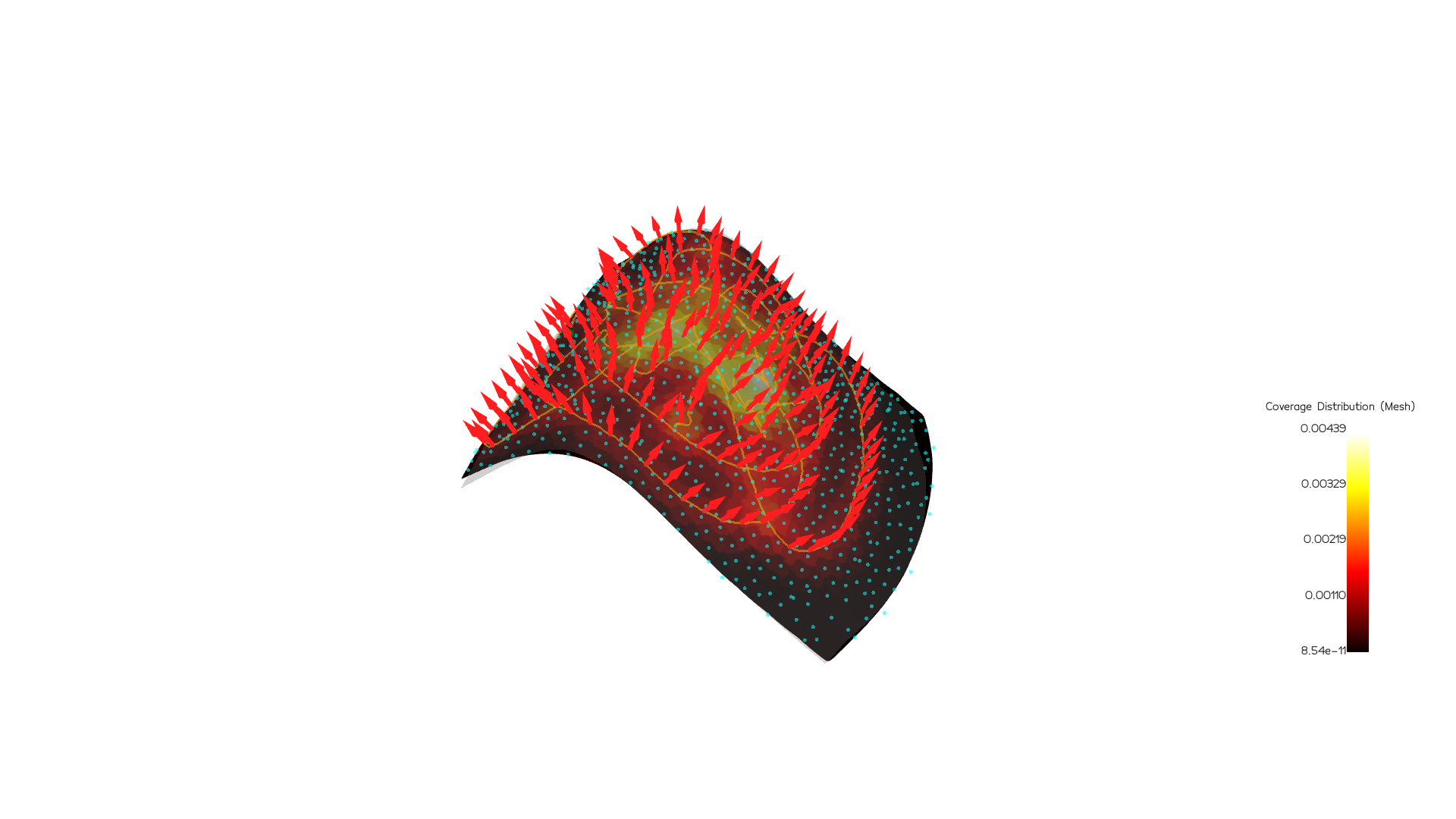} &
        \includegraphics[width=0.17\textwidth,clip,trim=600 200 600 230]{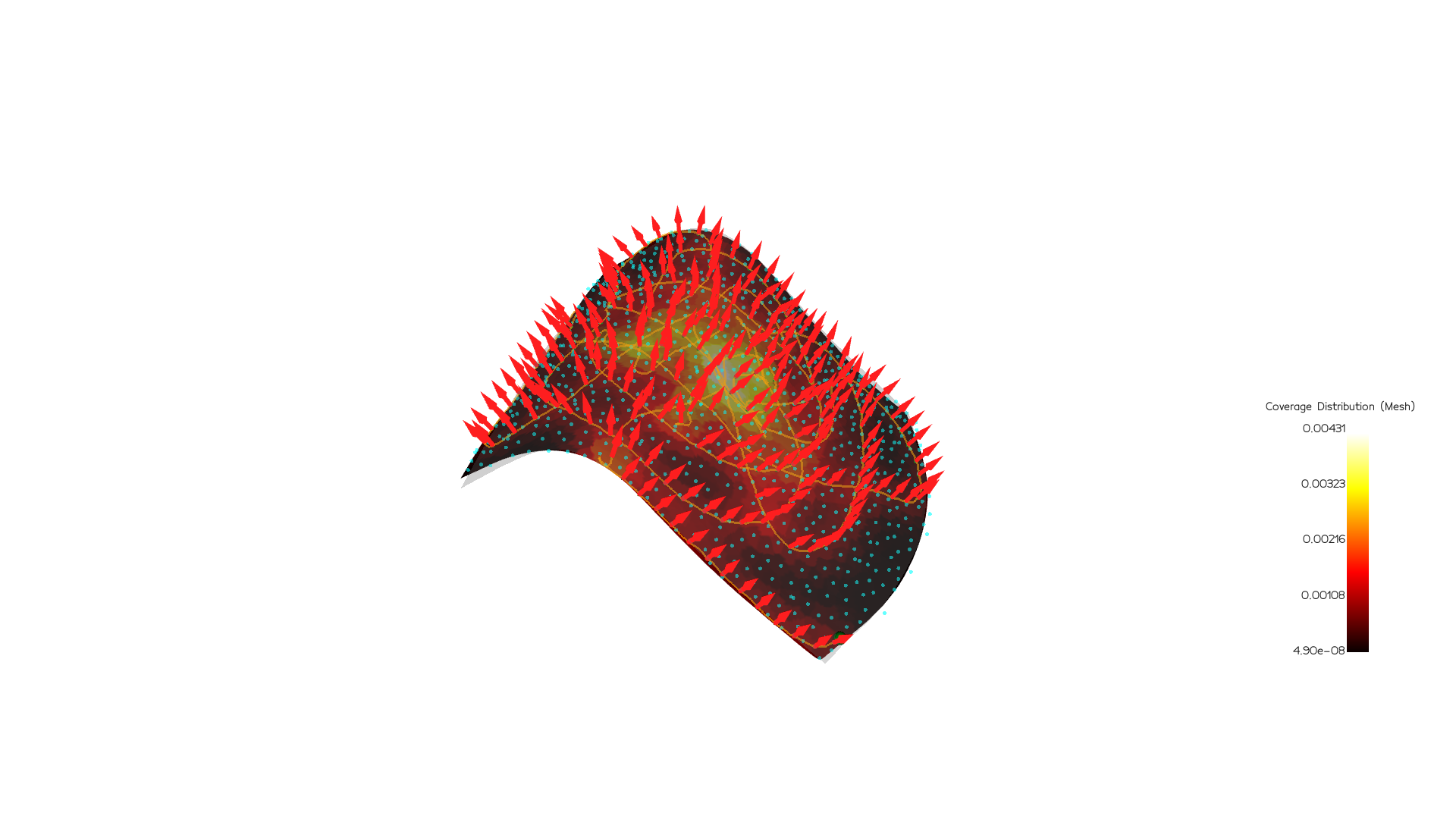} \\
        &
        \includegraphics[width=0.17\textwidth,clip,trim=600 200 600 230]{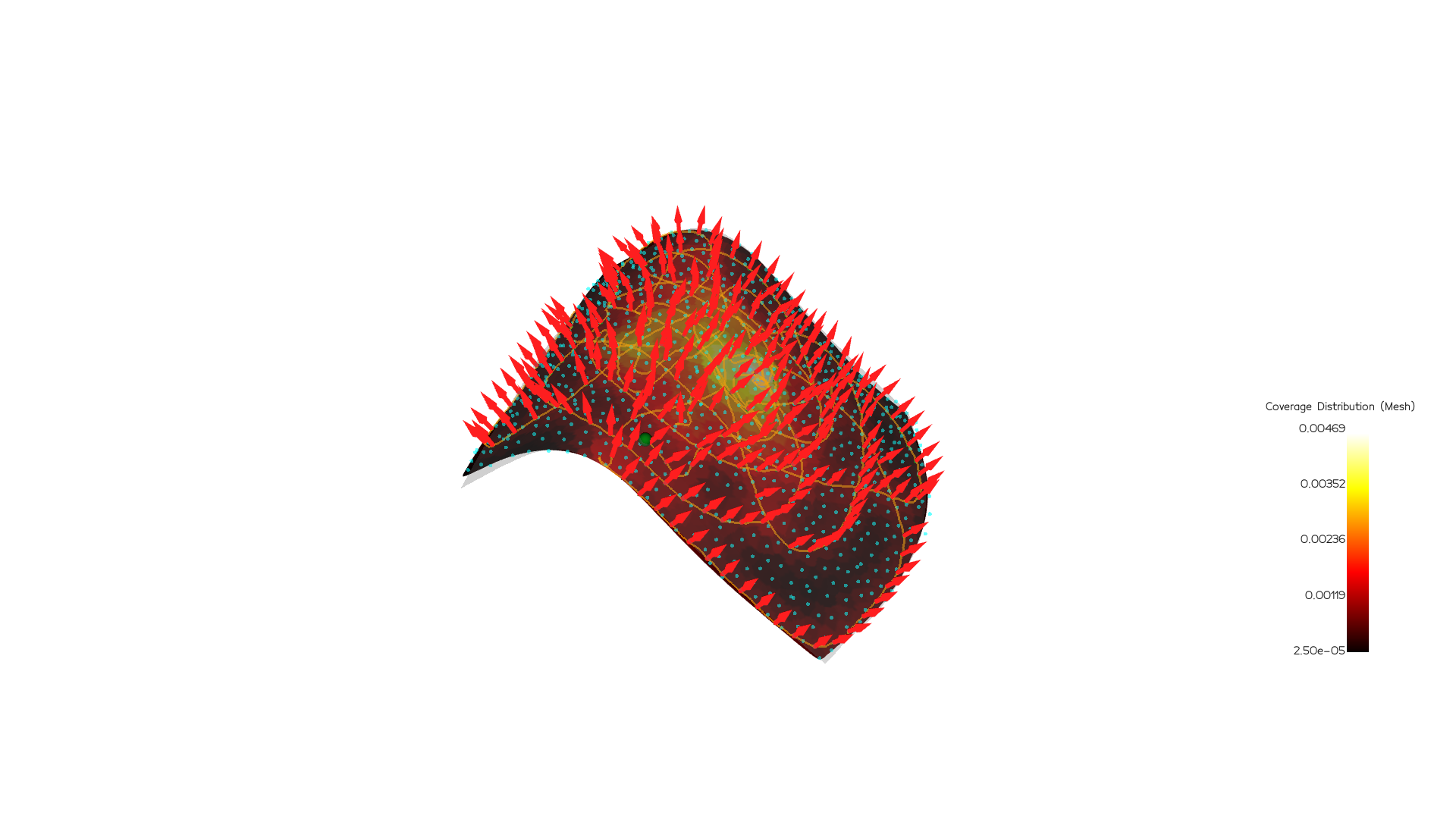} &
        \includegraphics[width=0.17\textwidth,clip,trim=600 200 600 230]{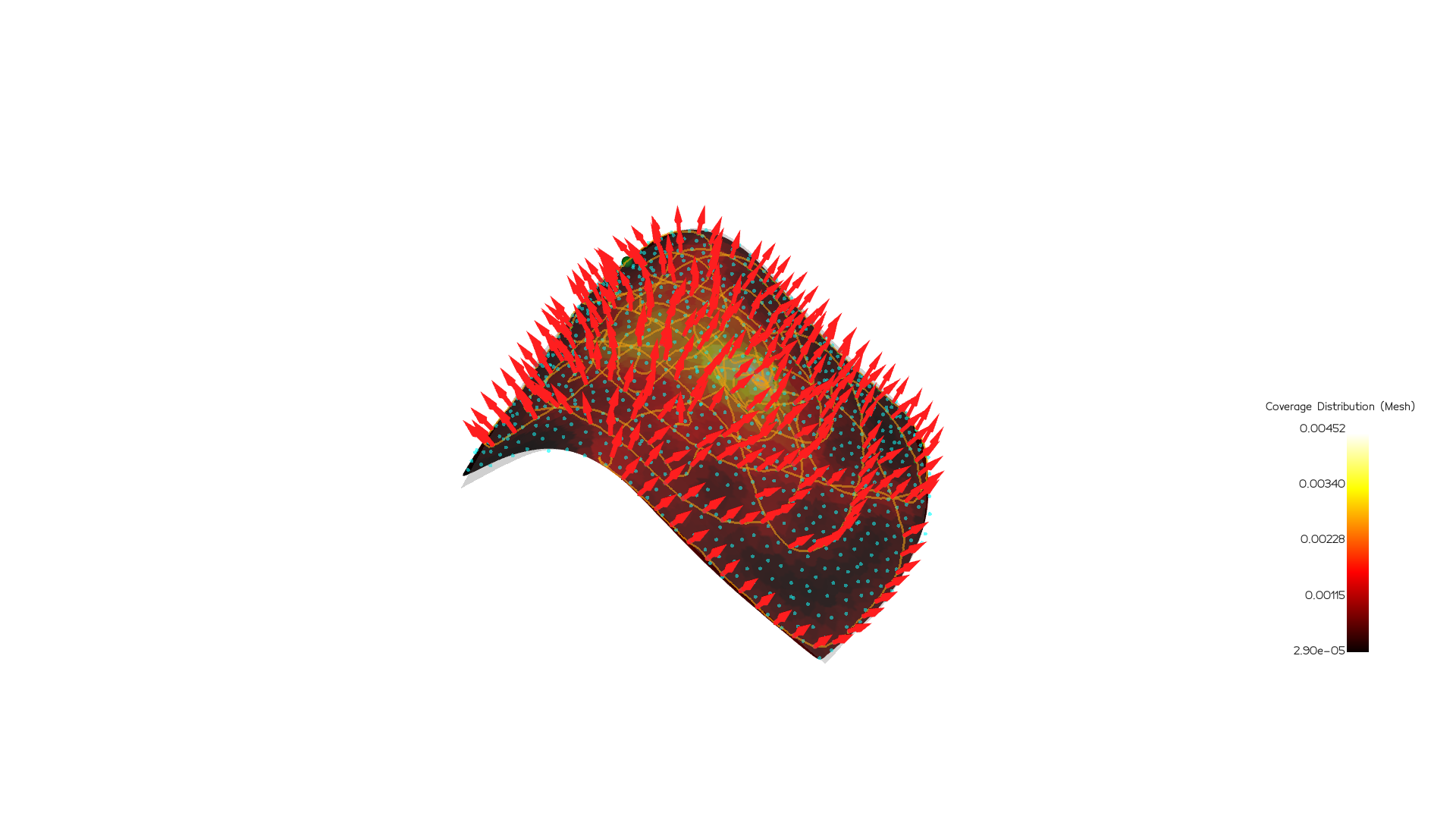} &
        \includegraphics[width=0.17\textwidth,clip,trim=600 200 600 230]{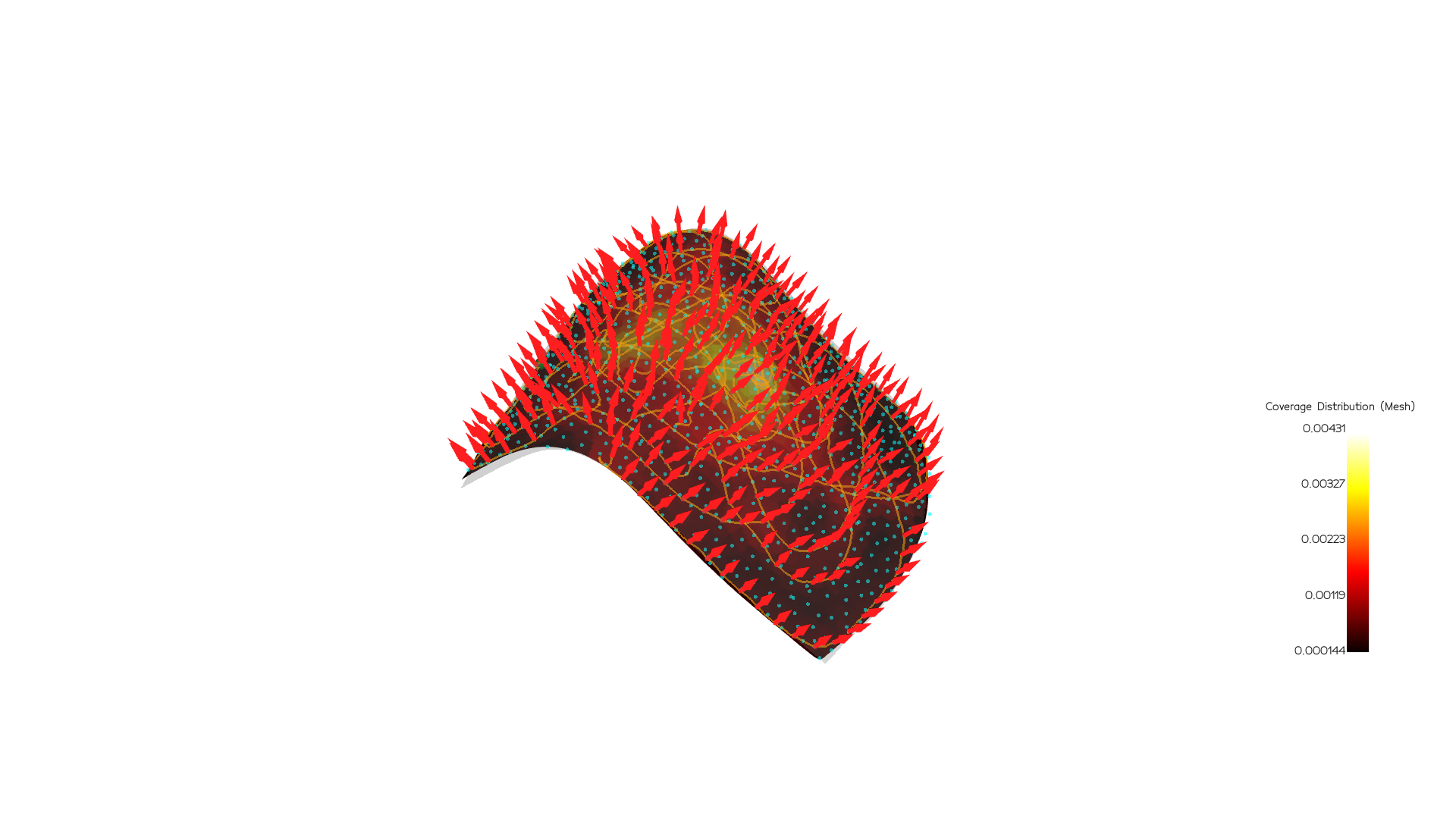} &
        \includegraphics[width=0.17\textwidth,clip,trim=600 200 600 230]{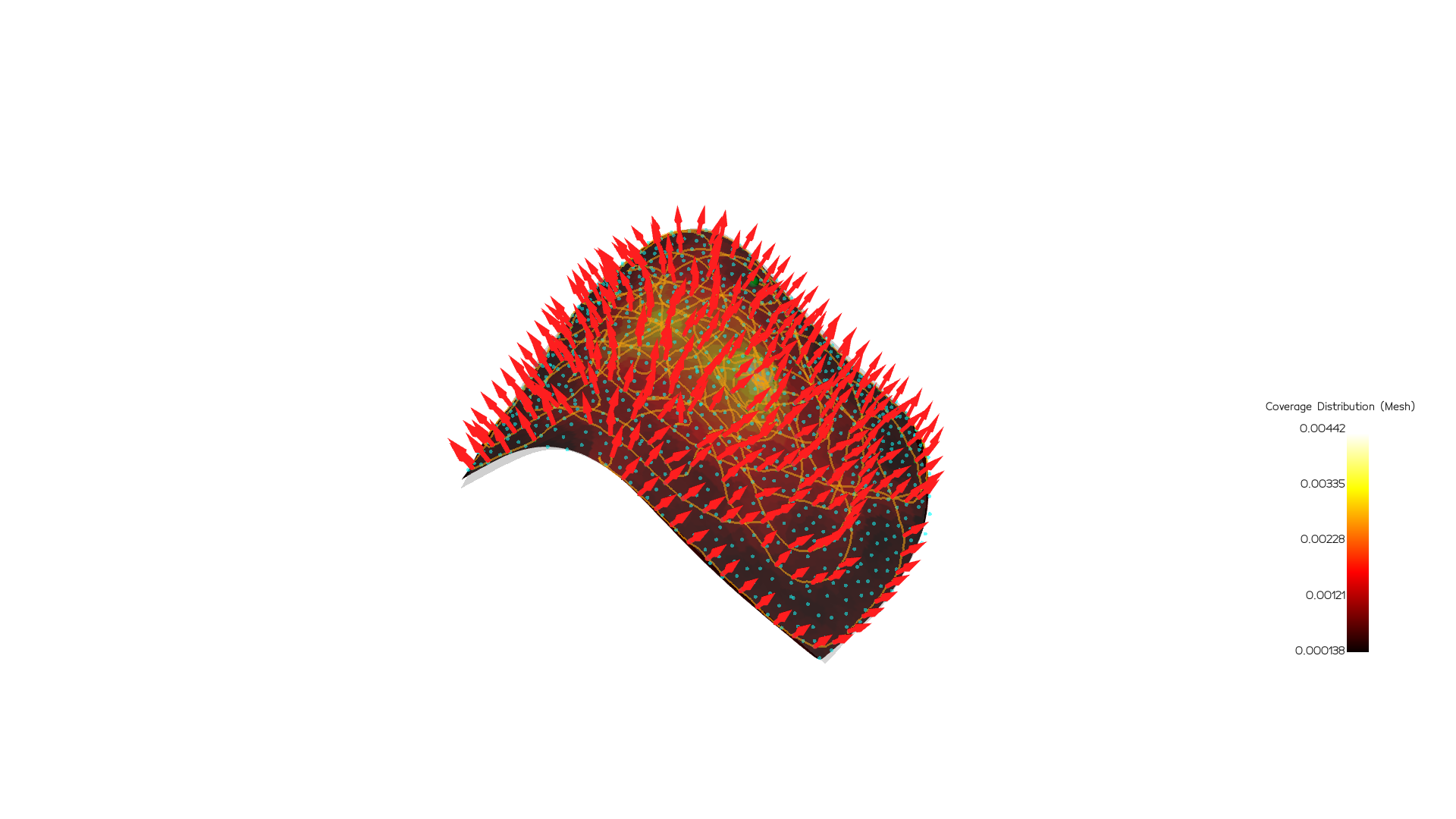} &
        \includegraphics[width=0.17\textwidth,clip,trim=600 200 600 230]{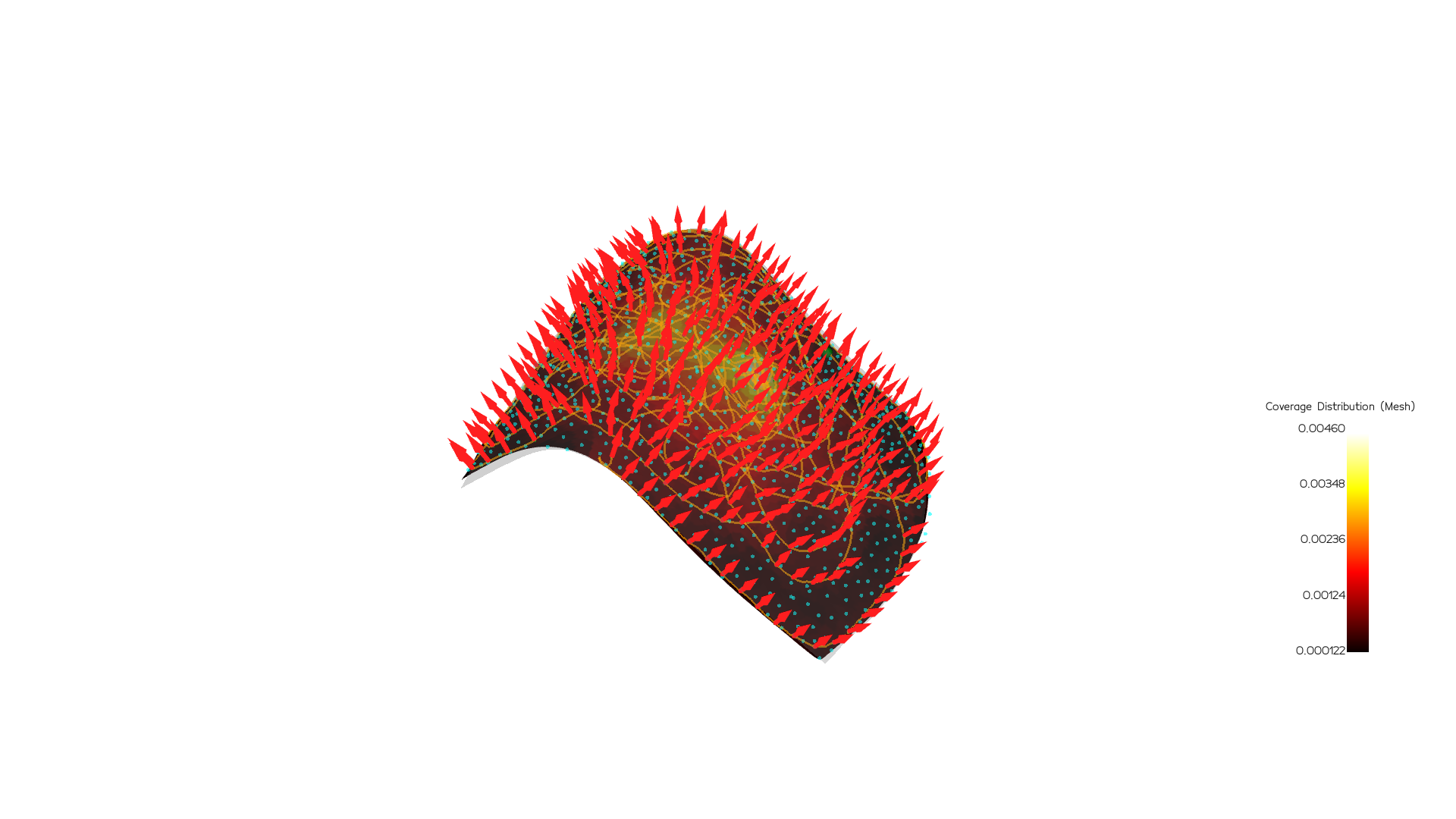} \\[0.1em]
        \cline{2-6}\\[-0.7em]
        \multirow{2}{*}[3em]{\rotatebox{90}{\small Target}} &
        \includegraphics[width=0.17\textwidth,clip,trim=600 200 600 230]{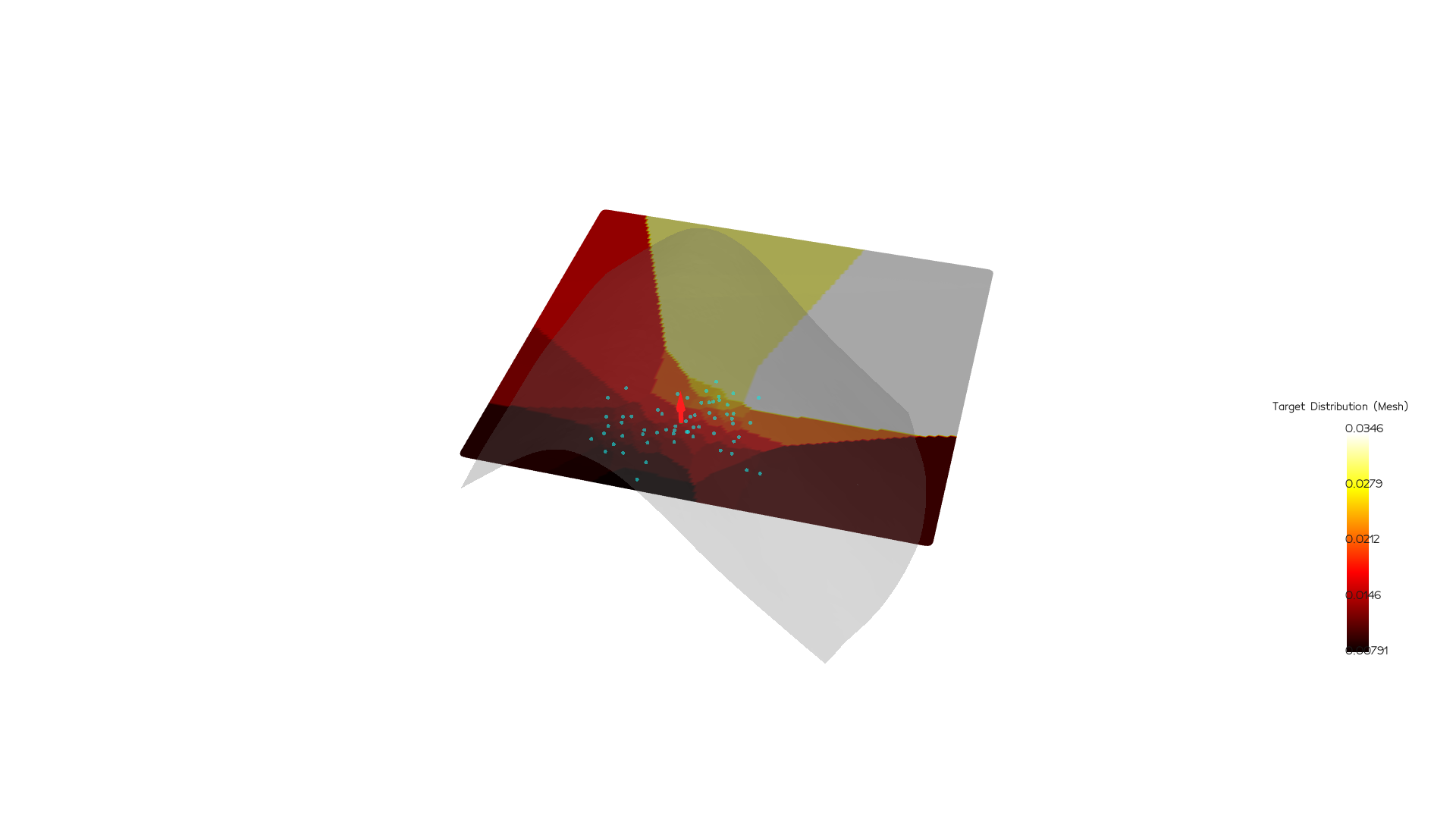} &
        \includegraphics[width=0.17\textwidth,clip,trim=600 200 600 230]{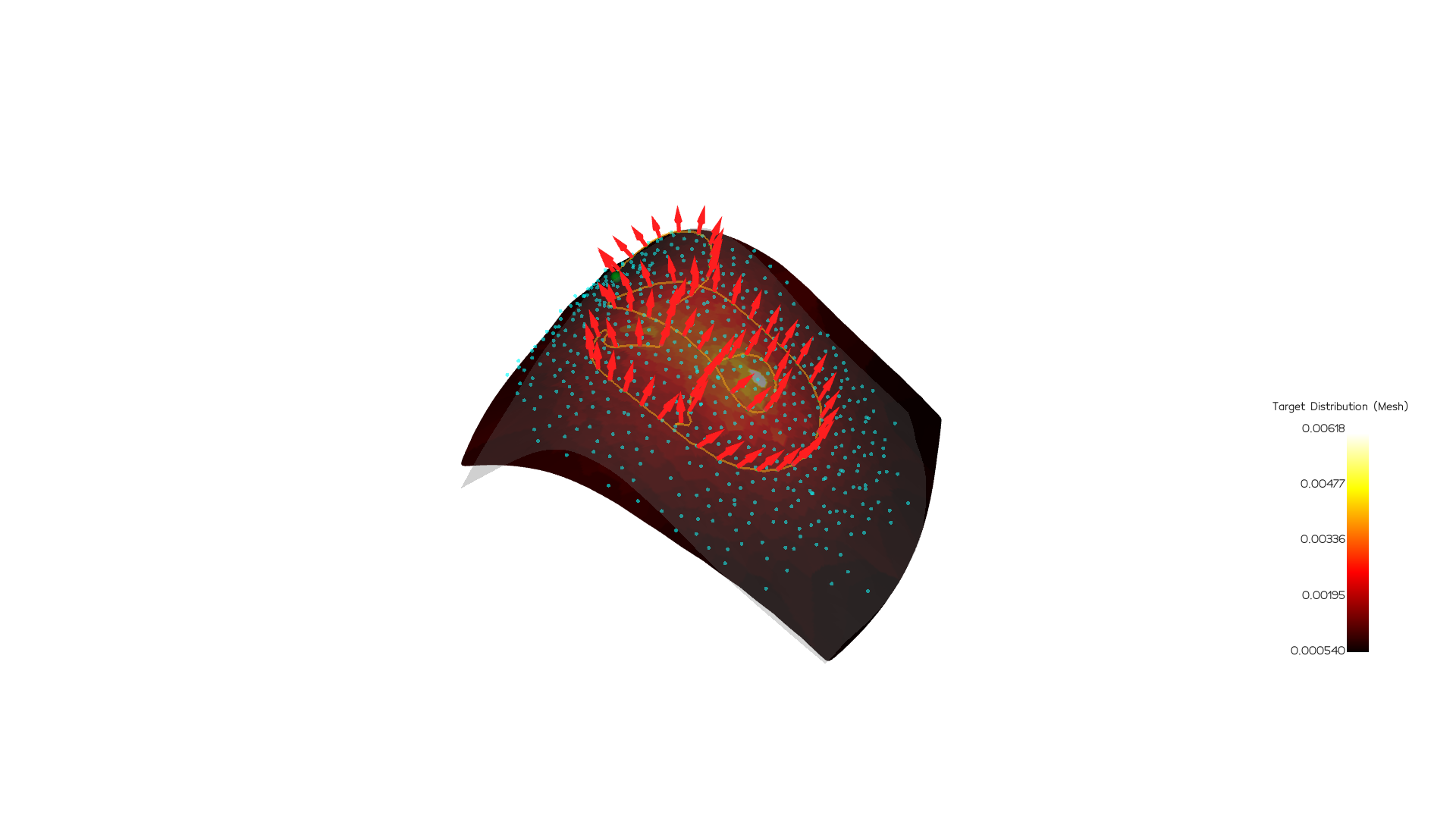} &
        \includegraphics[width=0.17\textwidth,clip,trim=600 200 600 230]{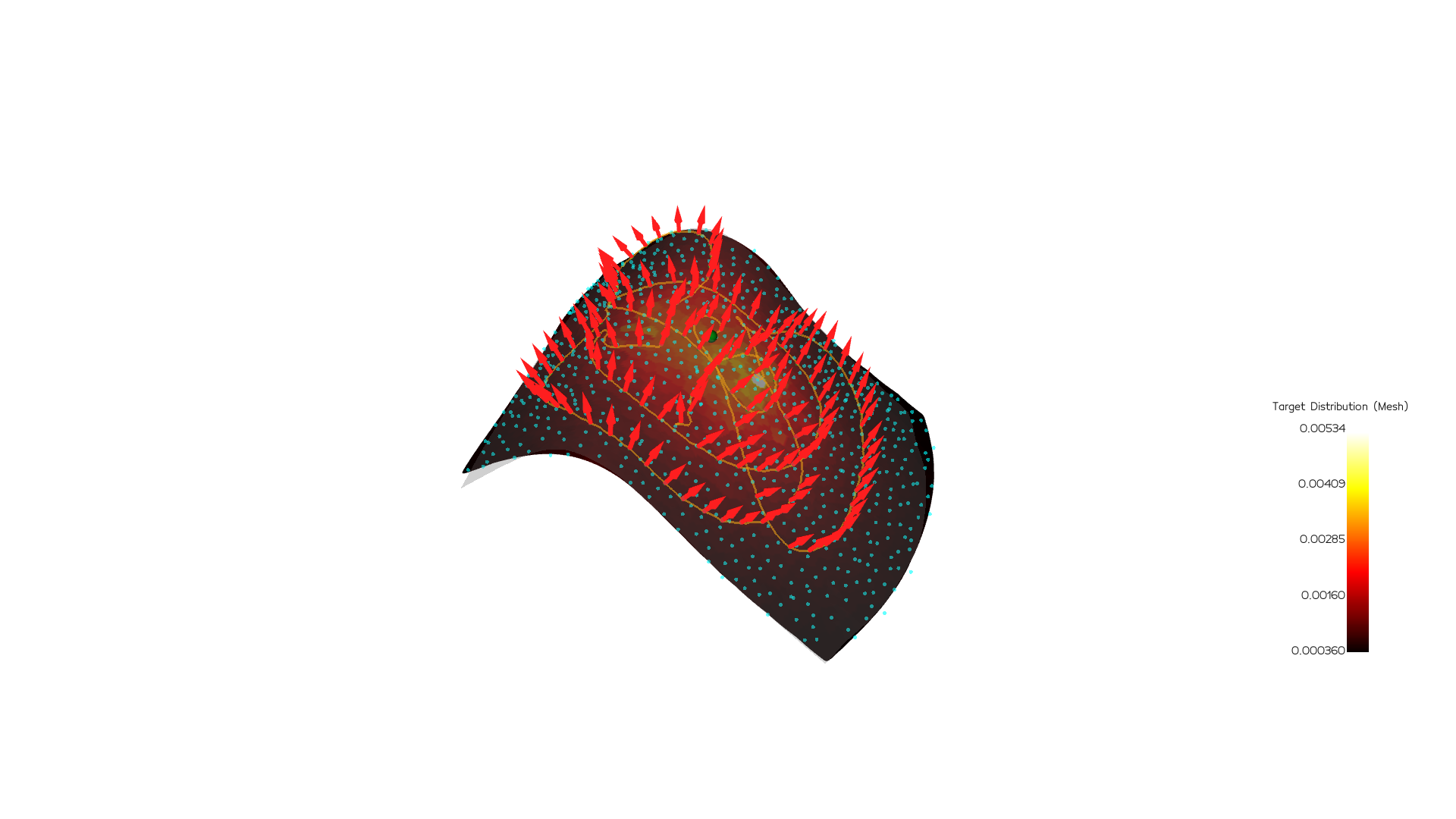} &
        \includegraphics[width=0.17\textwidth,clip,trim=600 200 600 230]{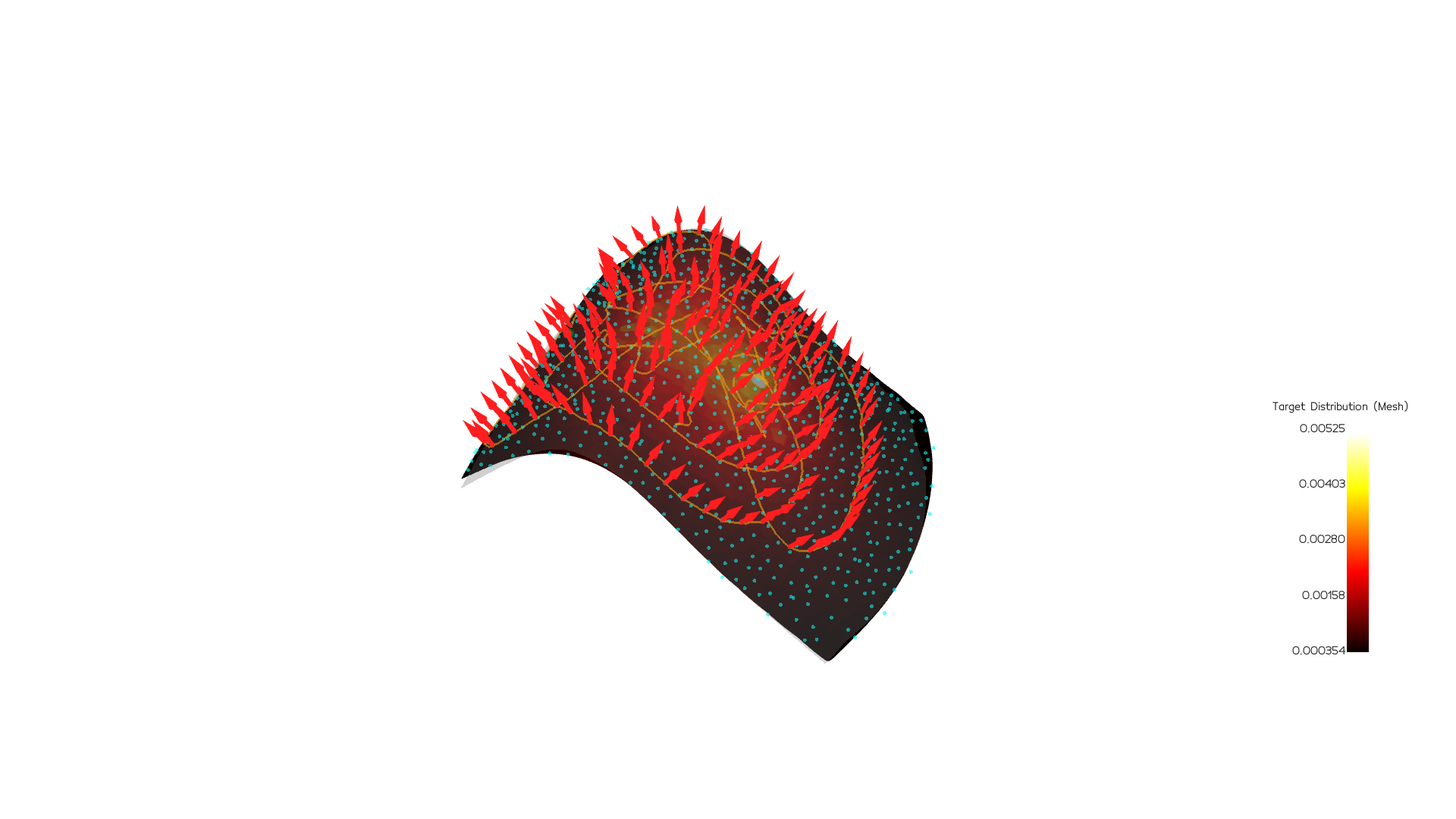} &
        \includegraphics[width=0.17\textwidth,clip,trim=600 200 600 230]{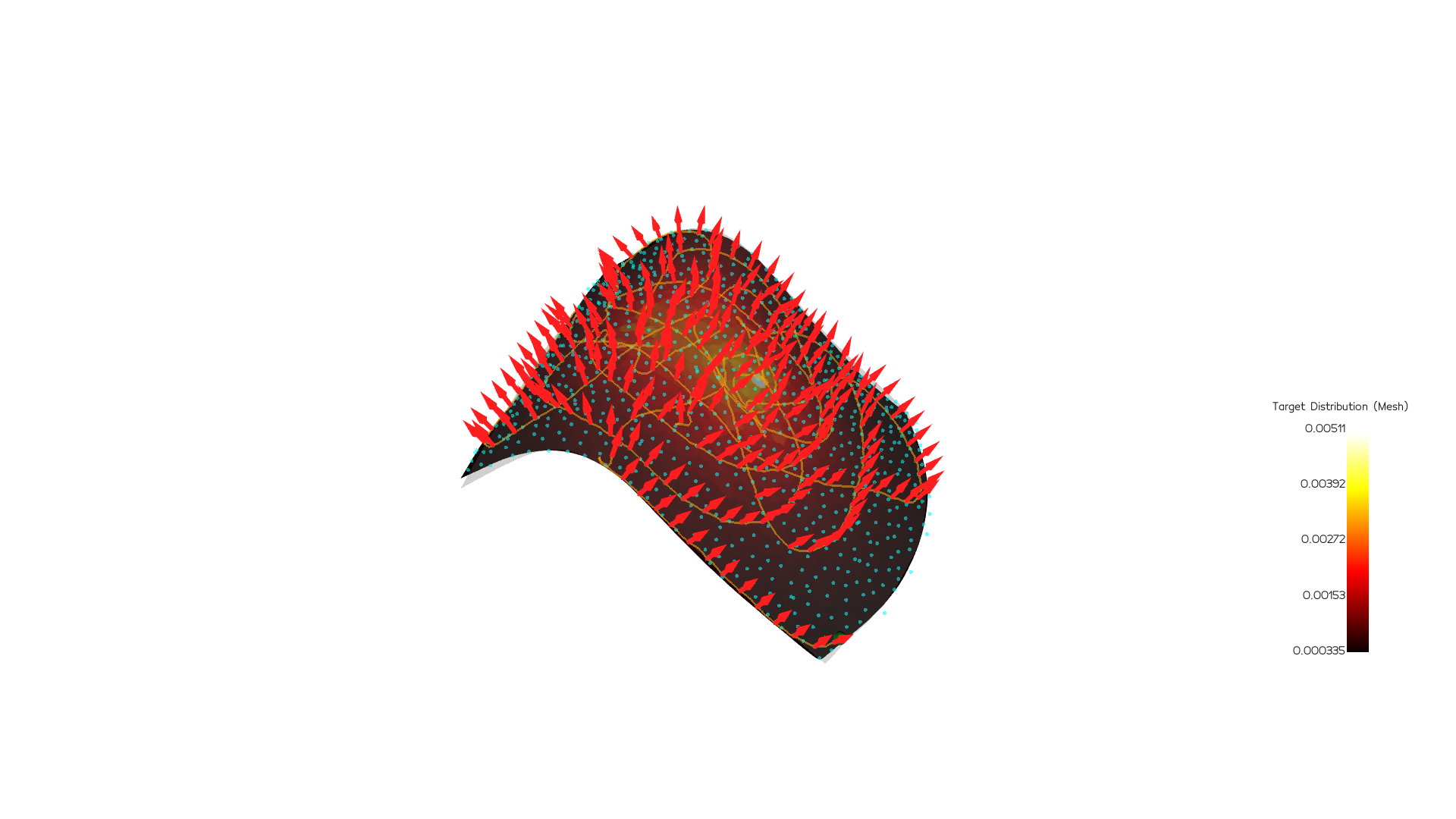} \\
        &
        \includegraphics[width=0.17\textwidth,clip,trim=600 200 600 230]{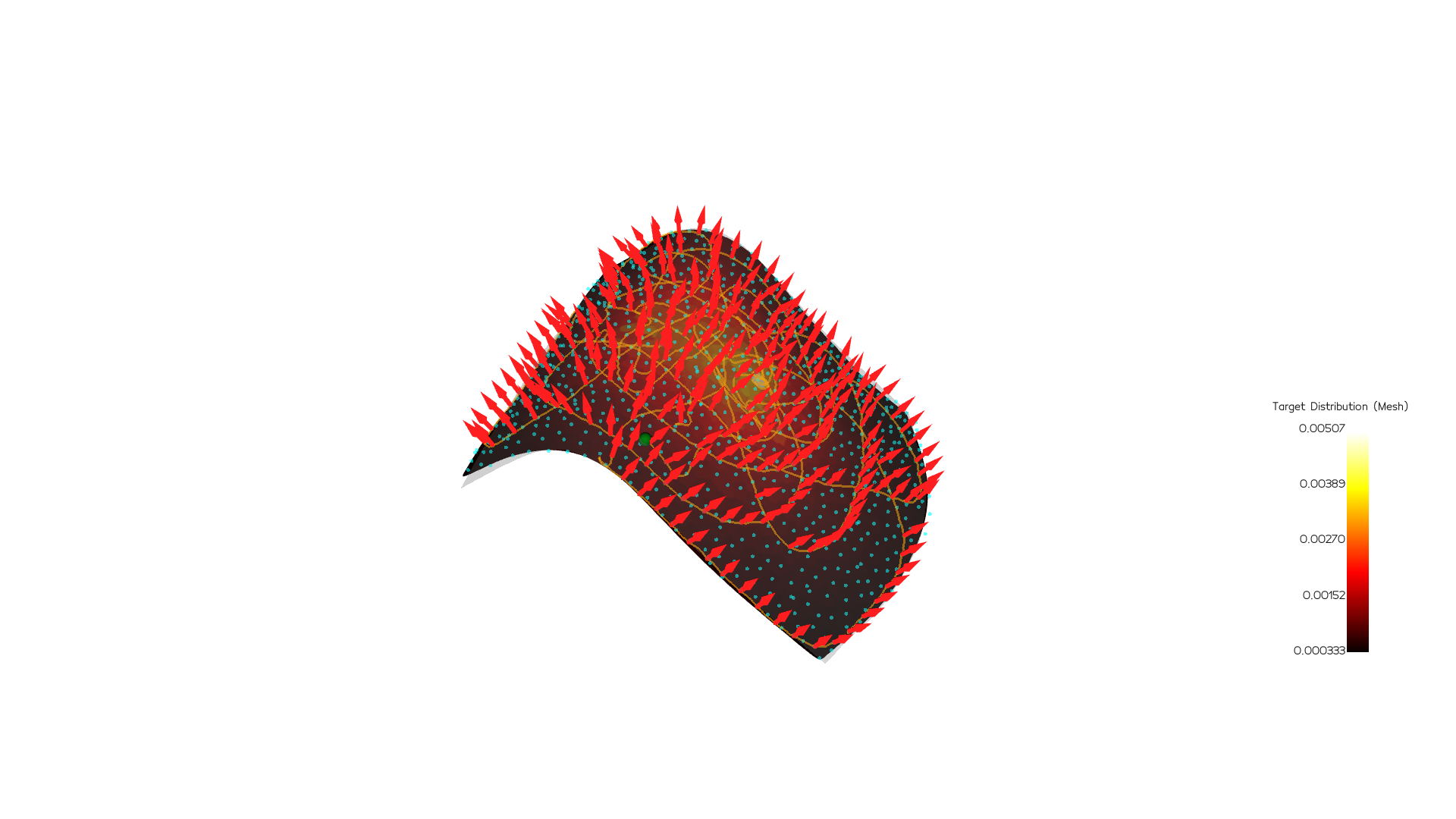} &
        \includegraphics[width=0.17\textwidth,clip,trim=600 200 600 230]{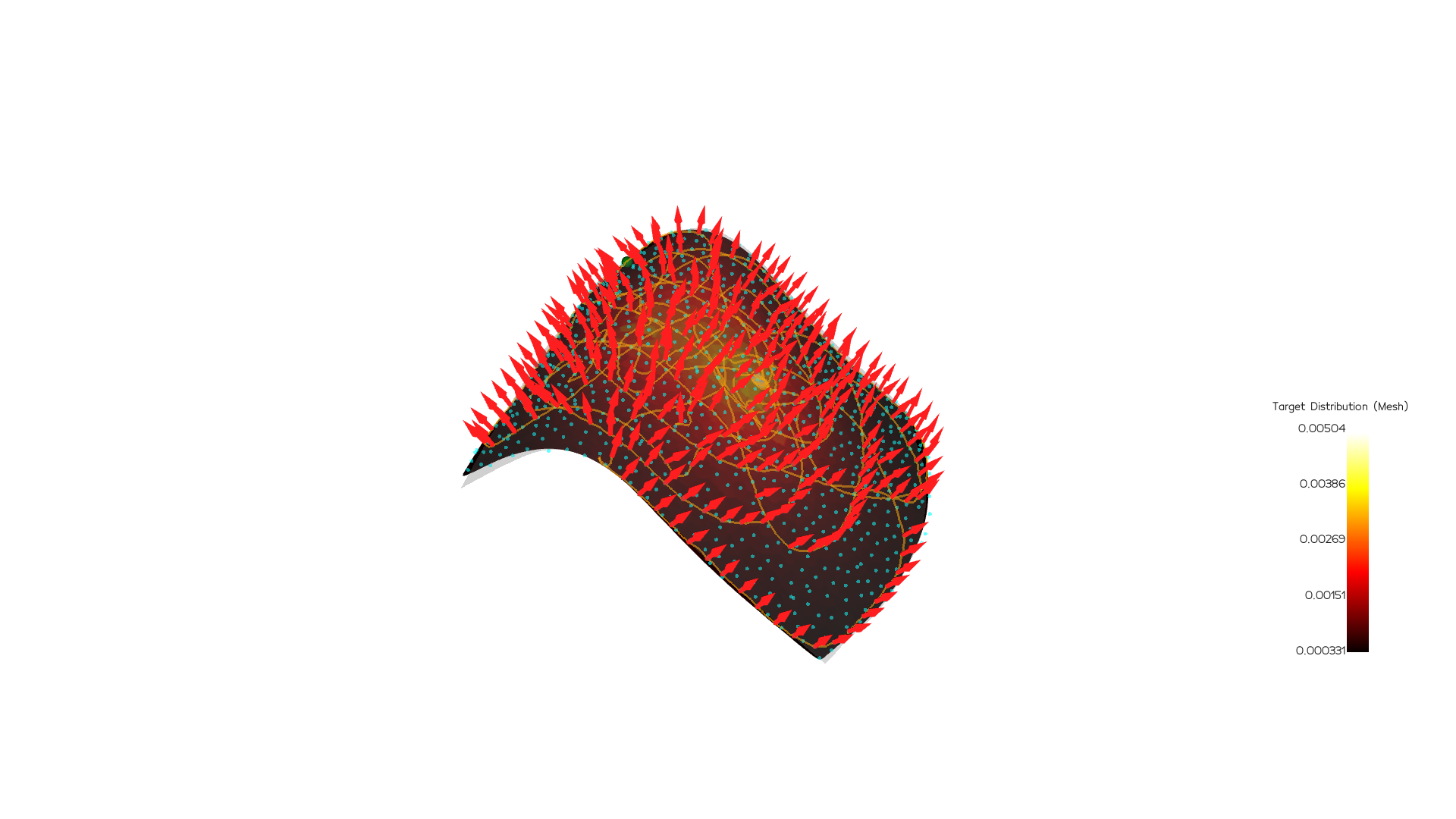} &
        \includegraphics[width=0.17\textwidth,clip,trim=600 200 600 230]{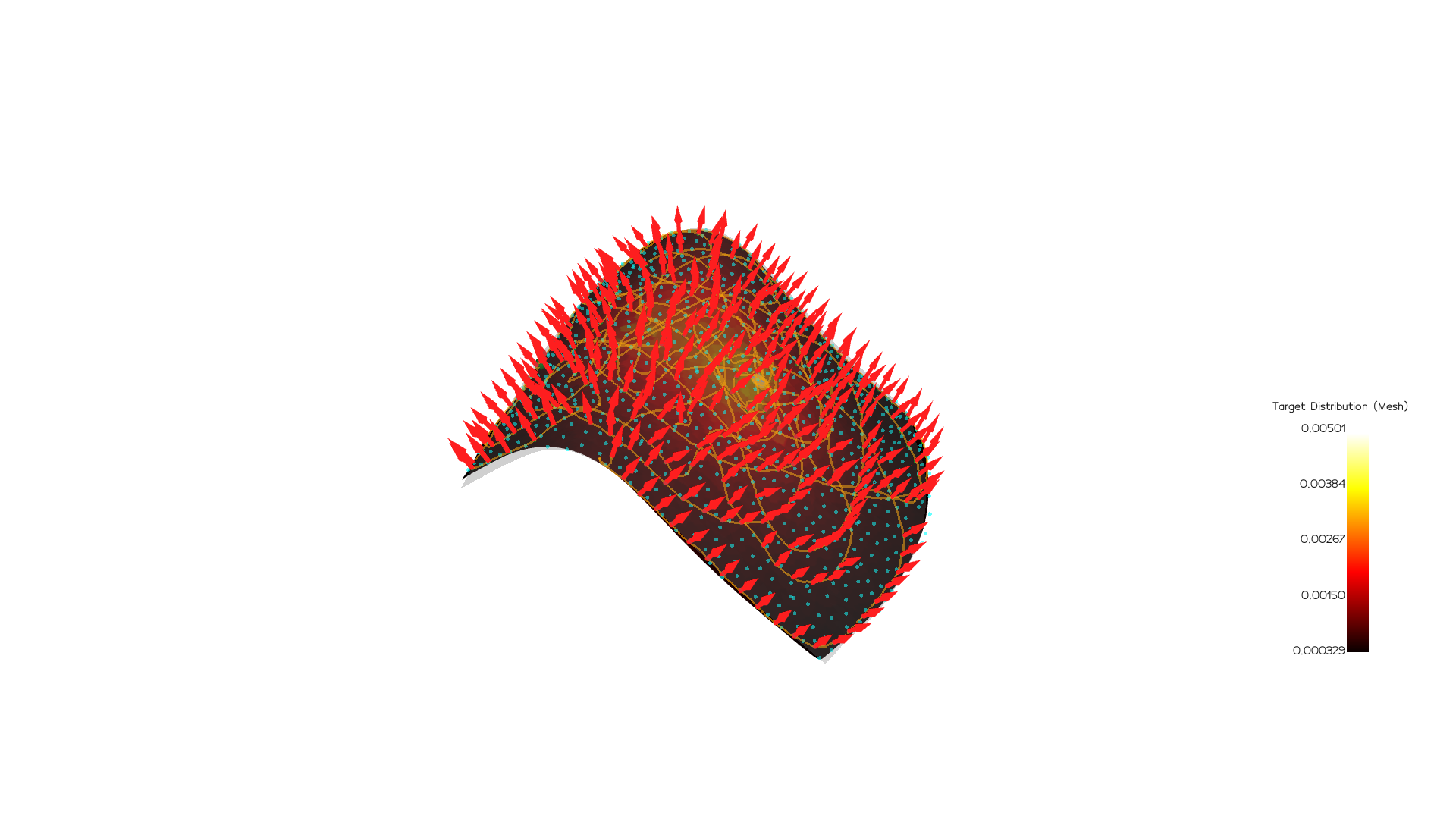} &
        \includegraphics[width=0.17\textwidth,clip,trim=600 200 600 230]{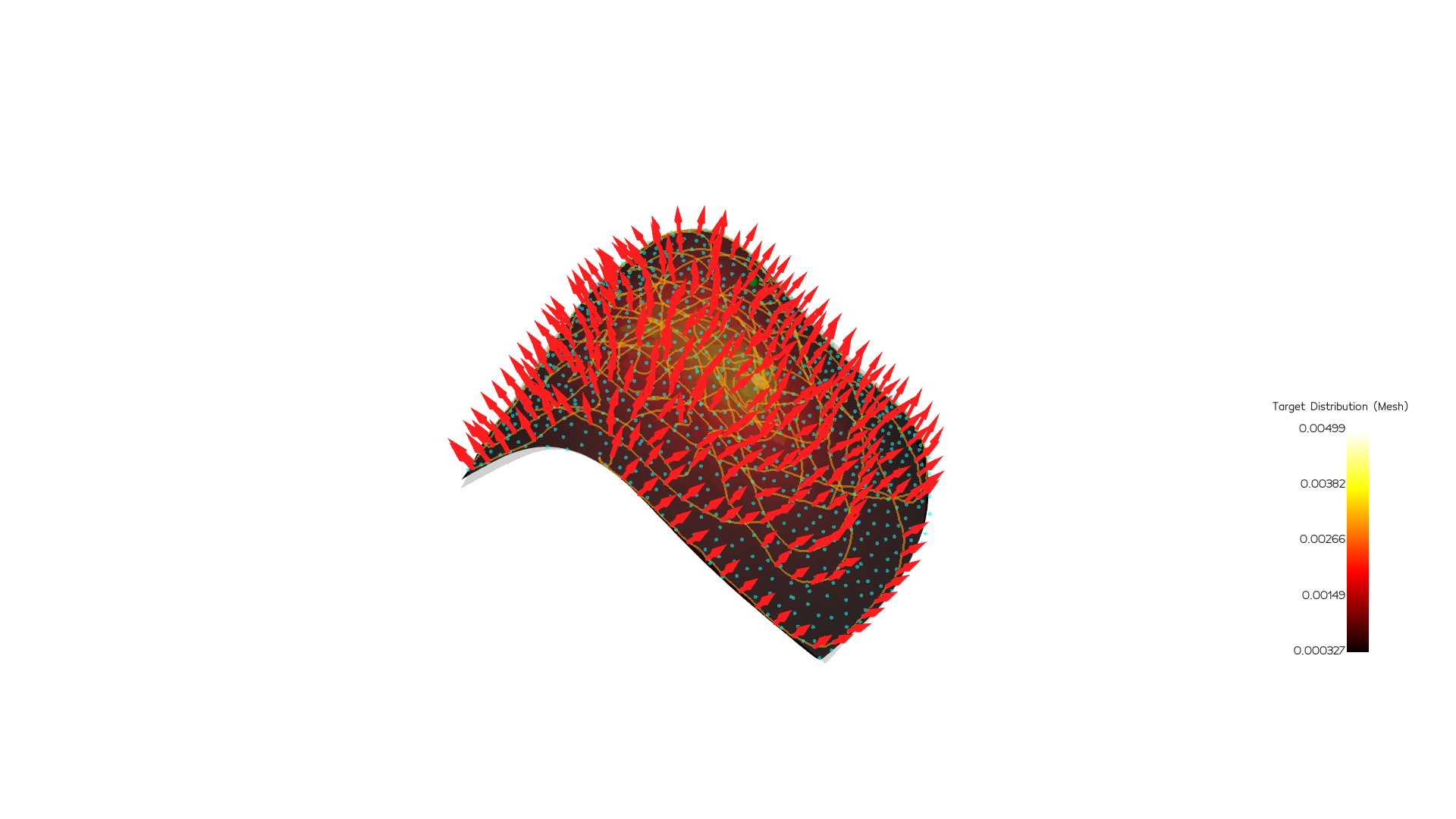} &
        \includegraphics[width=0.17\textwidth,clip,trim=600 200 600 230]{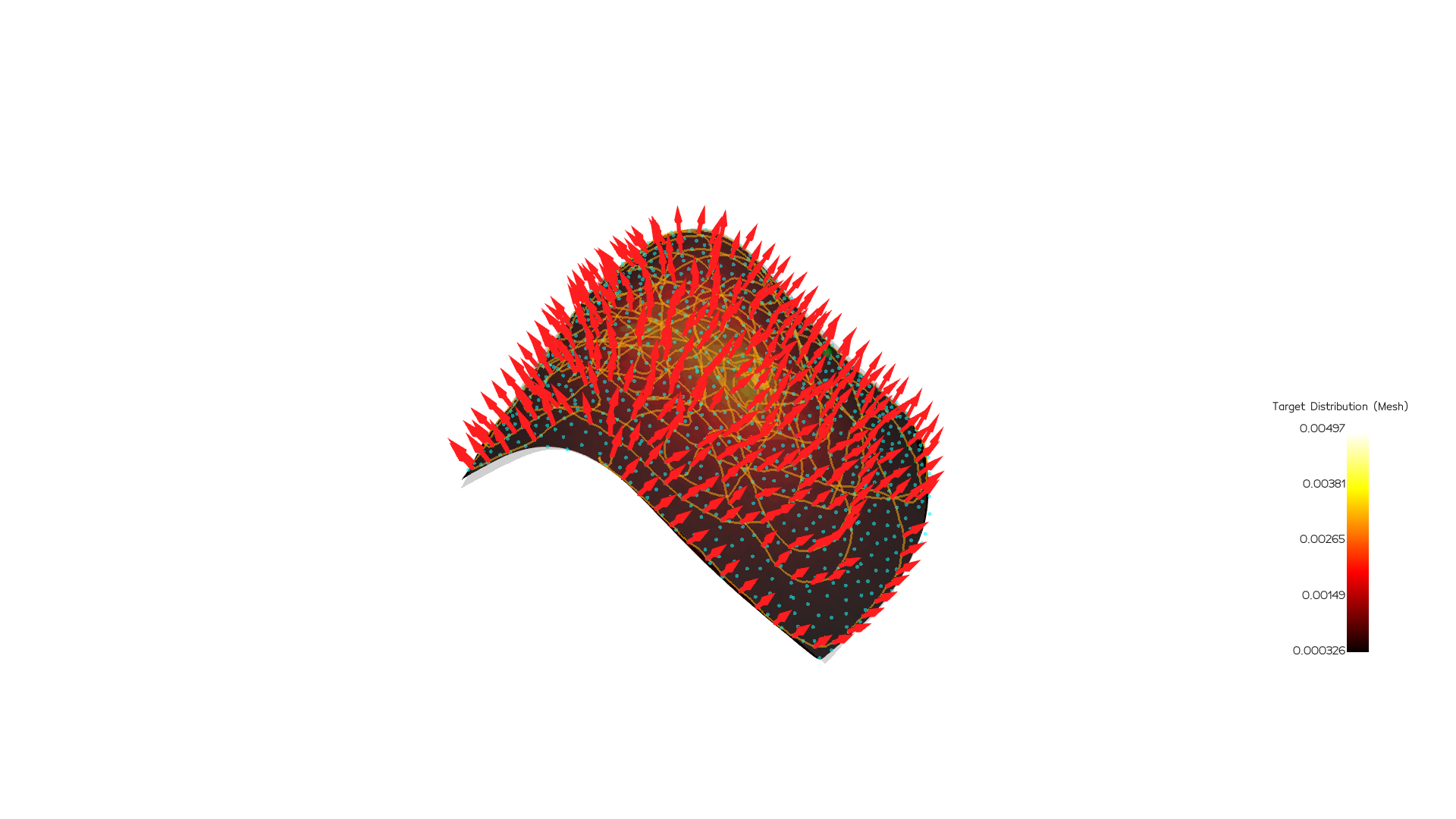} \\[0.1em]
        \cline{2-6}\\[-0.7em]
        \multirow{2}{*}[3em]{\rotatebox{90}{\small Uncertainty}} &
        \includegraphics[width=0.17\textwidth,clip,trim=600 200 600 230]{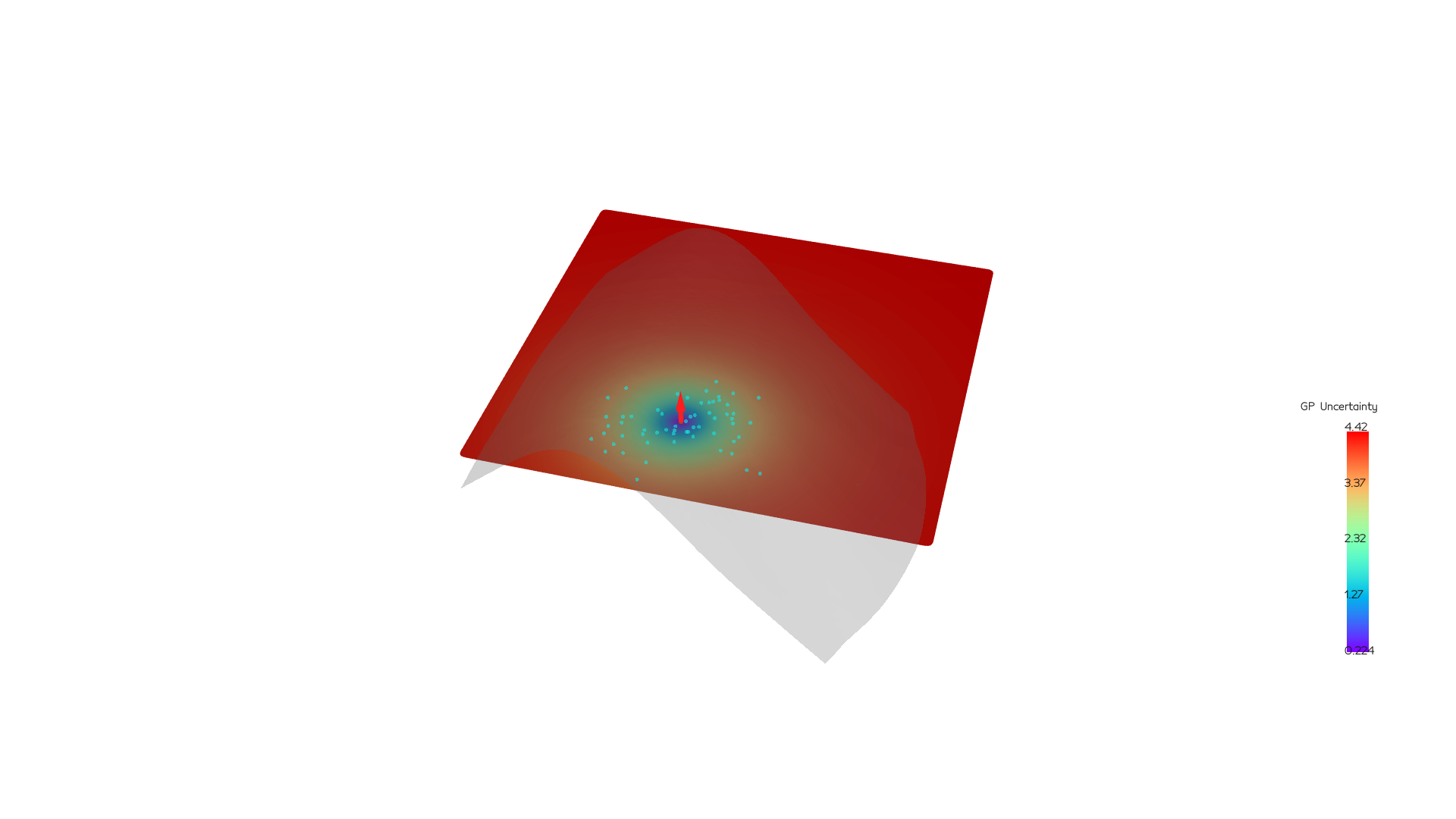} &
        \includegraphics[width=0.17\textwidth,clip,trim=600 200 600 230]{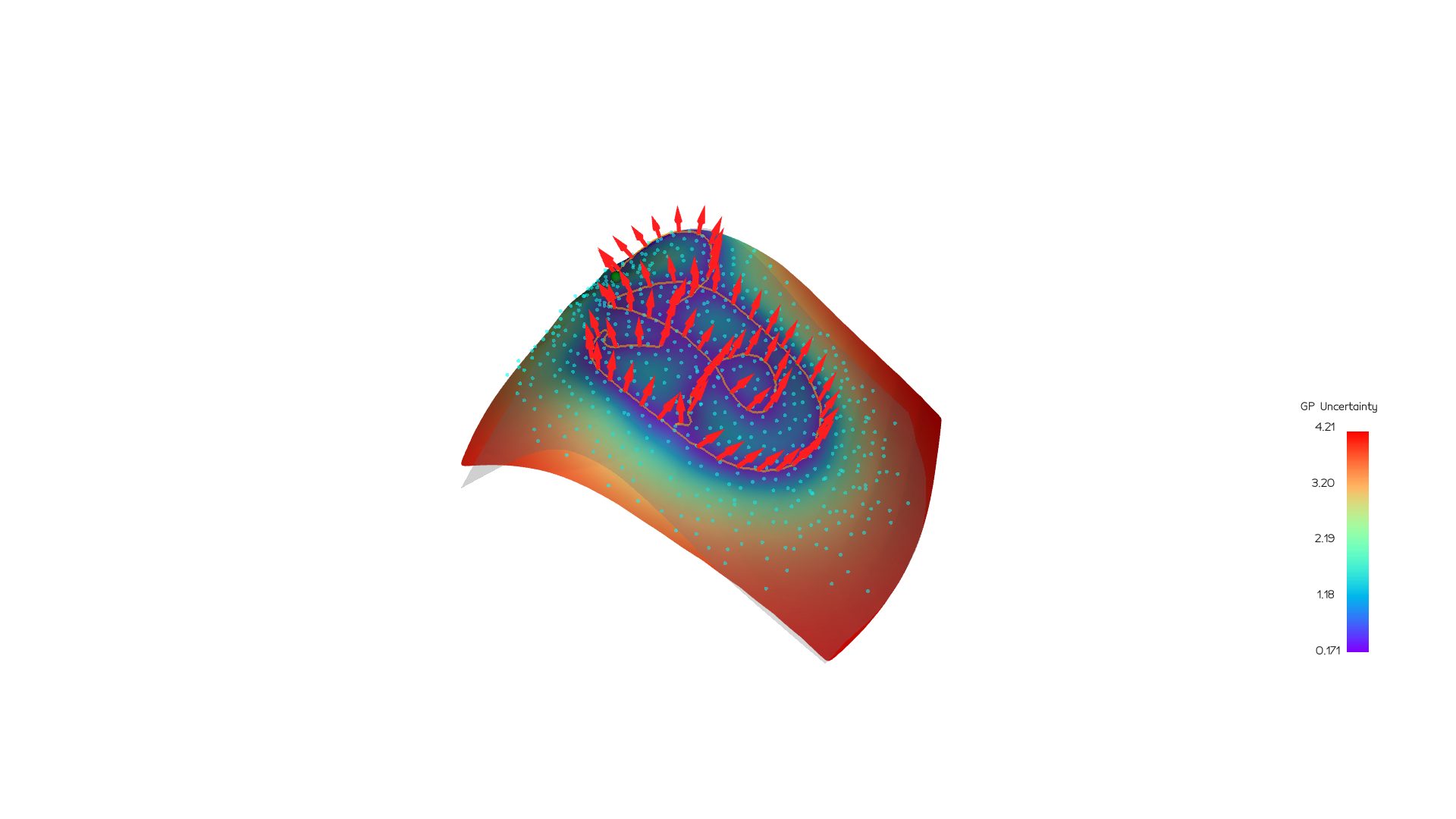} &
        \includegraphics[width=0.17\textwidth,clip,trim=600 200 600 230]{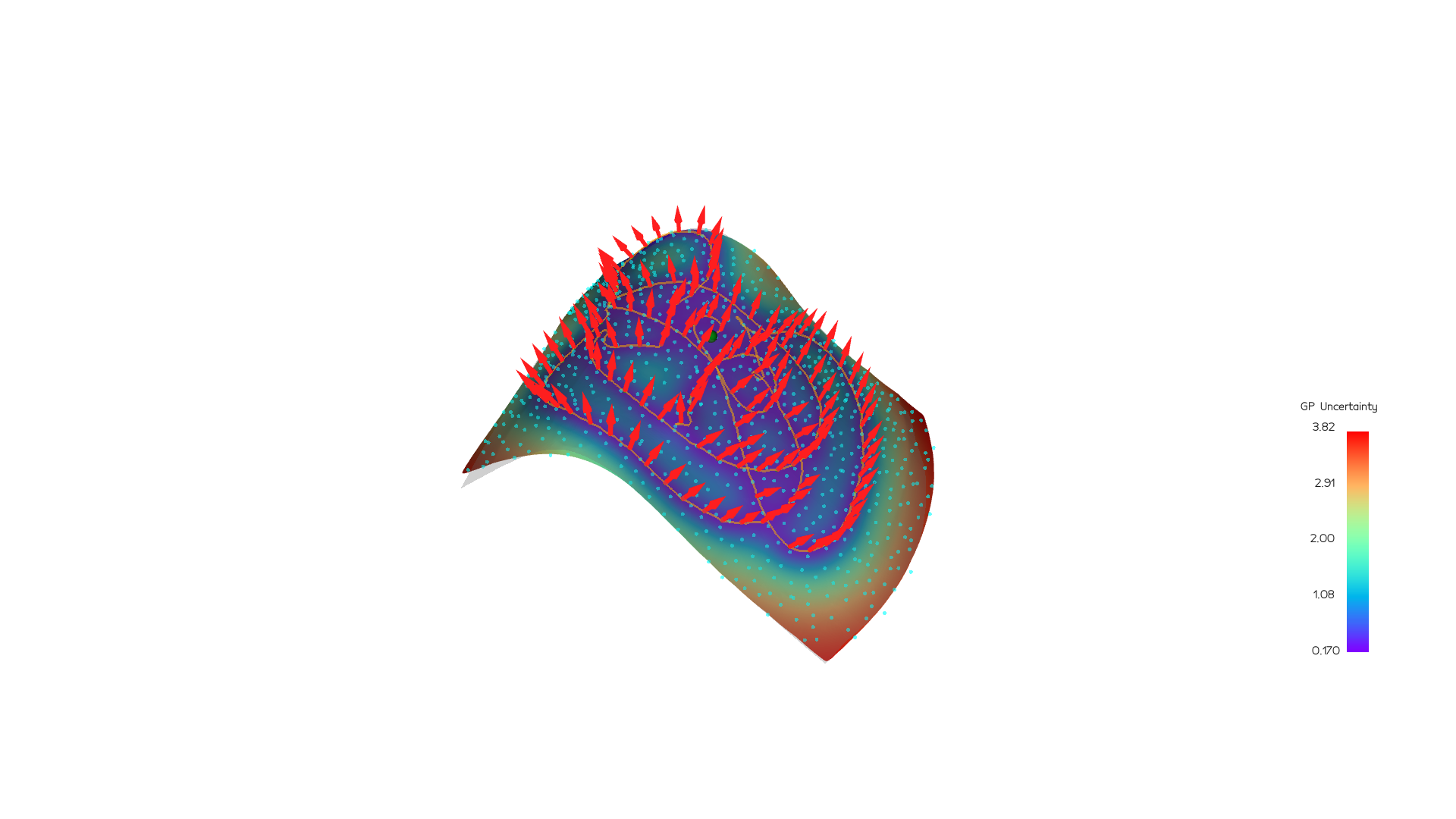} &
        \includegraphics[width=0.17\textwidth,clip,trim=600 200 600 230]{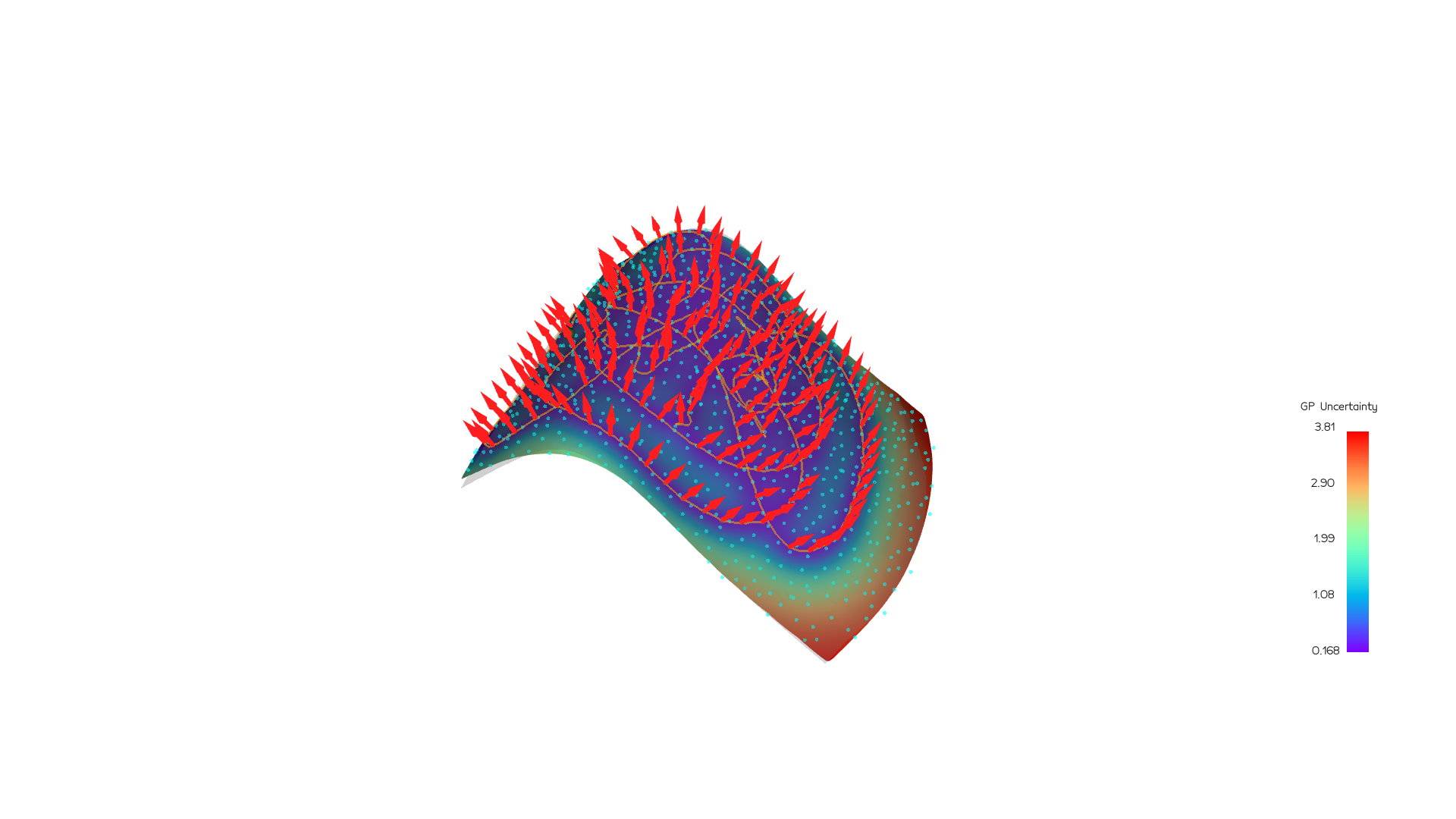} &
        \includegraphics[width=0.17\textwidth,clip,trim=600 200 600 230]{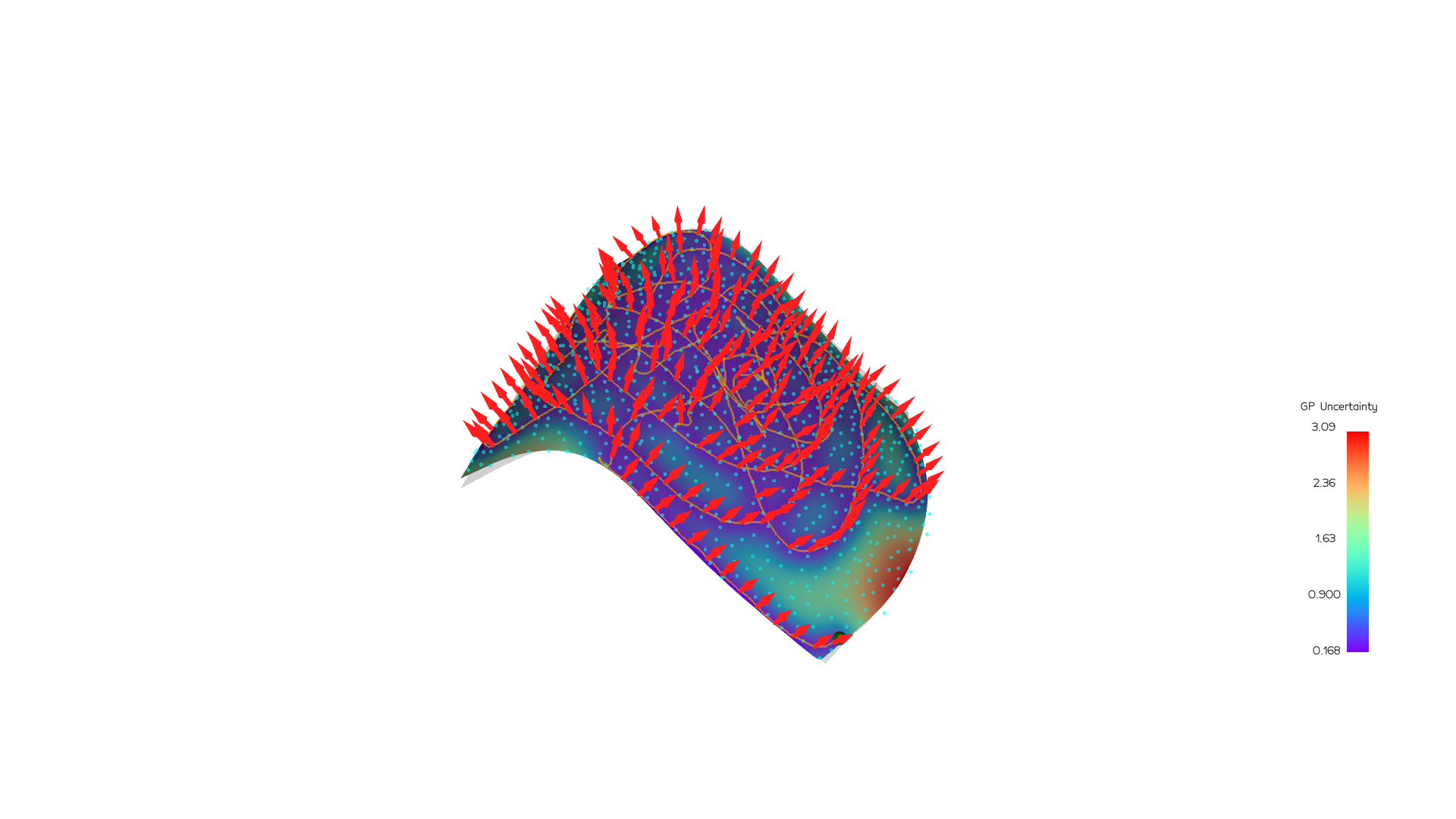} \\
        &
        \includegraphics[width=0.17\textwidth,clip,trim=600 200 600 230]{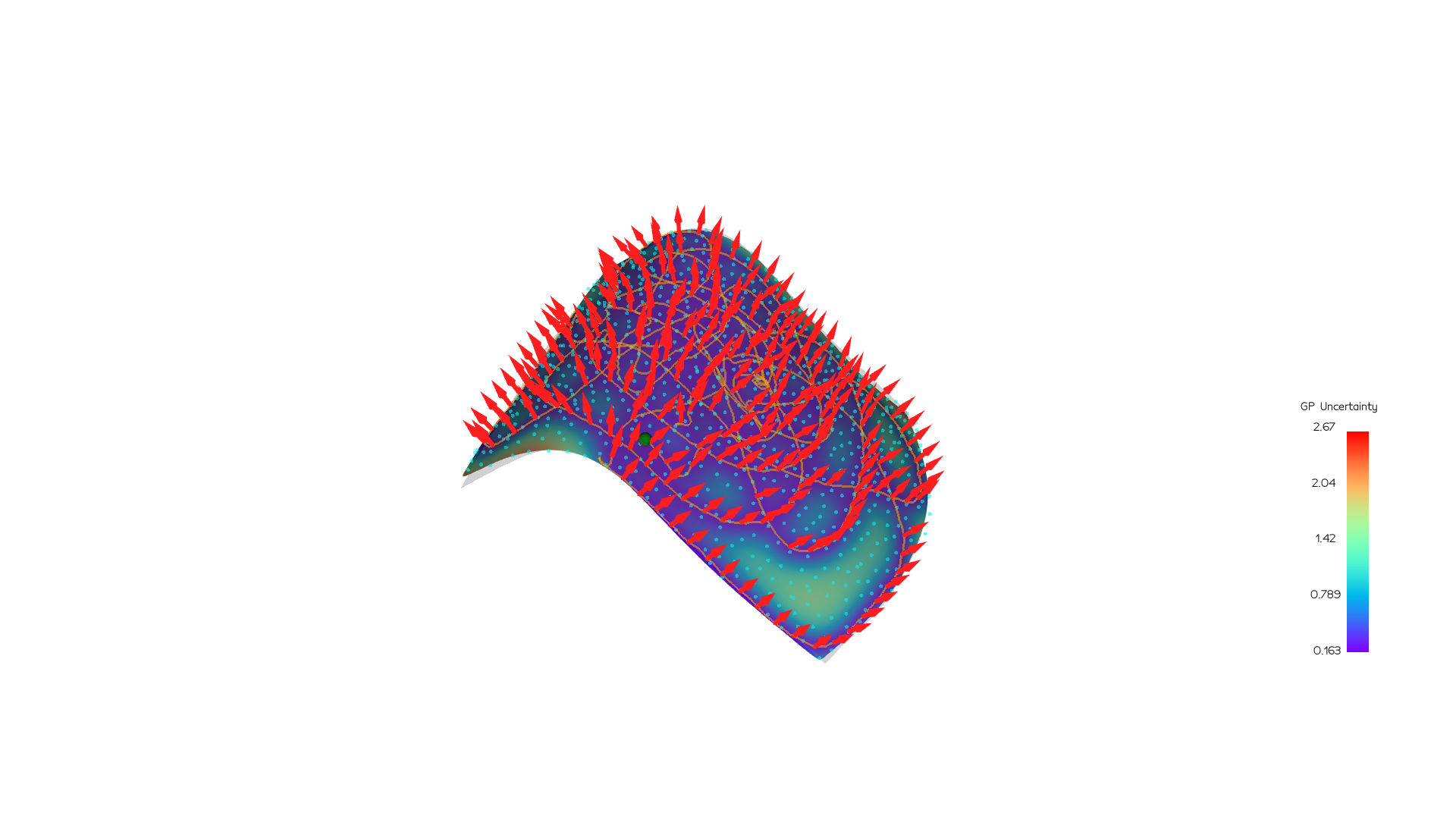} &
        \includegraphics[width=0.17\textwidth,clip,trim=600 200 600 230]{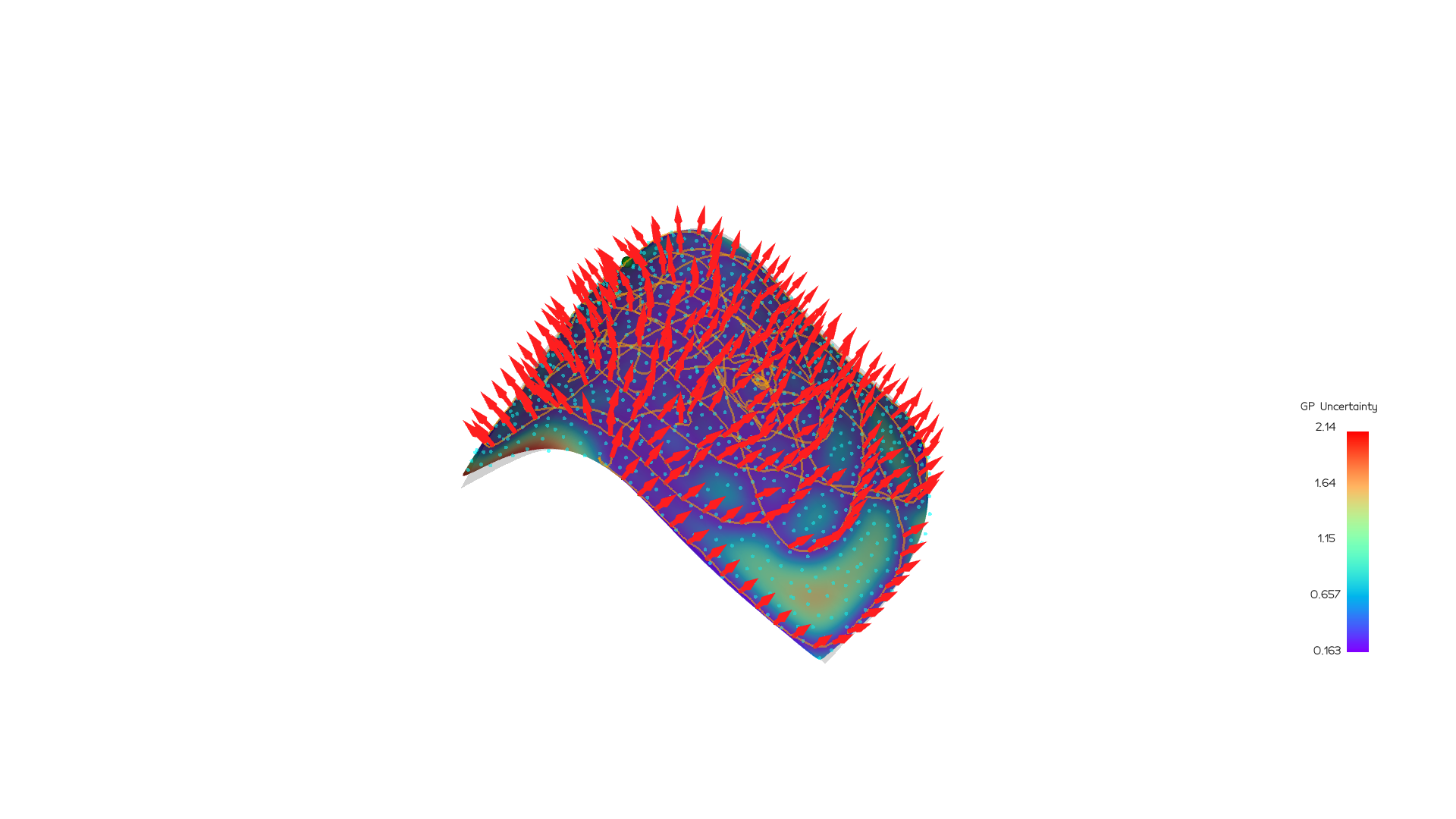} &
        \includegraphics[width=0.17\textwidth,clip,trim=600 200 600 230]{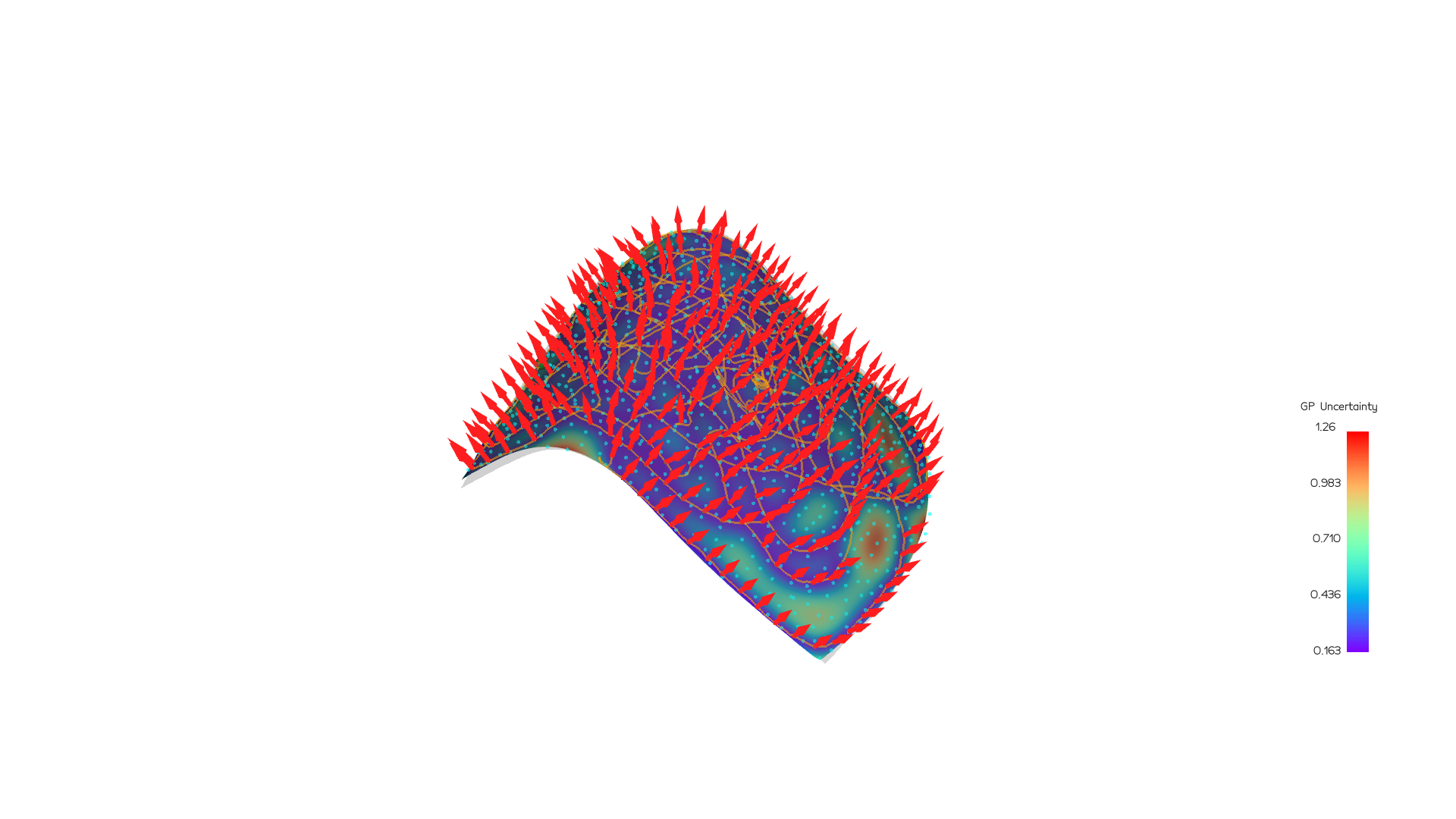} &
        \includegraphics[width=0.17\textwidth,clip,trim=600 200 600 230]{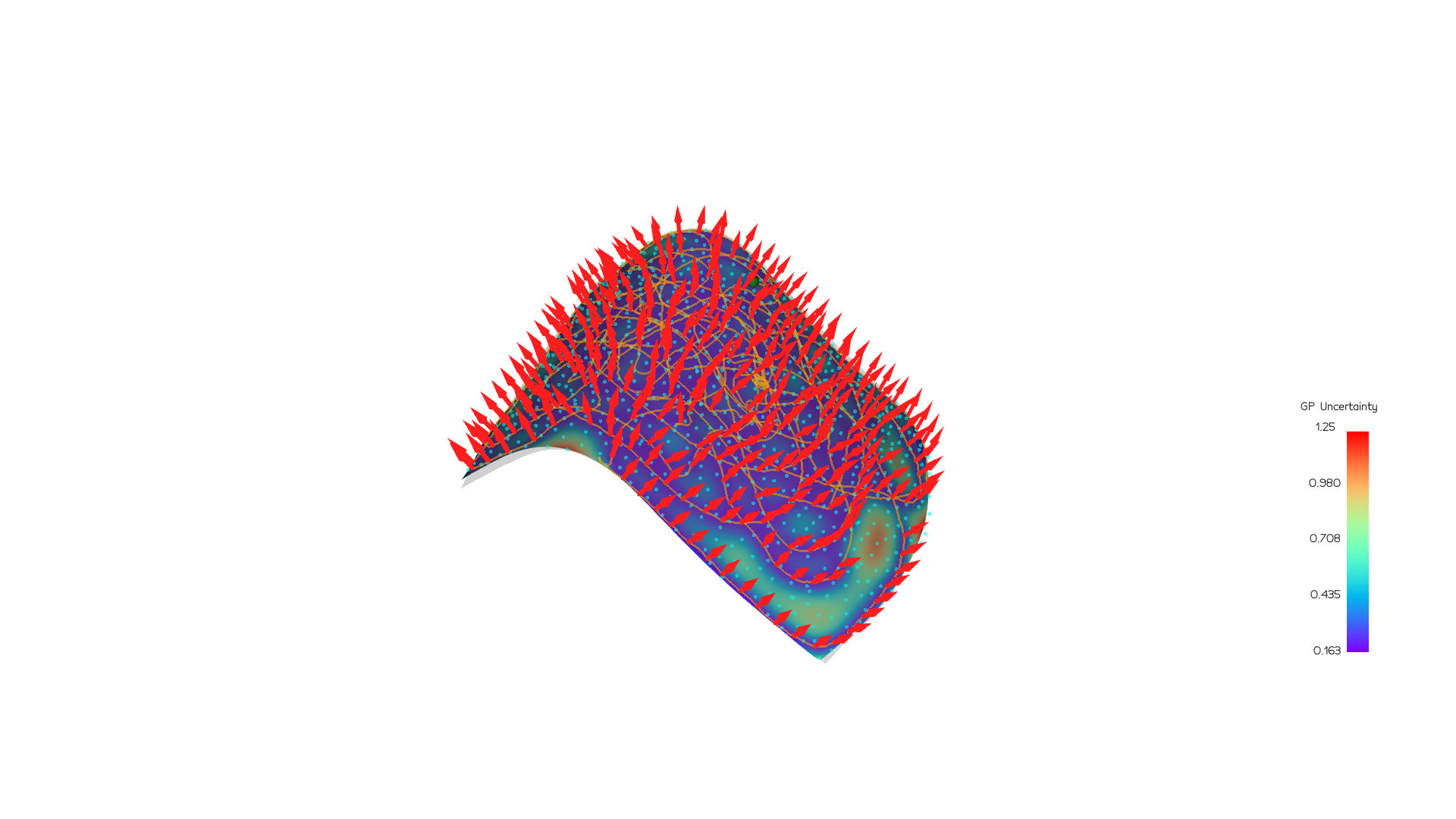} &
        \includegraphics[width=0.17\textwidth,clip,trim=600 200 600 230]{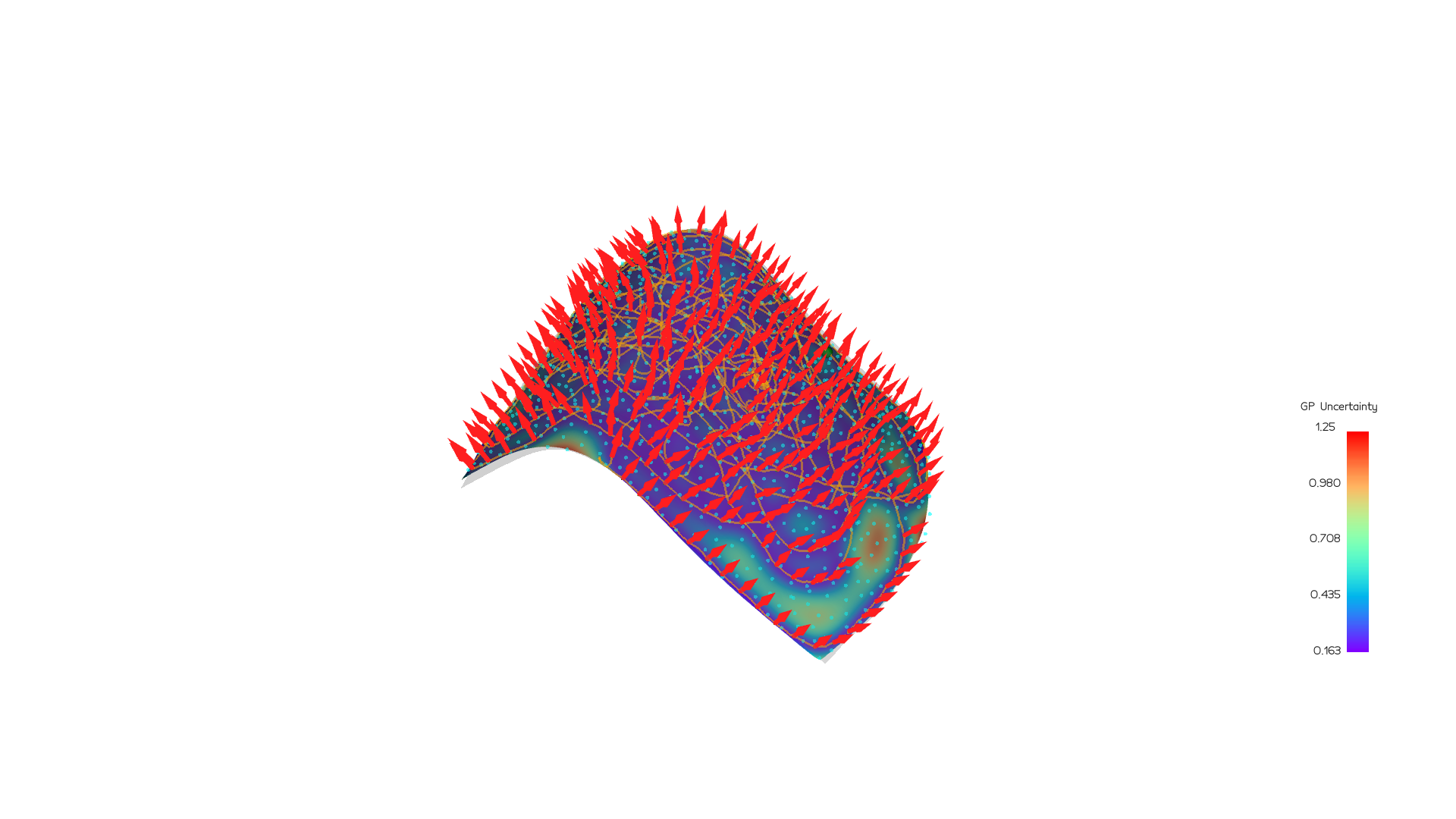} \\
        & \multicolumn{5}{c}{\scriptsize Time steps: 1, 334, 667, 1001, 1334 (top rows) | 1667, 2000, 2334, 2667, 3000 (bottom rows)} \\
    \end{tabular}
    \caption{Chair Experiment Progression over 3000 time steps (linearly interpolated). \textbf{Coverage:} Achieved exploration pattern showing systematic coverage expansion from sparse initial contact toward comprehensive surface exploration. \textbf{Target:} Task-specific target distribution from a single heat source at the chair's top. \textbf{Uncertainty:} GPIS prediction uncertainty, with warmer colors indicating higher uncertainty that progressively reduces in explored areas.}
    \label{fig:chair_coverage_progression}
    \label{fig:chair_target_progression}
    \label{fig:chair_uncertainty_progression}
\end{figure*}

\begin{figure*}[htbp]
    \centering
    \begin{tabular}{@{}r@{\hspace{0.3em}}ccccc@{}}
        \multirow{2}{*}[3em]{\rotatebox{90}{\small Coverage}} &
        \includegraphics[width=0.17\textwidth,clip,trim=600 180 600 380]{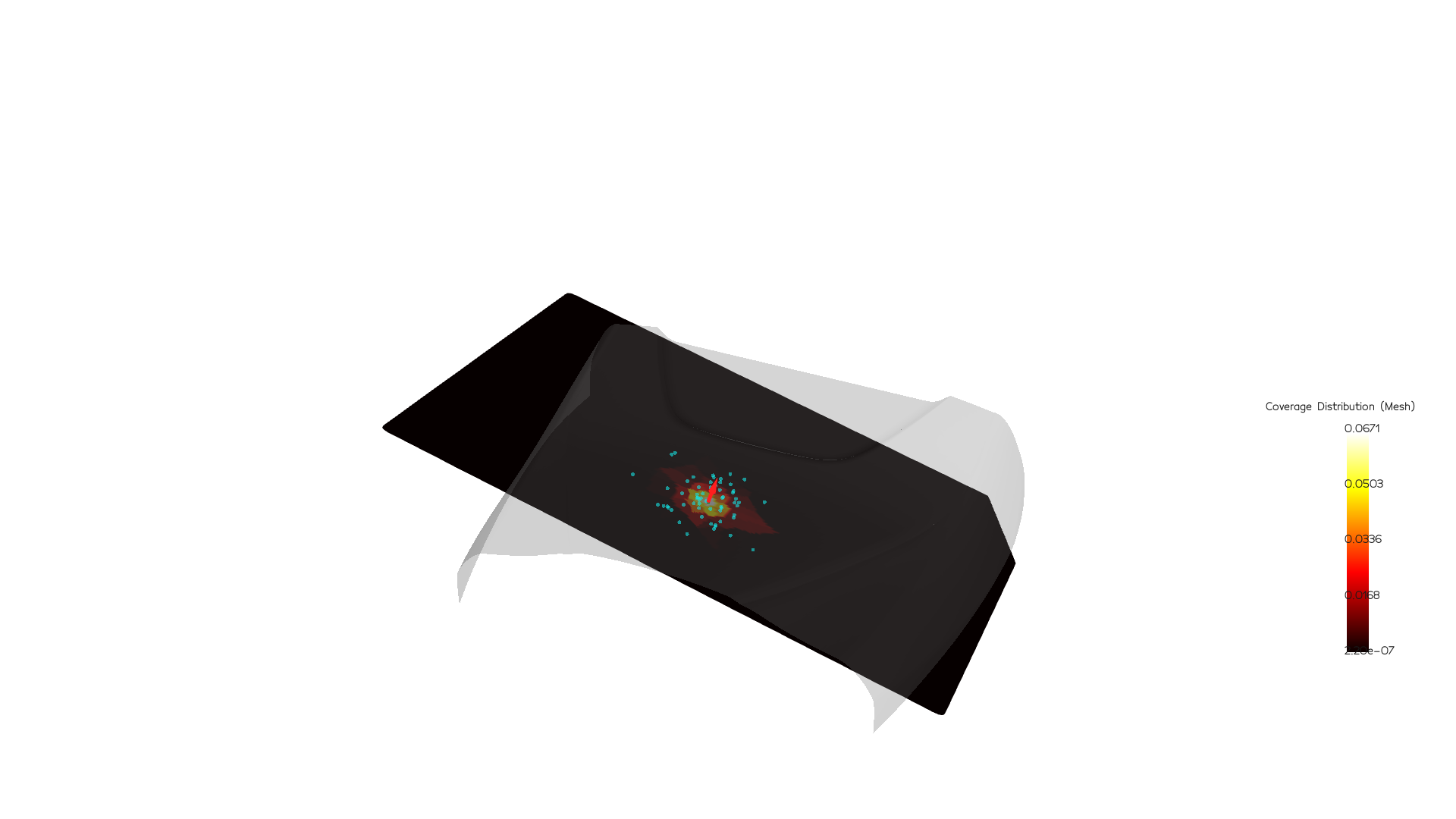} &
        \includegraphics[width=0.17\textwidth,clip,trim=600 180 600 380]{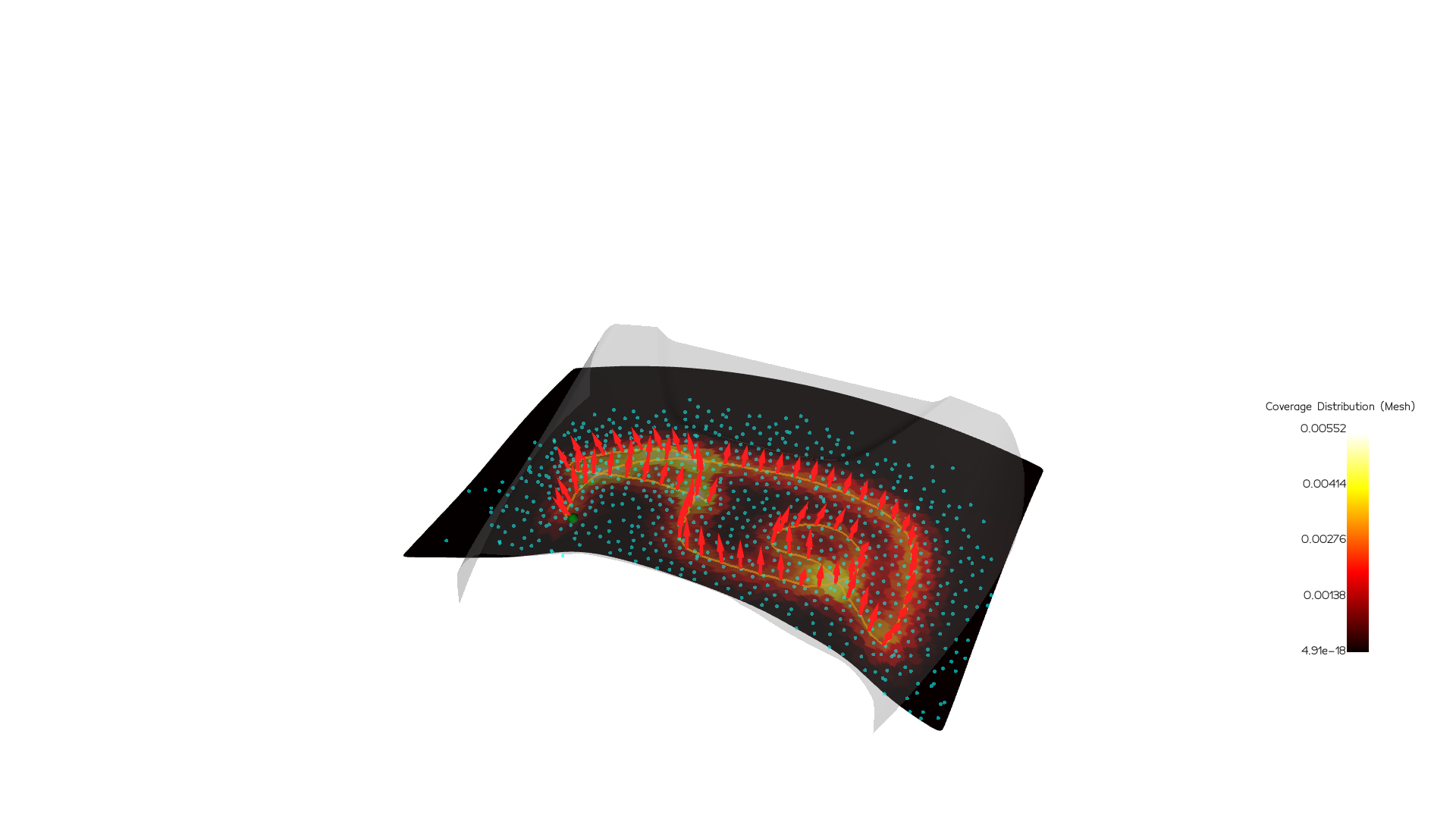} &
        \includegraphics[width=0.17\textwidth,clip,trim=600 180 600 380]{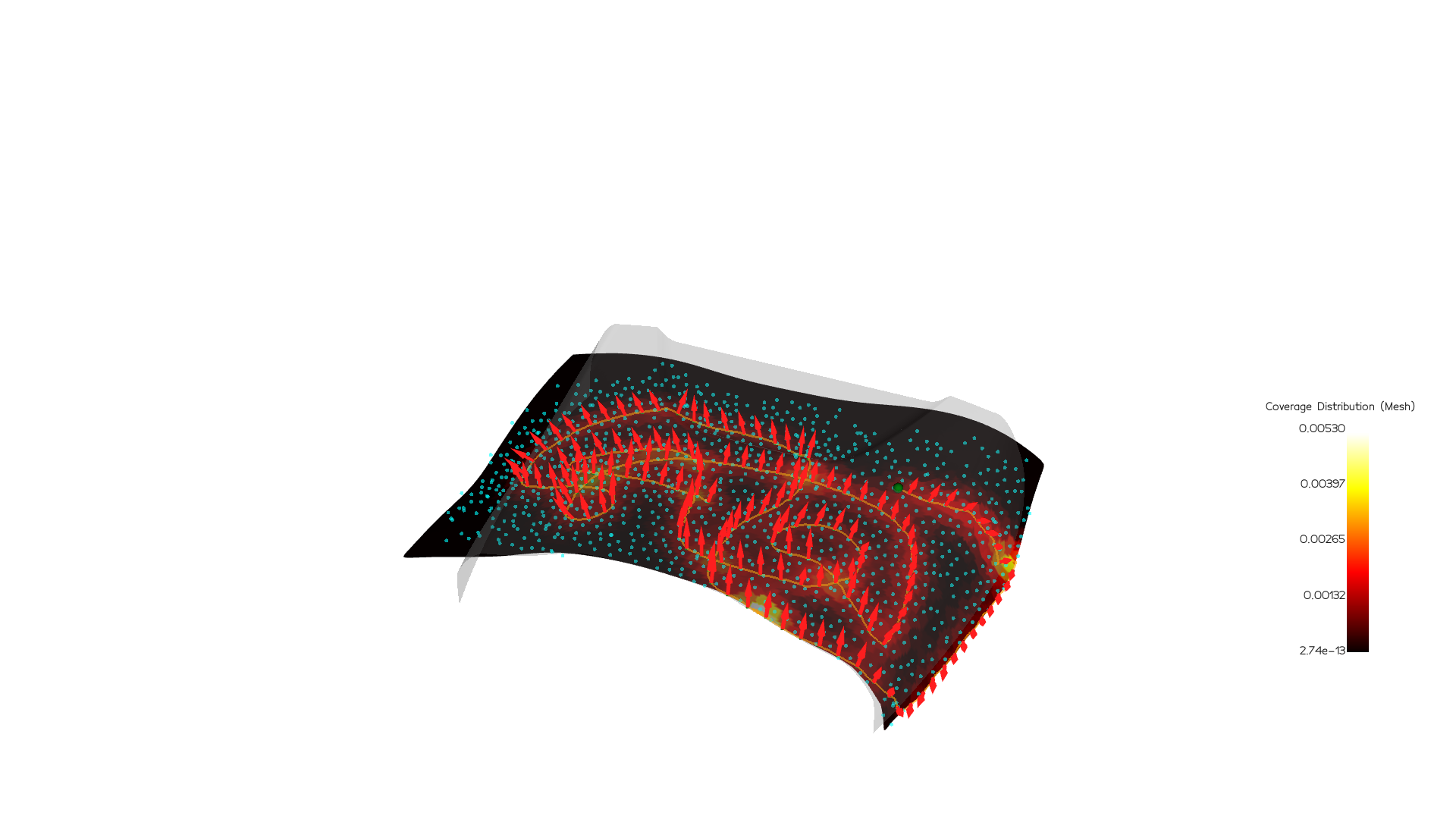} &
        \includegraphics[width=0.17\textwidth,clip,trim=600 180 600 380]{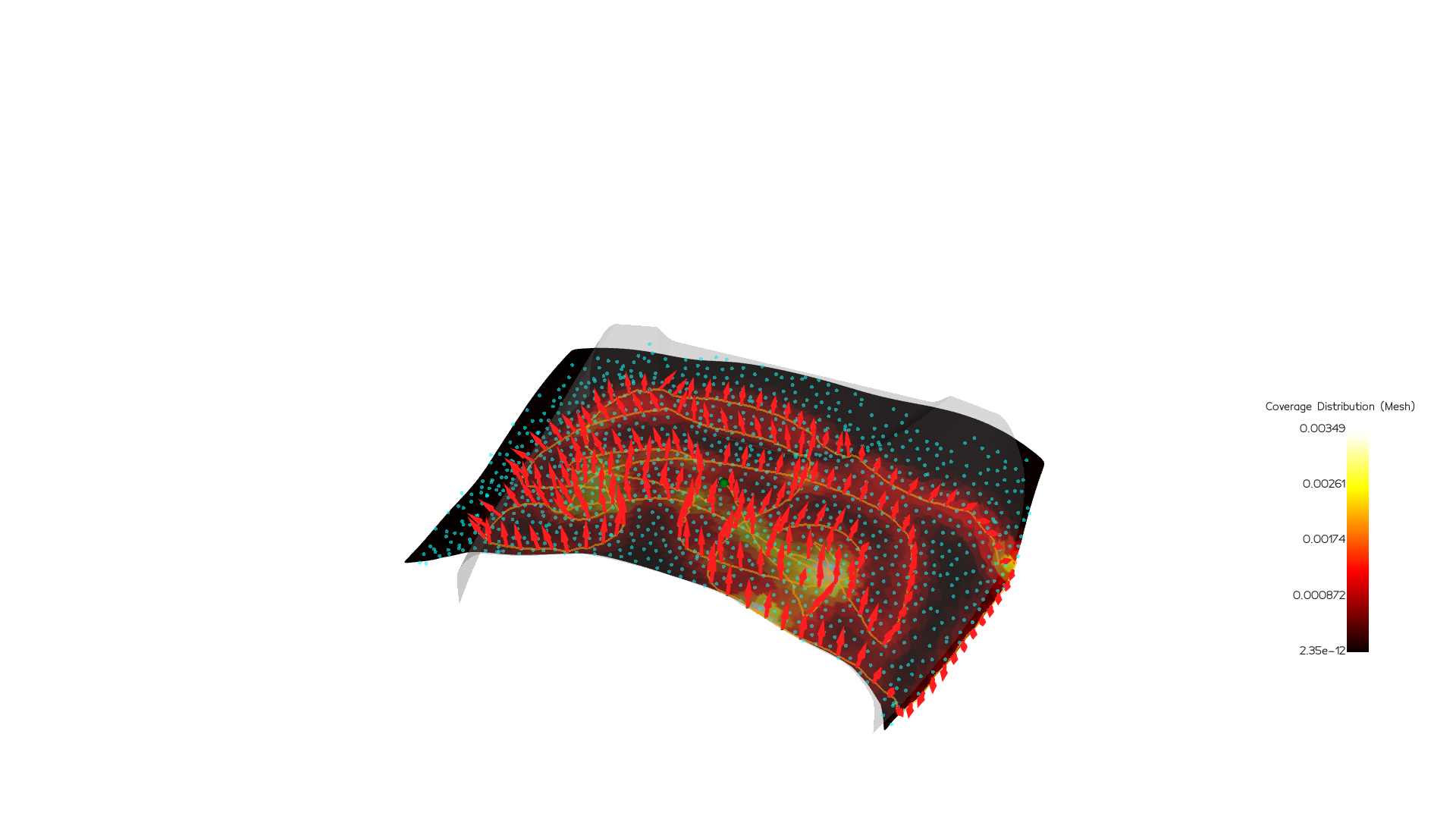} &
        \includegraphics[width=0.17\textwidth,clip,trim=600 180 600 380]{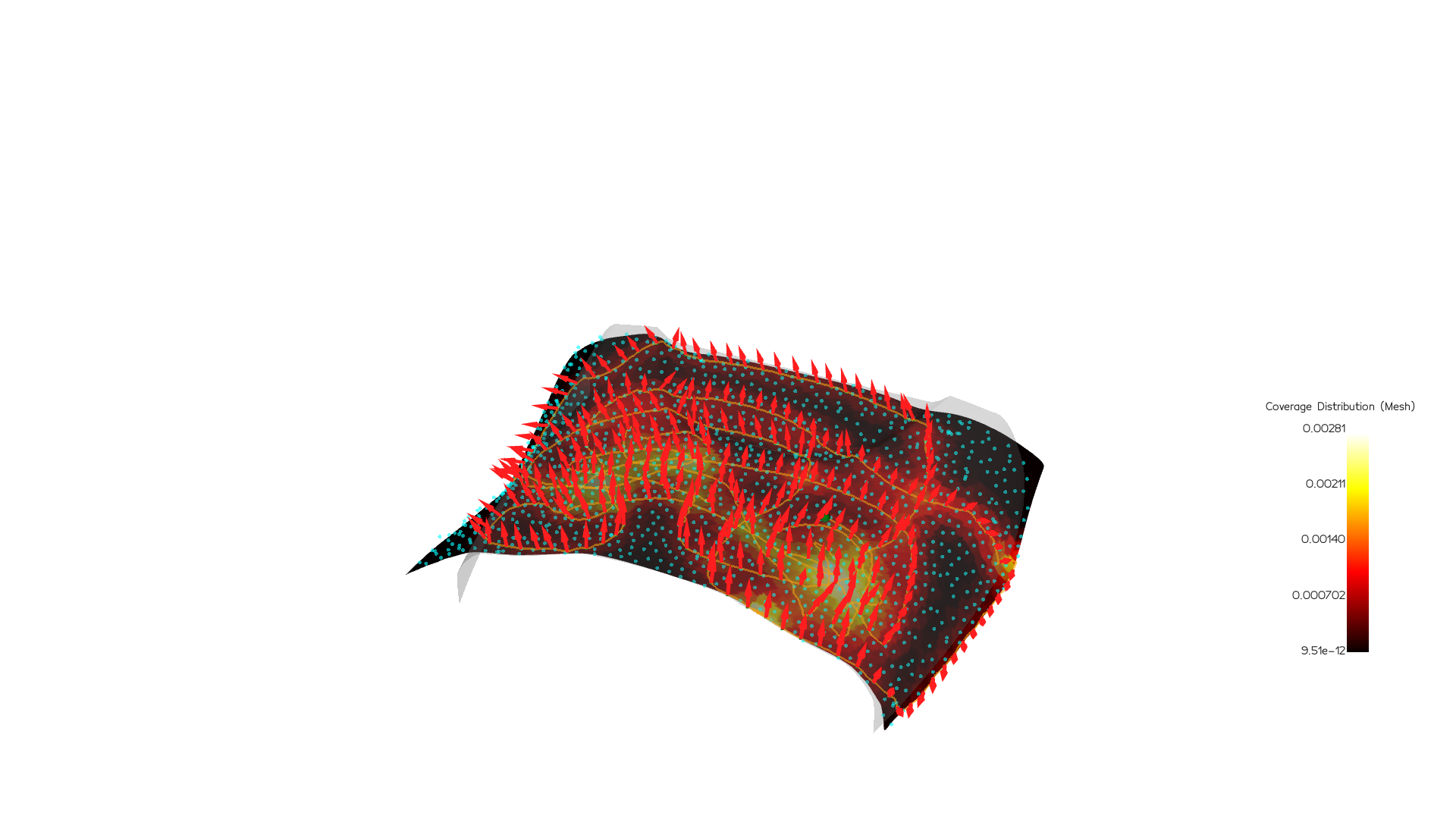} \\
        &
        \includegraphics[width=0.17\textwidth,clip,trim=600 180 600 380]{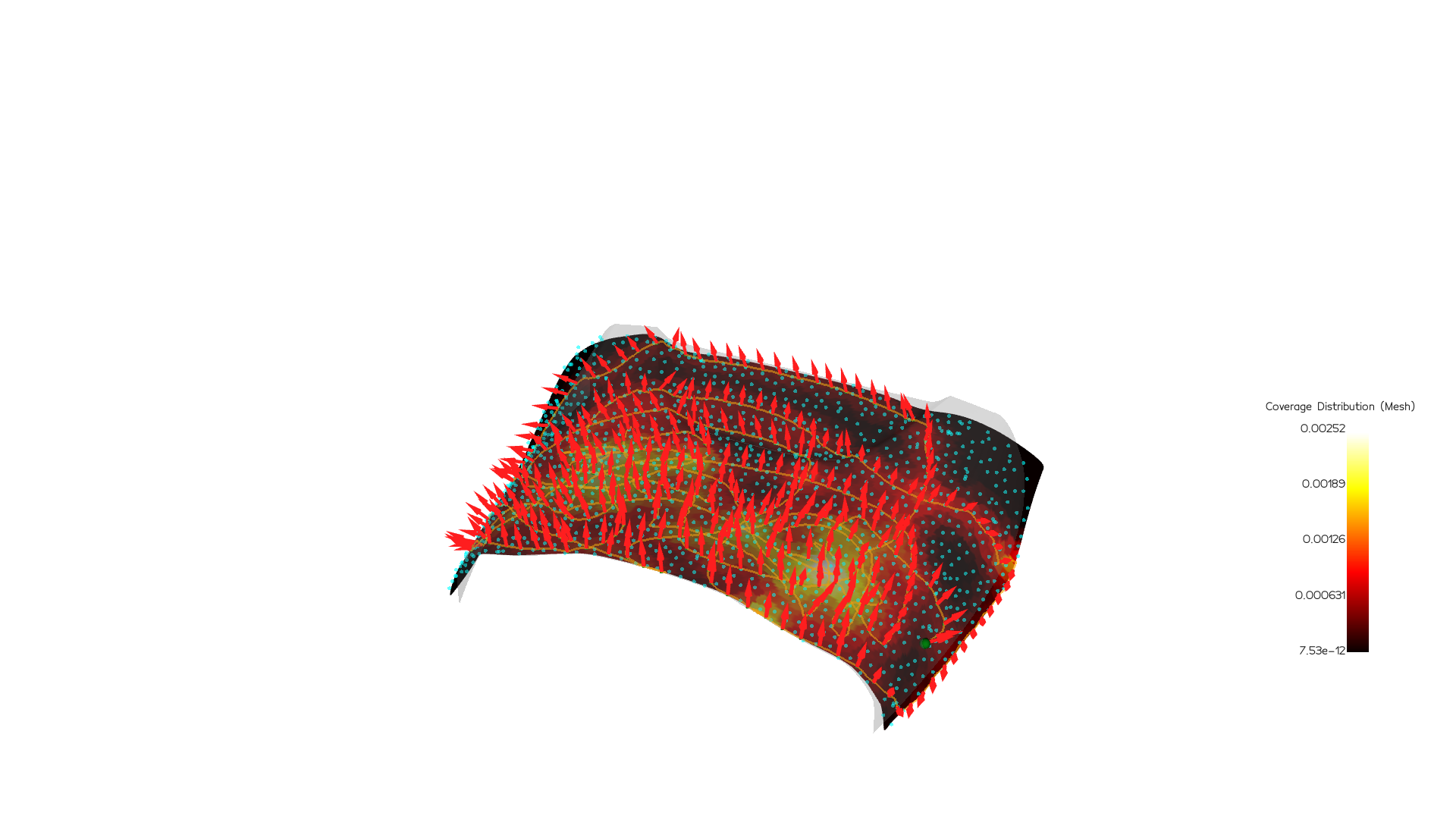} &
        \includegraphics[width=0.17\textwidth,clip,trim=600 180 600 380]{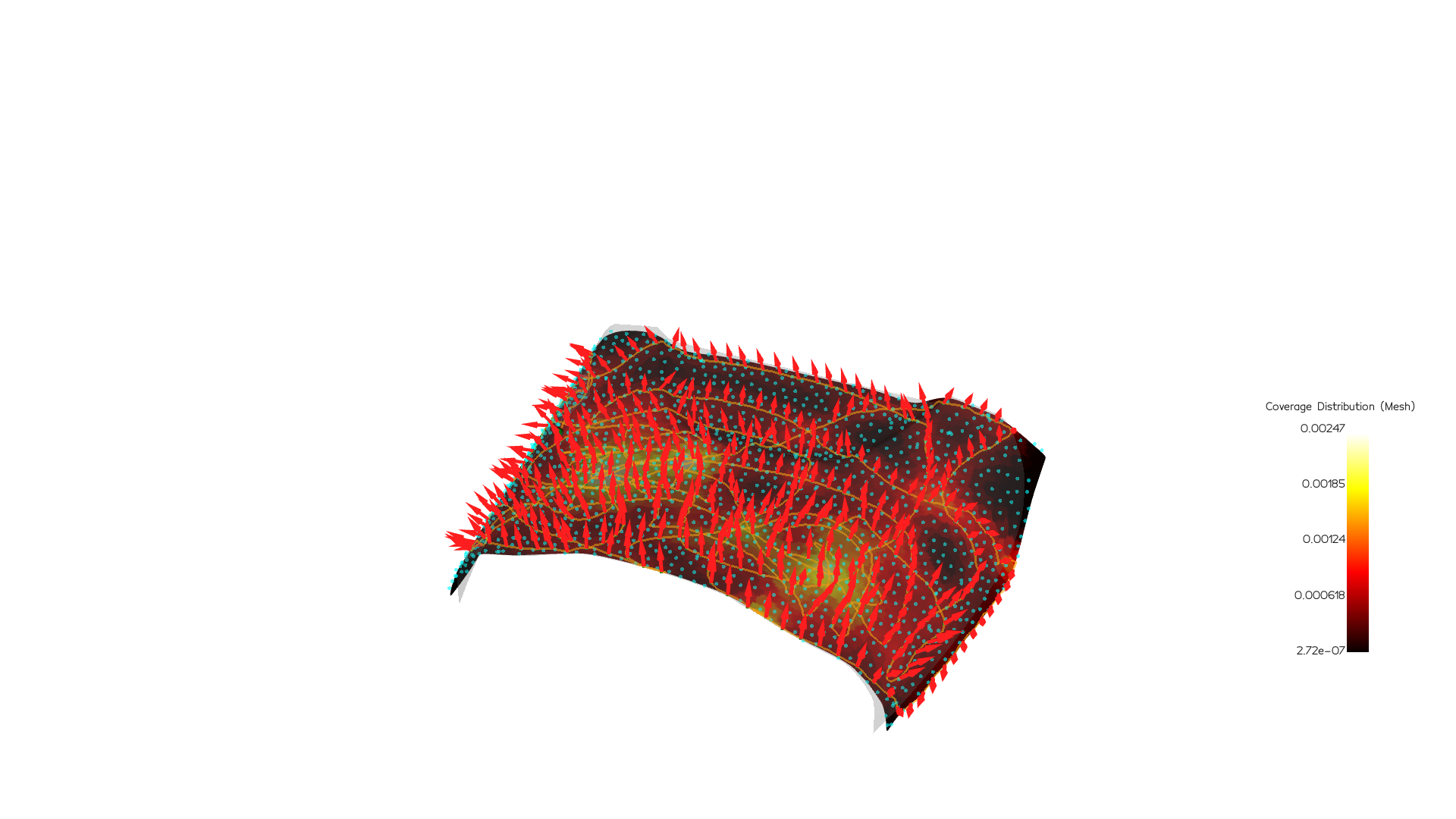} &
        \includegraphics[width=0.17\textwidth,clip,trim=600 180 600 380]{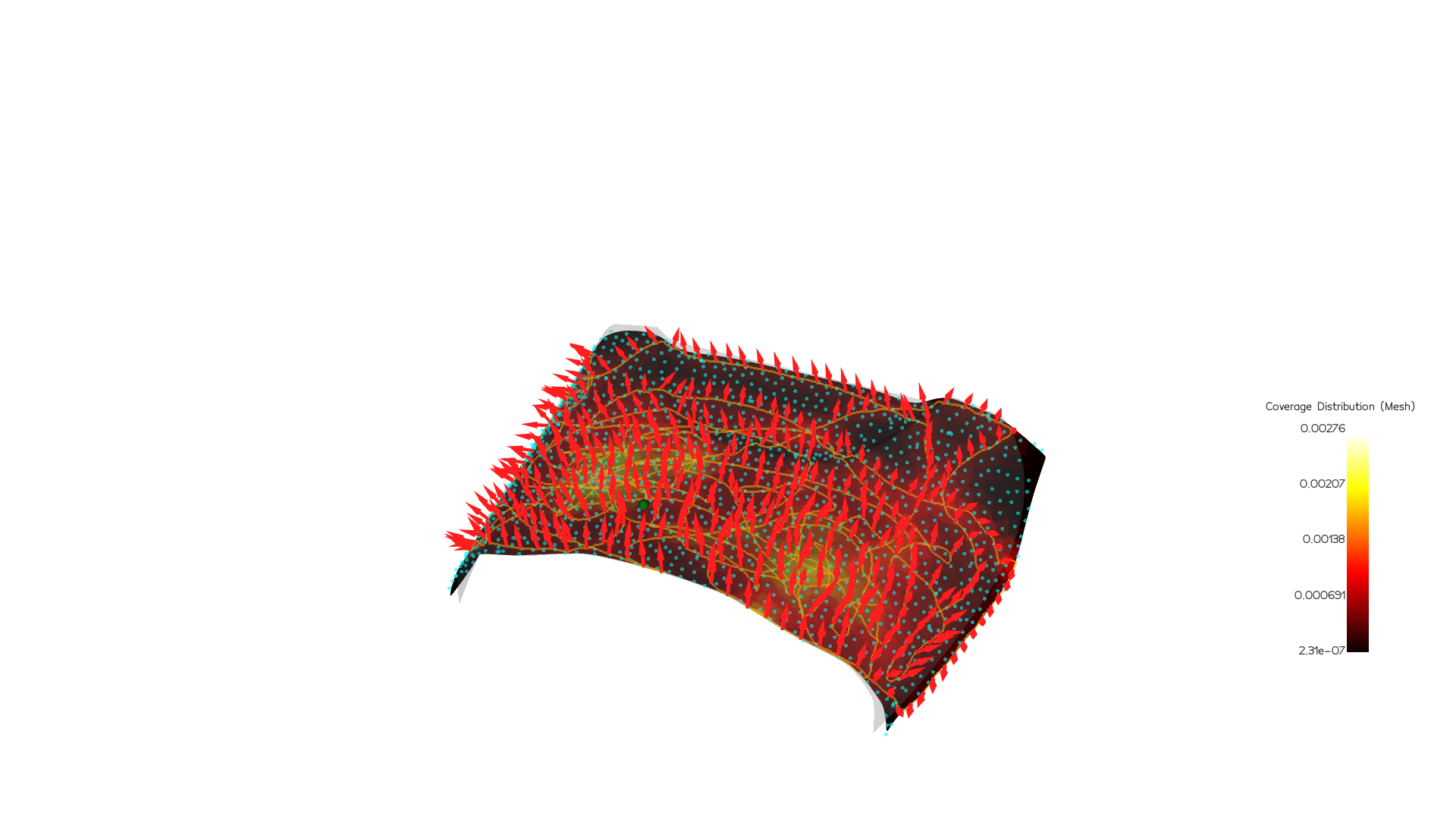} &
        \includegraphics[width=0.17\textwidth,clip,trim=600 180 600 380]{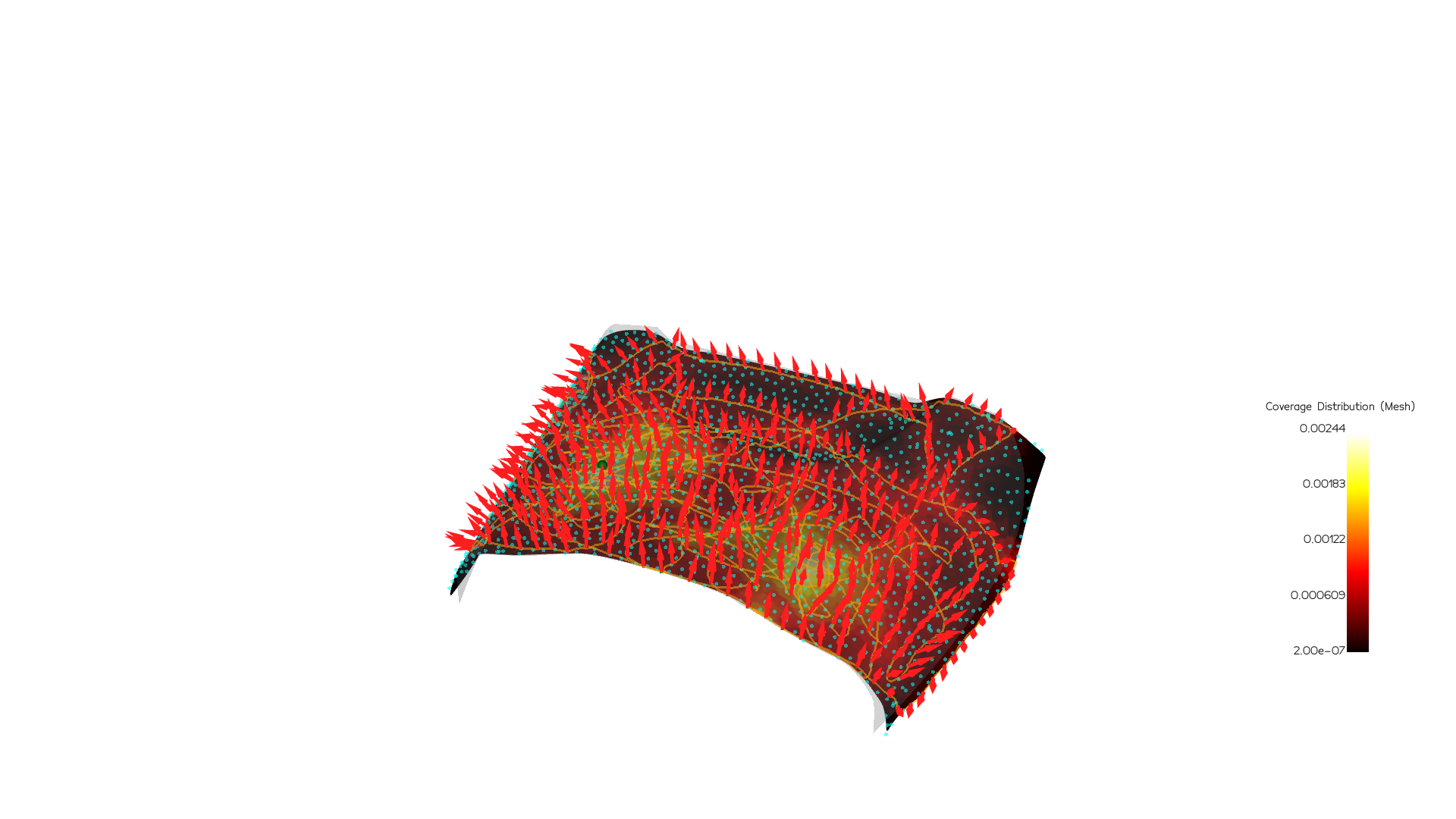} &
        \includegraphics[width=0.17\textwidth,clip,trim=600 180 600 380]{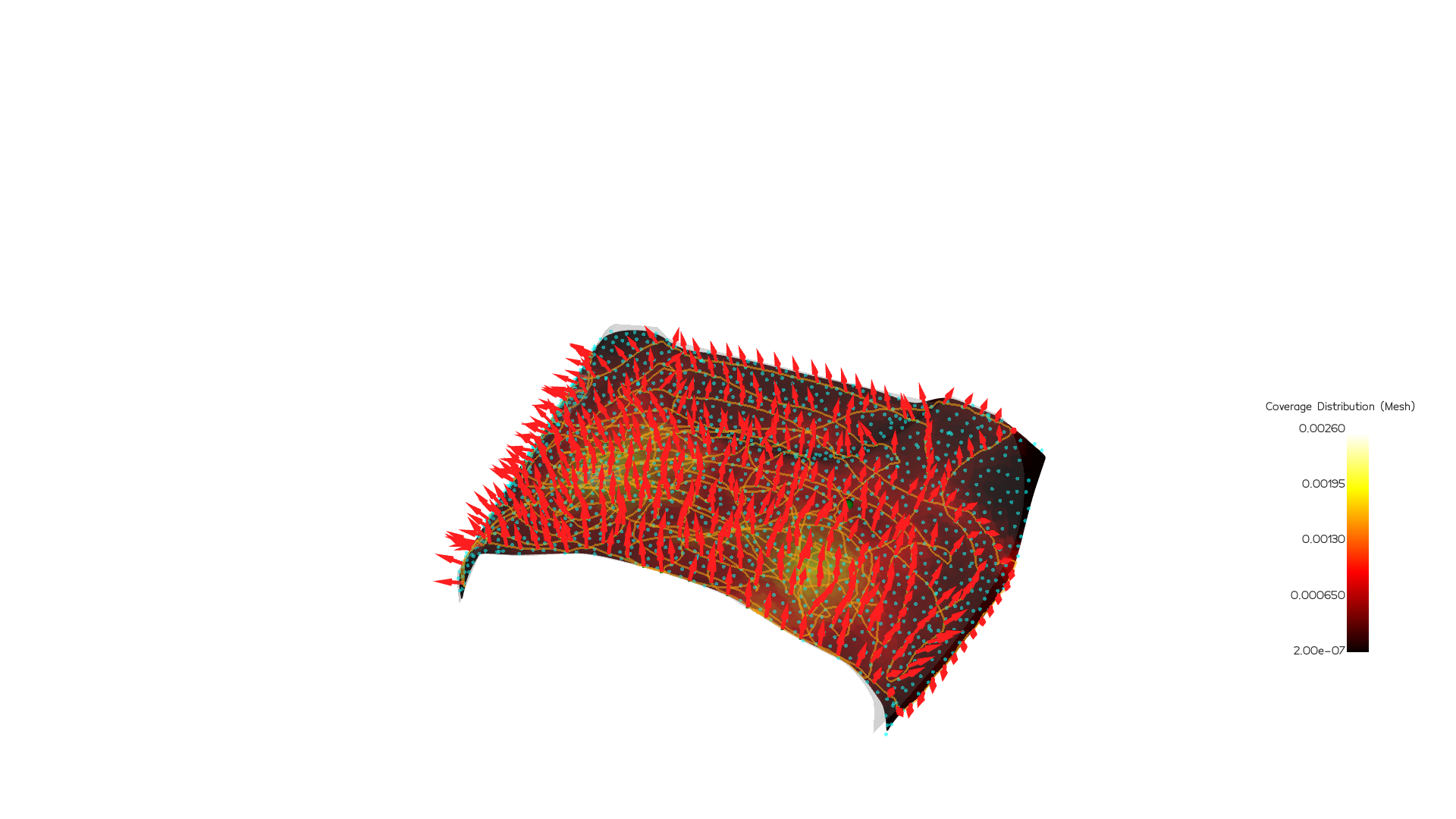} \\[0.1em]
        \cline{2-6}\\[-0.7em]
        \multirow{2}{*}[3em]{\rotatebox{90}{\small Target}} &
        \includegraphics[width=0.17\textwidth,clip,trim=600 180 600 380]{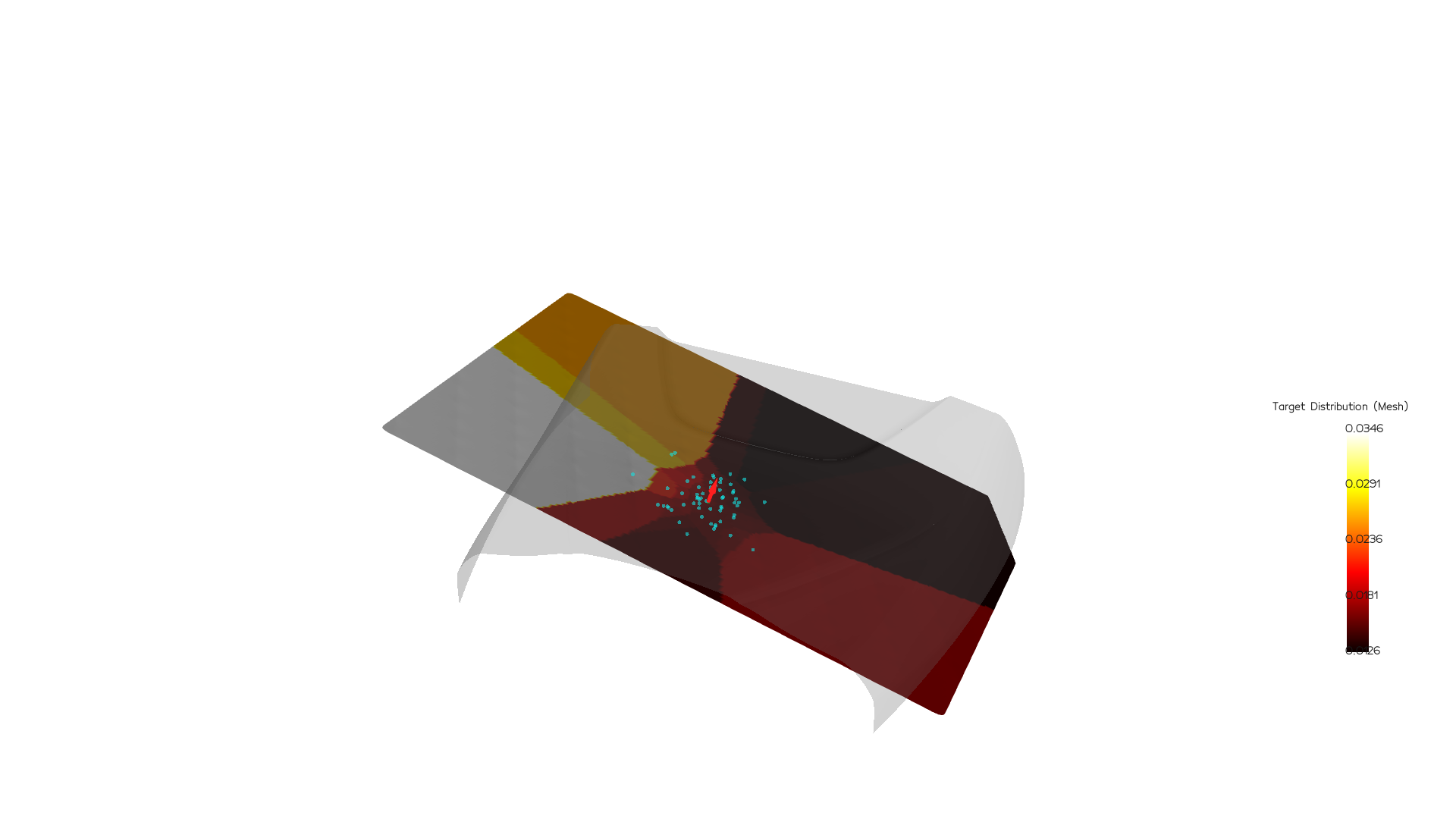} &
        \includegraphics[width=0.17\textwidth,clip,trim=600 180 600 380]{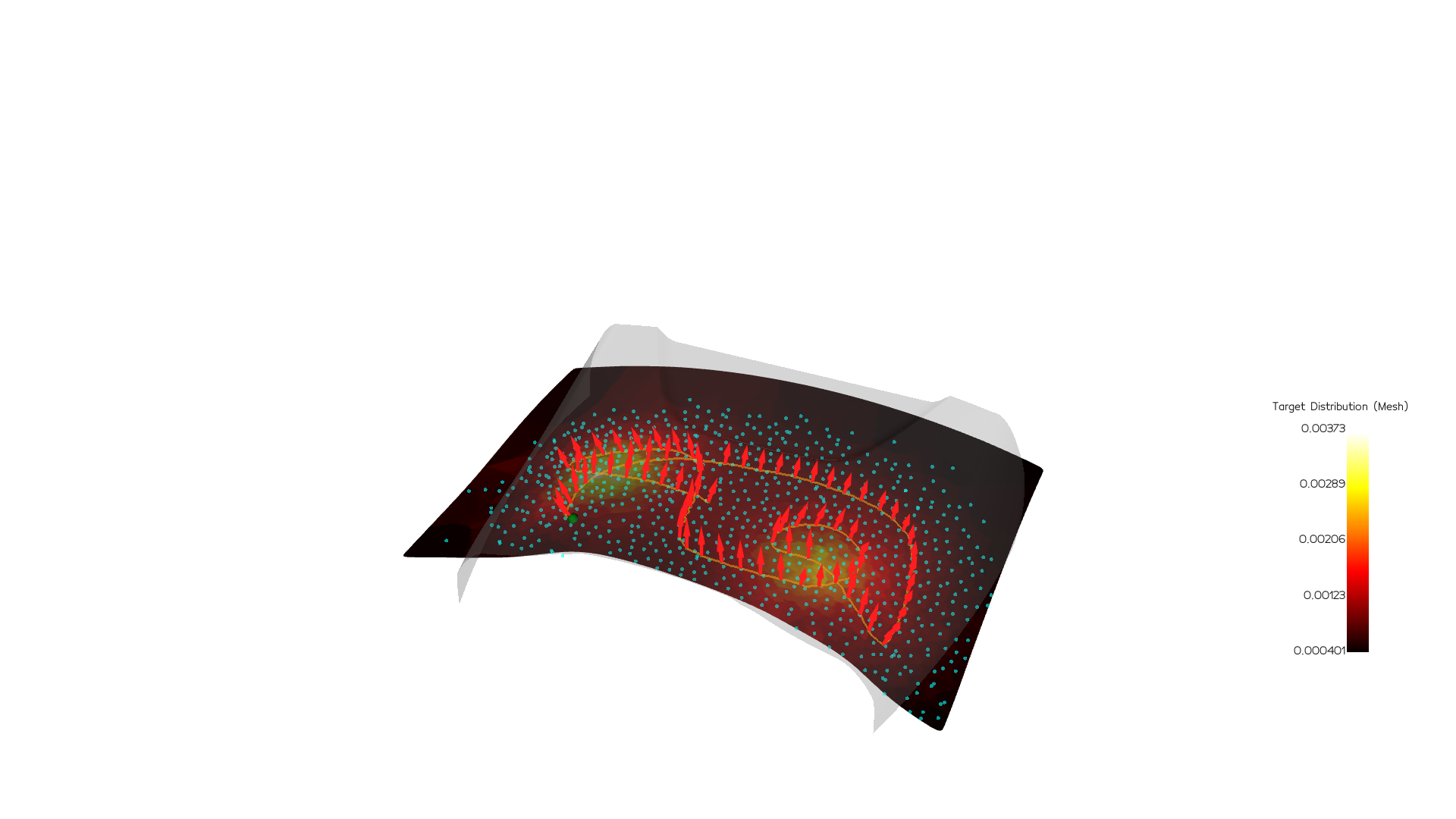} &
        \includegraphics[width=0.17\textwidth,clip,trim=600 180 600 380]{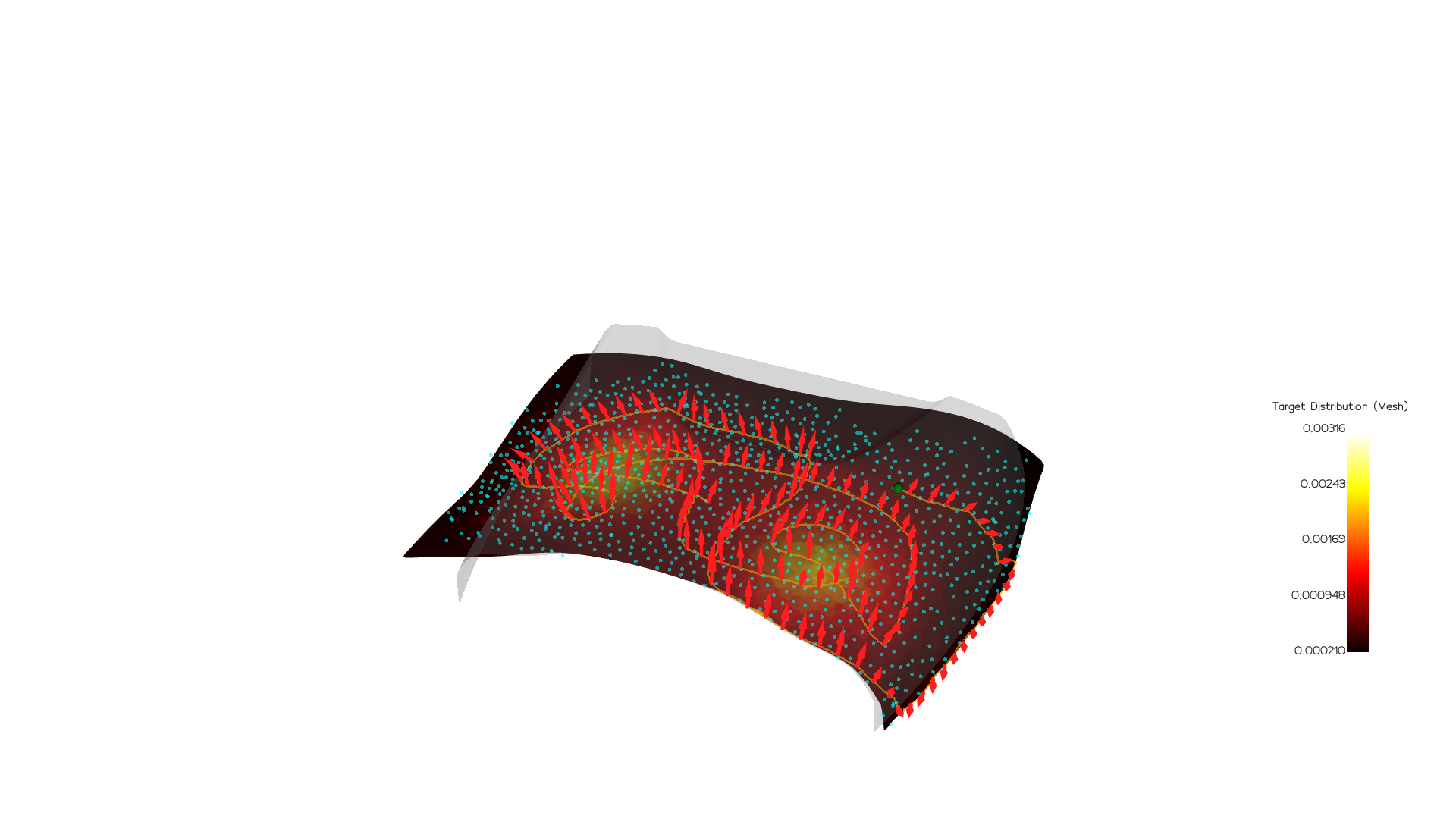} &
        \includegraphics[width=0.17\textwidth,clip,trim=600 180 600 380]{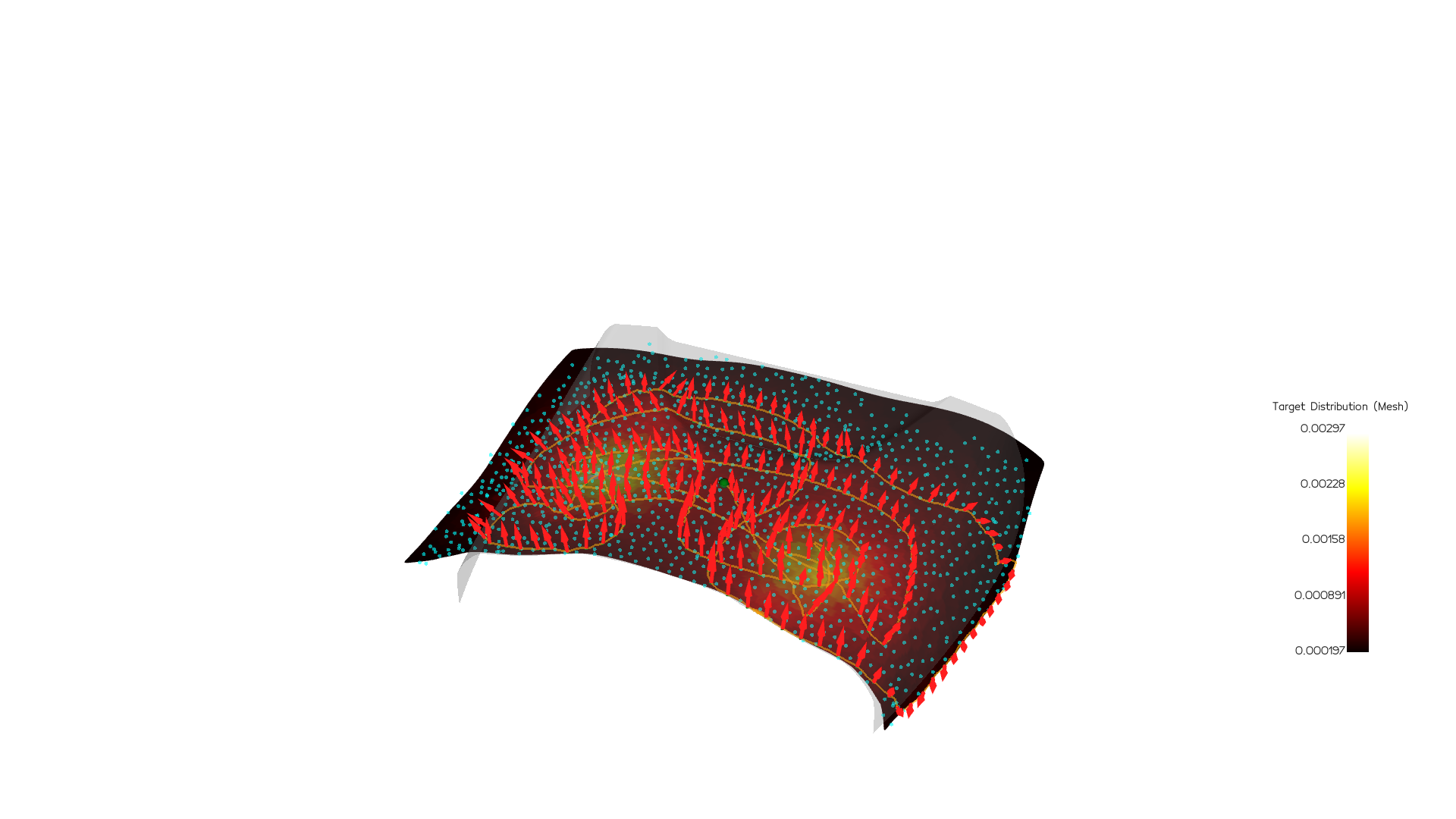} &
        \includegraphics[width=0.17\textwidth,clip,trim=600 180 600 380]{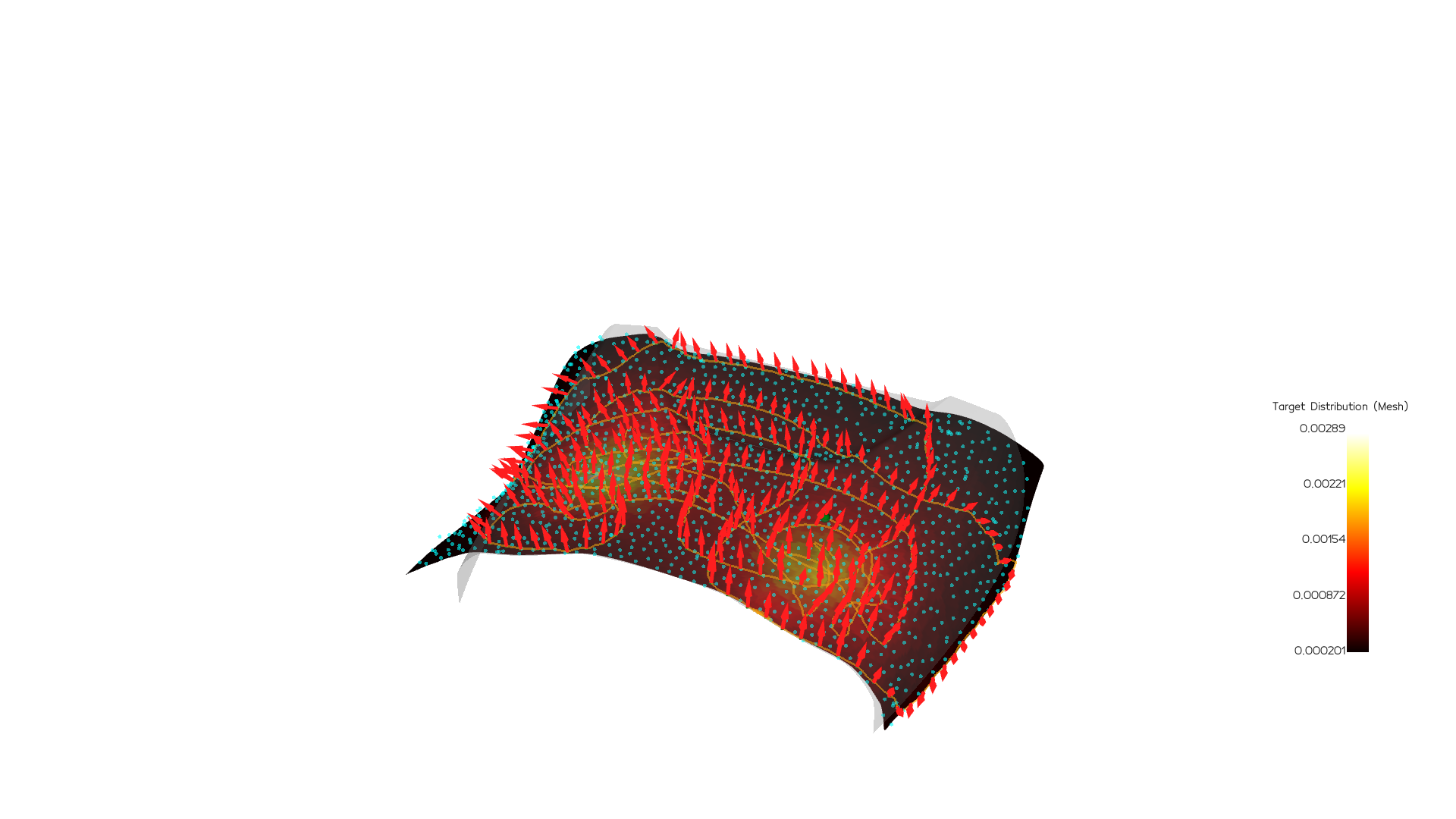} \\
        &
        \includegraphics[width=0.17\textwidth,clip,trim=600 180 600 380]{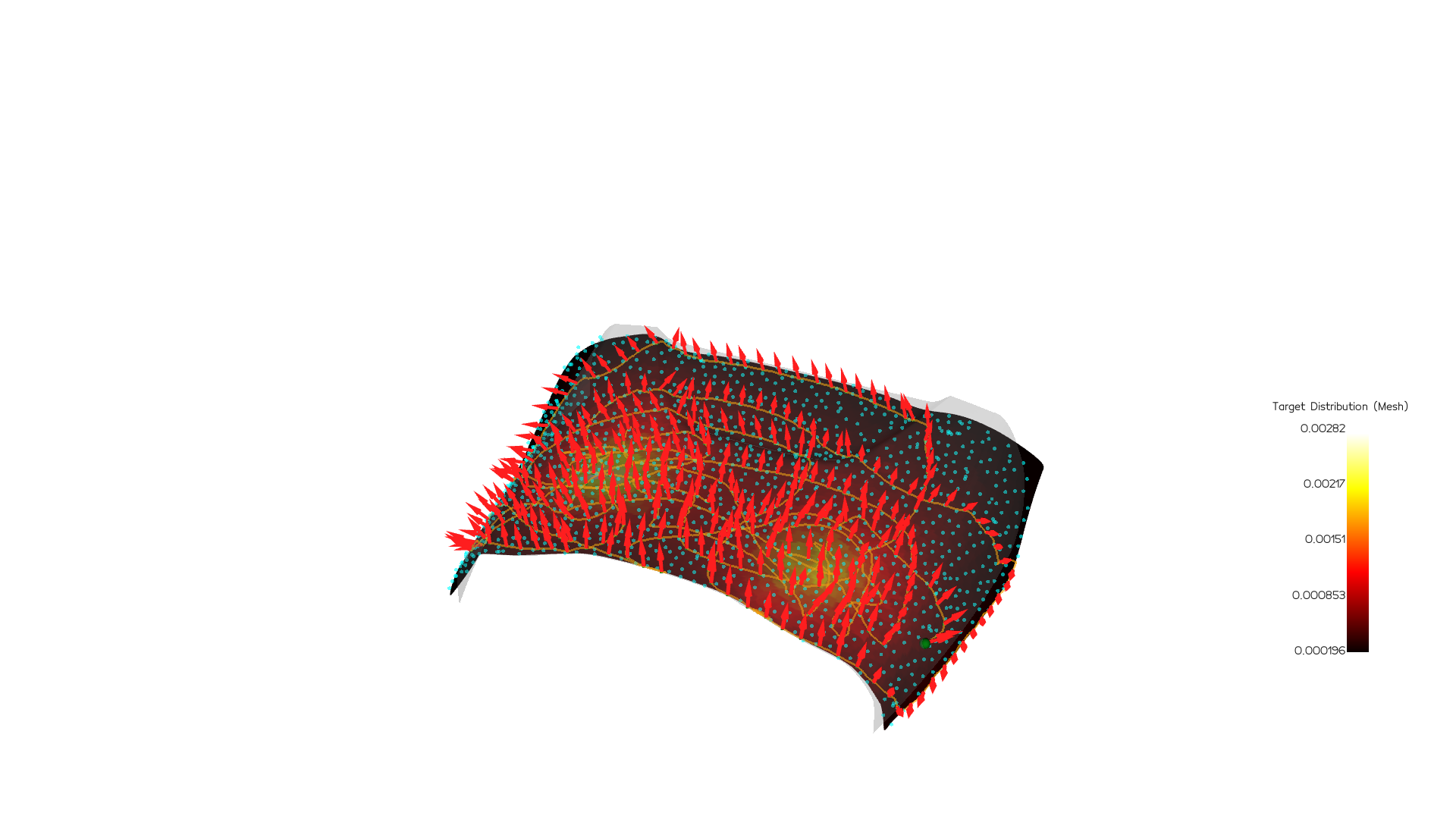} &
        \includegraphics[width=0.17\textwidth,clip,trim=600 180 600 380]{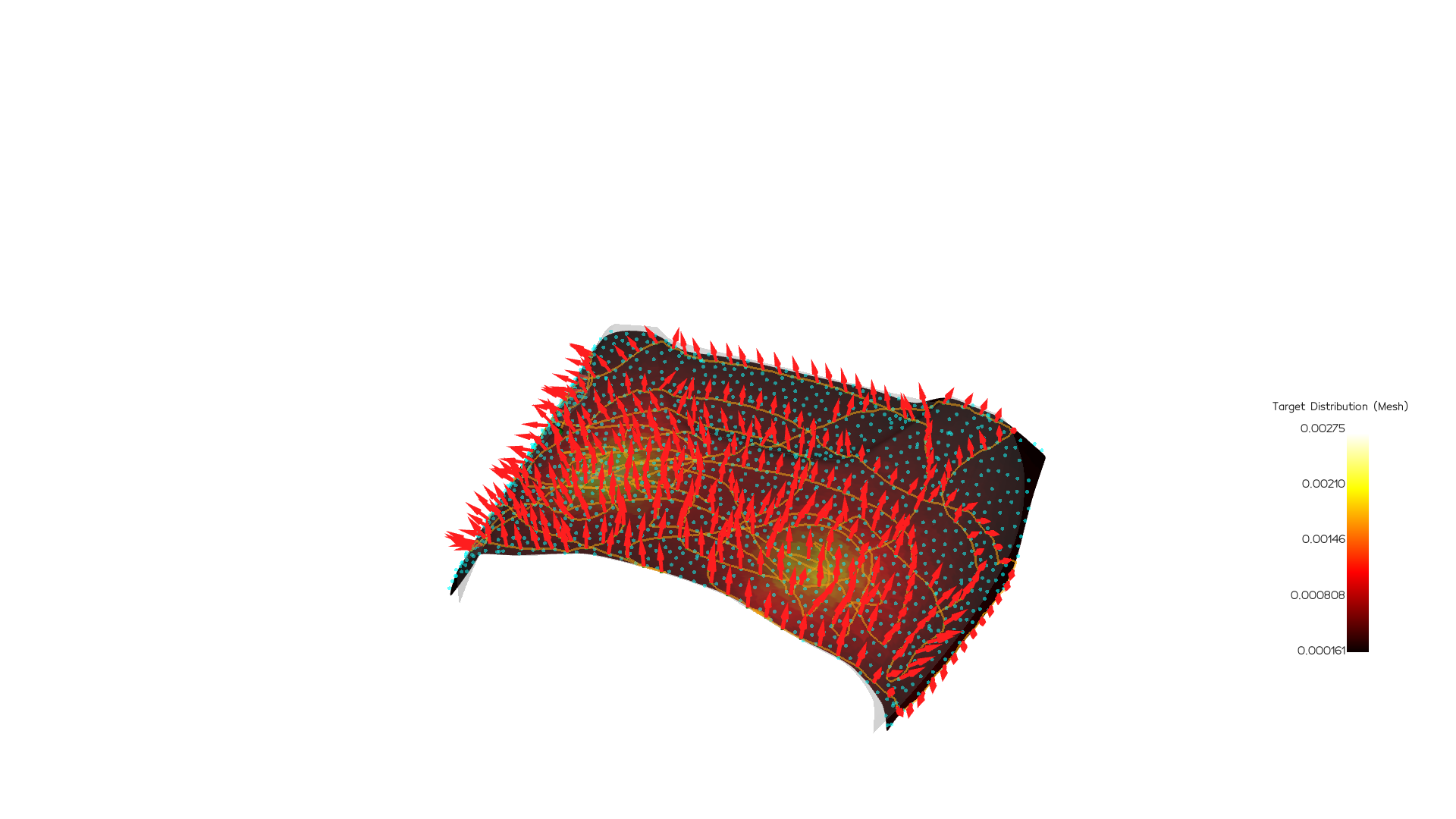} &
        \includegraphics[width=0.17\textwidth,clip,trim=600 180 600 380]{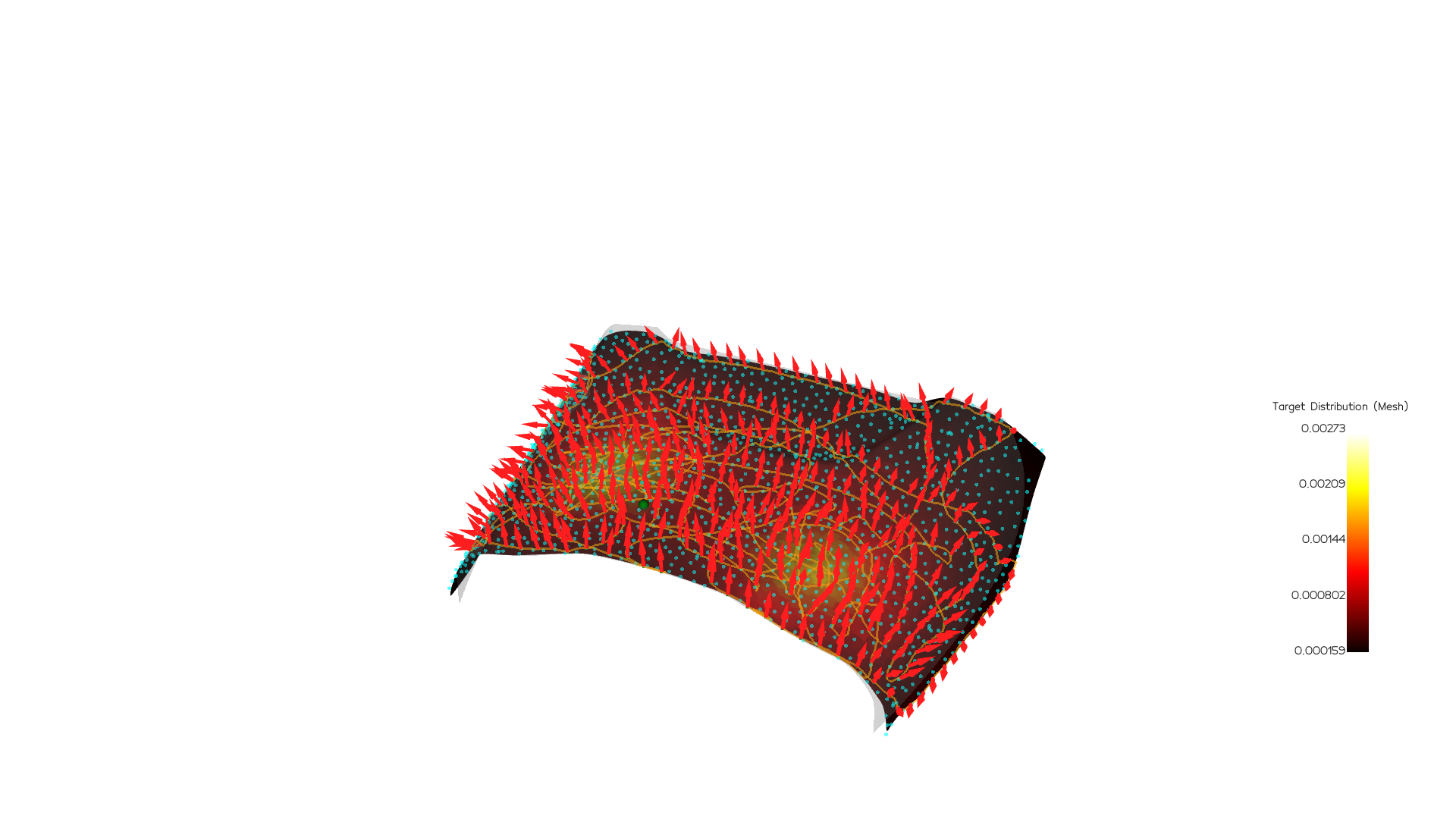} &
        \includegraphics[width=0.17\textwidth,clip,trim=600 180 600 380]{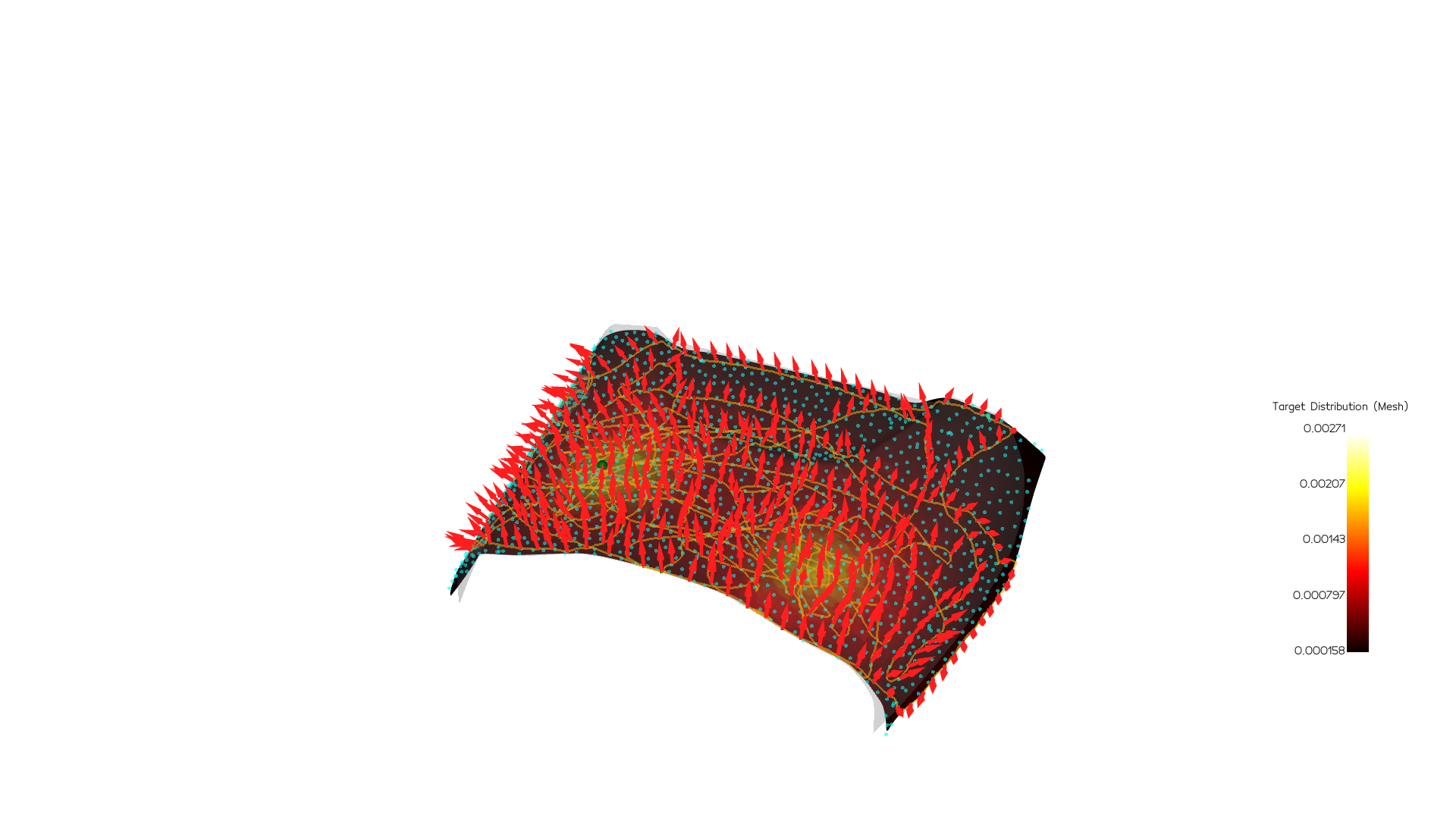} &
        \includegraphics[width=0.17\textwidth,clip,trim=600 180 600 380]{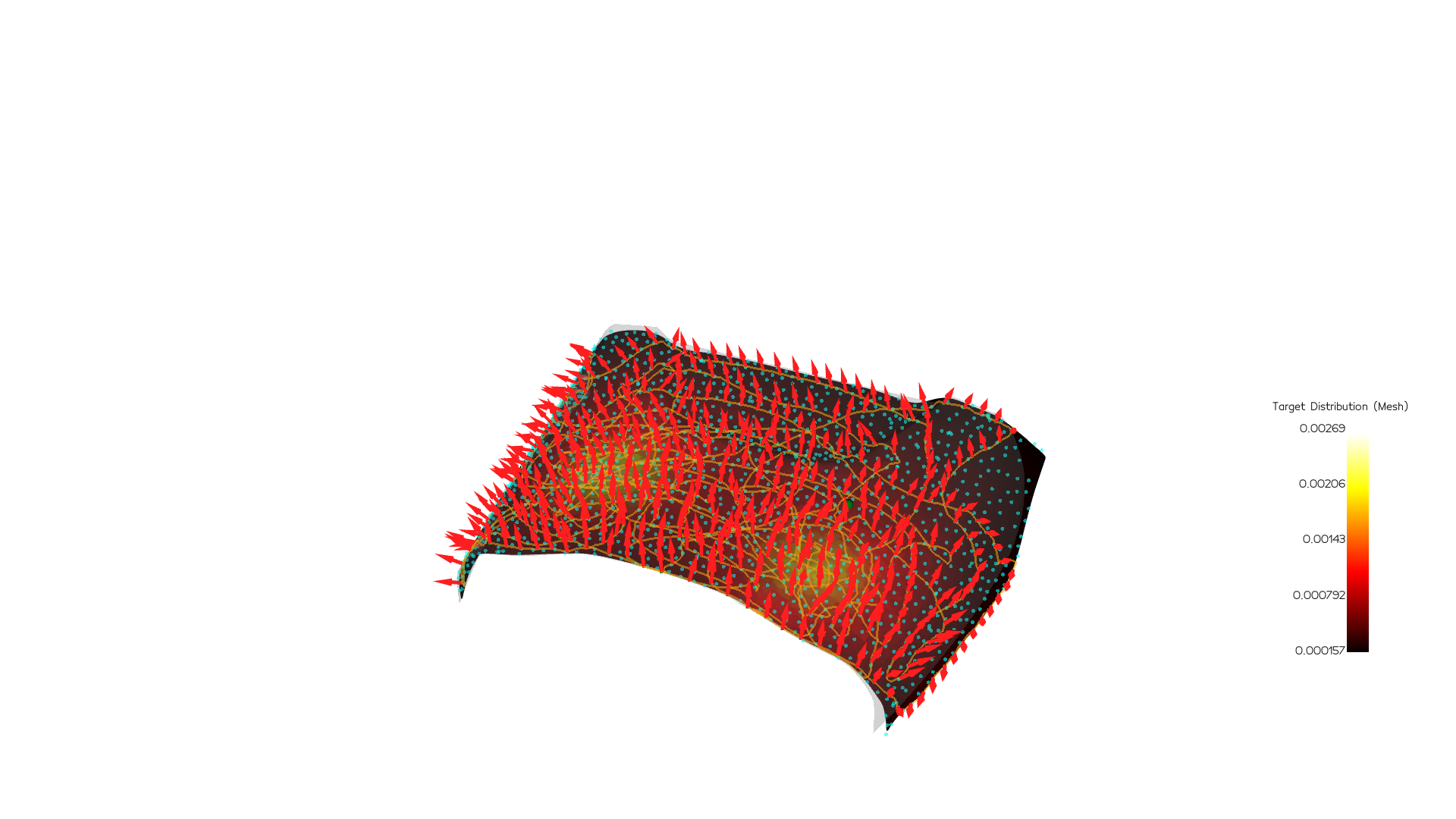} \\[0.1em]
        \cline{2-6}\\[-0.7em]
        \multirow{2}{*}[3em]{\rotatebox{90}{\small Uncertainty}} &
        \includegraphics[width=0.17\textwidth,clip,trim=600 180 600 380]{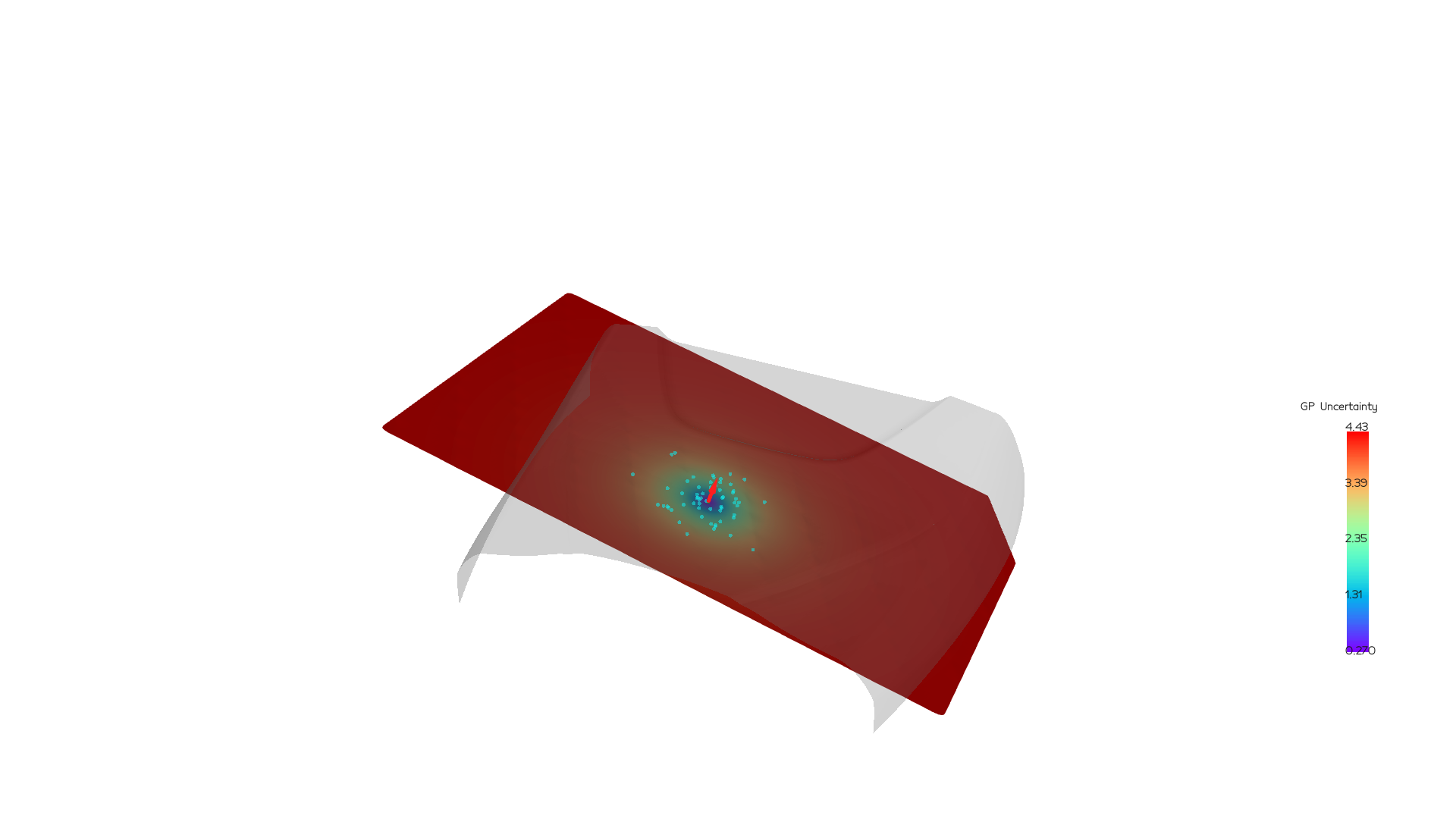} &
        \includegraphics[width=0.17\textwidth,clip,trim=600 180 600 380]{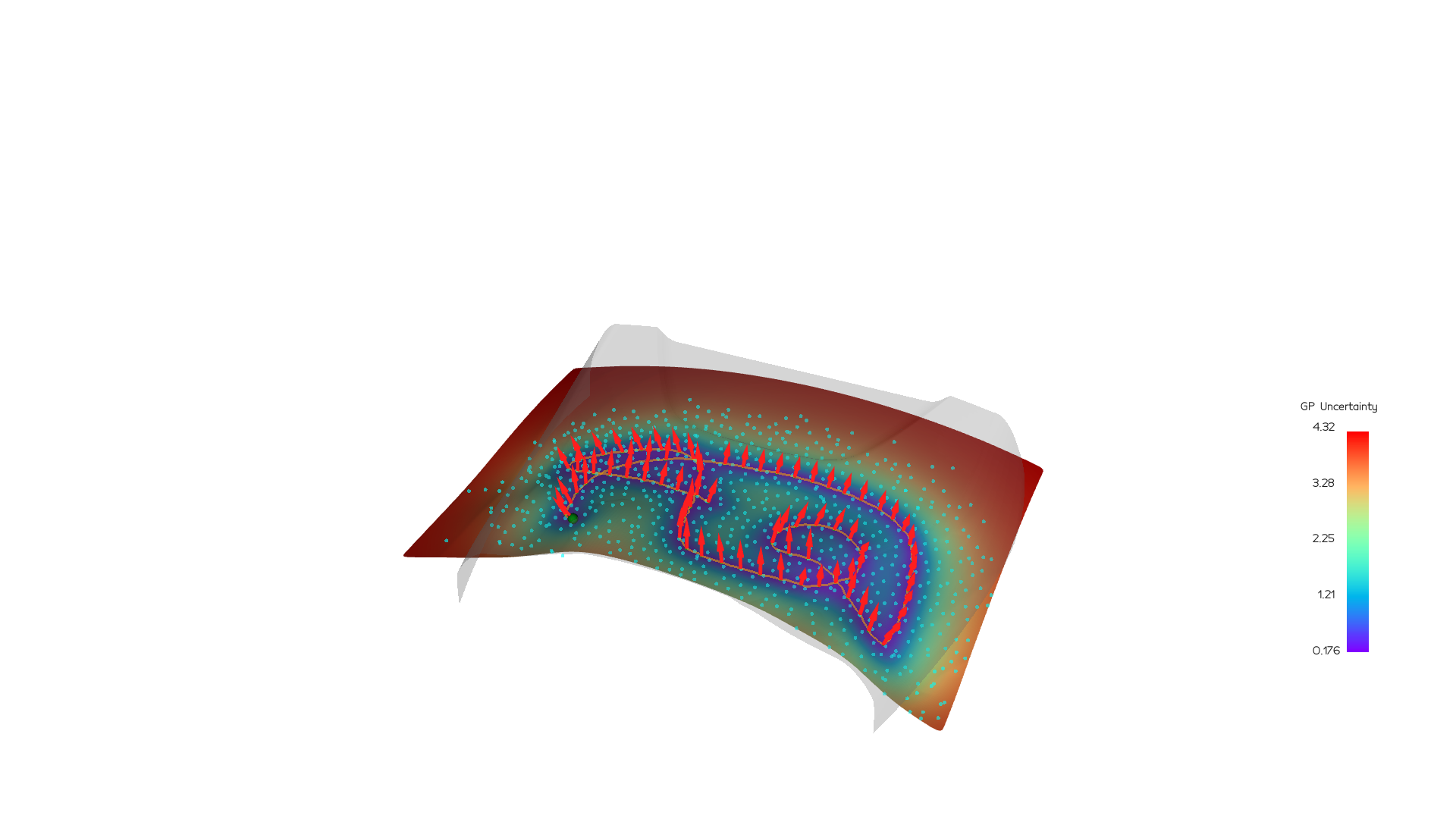} &
        \includegraphics[width=0.17\textwidth,clip,trim=600 180 600 380]{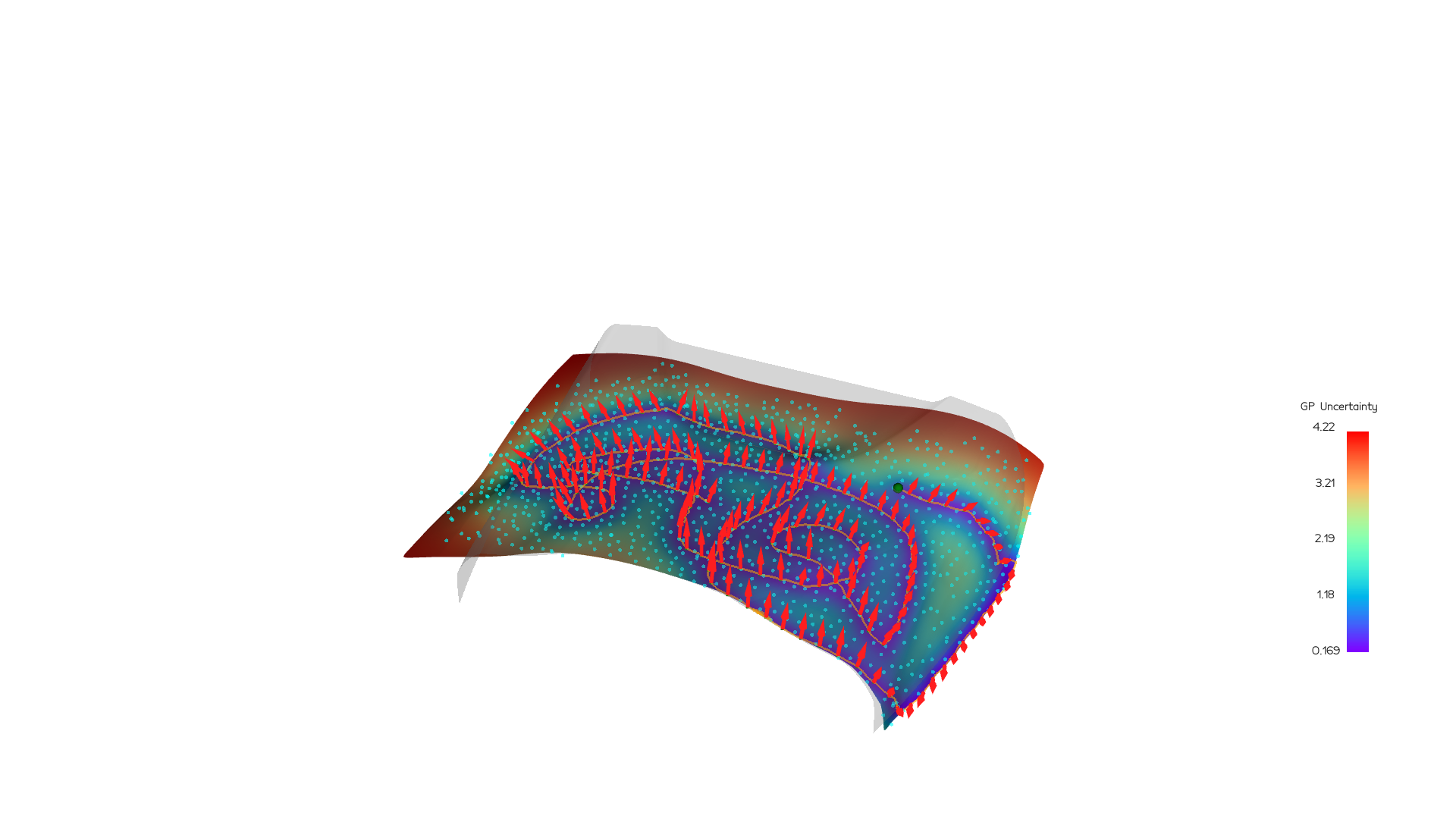} &
        \includegraphics[width=0.17\textwidth,clip,trim=600 180 600 380]{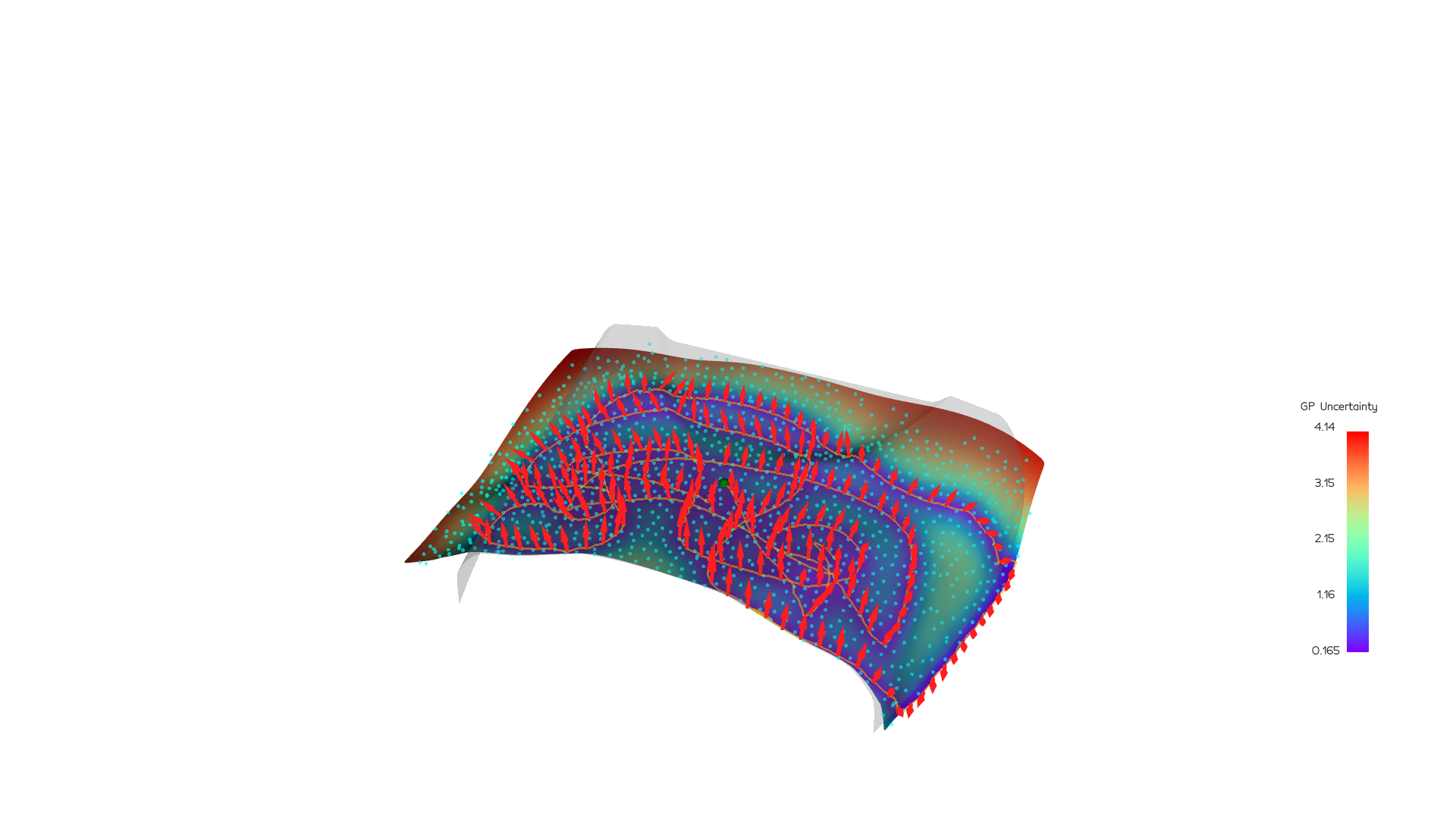} &
        \includegraphics[width=0.17\textwidth,clip,trim=600 180 600 380]{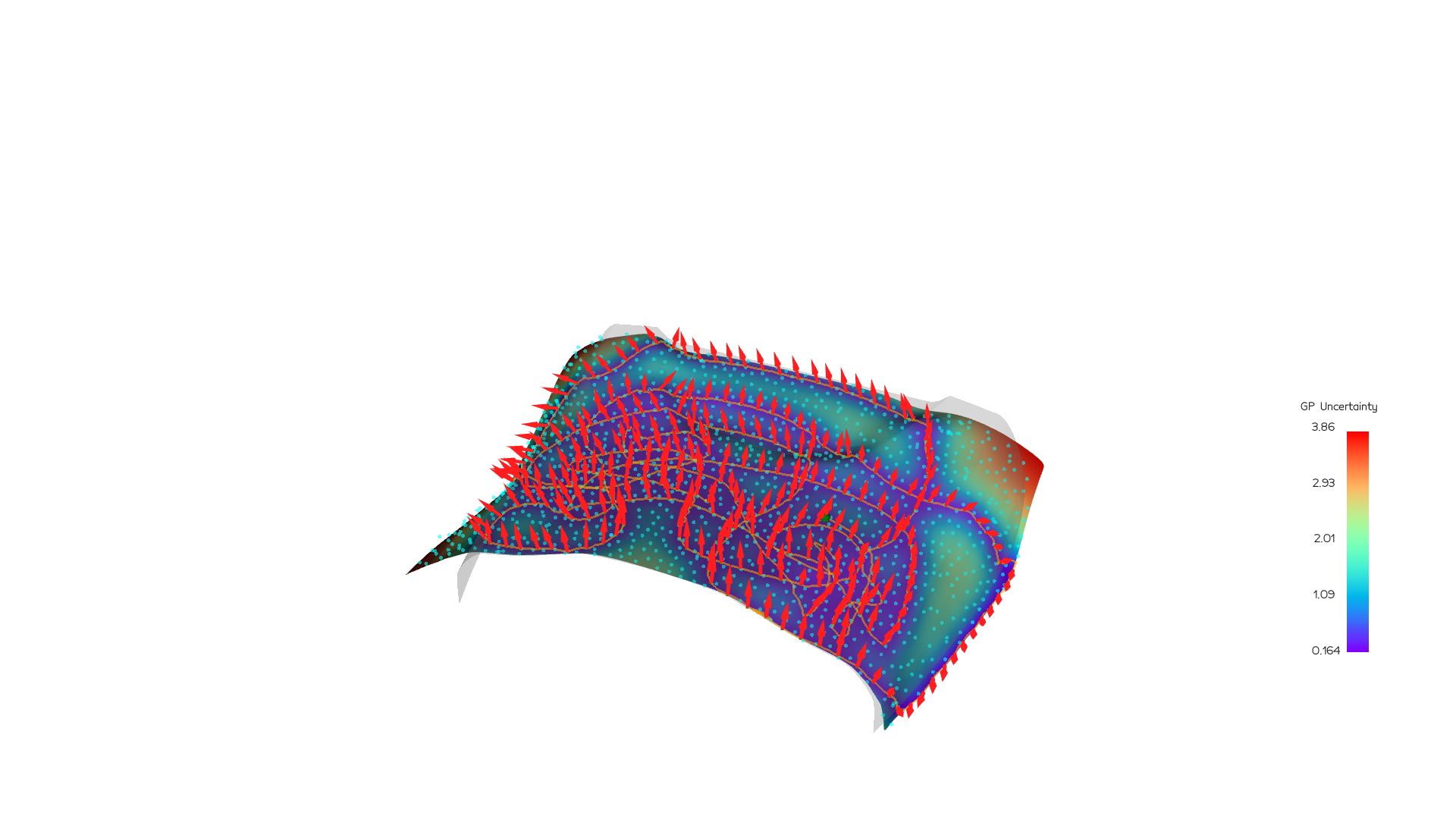} \\
        &
        \includegraphics[width=0.17\textwidth,clip,trim=600 180 600 380]{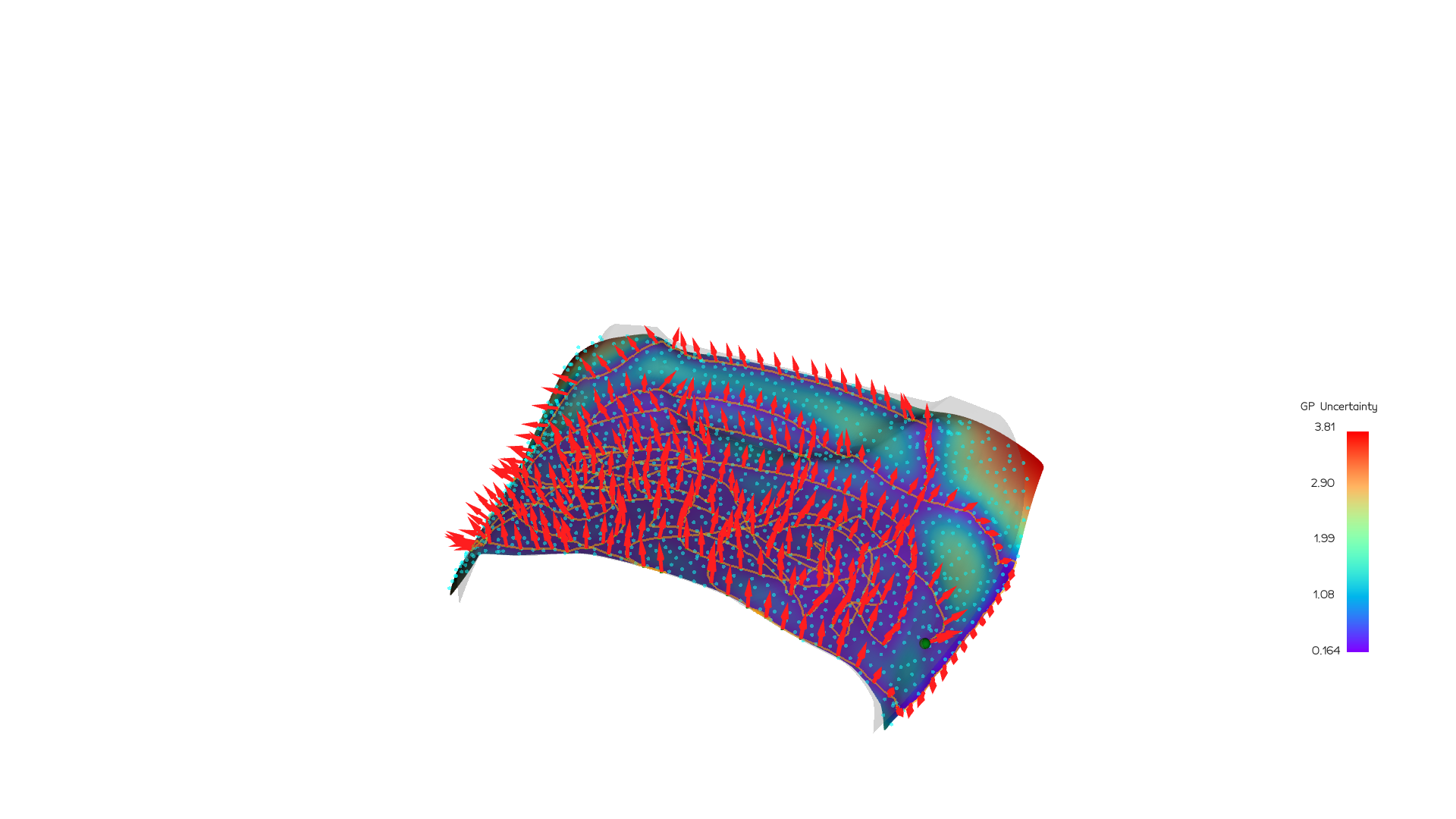} &
        \includegraphics[width=0.17\textwidth,clip,trim=600 180 600 380]{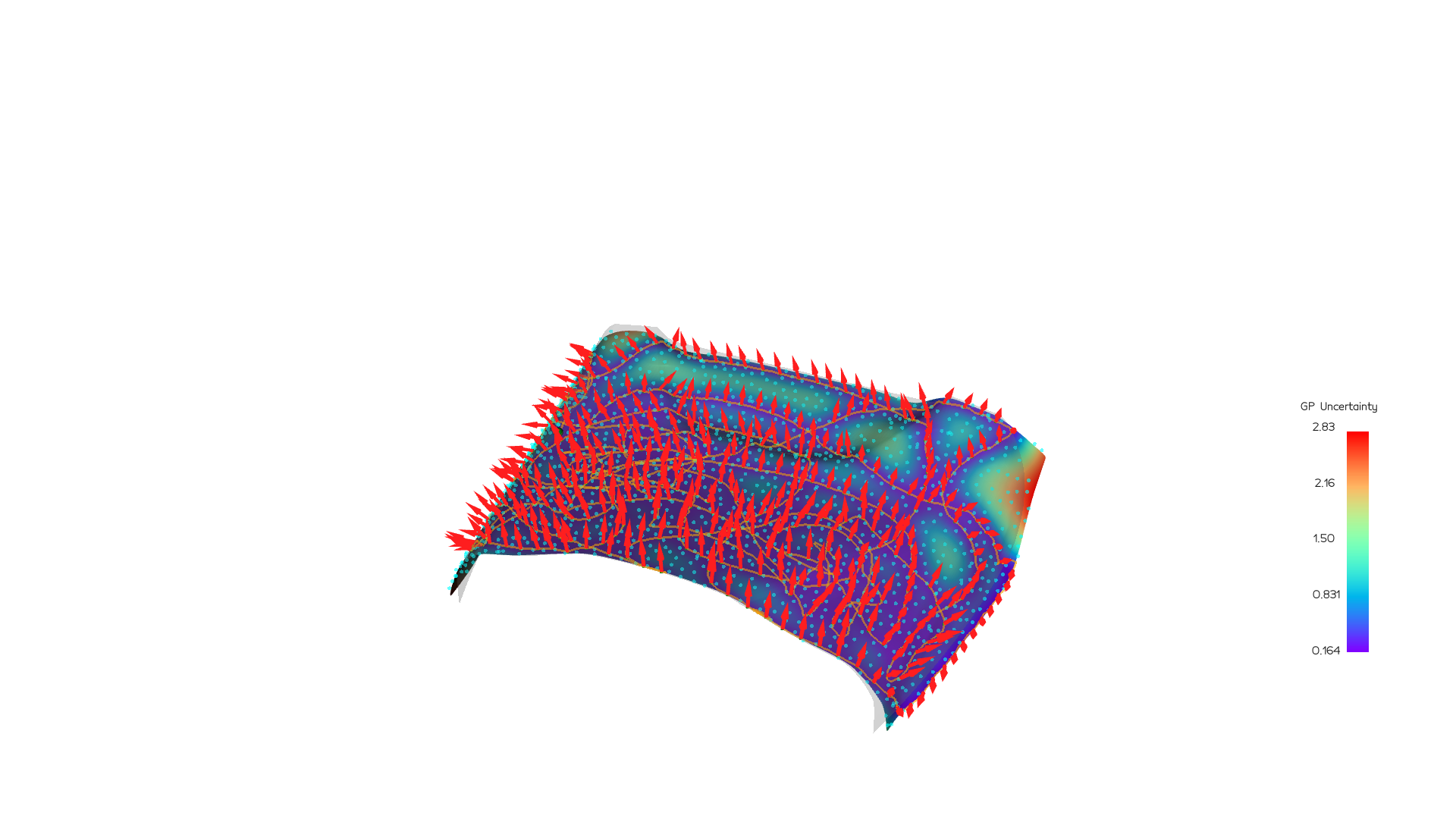} &
        \includegraphics[width=0.17\textwidth,clip,trim=600 180 600 380]{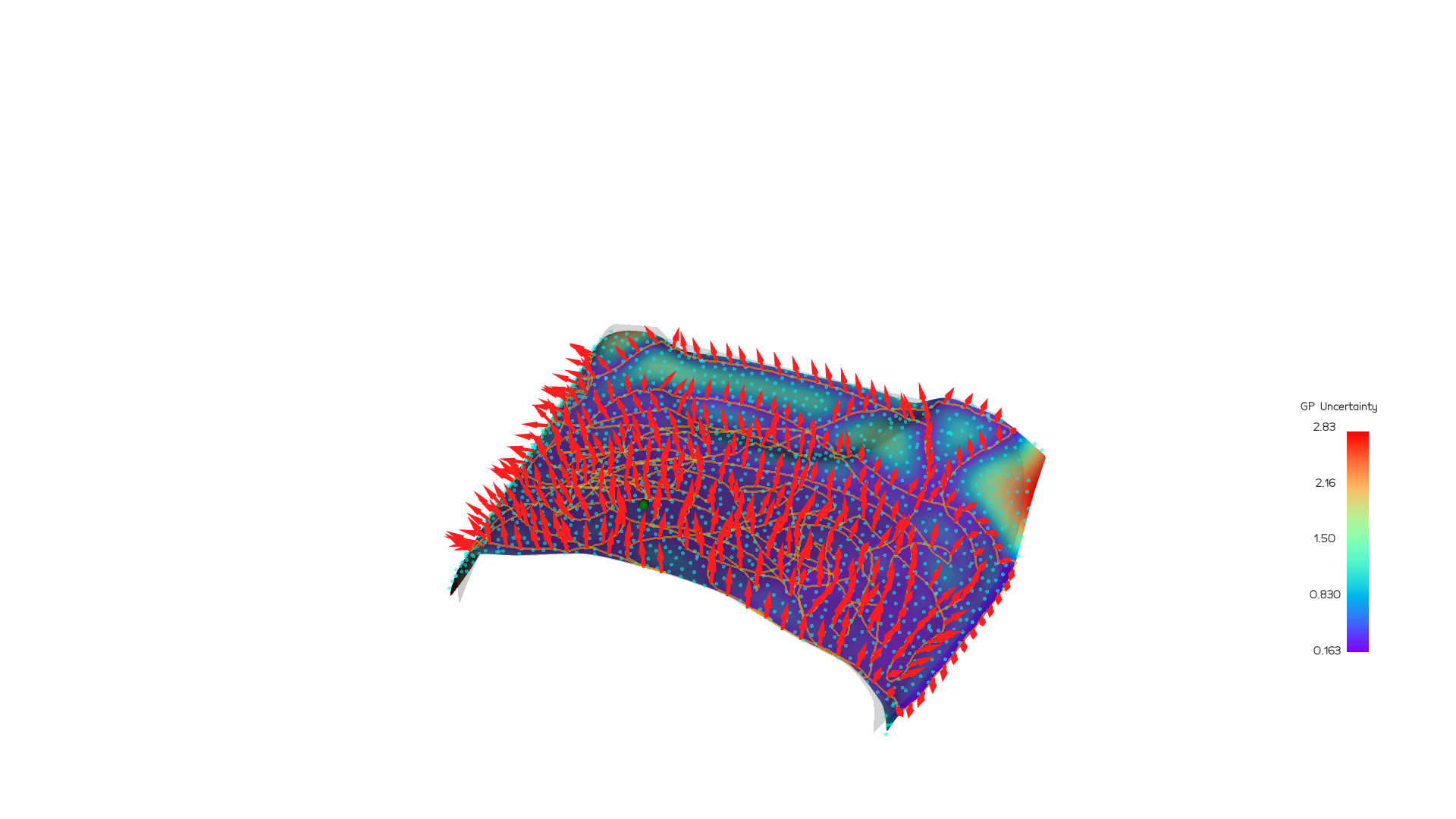} &
        \includegraphics[width=0.17\textwidth,clip,trim=600 180 600 380]{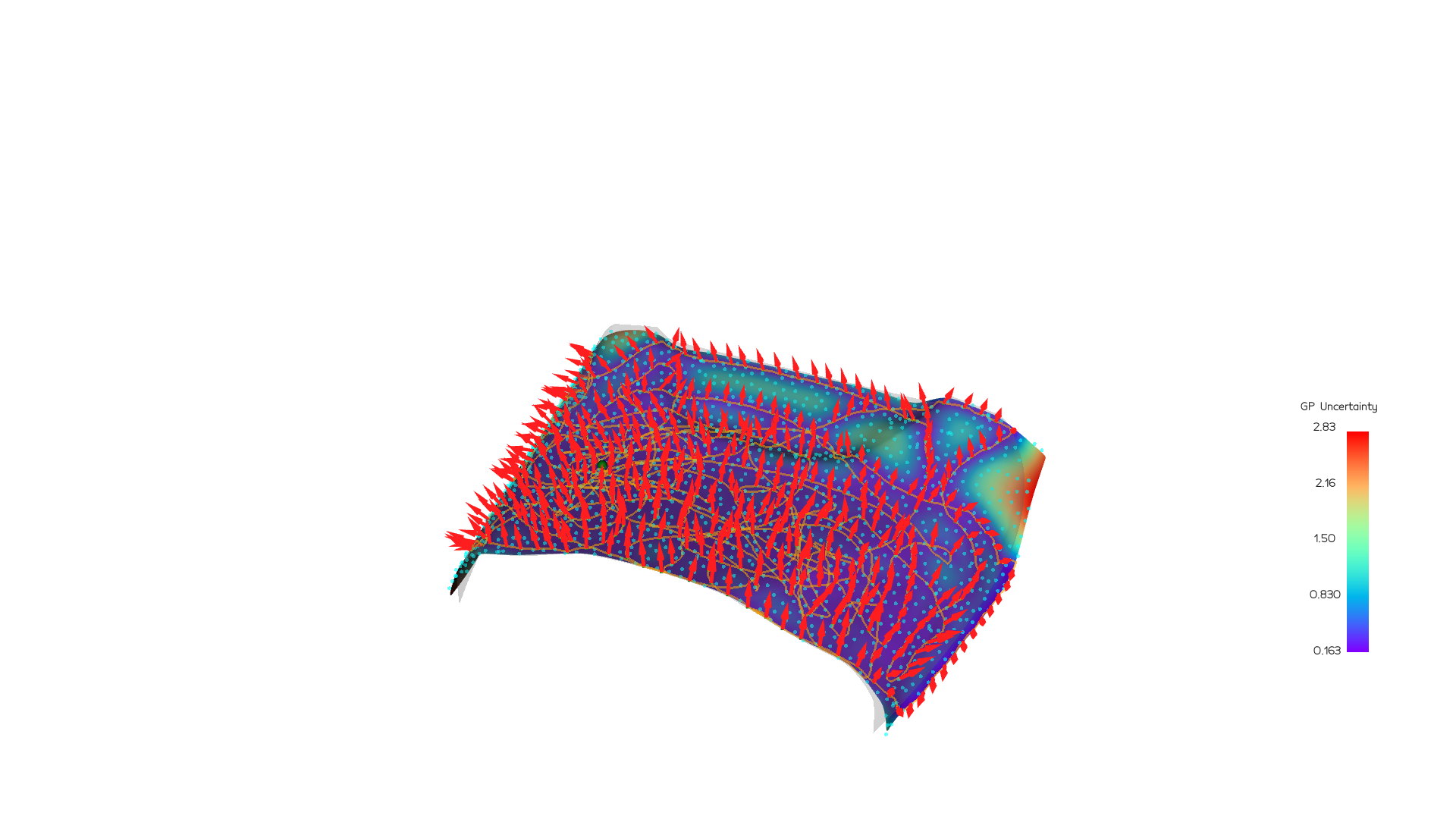} &
        \includegraphics[width=0.17\textwidth,clip,trim=600 180 600 380]{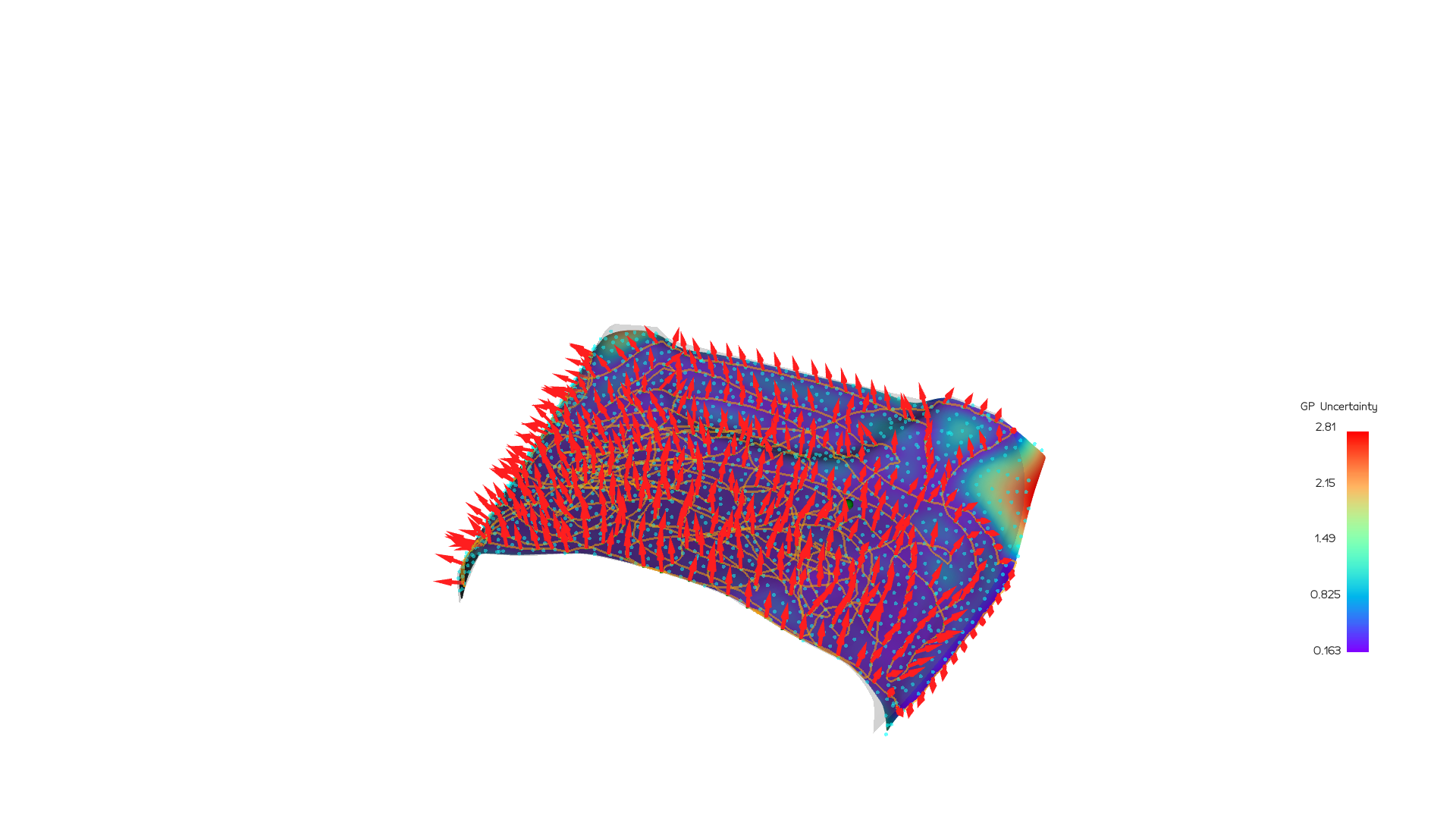} \\
        & \multicolumn{5}{c}{\scriptsize Time steps: 1, 334, 667, 1001, 1334 (top rows) | 1667, 2000, 2334, 2667, 3000 (bottom rows)} \\
    \end{tabular}
    \caption{Backpanel Experiment Progression over 3000 time steps (linearly interpolated). \textbf{Coverage:} Achieved exploration pattern showing systematic coverage across the complex backpanel surface with two target regions. \textbf{Target:} Task-specific target distribution from two heat sources, one on each side of the panel. \textbf{Uncertainty:} GPIS prediction uncertainty, with warmer colors indicating higher uncertainty that progressively reduces in explored areas.}
    \label{fig:backpanel_coverage_progression}
    \label{fig:backpanel_target_progression}
    \label{fig:backpanel_uncertainty_progression}
\end{figure*}

This section presents the visual progression of the coverage distribution, target distribution, and GPIS prediction uncertainty for the bunny (\Cref{fig:bunny_coverage_progression}), chair (\Cref{fig:chair_coverage_progression}), and backpanel (\Cref{fig:backpanel_coverage_progression}) experiments, over 10000, 3000, and 3000 time steps, respectively. For the bunny and backpanel experiments, \Cref{fig:ammount_plots} additionally shows the evolution of observation points, point cloud, and trajectory length.

\end{document}

%% file: preamble_ijrr.tex
\usepackage{amsmath,amssymb,amsfonts}
\usepackage{bm}
\usepackage{times}
\usepackage{multicol}
\usepackage{algorithmicx}
\usepackage{algorithm}
\usepackage{algpseudocode}
\usepackage{multirow}
\usepackage{pgfplots}
\pgfplotsset{compat=1.18}
\usepgfplotslibrary{groupplots}
\usepgfplotslibrary{fillbetween}
\usepackage{float}
\usepackage[x11names]{xcolor}

\usepackage{cleveref}
\Crefname{equation}{Eq.}{Eqs.}%
\Crefname{figure}{Fig.}{Figs.}

\graphicspath{{images-src/}{images-bin/}}%

\input{./style/dlr_colors}

\usepackage{tikz}
\usetikzlibrary{arrows.meta, positioning, shapes.geometric, fit}
\usepackage{subfigure}

\usepackage[textsize=scriptsize]{todonotes} 

%% file: style/dlr_colors.tex
\definecolor{dlrblue5}{HTML}{d1e8fa}
\definecolor{dlrgreen1}{HTML}{82a043}
\definecolor{dlrgreen5}{HTML}{e6eaaf}